%% file: main.tex
\documentclass[10pt,letterpaper,twocolumn]{article}

\usepackage[pagenumbers]{config/cvpr} 

\input{config/packages.tex}
\input{config/preamble}

\def\paperID{4195} 
\def\confName{CVPR}
\def\confYear{2024}

\title{\ourmethod: Neural Surface Reconstruction \\ via Multi-View Normal Integration}
\input{config/authors}

\begin{document}
\maketitle
\input{sections/00_abstract.tex}
\input{sections/01_intro.tex}

\input{sections/02_related_work.tex}

\input{sections/03_method.tex}
\input{sections/04_exp}

\clearpage
{
    \small
    \bibliographystyle{config/ieeenat_fullname}
    \bibliography{config/egbib}
}

\input{sections/X_suppl}

\end{document}

%% file: config/packages.tex
\usepackage{graphicx}
\usepackage{amsmath, mathtools, empheq, esint}
\usepackage{amssymb}
\usepackage{booktabs}
\usepackage{wrapfig}
\usepackage{lipsum}
\usepackage{xspace}
\usepackage{amsthm}
\usepackage{arydshln}
\usepackage{siunitx}
\usepackage{xcolor,colortbl}
\usepackage{multirow}
\usepackage{tabularx}
\graphicspath{{images/}}

\usepackage{bm}
\usepackage{url}
\usepackage{pgf}
\usepackage{collcell}
\usepackage{xstring} 
\usepackage{xfp}     
\usepackage[toc,page]{appendix}

\definecolor{cvprblue}{rgb}{0.21,0.49,0.74}
\usepackage[pagebackref,breaklinks,colorlinks,citecolor=cvprblue]{hyperref}

\usepackage[capitalize]{cleveref}
\crefname{section}{Sec.}{Secs.}
\Crefname{section}{Section}{Sections}
\Crefname{table}{Table}{Tables}
\crefname{table}{Tab.}{Tabs.}

%% file: config/preamble.tex
\newcounter{todos}
\AtEndDocument{\ifnum\value{todos}>0 \PackageWarning{TODOS}{There are \arabic{todos} todos left in this paper! Fix them before submitting the paper!} \fi}

\newcommand{\camera}[1]{{\color{black} #1}}
\newcommand{\suppl}{\camera{SupMat.}\xspace}

\newcommand{\norm}[1]{\lVert#1\rVert}

\DeclareMathOperator{\bce}{BCE}

\newcommand{\V}[1]{\ensuremath{\mathbf{#1}}}

\newcommand{\defeq}{\vcentcolon=}
\newcommand{\abs}[1]{\lvert#1\rvert}

\newcommand{\opacity}{\ensuremath{\alpha}\xspace}
\newcommand{\transmittance}{\ensuremath{T}\xspace}
\newcommand{\sharpness}{\ensuremath{s}\xspace}
\newcommand{\sigmoid}{\ensuremath{\phi_\sharpness}\xspace}

\newcommand{\pixel}{\V{p}\xspace}

\newcommand{\surface}{\ensuremath{\mathcal{M}}\xspace}

\newcommand{\point}{\ensuremath{\V{x}}\xspace}
\newcommand{\normal}{\ensuremath{\V{n}}\xspace}

\newcommand{\distToCameraCenter}{\ensuremath{t}\xspace}

\newcommand{\cameraCenter}{\ensuremath{\V{o}}\xspace}
\newcommand{\viewDirection}{\ensuremath{\V{v}}\xspace}

\newcommand{\loss}{\mathcal{L}\xspace}

\newcommand{\thresholdFscore}{\ensuremath{\tau_F}\xspace}
\newcommand{\thresholdOcc}{\ensuremath{\tau_o}\xspace}

\newcommand{\pointOne}{\ensuremath{\point_{1}}\xspace}
\newcommand{\pointTwo}{\ensuremath{\point_{2}}\xspace}
\newcommand{\pointsetOne}{\ensuremath{\chi_{1}}\xspace}
\newcommand{\pointsetTwo}{\ensuremath{\chi_{2}}\xspace}
\newcommand{\fscoreThreshold}{\ensuremath{\tau}\xspace}
\newcommand{\chamferDist}{\ensuremath{d(\pointsetOne, \pointsetTwo)}\xspace}
\newcommand{\precision}{\ensuremath{\mathcal{P}}\xspace}
\newcommand{\recall}{\ensuremath{\mathcal{R}}\xspace}
\newcommand{\fscore}{\ensuremath{\mathcal{F}}\xspace}

\newcommand{\nerf}{\mbox{NeRF}~\cite{mildenhall2020nerf}\xspace}
\newcommand{\neustwo}{\mbox{NeuS2}~\cite{neus2Wang2023}\xspace}
\newcommand{\neus}{\mbox{NeuS}~\cite{wang2021neus}\xspace}
\newcommand{\idr}{\mbox{IDR}~\cite{idr2020multiview}\xspace}

\newcommand{\psnerf}{\mbox{PS-NeRF}~\cite{yang2022psnerf}\xspace}
\newcommand{\sdps}{\mbox{SDPS}~\cite{chen2019SDPS_Net}\xspace}
\newcommand{\uanet}{\mbox{UA-MVPS}~\cite{kaya2022uncertainty}\xspace}
\newcommand{\rmvps}{\mbox{R-MVPS}~\cite{park2016robust}\xspace}
\newcommand{\bmvps}{\mbox{B-MVPS}~\cite{li2020multi}\xspace}
\newcommand{\volsdf}{\mbox{VolSDF}~\cite{volsdf2021yariv}\xspace}
\newcommand{\unisurf}{\mbox{UNISURF}~\cite{unisurf2021ICCV}\xspace}
\newcommand{\diligentmv}{\mbox{DiLiGenT-MV}~\cite{li2020multi}\xspace}

\newcommand{\mvpsnet}{\mbox{MVPSNet}~\cite{Zhao_2023_ICCV}\xspace}
\newcommand{\mvas}{\mbox{MVAS}~\cite{Cao_2023_CVPR}\xspace}
\newcommand{\nerfacc}{\mbox{NerfAcc}~\cite{nerfacc_2023_ICCV}\xspace}

\newcommand{\sdmunips}{\mbox{SDM-UniPS}~\cite{ikehata2023sdmunips}\xspace}
\newcommand{\ourmethod}{\mbox{SuperNormal}\xspace}

\newcommand{\colorbar}[2]{
	\begin{tabular}[b]{@{}l}
        #2 \\
		\includegraphics[height=#1\linewidth, width=0.5em]{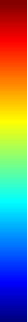} \\
        $0$
	\end{tabular}
}

\newcommand{\ColCell}[1]{%
  \IfDecimal{#1}{%
    \pgfmathparse{100-#1*100/15} 
    \edef\opacity{\pgfmathresult}
    \cellcolor{orange!\opacity!white}{#1}%
  }{%
    #1
  }%
}

\newcolumntype{R}{>{\collectcell\ColCell}r<{\endcollectcell}}

%% file: config/authors.tex
\author{
Xu Cao 
\quad
Takafumi Taketomi
\\
CyberAgent
\\
{\tt \small \{xu\_cao, taketomi\_takafumi\}@cyberagent.co.jp}
}

%% file: sections/00_abstract.tex
\begin{abstract}
We present \ourmethod, a fast, high-fidelity approach to multi-view 3D reconstruction using surface normal maps. 
With a few minutes, \ourmethod produces detailed surfaces on par with 3D scanners.
We harness volume rendering to optimize a neural signed distance function~(SDF) powered by multi-resolution hash encoding.
To accelerate training, we propose directional finite difference and patch-based ray marching to approximate the SDF gradients numerically.
While not compromising reconstruction quality, this strategy is nearly twice as efficient as analytical gradients and about three times faster than axis-aligned finite difference.
Experiments on the benchmark dataset demonstrate the superiority of \ourmethod in efficiency and accuracy compared to existing multi-view photometric stereo methods.
On our captured objects, \ourmethod produces more fine-grained geometry than recent neural 3D reconstruction methods.
\end{abstract}

%% file: sections/01_intro.tex
\section{Introduction}
\label{sec.intro}
\begin{figure}
    \centering
    \includegraphics[width=\linewidth]{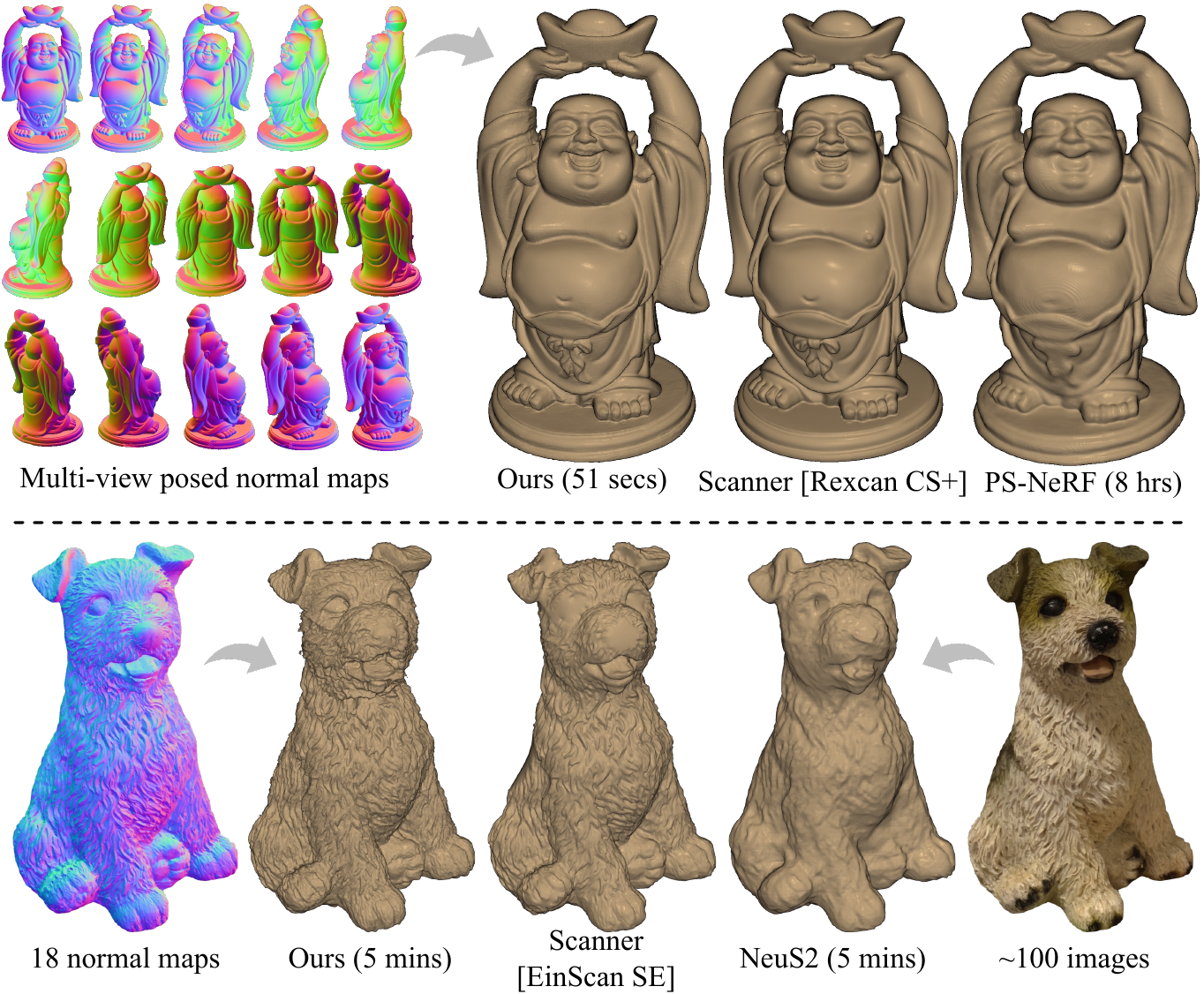}
    \caption{
    \textbf{(Top)} From multi-view surface normal maps, \ourmethod recovers fine-grained surface details comparable to 3D scanners while being orders of magnitude faster than the existing MVPS method~\cite{yang2022psnerf}. \textbf{(Bottom)} Using fewer normal maps produces more faithful high-frequency surface detail than the MVS-based method~\cite{neus2Wang2023}, although both use multi-resolution hash encoding~\cite{muller2022ingp}.}
\label{fig.teaser1}
\end{figure}

Recovering high-quality 3D geometry of real-world objects from their multi-view images is a long-standing challenge in computer vision.
Recently, neural implicit surface-based methods have shown remarkable reconstruction results. 
Compared to traditional multi-view stereo (MVS) methods~\cite{schoenberger2016mvs}, neural methods are more robust to view-dependent observations and textureless surfaces~\cite{wang2021neus}. Furthermore, the reconstruction procedure has become highly efficient~\cite{neus2Wang2023} by introducing multi-resolution hash coding~\cite{muller2022ingp}. 
However, even though this spatial encoding allows fine-grained geometry to be represented, shape reconstruction from multi-view images tends to smooth out high-frequency surface details, as shown in \cref{fig.teaser1} bottom right.

Multi-view photometric stereo (MVPS), on the other hand, aims to recover pixel-level high-frequency surface detail by introducing additional lighting conditions during the image acquisition~\cite{hernandez2008multiview}. 
In a typical workflow, a surface normal map is first recovered for each view, recording per-pixel surface orientation information. 
The normal maps estimated at different viewpoints are then fused into a 3D model, also known as multi-view normal integration~\cite{chang2007multiview}. 
However, existing normal fusion methods struggle to reflect the details of the normal maps in the shape of the recovered 3D model. 
Moreover, the reconstruction process takes hours even when only a few low-resolution normal maps are used~\cite{yang2022psnerf,Cao_2023_CVPR}. 
Due to the lack of a fast and accurate multi-view normal fusion method, current multi-view photometric stereo results remain unsatisfactory.

This paper presents \ourmethod to unite the best of both worlds.
Normal maps with pixel-level surface details are utilized to exploit the expressive power and efficiency of multi-resolution hash encoding-powered neural implicit surface.
Using normal maps alone avoids per-scene reflectance modeling and optimization, thus improving training efficiency compared to using color images.
In a few minutes, our method can produce fine-grained geometry on par with commercial 3D scanners. 

\begin{figure}
    \centering
    \includegraphics[width=\linewidth]{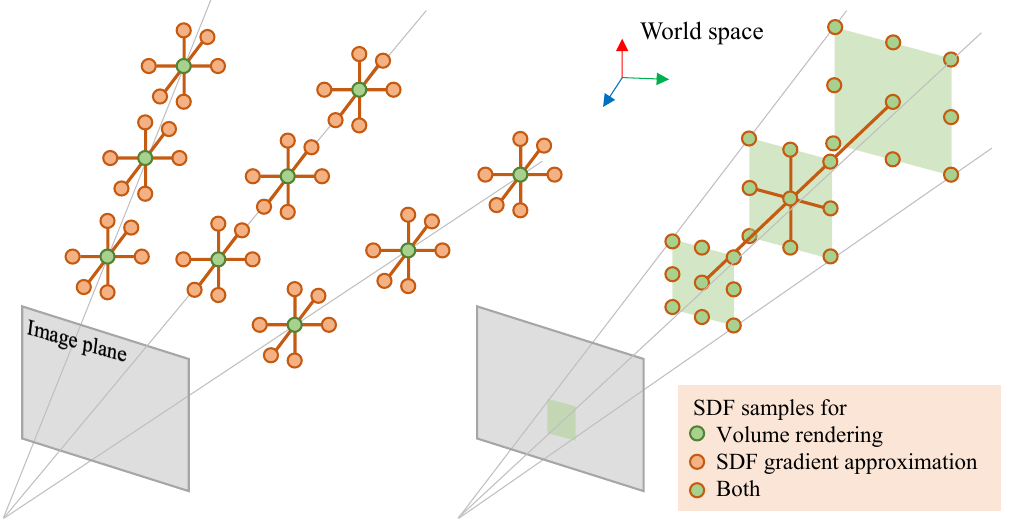}
    \caption{\textbf{(Left)} In SDF-based volume rendering, approximating SDF gradients using axis-aligned finite difference~\cite{neuralangelo2023CVPR,instantNSR} requires additional SDF samples at neighboring positions of volume rendering points, increasing computational complexity and training time. \textbf{(Right)} Our strategy of patch-based ray marching and directional finite difference effectively uses all SDF samples for both SDF gradient approximation and volume rendering.}
    \label{fig.FD_vs_DFD}
\end{figure}

Specifically, we use volume rendering to optimize a neural signed distance function (SDF) such that its rendered normal maps are consistent with input normal maps.
To further improve the training efficiency, we propose a directional finite difference and patch-based ray marching strategy to approximate the SDF gradients numerically.
This strategy kills two birds with one stone: SDF values evaluated at sampled points are \emph{all} used both for SDF-based volume rendering and SDF gradient approximation.
This way, we avoid second-order derivative computation during training.
Unlike traditional axis-aligned finite difference, directional finite difference avoids redundant SDF evaluations (\cref{fig.FD_vs_DFD}).
Without compromising reconstruction quality, our method is nearly twice as fast as using analytical gradients and about three times faster than finite difference.

Compared to existing neural MVPS methods, our method is two orders of magnitude faster and recovers better surface details (\cref{fig.teaser1} above).
We also capture objects with complex geometry and show that our method recovers more faithful surface details than the recent neural 3D reconstruction method~\cite{neus2Wang2023} (\cref{fig.teaser1} bottom).
Taking advantage of recent advances in photometric stereo, we can estimate high-quality normal maps from images captured under common lighting conditions with inexpensive equipment~\cite{ikehata2023sdmunips}.
Combined with \ourmethod, high-fidelity 3D objects can be reconstructed using inexpensive equipment in widely accessible environments.

In summary, this paper's key contributions are:
\begin{itemize}
\setlength{\itemsep}{0.2em}
\setlength{\parskip}{0.2em}
\item An effective neural 3D reconstruction approach using multi-view normal maps, enhancing the reconstruction quality to near 3D scanner levels;
\item An efficient method for computing numerical gradients in SDF-based neural rendering, speeding up the reconstruction process; and
\item A comprehensive evaluation using benchmark objects with MVPS approaches and an assessment using our captured objects with MVS approaches.
\end{itemize}

%% file: sections/02_related_work.tex
\section{Related work}
\label{sec.related_work}

\subsection{Multi-view photometric stereo (MVPS)}
MVPS aims at high-fidelity 3D reconstruction using multi-view observations under varying light conditions~\cite{hernandez2008multiview} (usually one-light-at-time (OLAT) images~\cite{li2020multi,rene2023}).

Existing methods differ in what geometry-related information is inferred in each view and how information from different views is fused into a complete 3D model.
Chang~\etal~\cite{chang2007multiview} proposes a level set method~\cite{osher2004level} to fuse multi-view normal maps into a volumetric representation of the shape.
\rmvps uses the multi-view normal maps to refine a coarse mesh estimated by multi-view stereo.
\bmvps initialize the shape as a sparse point cloud estimated by structure from motion (SfM), then propagates the known point spatially using the normal maps.

Recent MVPS methods have undergone a paradigm shift since the advent of \nerf by using a neural implicit scene representation, usually parameterized as a multi-layer perception (MLP).
\psnerf recovers a neural field based on \unisurf by regularizing the gradients of the neural field using the normal maps.
\uanet fuses the multi-view depth and normal maps into a neural SDF using neural inverse rendering while considering per-pixel confidence.
\mvas shows utilizing the azimuth components of multi-view normal maps is sufficient to recover the neural implicit surface.
While most methods only use images in the same view for normal map estimation, \mvpsnet utilizes neighboring views to infer per-view depth and normal maps jointly and applies screened Poisson surface reconstruction (sPSR)~\cite{kazhdan2013screened} to obtain the shape.

However, the reconstructed surfaces of these methods tend to be smoothed and do not reflect the high-frequency details of the input normal maps.
Our approach introduces multi-resolution hash encoding and directional finite difference to improve the implicit neural field's expressive power and training efficiency.

\subsection{Neural 3D reconstruction}
The seminal work \nerf and follow-up works~\cite{barron2021mipnerf,verbin2022refnerf,barron2023zipnerf} have demonstrated remarkable novel view synthesis quality by volume rendering a neural implicit field. 
For better geometry reconstruction quality, \idr imposes eikonal regularization~\cite{igr2020icml} such that the neural field approximates a valid SDF.
Later, the concurrent works \volsdf and \neus develop SDF-based volume rendering, which bridges SDF values to volume rendering weights and improves the geometry fidelity.
However, these methods require several hours to recover one object due to optimizing a deep coordinate-based MLP.

Multi-resolution hash encoding has been proposed to improve expressive power and accelerate the learning of neural implicit field~\cite{muller2022ingp}.
Several efforts have been made to incorporate this expressive spatial encoding into neural SDF learning.
Since the official CUDA implementation does not support the double propagation deduced by the eikonal regularization, PermutoSDF~\cite{rosu2023permutosdf} and NeuS2~\cite{neus2Wang2023} re-implement the CUDA programming.
InstantNSR~\cite{instantNSR} and Neuralangelo~\cite{neuralangelo2023CVPR} use finite difference instead of automatic differentiation to avoid double propagation but trade off the training speed.

We propose directional finite difference, a numerical way to compute SDF gradients that avoids double propagation and is even faster than automatic differentiation when combined with a tailored ray marching strategy.

\subsection{Neural 3D reconstruction with normals}
Normal maps have been used as auxiliary information to facilitate neural scene reconstruction in MonoSDF~\cite{Yu2022MonoSDF} and NeuRIS~\cite{wang2022neuris}.
These works target scene-level reconstruction, and normal maps help infer the geometry of textureless smooth regions (\eg, walls) where using color images is likely to break down.

Instead, our method is particularly useful for single objects with complex surface details.
We use photometric stereo~\cite{ikehata2023sdmunips} to estimate normal maps from images under varying light conditions.
This produces high-quality normal maps that can be used alone for shape reconstruction.

%% file: sections/03_method.tex
\begin{figure*}
    \centering
    \includegraphics[width=\textwidth]{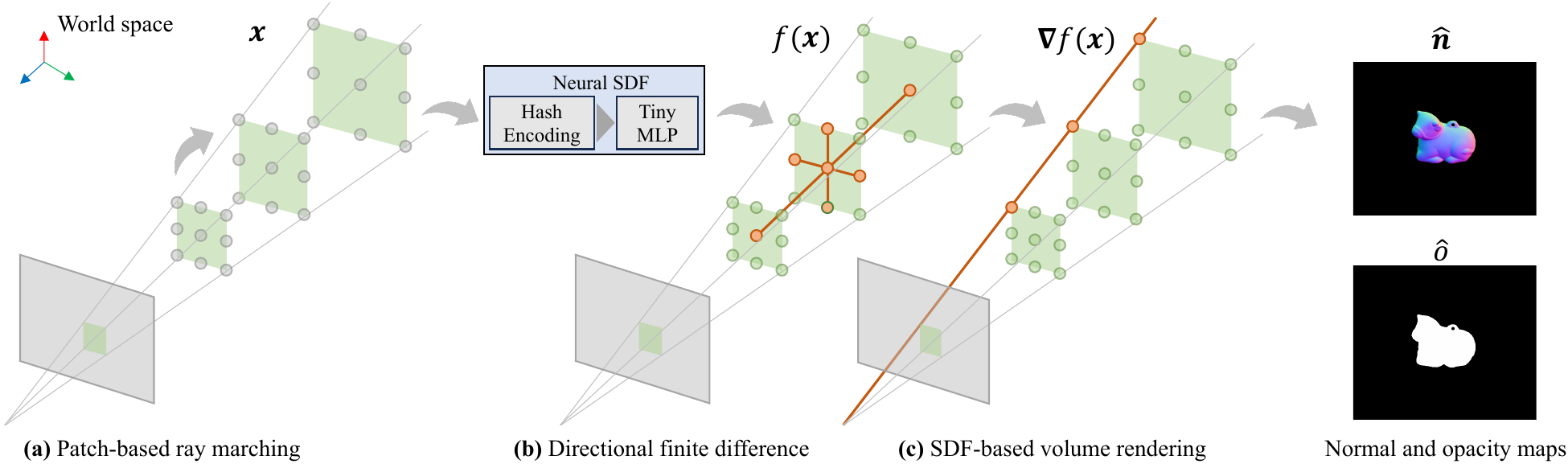}
    \vspace{-2em}
    \caption{\textbf{Approach overview.} \textbf{(a)} Given input views, we randomly sample patches of pixels, cast rays from those pixels, and march a plane along the patch of rays (\cref{sec.patch_based_ray_marching}).
    We treat the ray-plane intersections as sampled points and evaluate their SDF values via the neural SDF.
    \textbf{(b)} SDF samples neighbor to a point are used to compute the SDF gradients using directional finite difference (\cref{sec.DFD}).
    \textbf{(c)} SDF samples on the same rays are used for SDF-based volume rendering (\cref{sec.pipeline}).
    }
    \label{fig.pipeline}
\end{figure*}

\section{Approach}
\label{sec.method}
We aim to recover the surface given a set of normal maps, object masks, and camera intrinsic and extrinsic parameters.
Since photometric stereo methods estimate the normal maps in camera space, we first apply the camera-to-world rotation to normal maps to obtain world-space normal maps.
\Cref{fig.pipeline} depicts our pipeline.
We represent the geometry as a parametric SDF and optimize SDF parameters such that its rendered gradient and opacity values are consistent with input normal vectors and mask values, respectively. 

\subsection{Pipeline}
\label{sec.pipeline}
\paragraph{SDF-based volume rendering (Forward pass)}
A signed distance function (SDF) implicitly captures the geometry by assigning each spatial point with the signed distance to its closest surface point.
The surface \surface can then be represented as the zero-level-set of the SDF:
\begin{equation}
\label{eq.sdf}
 \surface = \{\point \mid f(\point)=0\}.
\end{equation}
We parameterize the SDF using multi-resolution hash encoding followed by a shallow MLP. 
In multi-resolution hash encoding, the 3D space is discretized into a set of multi-resolution 3D grids. Each grid cell is associated with a unique hash value mapping the 3D coordinates to a learnable feature vector. 
Specifically, $h(\point; \phi)$ is the feature vector obtained by concatenating feature vectors of different levels $h(\point) = [\point, h_1(\point),..., h_L(\point)]$.
Then, the SDF can be parameterized as 
\begin{equation}
    f(\point) = f(h(\point; \phi); \theta),
\end{equation}
where $\phi$ and $\theta$ are learnable hash-encoding feature vectors and MLP parameters. Multi-resolution hash encoding has proven to speed up training and improve the expressive power of the neural SDF~\cite{muller2022ingp,instantNSR,neus2Wang2023}. 

During training, we evaluate the SDF values at points sampled on the rays cast from the camera centers toward the scene.
\neus proposes an unbiased and occlusion-aware way to transfer the SDF samples to volume rendering opacities.
Given $N$ ordered 3D points $\{\point_i\}_{i=0}^{N}$ on a ray and their SDF values $\{f(\point_i)\}_{i=0}^{N}$, the opacity at each point is designed as
\begin{equation}
\label{eq.sdf_opacity}
    \opacity_i = \max\left(\frac{\sigmoid(f(\point_i))-\sigmoid(f(\point_{i+1}))}{\sigmoid(f(\point_i))}, 0\right),
\end{equation}
where $\sigmoid(x)=\frac{1}{1+\exp(-\sharpness x)} $ is the sigmoid function with a trainable sharpness \sharpness.

With this SDF-based opacity, we can render the normal and opacity for the pixel \pixel as:
\begin{equation}
\begin{aligned}
       \hat{\normal}(\pixel) = \sum_{i=0}^{N} \transmittance_i \opacity_i \nabla f (\point_i), \quad \hat{o}(\pixel) = \sum_{i=0}^{N} \transmittance_i \opacity_i \\ \text{with} \quad \transmittance_i = \prod_{j=0}^{i-1} (1-\opacity_j). 
\end{aligned}
\label{eq.vol_rendering}
\end{equation}
From \cref{eq.sdf_opacity,eq.vol_rendering}, the neural SDF must be evaluated at least once at sampled points for volume rendering. 
In \cref{sec.DFD,sec.patch_based_ray_marching}, we show that these SDF samples can be reused for SDF gradient approximation, thus reducing the computational amount and speeding up the training.

\paragraph{Neural SDF Training (Backward pass)}
To train the neural SDF, we supervise the rendered normals and opacities using input ones $\normal(\pixel)$ and $\hat{o}(\pixel)$.
The loss function consists of three terms: The normal, the mask, and the eikonal term:
\begin{equation}
\begin{aligned}
       \loss = \overbrace{\sum_{\pixel}(\hat{\normal}\left(\pixel) - \normal(\pixel)\right)^2}^{normal}  + \lambda_1 \overbrace{\sum_{\pixel} \bce\left(\hat{o}(\pixel), o(\pixel)\right)}^{mask} \\ 
   + \lambda_2 \underbrace{\sum_{\point} \left(\norm{\nabla f(\point)}_2 - 1\right)^2}_{eikonal}, 
\end{aligned}
\end{equation}
where \pixel is the sampled pixels in a batch, \point is the sampled points on rays, and $\bce$ is the binary cross entropy function.
The normal term encourages the rendered SDF gradients to be consistent with the input normal vectors so that accurate geometry details are recovered.
The mask term aligns the shape's silhouette with input masks and can be considered the boundary condition.
The eikonal term enforces the SDF gradient norm to be close to $1$ almost everywhere so that the neural SDF is approximately valid~\cite{osher2004level}.

\subsection{Directional Finite Difference}
\label{sec.DFD}

As seen, both the normal and eikonal terms backpropagate the loss gradient via the SDF gradient to neural SDF parameters (\ie, double backpropagation $\frac{\partial \loss}{\partial \nabla f} \frac{\partial \nabla f}{\partial \theta}$).
The most common way is to use Pytorch's automatic differentiation~(AD) due to implementation simplicity.
However, the multi-resolution hash encoding package does not officially support double propagation.
Common workarounds include CUDA programming~\cite{neus2Wang2023,rosu2023permutosdf} or use finite difference (FD)~\cite{neuralangelo2023CVPR,instantNSR} to bypass double propagation.
However, CUDA programming can be inflexible in that it shrinks the design space (\eg, NeuS2~\cite{neus2Wang2023} requires ReLU activation), and FD will slow down the training.
Consider the ordinary axis-aligned FD:
\begin{equation}
    \nabla f(\point) \approx \left[\begin{matrix}
        \frac{f(\point+ \boldsymbol{\epsilon}_x) - f(\point-\boldsymbol{\epsilon}_x)}{2\epsilon}\\
        \frac{f(\point+\boldsymbol{\epsilon}_y) - f(\point-\boldsymbol{\epsilon}_y)}{2\epsilon}\\
        \frac{f(\point+\boldsymbol{\epsilon}_z) - f(\point-\boldsymbol{\epsilon}_z)}{2\epsilon}
    \end{matrix}\right],
\end{equation}
where $\boldsymbol{\epsilon}_{* \in \{x, y, z\}}$ is an offset vector along $x$, $y$, or $z$ axis with a tiny step size $\epsilon$.
Computing the gradient at one point requires evaluating the SDF at its nearby points. 
These additional SDF samples are also included in the computation graph in the forward pass. 
Consequently, using FD significantly slows the forward rendering and backward optimization even if the efficient spatial encoding is used~\cite{neuralangelo2023CVPR}.

To avoid redundant SDF evaluations and double backpropagation, 
we propose directional finite difference~(DFD) to reuse the volume rendering SDF samples for SDF gradients approximation.
Consider the SDF value at a 3D point \point parameterized by the ray origin \cameraCenter and direction \viewDirection:
\begin{equation}
    f(\point)=f(\cameraCenter + \distToCameraCenter\viewDirection),
\label{eq.sdf_parameterization}
\end{equation}
where \distToCameraCenter is the distance to the ray origin.
From \cref{eq.sdf_parameterization}, we have
\begin{equation}
\label{eq.directional_derivative}
    \frac{df}{d\distToCameraCenter} = \nabla f(\point)^\top \viewDirection \defeq \nabla_{\viewDirection} f .
\end{equation}
\Cref{eq.directional_derivative} is exactly the directional derivative of the SDF, which projects the SDF gradient onto the ray direction. 
Suppose we parameterize the same point along three different directions as 
\begin{equation}
    \point = \cameraCenter_i + \distToCameraCenter_i \viewDirection_i, \quad i \in \{1, 2, 3\}.
\end{equation}
Applying \cref{eq.directional_derivative} then yields three linear equations:
\begin{equation}
\label{eq.dfd_linear_system}
\underbrace{\begin{bmatrix}
       - \viewDirection_1^\top - \\
       - \viewDirection_2^\top - \\
       - \viewDirection_3^\top - 
    \end{bmatrix}}_{\mathbf{V}}
     \nabla f
= \underbrace{\begin{bmatrix}
    \frac{df}{d\distToCameraCenter_1}\\
    \frac{df}{d\distToCameraCenter_2}\\
    \frac{df}{d\distToCameraCenter_3}
\end{bmatrix}}_{\nabla_{\V{V}} f}.
\end{equation}
From \cref{eq.dfd_linear_system}, we can compute the SDF gradient by
\begin{equation}
\label{eq.dfd_solution}
    \nabla f = \V{V}^{-1} \nabla_{\V{V}} f
\end{equation}
Numerically, we can approximate the directional derivative by
\begin{equation}
\label{eq.directional_derivative_fd}
\begin{split}
      \frac{df}{dt_i} &\approx \frac{f(\cameraCenter_i + (t_i+\Delta t)\viewDirection_i) - f(\cameraCenter_i + (t_i-\Delta t)\viewDirection_i)}{2\Delta t} \\
    &=  \frac{f(\point_i+\Delta t_i\viewDirection_i) - f(\point_i-\Delta t_i\viewDirection_i)}{2\Delta t}, \quad i \in \{1, 2, 3\}
\end{split}
\end{equation}
\Cref{eq.dfd_solution,eq.directional_derivative_fd} imply that,
to numerically approximate the SDF gradient, we can first compute the finite difference along three directions $\V{V}$, then linearly transfer it by $\V{V}^{-1}$.
We call this SDF gradient approximation directional finite difference.
\Cref{eq.dfd_solution} has a unique solution if the three directions are non-coplanar.

The ordinary finite difference can be viewed as the special case of \cref{eq.dfd_solution} by choosing the three coordinate axes as the ray directions.
In this case, $\V{V}$ becomes the identity matrix.
Both DFD and FD require nearby SDF samples to approximate the SDF gradient. 
However, by tailoring the sampling strategy, we can reuse the SDF samples in volume rendering for DFD without additional SDF samples, as described in the following.

\subsection{Patch-based Ray Marching}
\label{sec.patch_based_ray_marching}
DFD requires computing the finite difference in three directions.
In volume rendering, we naturally have SDF samples along the viewing ray direction that can be used for finite difference.
For the other two directions, we propose to sample patches of pixels instead of single pixels and march a plane on each patch of rays to locate the sampled points on parallel planes.

Specifically, we perform ordinary ray marching for the center pixel/ray of the patch and treat the sampled points as the intersections between a marching plane and the center ray.
The points to be sampled on the remaining rays within the patch can then be found as the intersection of the planes and the rays:
\begin{equation}
    t_j = \frac{t_i \viewDirection_i^\top \V{m}}{\viewDirection_j^\top \V{m}},
\end{equation}
where $\V{m}$ is the unit normal direction of the marching plane, $t_i$ is the ray marching distance on the center ray with direction $\viewDirection_i$, and $t_j$ is the distance from the intersection point to the ray origin (\ie, camera center) along the direction $\viewDirection_j$; for derivation see \cref{sec.ray_marching} in \suppl

We choose the marching plane to parallel the image plane where the patch is located (\cref{fig.pipeline}).
With this sampling strategy, the three directions for DFD computation are the viewing ray direction and the world-space directions of $x$- and $y$-axis of the camera coordinates.
The marching plane normal $\V{m}$ becomes the world-space direction of the $z$-axis of the camera coordinates and is the same for all pixels/rays from the same view.
This way, the SDF samples on the same ray can be used for volume rendering as \cref{eq.vol_rendering}, and the SDF samples neighbor to a point can be used for SDF gradient computation as \cref{eq.dfd_solution}.
We use the central difference~(\ie, \cref{eq.directional_derivative_fd}) for points with neighbors in both directions and forward/backward difference for boundary points.

This sampling strategy is beneficial in three aspects.
First, it allows us to reuse the SDF samples from neighboring rays to compute DFD. 
Second, it stabilizes training by ensuring the existence of $\V{V}^{-1}$, since the viewing ray direction will never co-planer to the image plane.
Third, it reduces the computation overhead during training by fixing the three directions for samples from the same pixel/ray. 
$\V{V}^{-1}$ need be computed only once for each pixel before training. 
In our experiments, this takes about \SI{0.6}{\sec} for two million pixels. 
During training, only the $3\times3$ matrix-vector multiplication~\cref{eq.dfd_solution} is required, which can almost be ignored compared to AD or FD.

%% file: sections/04_exp.tex
\section{Experiments}
We evaluate our method quantitatively on the MVPS benchmark in \cref{sec.exp_mvps}, investigate different components of our method in \cref{sec.ablation_study} and evaluate our method qualitatively on our captured real-world data in \cref{sec.exp_real_world}.

\paragraph{Implemetation details}
We apply \sdmunips to images in each view to obtain normal maps.
In each batch, we randomly sample $2048$ patches of $3\times3$ pixels from all input normal maps.
Like INGP~\cite{muller2022ingp}, we skip the empty and occluded space during ray marching using \nerfacc.
The scene is bounded within a unit sphere, and we use a coarse-to-fine ray marching step, which decreases log-linearly from $1e^{-2}$ to $5e^{-4}$.
We optimize the shape of \diligentmv benchmark objects for $5000$ batches (roughly equal to $20$ epochs), and $30000$ batches for our captured data.
The weights of loss terms are set as $1$.
We use the Adam optimizer with an initial learning rate $5e^{-3}$.
More details can be found in \cref{sec.implementation_details} in \suppl

\subsection{Quantitative evaluation}
\label{sec.exp_mvps}

\paragraph{Dataset}
\diligentmv captures five objects from $20$ views with $96$ OLAT images, yielding a total of $1920$ images per object.
Each image is $512 \times 612$, and the total number of foreground object pixels is about $2.2$ million. 
The camera is fixed during capture, and the object is placed on a turntable in a darkroom.
The ``ground truth'' meshes are created using a 3D scanner.

\paragraph{Baselines} 
We benchmark \ourmethod against several MVPS methods grouped into three categories. 
(1) Those that refines an initial point cloud or mesh and require manual effort, including \rmvps and \bmvps.
(2) Those that use Poisson surface reconstruction (PSR)~\cite{kazhdan2013screened} to fuse multi-view depth and normal maps, including \mvpsnet.
(3) Those that optimize an implicit neural surface representation, including \uanet, \psnerf, \mvas, and our method.

\begin{table*}
\newcommand{\Frst}{\cellcolor{red!25}}
\newcommand{\Scnd}{\cellcolor{orange!25}}
\scriptsize
	\definecolor{Gray}{gray}{0.85}
	\newcolumntype{g}{>{\columncolor{Gray}}r}
	\centering
	\caption{Quantitative evaluation on \diligentmv benchmark. Red and orange cells indicate \colorbox{red!25}{\textbf{the best}} and \colorbox{orange!25}{the second best} results respectively. On average, our approach achieves the best reconstruction quality at the second fastest speed.}
    \vspace{-1em}
	\resizebox{\linewidth}{!}{ 
		\begin{tabular}[c]{@{}l||rrrrrr||rrrrrr||r}
			\toprule
             & \multicolumn{6}{c||}{L2 Chamfer distance [\si{mm}] ($\downarrow$)} & \multicolumn{6}{c||}{F-score  ($\thresholdFscore = \SI{0.5}{mm}$) ($\uparrow$)} & Runtime ($\downarrow$)\\
			Methods & Bear & Buddha & Cow & Pot2 & Reading & Average & Bear & Buddha & Cow & Pot2 & Reading & Average & Average\\
			\midrule
			\rmvps  & 1.070 & 0.397 & 0.440 & 1.504 & 0.561 & 0.794 & 0.262 & 0.698 & 0.760 & 0.198 & 0.519 & 0.487 & NA\\
			\bmvps & \Scnd 0.212 & \Scnd 0.254 & \Frst \textbf{0.091} & 0.201 & \Scnd 0.259 &  \Scnd 0.203 & \Scnd 0.958 & \Scnd 0.902 & \Frst \textbf{0.986} & 0.946 & 0.914 & \Scnd 0.941& NA\\ 
            \mvpsnet &0.317&0.279&0.255&0.310&0.248& 0.282 & 0.813 &0.866&0.889&0.838& \Frst \textbf{0.918} & 0.865 & \Frst \textbf{37 secs}\\
			\uanet &  0.414 & 0.452 & 0.326 & 0.414 & 0.382 & 0.398 & 0.707 & 0.669 & 0.798 & 0.731 & 0.762  & 0.733& several hrs\\
			\psnerf & 0.260 & 0.314 & 0.287 & 0.254 & 0.352 & 0.293 & 0.898  & 0.806 & 0.856 & 0.919 & 0.785 & 0.853& 8 hrs\\
			\mvas & 0.243  & 0.357 & 0.216 & \Scnd 0.197 & 0.522 & 0.307 &0.909 & 0.754 & 0.907 & \Scnd 0.962 & 0.546 & 0.816& 70 mins\\
            Ours & \Frst  \textbf{0.184} & \Frst \textbf{0.218} & \Scnd 0.193 & \Frst \textbf{0.150} & \Frst \textbf{0.223} & \Frst \textbf{0.194} & \Frst \textbf{0.983} & \Frst \textbf{0.950} & \Scnd 0.977 & \Frst \textbf{0.994} & \Scnd 0.916 & \Frst \textbf{0.962}& \Scnd 48 secs\\
			\bottomrule
		\end{tabular}
	}
\label{tab.mvps_quan}
\end{table*} 

\begin{figure*}
\footnotesize
    \newcommand{\figwidthMeshComparison}{0.125}
    \centering
    \resizebox{0.98\linewidth}{!}{ 
    \begin{tabular}{@{}c@{}c@{}c@{}c@{}c@{}c@{}c@{}c@{}}
    \rmvps &
    \bmvps &
    \uanet &
    \psnerf &
    \mvas &
    \mvpsnet &
    Ours &
    3D Scanner
    \\
     \includegraphics[width=\figwidthMeshComparison\linewidth]{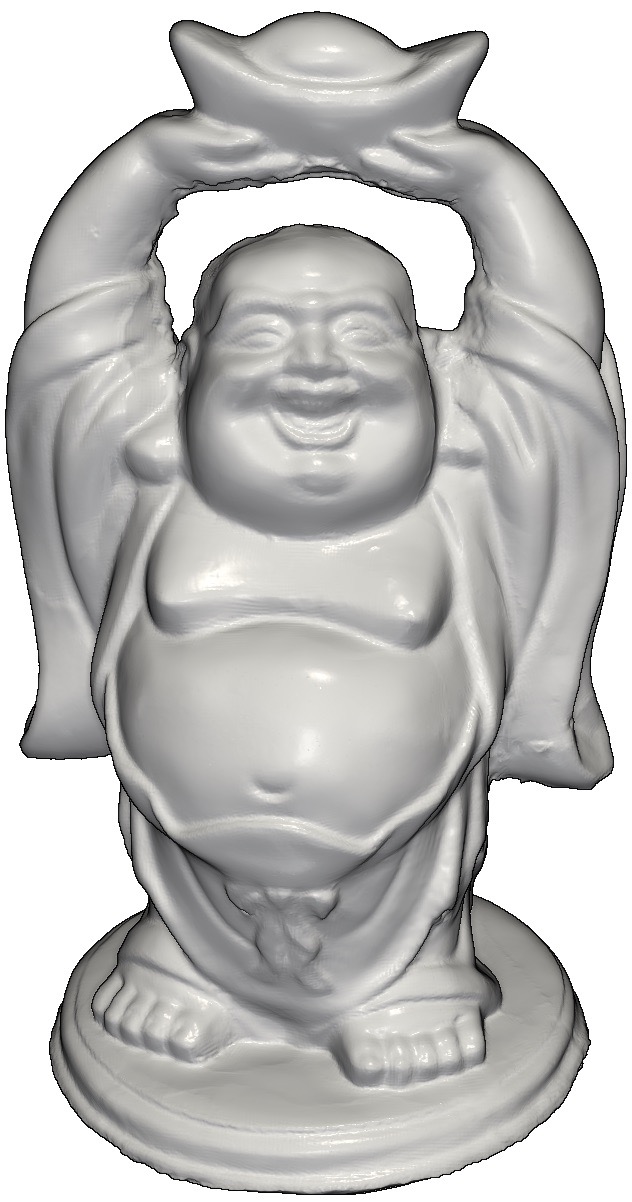} &
     \includegraphics[width=\figwidthMeshComparison\linewidth]{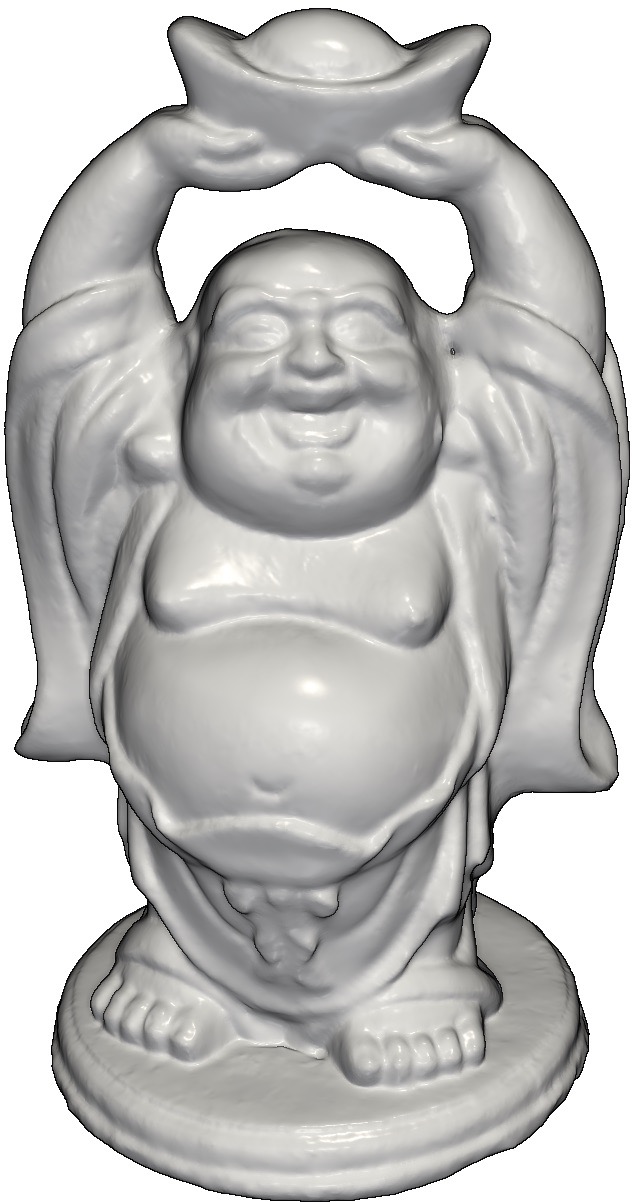} &
    \includegraphics[width=\figwidthMeshComparison\linewidth]{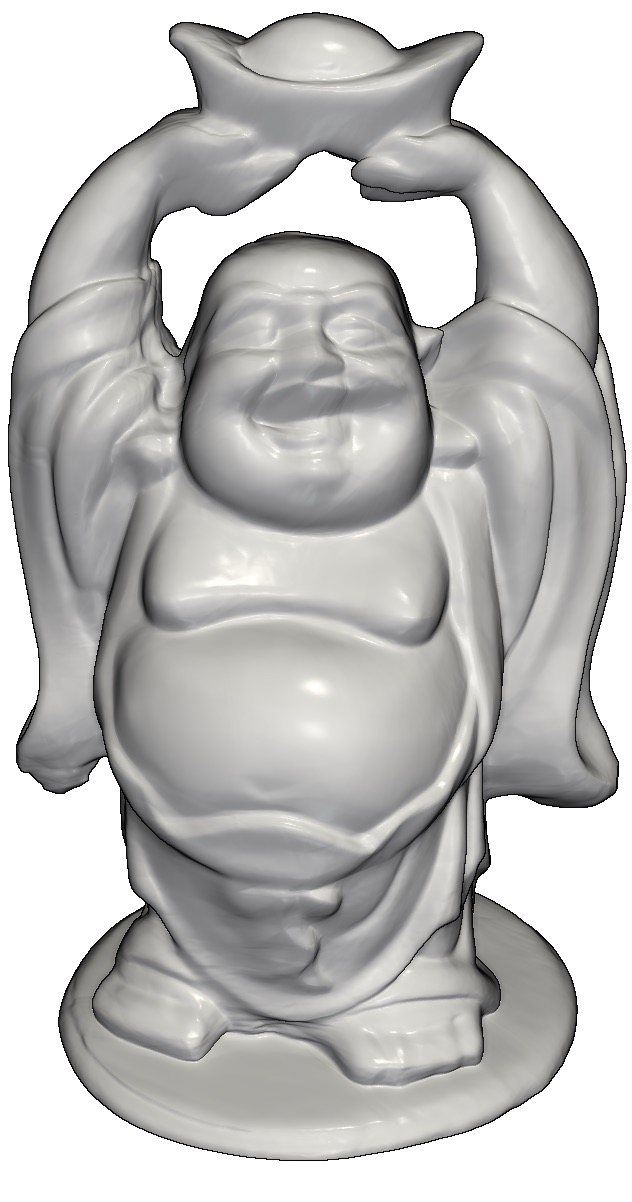} &
    \includegraphics[width=\figwidthMeshComparison\linewidth]{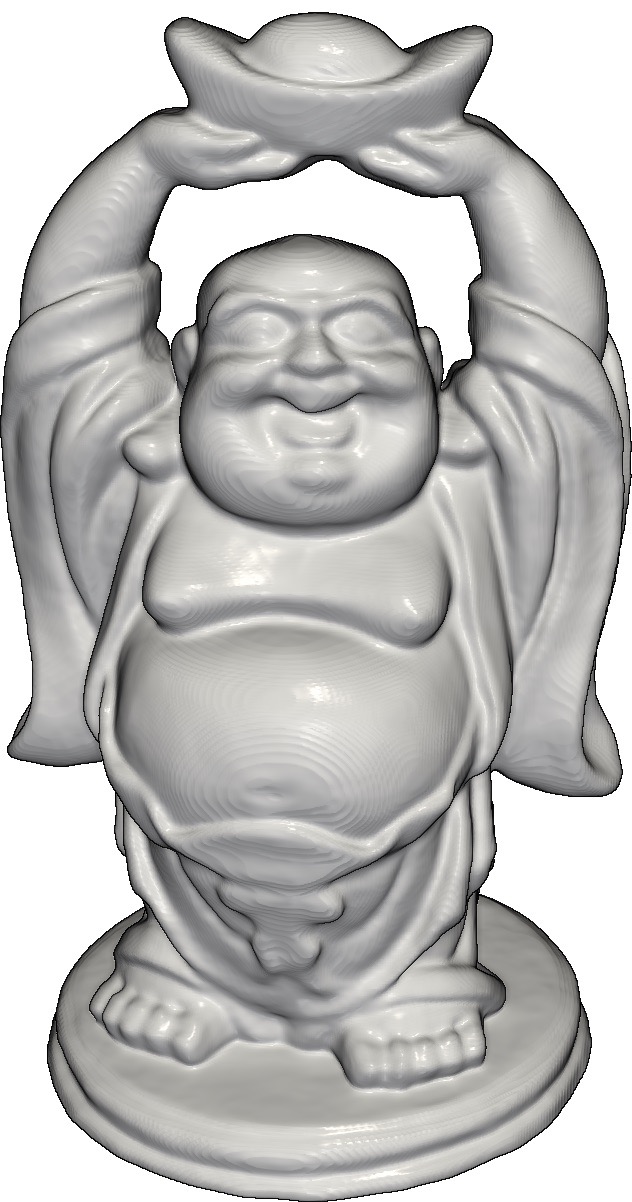} &
    \includegraphics[width=\figwidthMeshComparison\linewidth]{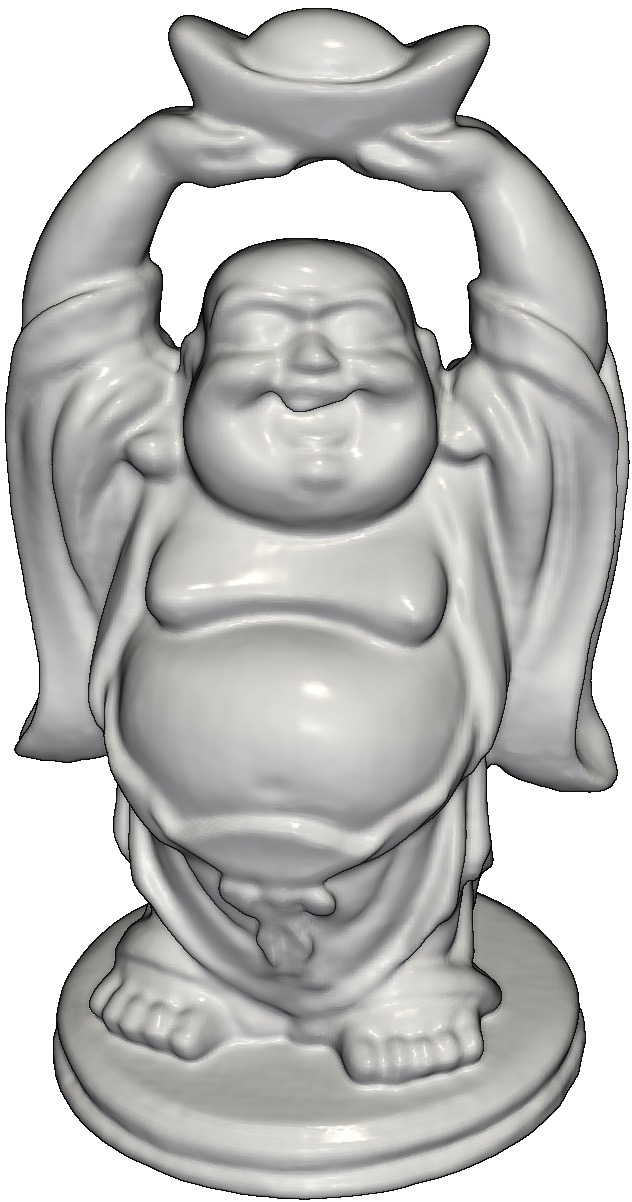} &
    \includegraphics[width=\figwidthMeshComparison\linewidth]{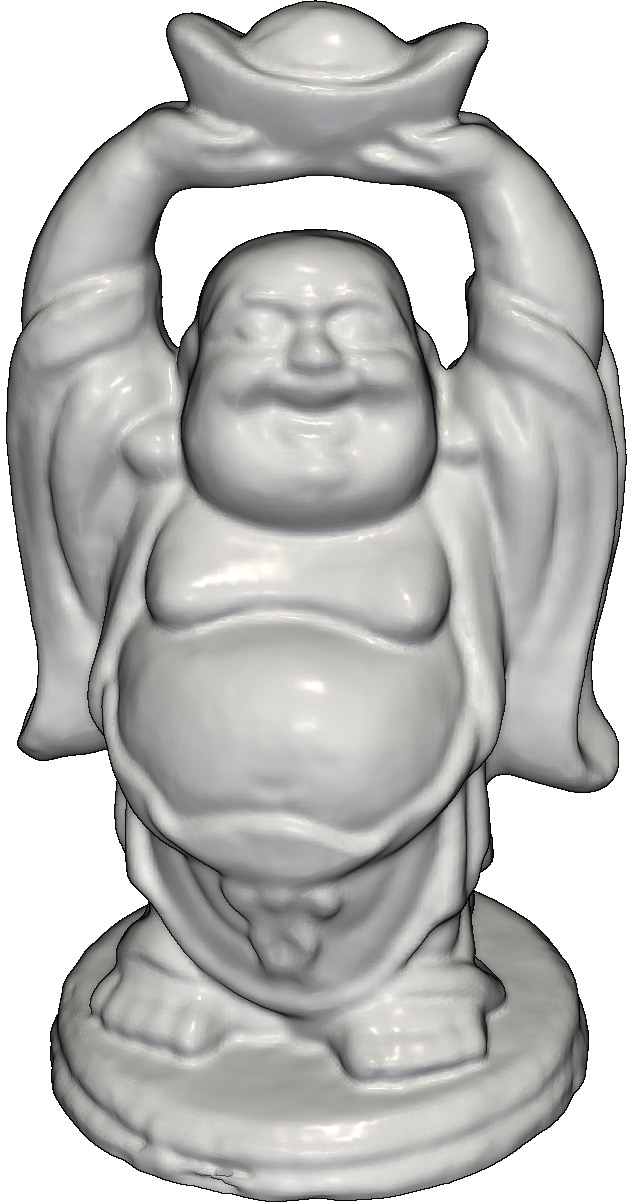}&
    \includegraphics[width=\figwidthMeshComparison\linewidth]{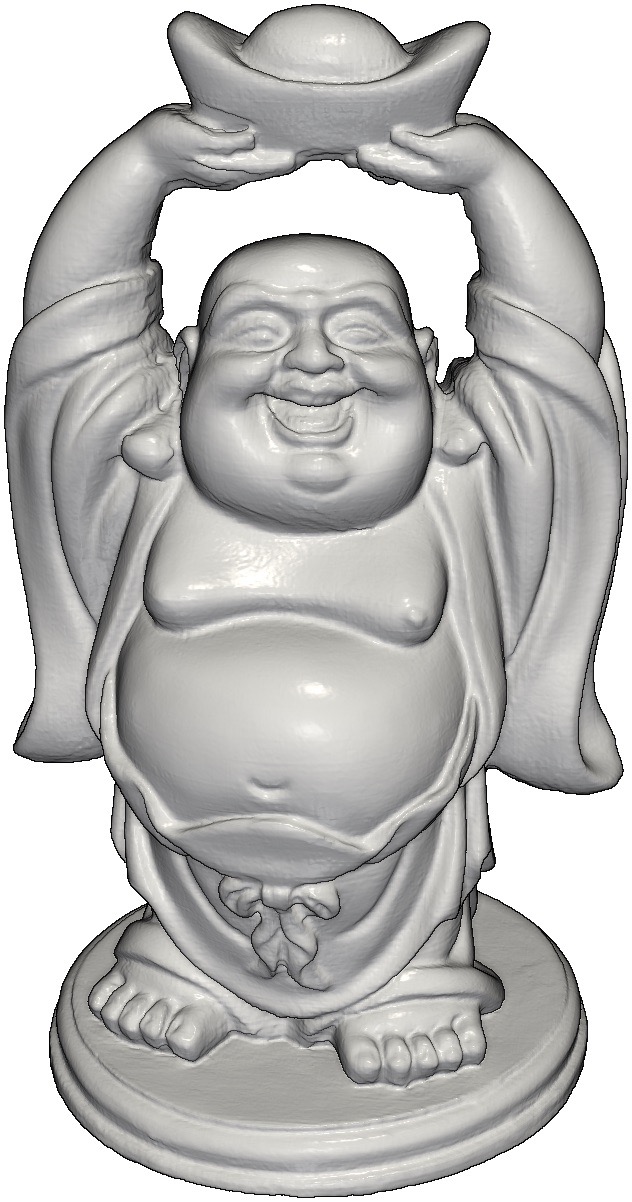} &
    \includegraphics[width=\figwidthMeshComparison\linewidth]{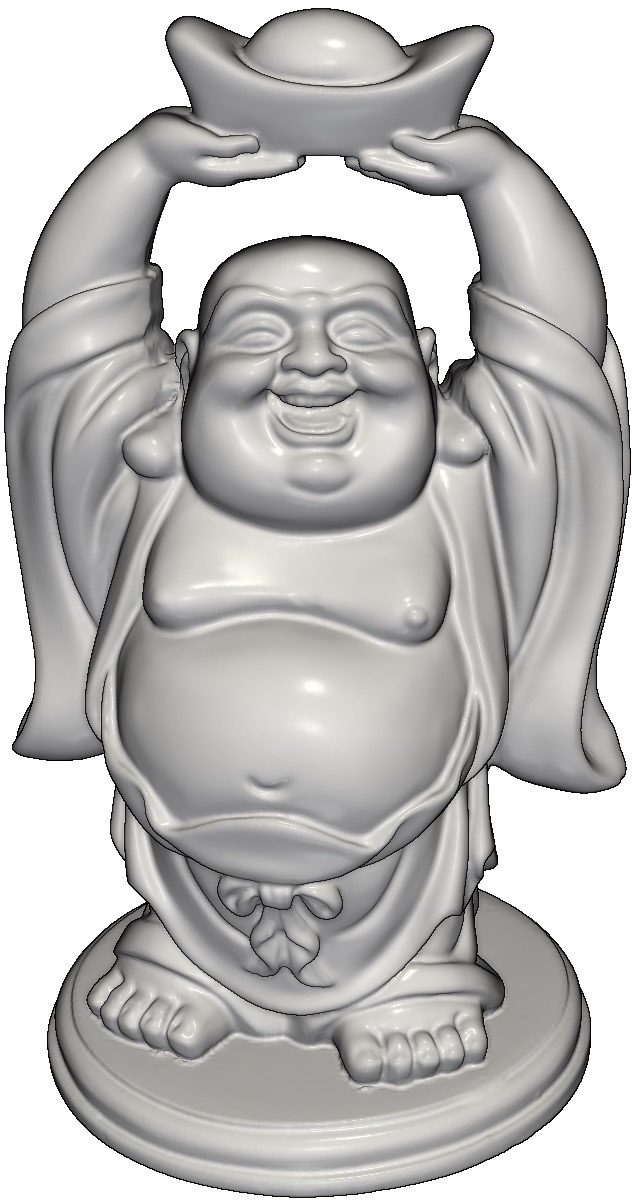}
    \\
    \includegraphics[width=\figwidthMeshComparison\linewidth]{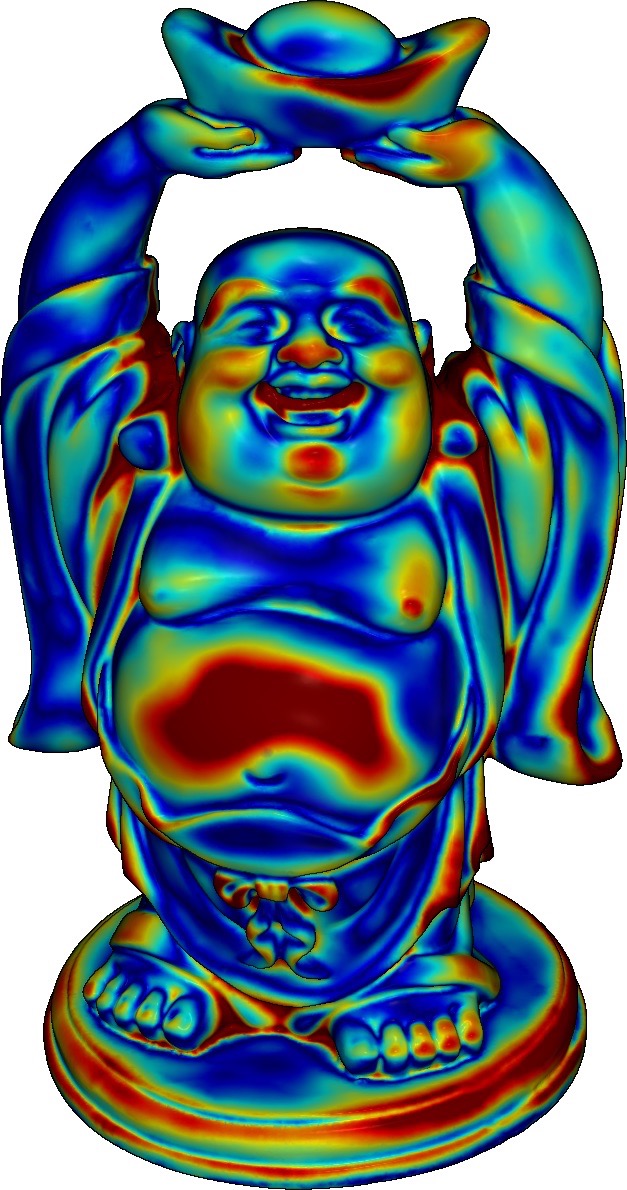} &
     \includegraphics[width=\figwidthMeshComparison\linewidth]{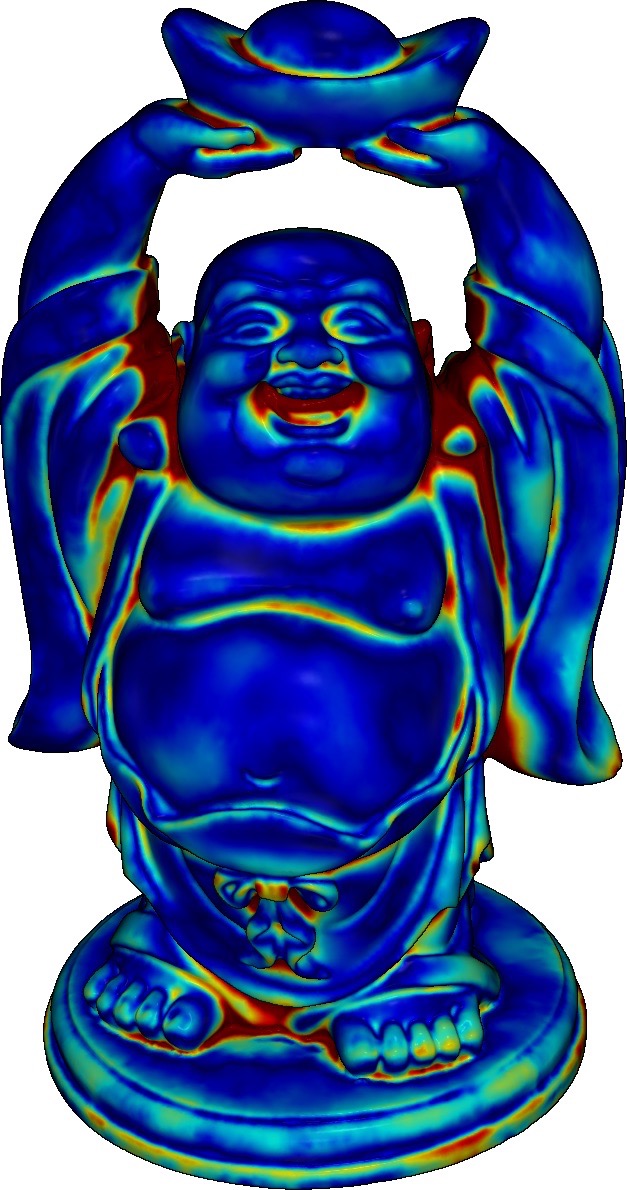} &
    \includegraphics[width=\figwidthMeshComparison\linewidth]{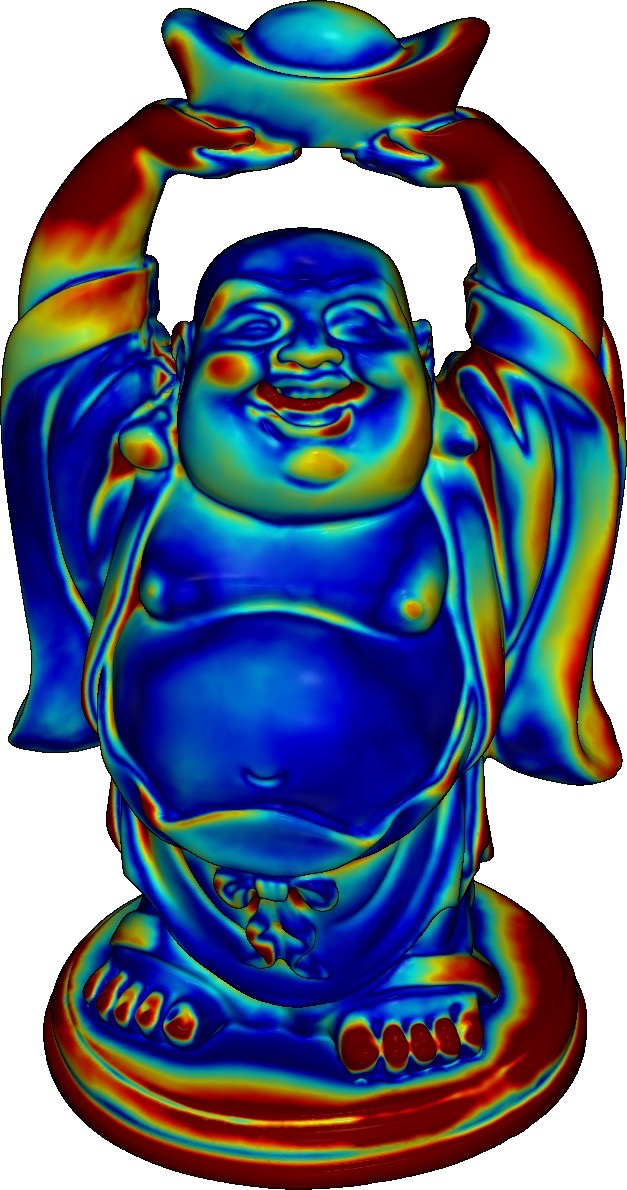} &
    \includegraphics[width=\figwidthMeshComparison\linewidth]{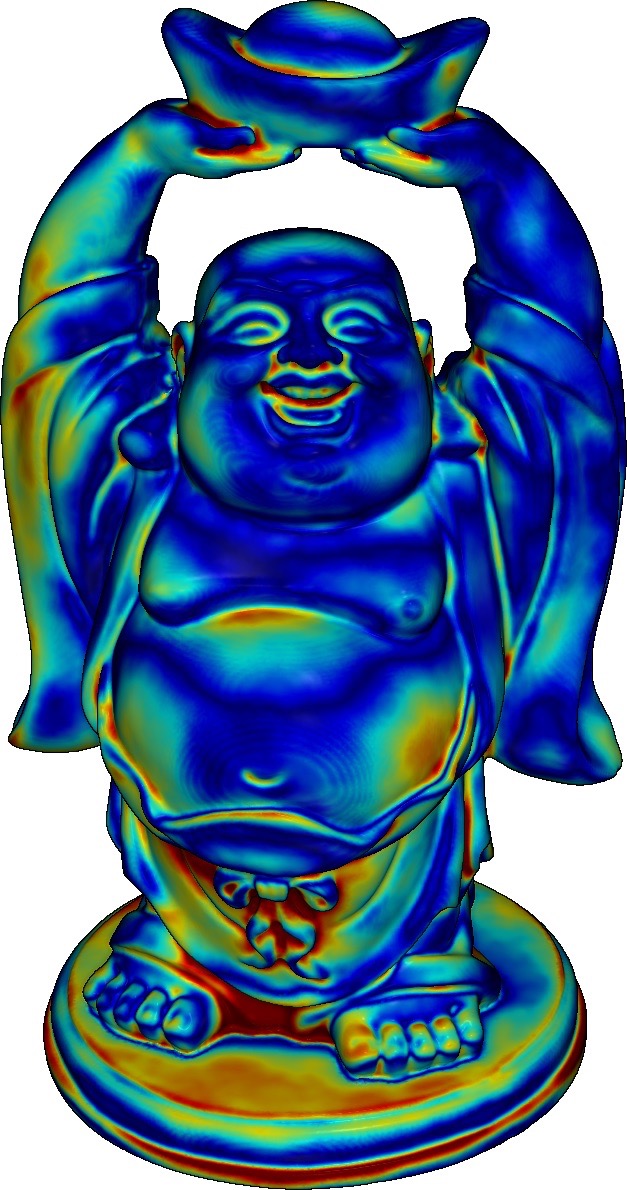} &
    \includegraphics[width=\figwidthMeshComparison\linewidth]{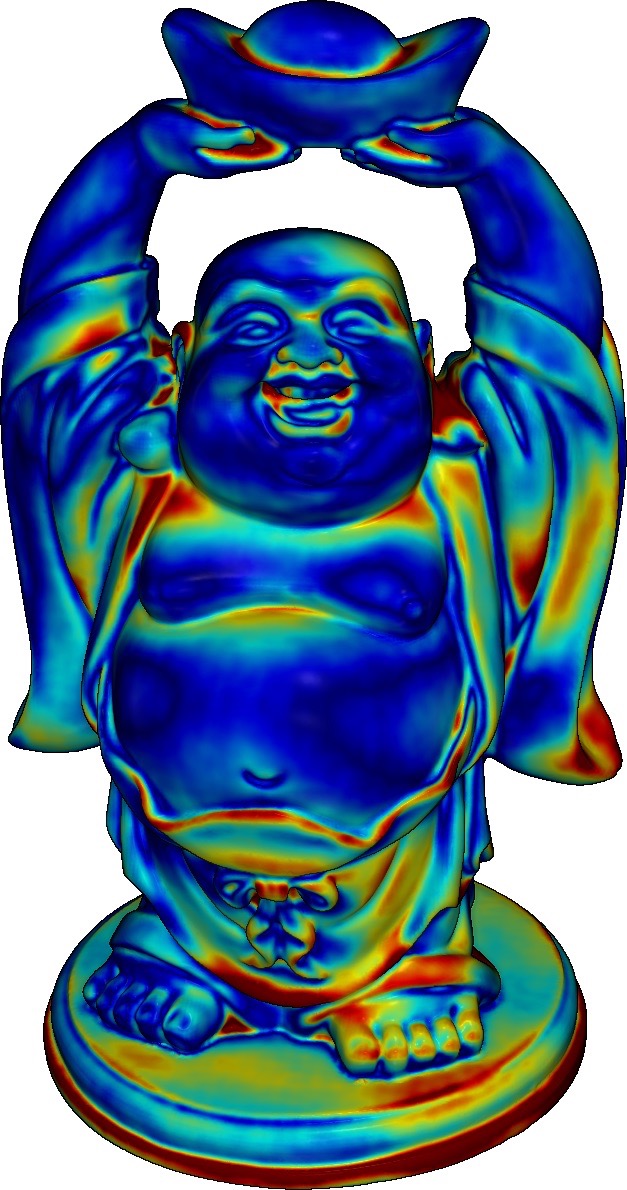} &
    \includegraphics[width=\figwidthMeshComparison\linewidth]{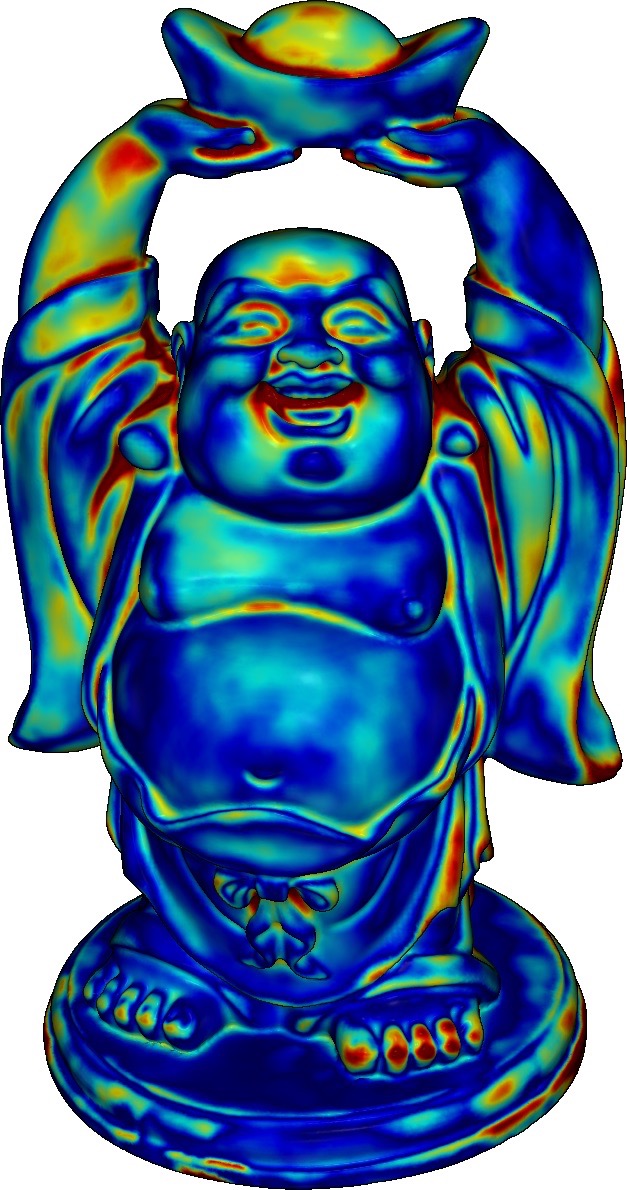}&
    \includegraphics[width=\figwidthMeshComparison\linewidth]{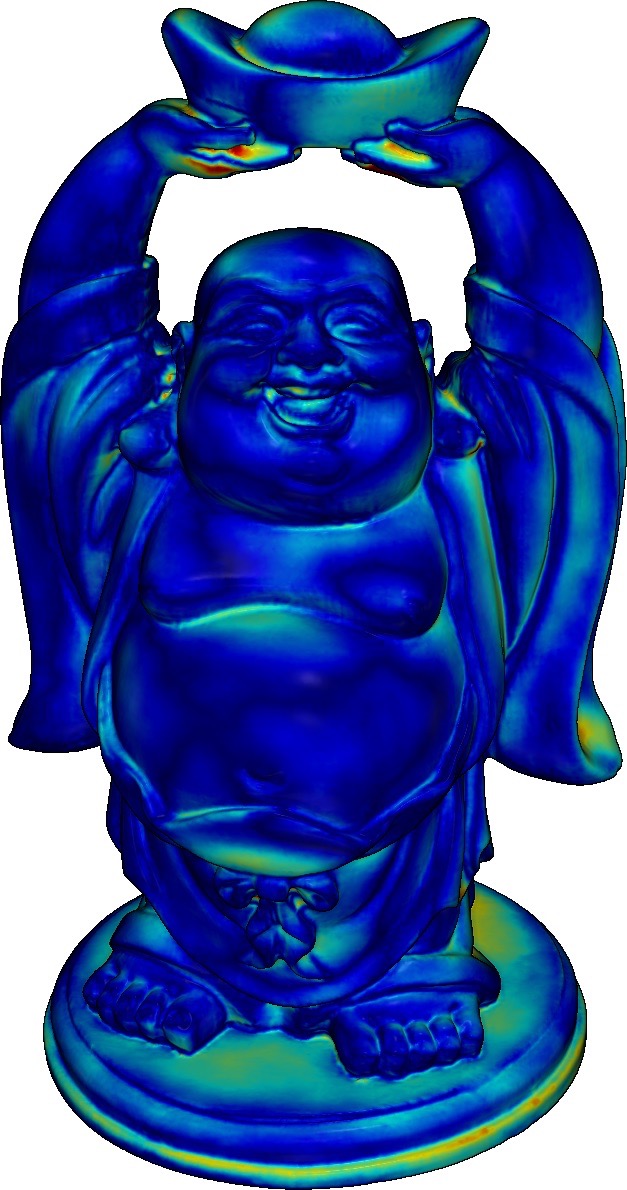} &
    \colorbar{0.14}{$\geq \SI{1}{\mm}$}
\\
    \includegraphics[width=\figwidthMeshComparison\linewidth]{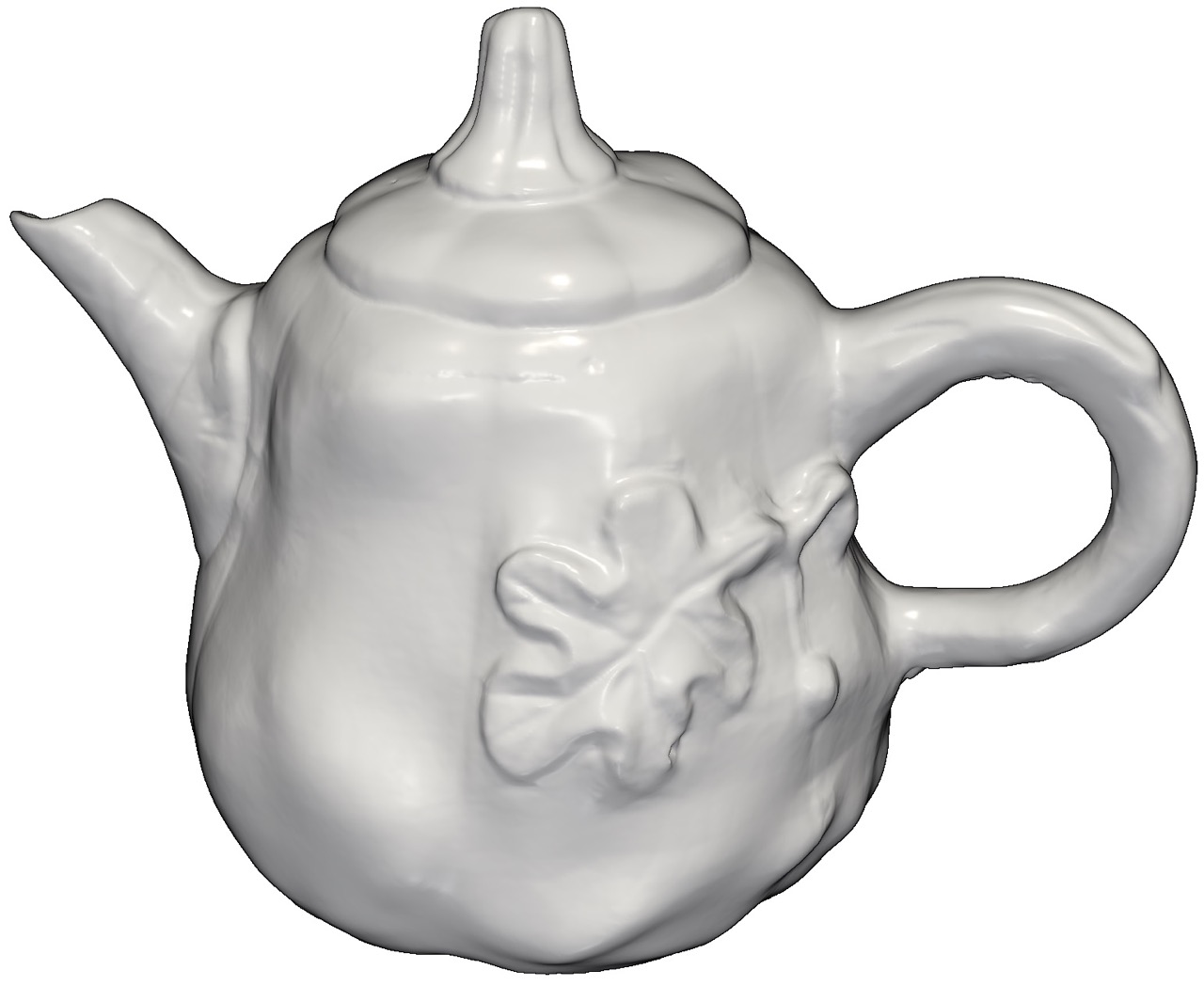} &
     \includegraphics[width=\figwidthMeshComparison\linewidth]{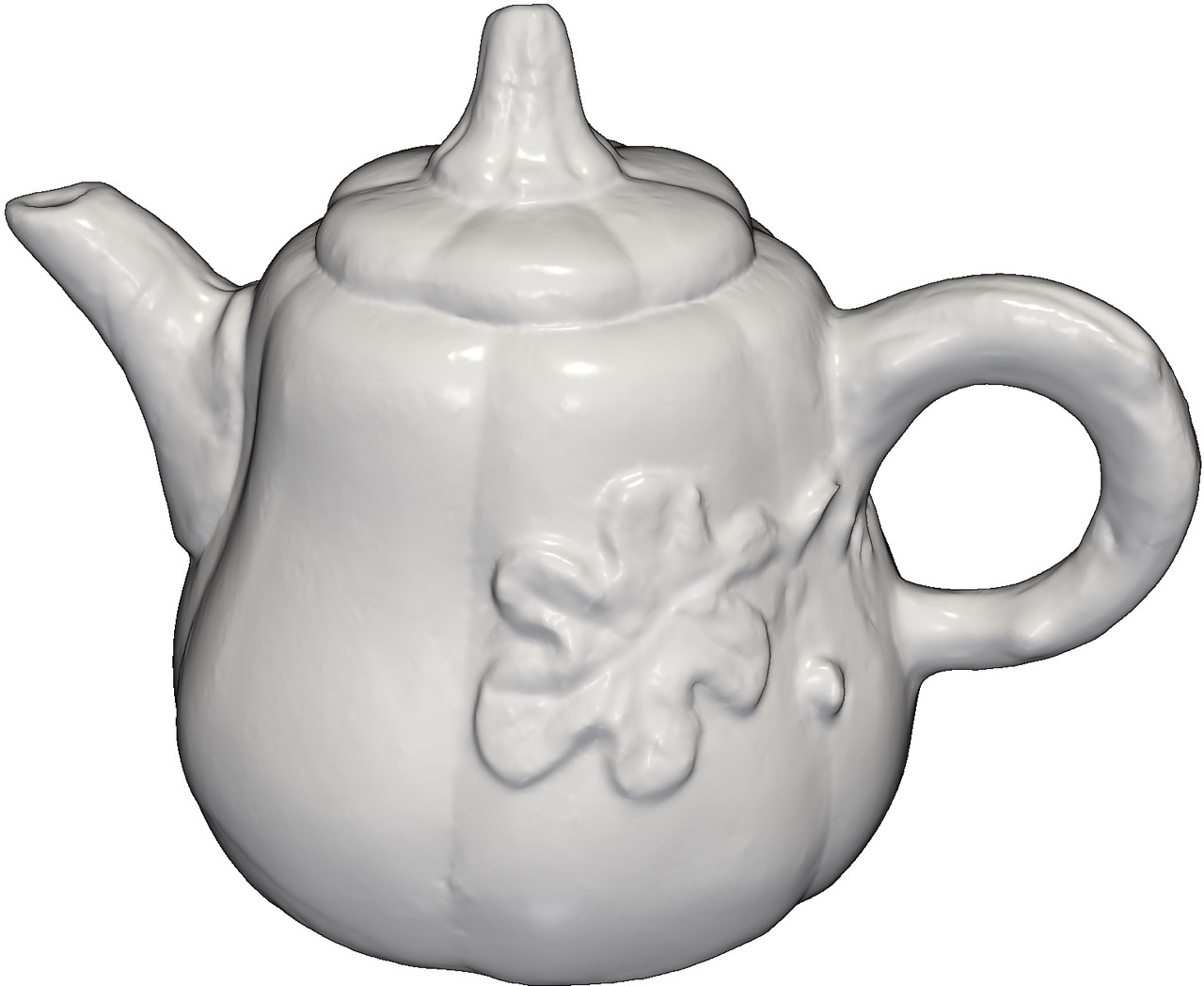} &
    \includegraphics[width=\figwidthMeshComparison\linewidth]{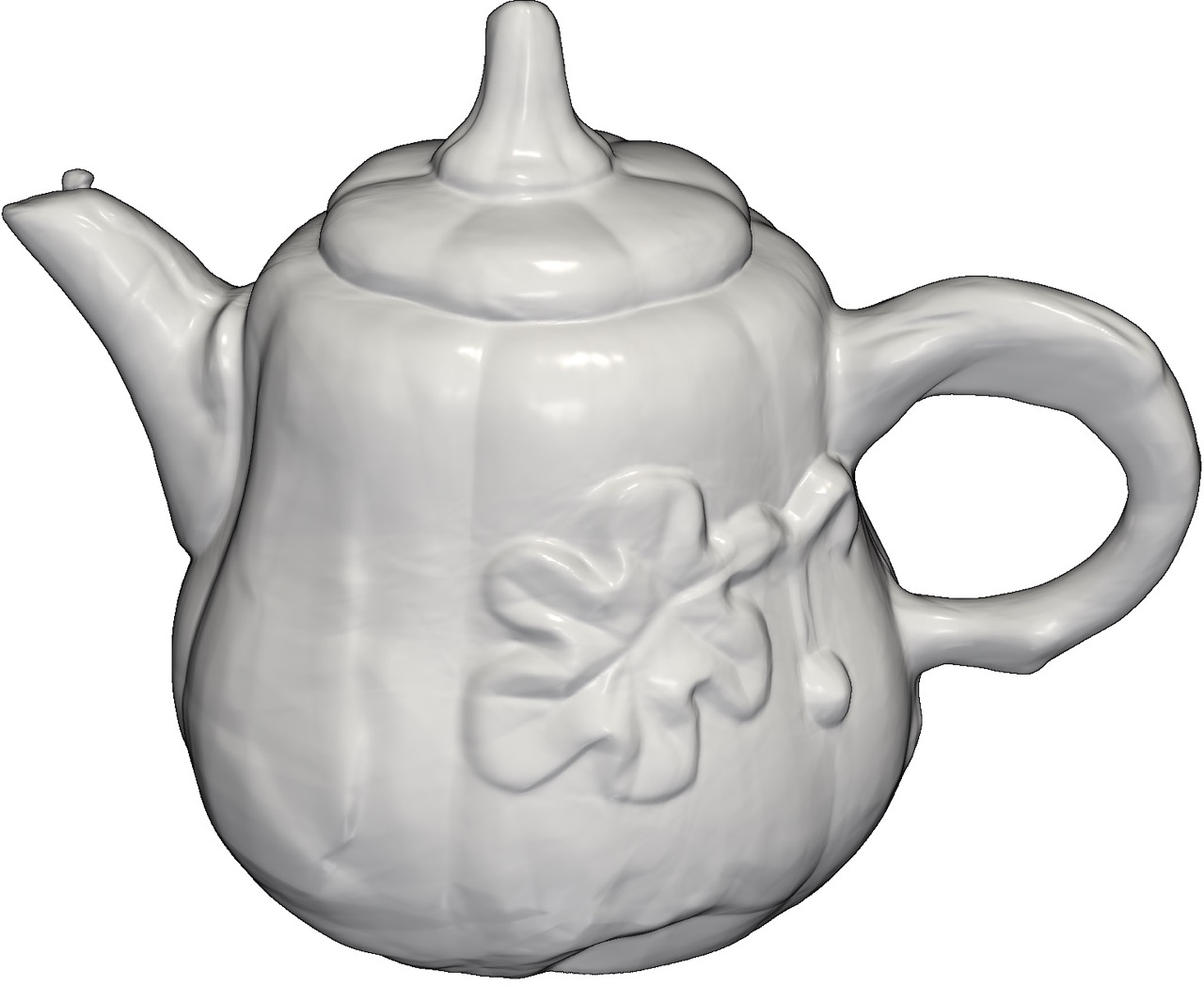} &
    \includegraphics[width=\figwidthMeshComparison\linewidth]{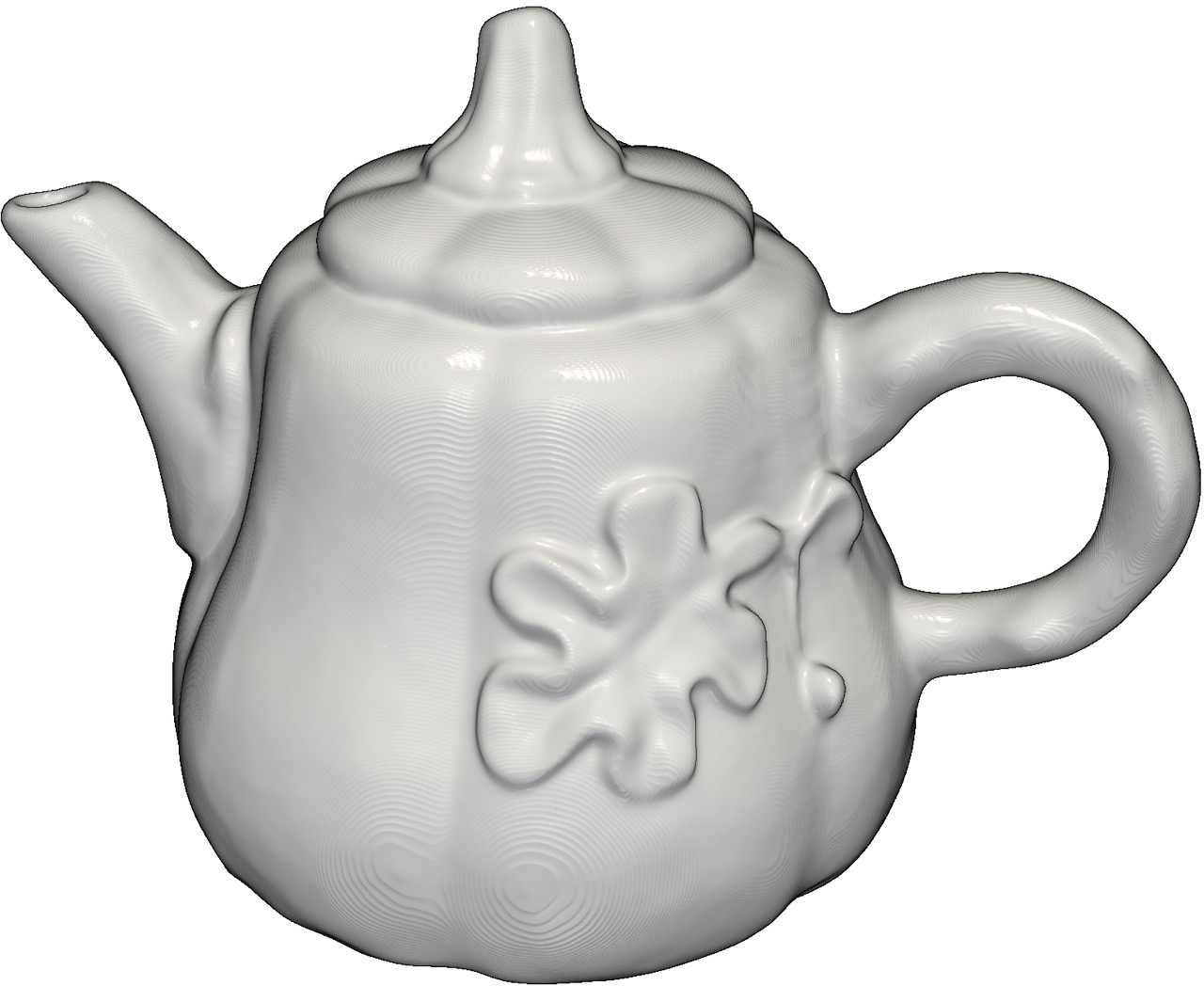} &
    \includegraphics[width=\figwidthMeshComparison\linewidth]{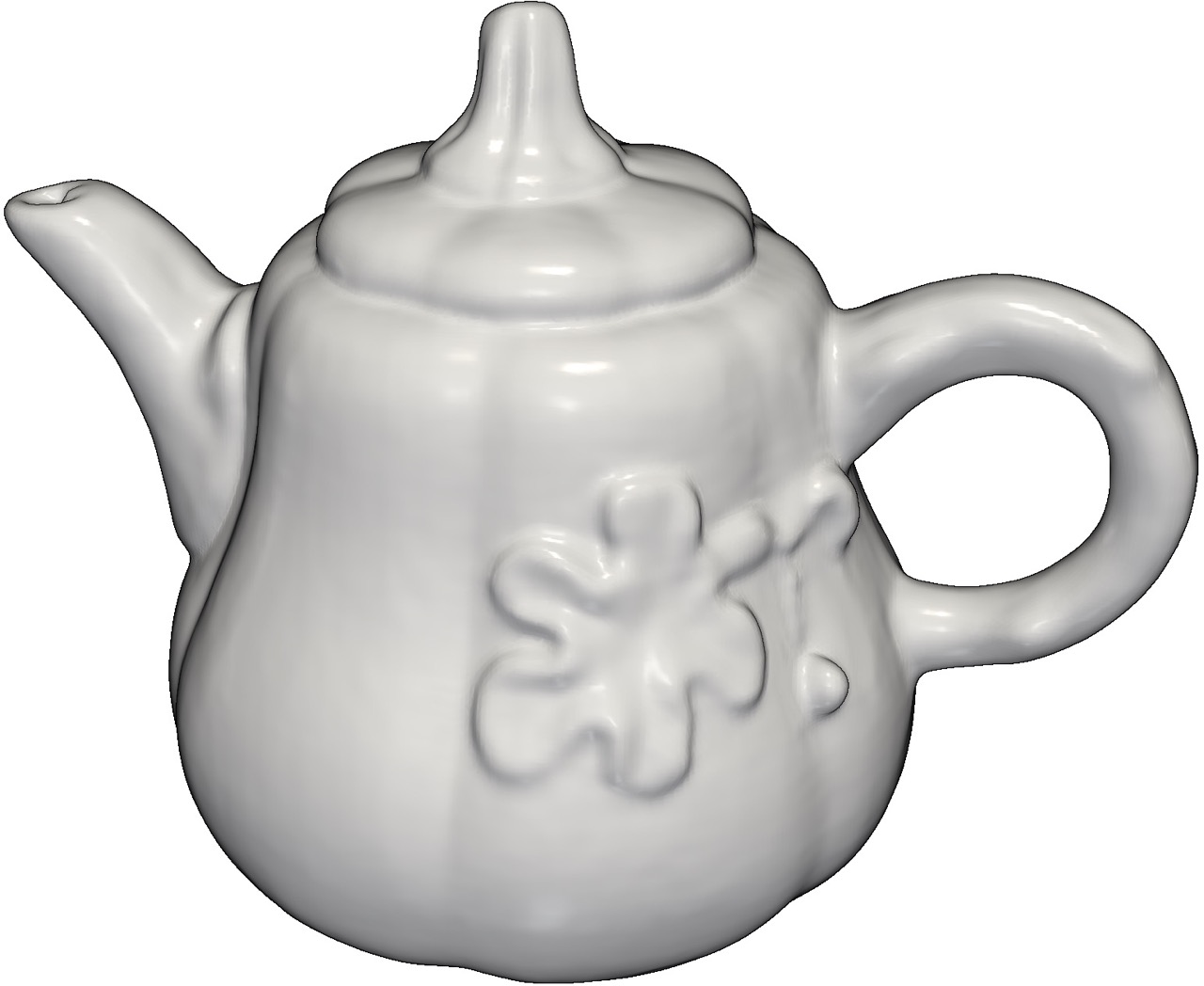} &
    \includegraphics[width=\figwidthMeshComparison\linewidth]{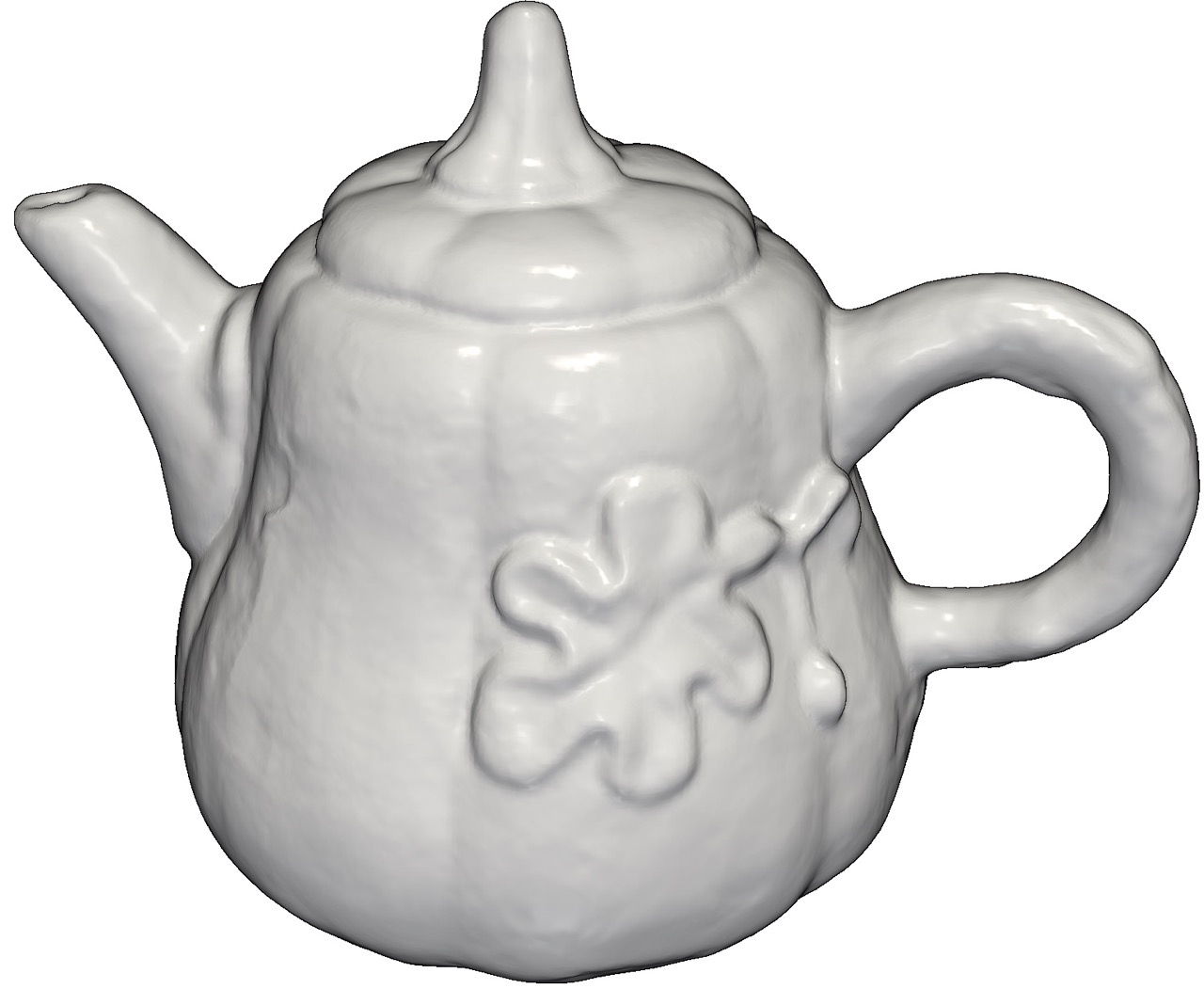}&
    \includegraphics[width=\figwidthMeshComparison\linewidth]{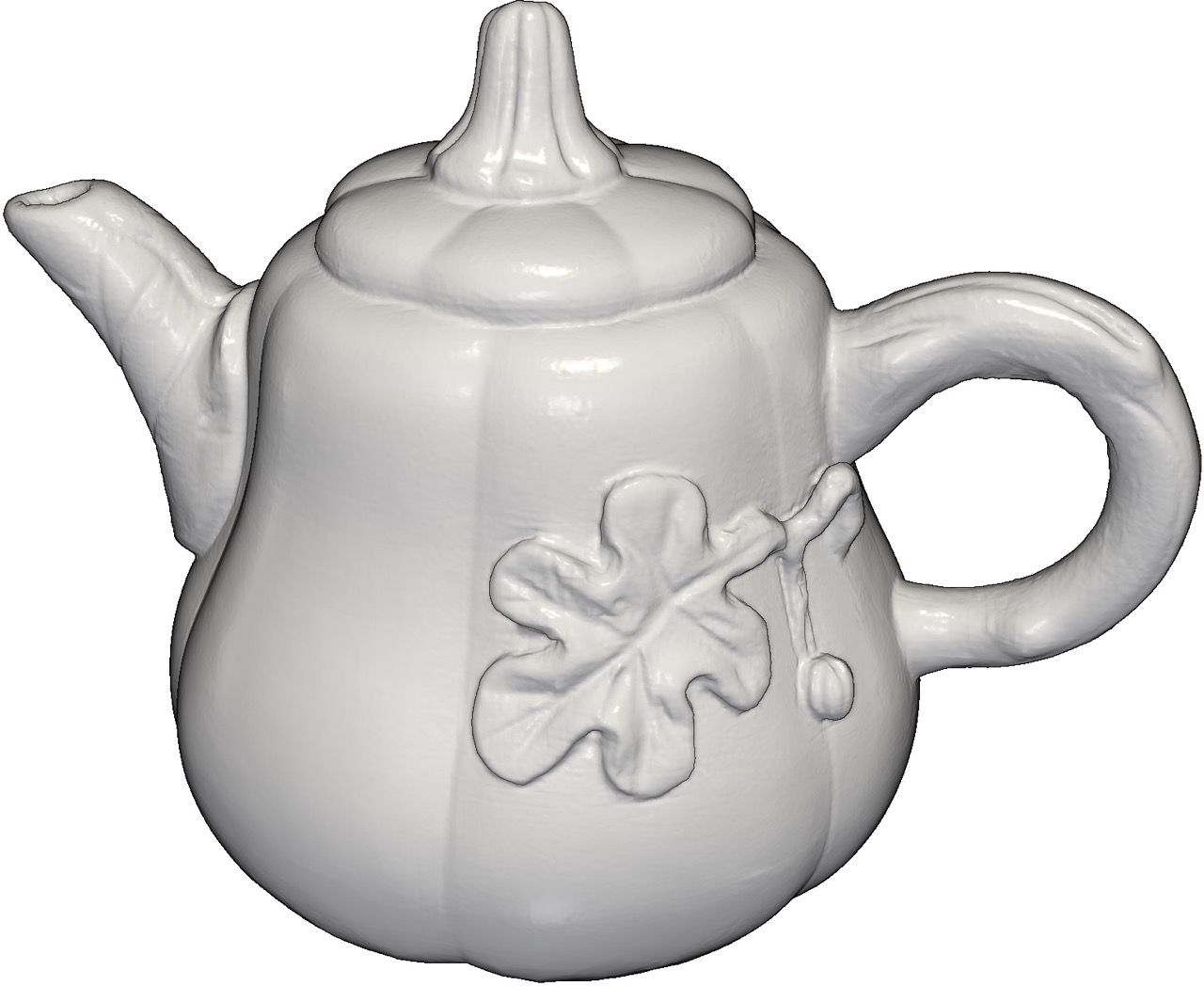} &
    \includegraphics[width=\figwidthMeshComparison\linewidth]{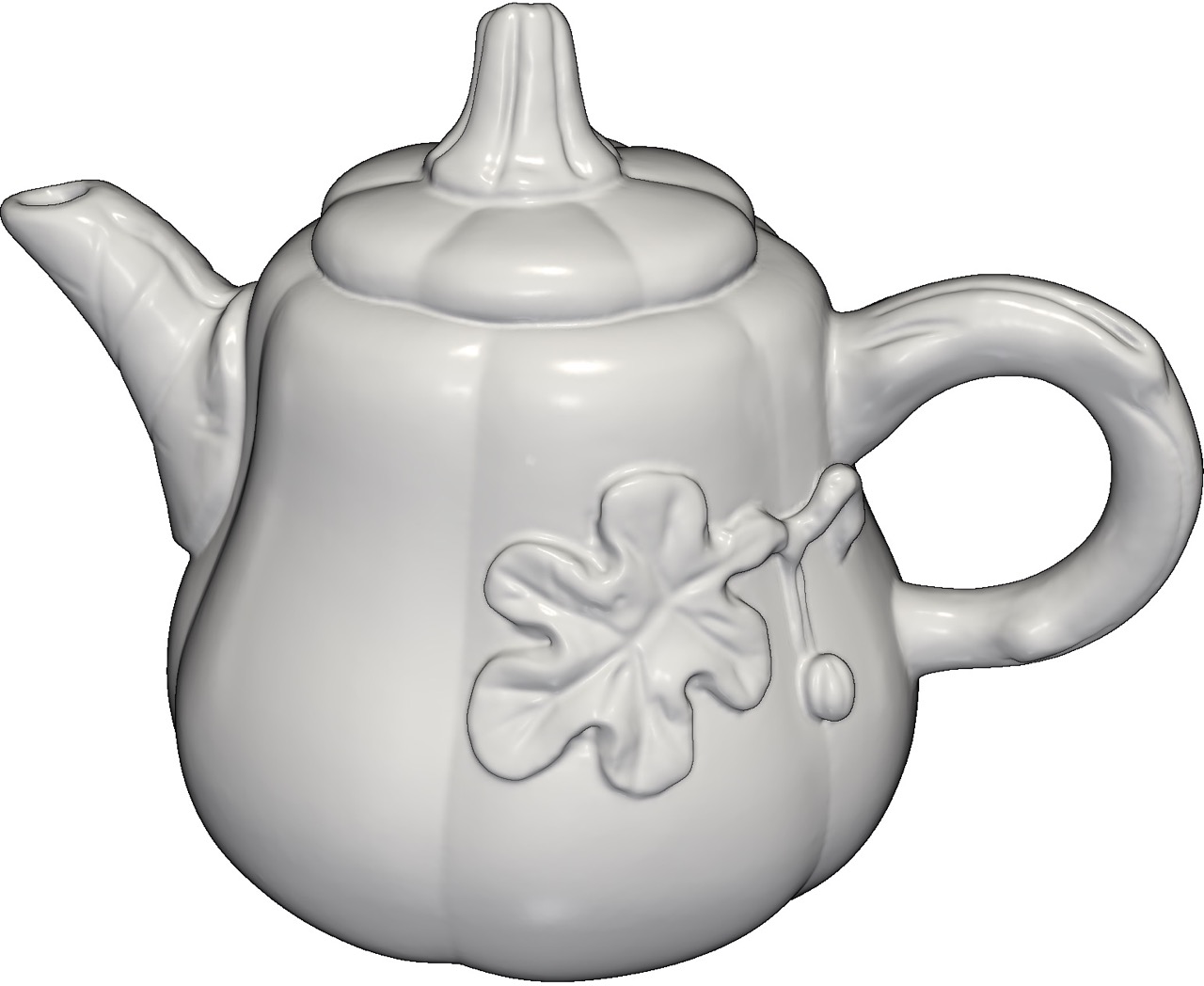}
    \\
    \includegraphics[width=\figwidthMeshComparison\linewidth]{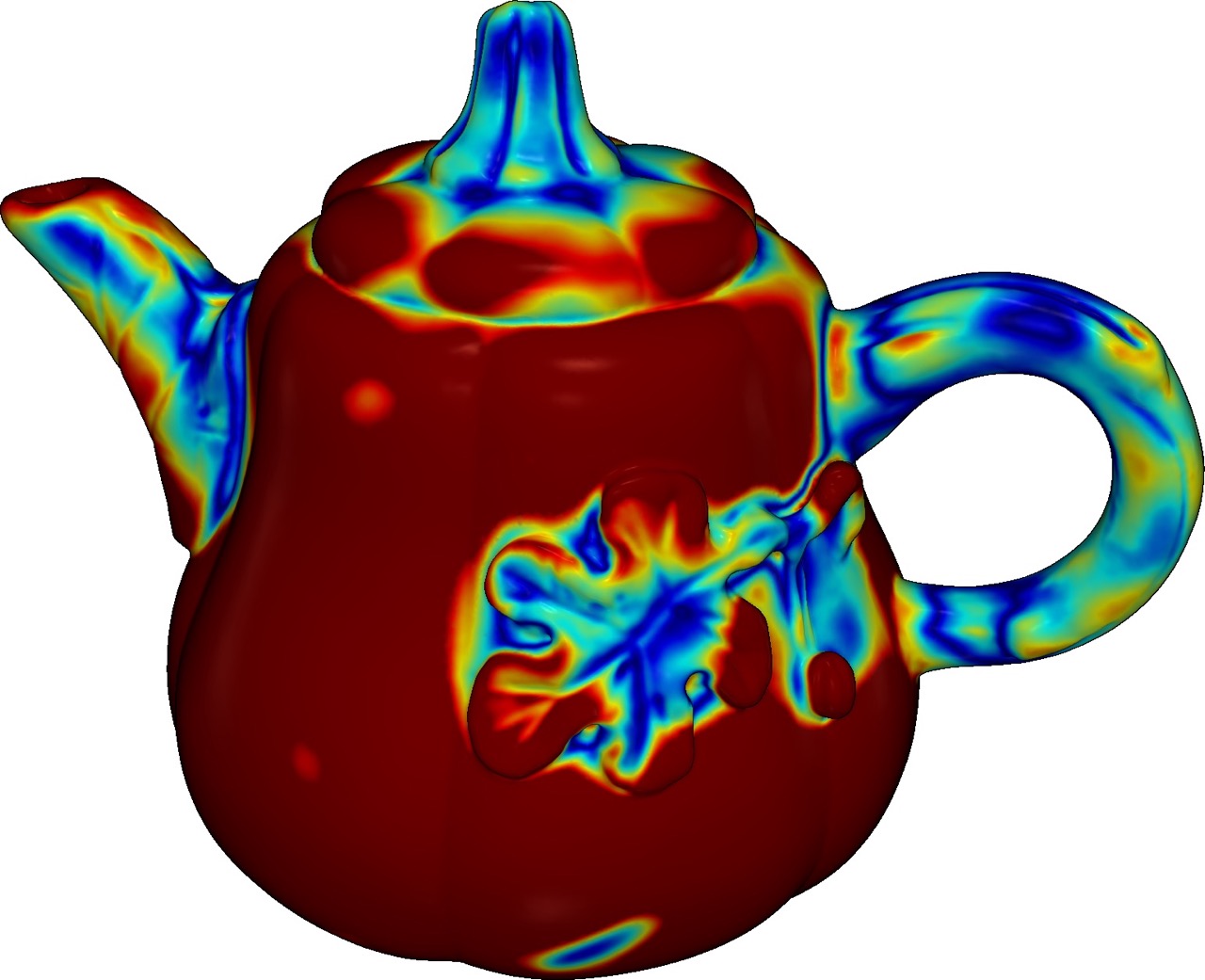} &
    \includegraphics[width=\figwidthMeshComparison\linewidth]{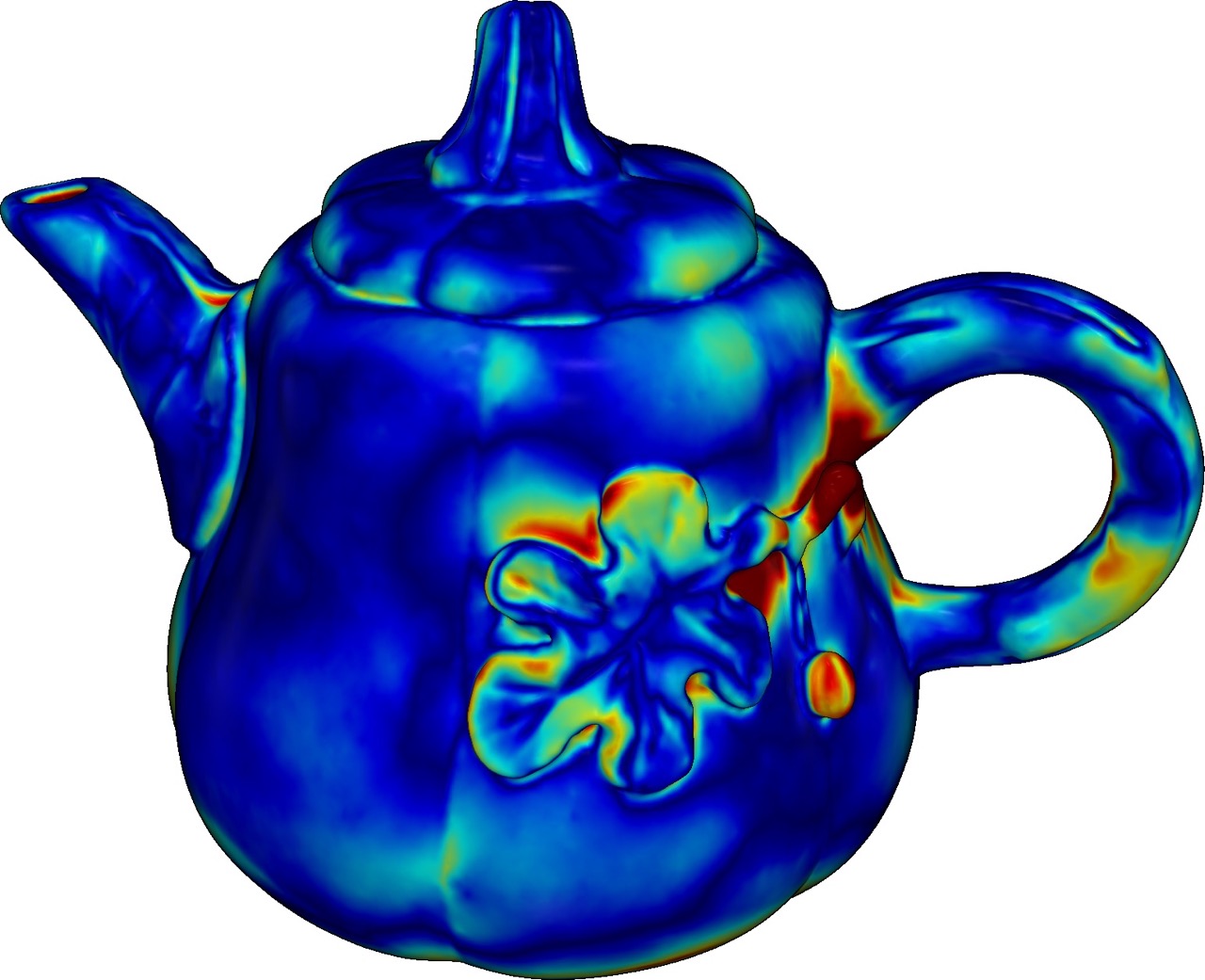} &
    \includegraphics[width=\figwidthMeshComparison\linewidth]{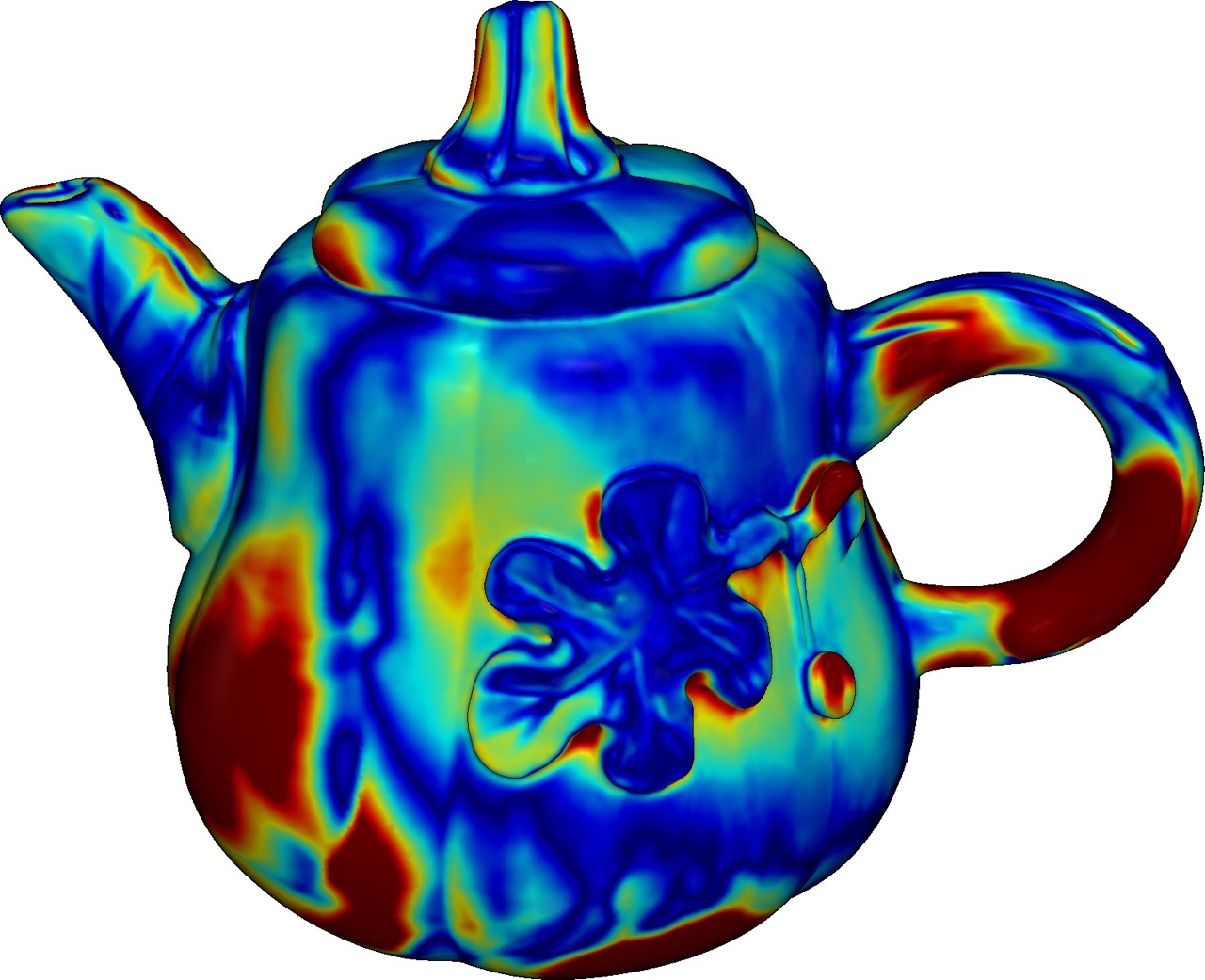} &
    \includegraphics[width=\figwidthMeshComparison\linewidth]{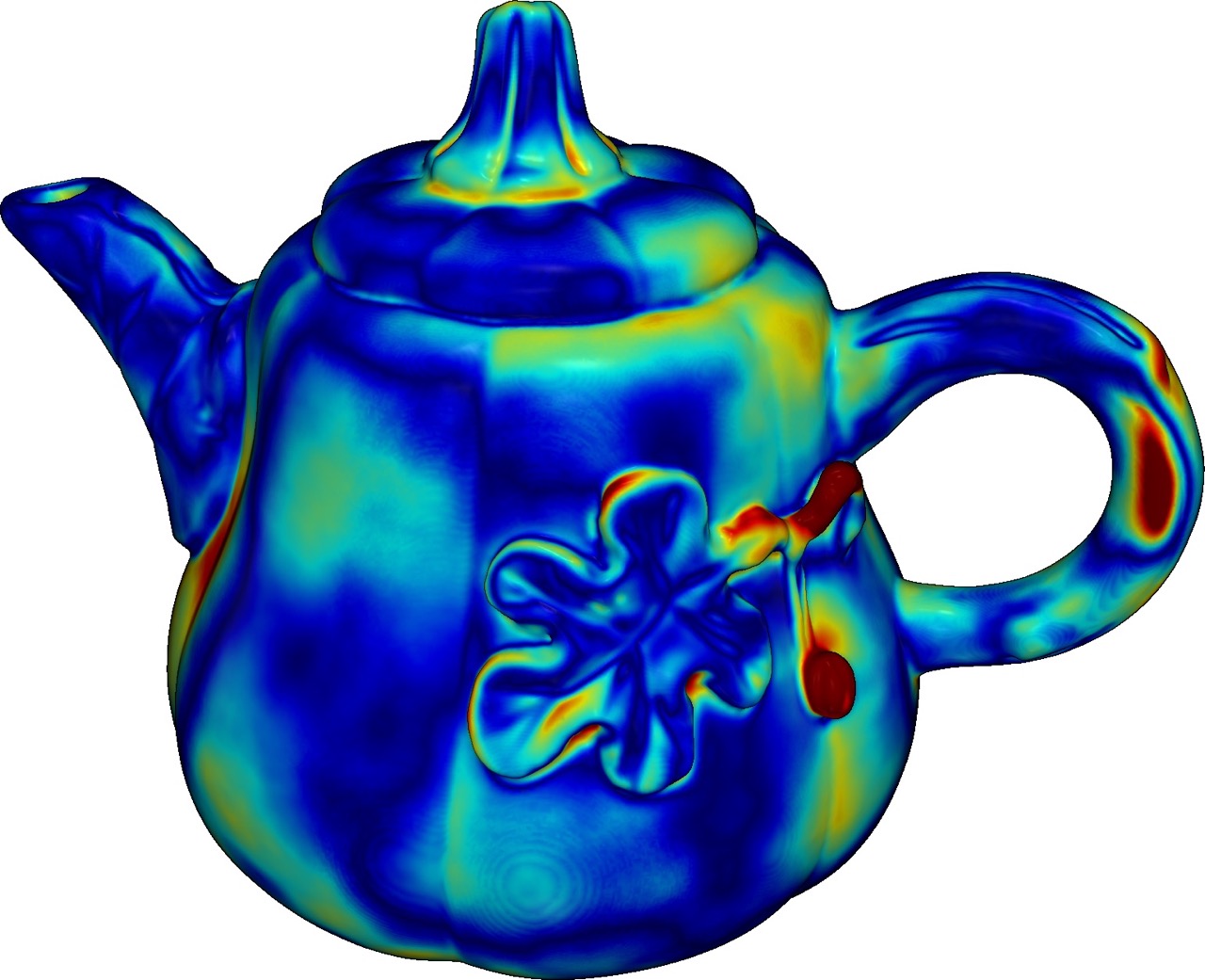} &
    \includegraphics[width=\figwidthMeshComparison\linewidth]{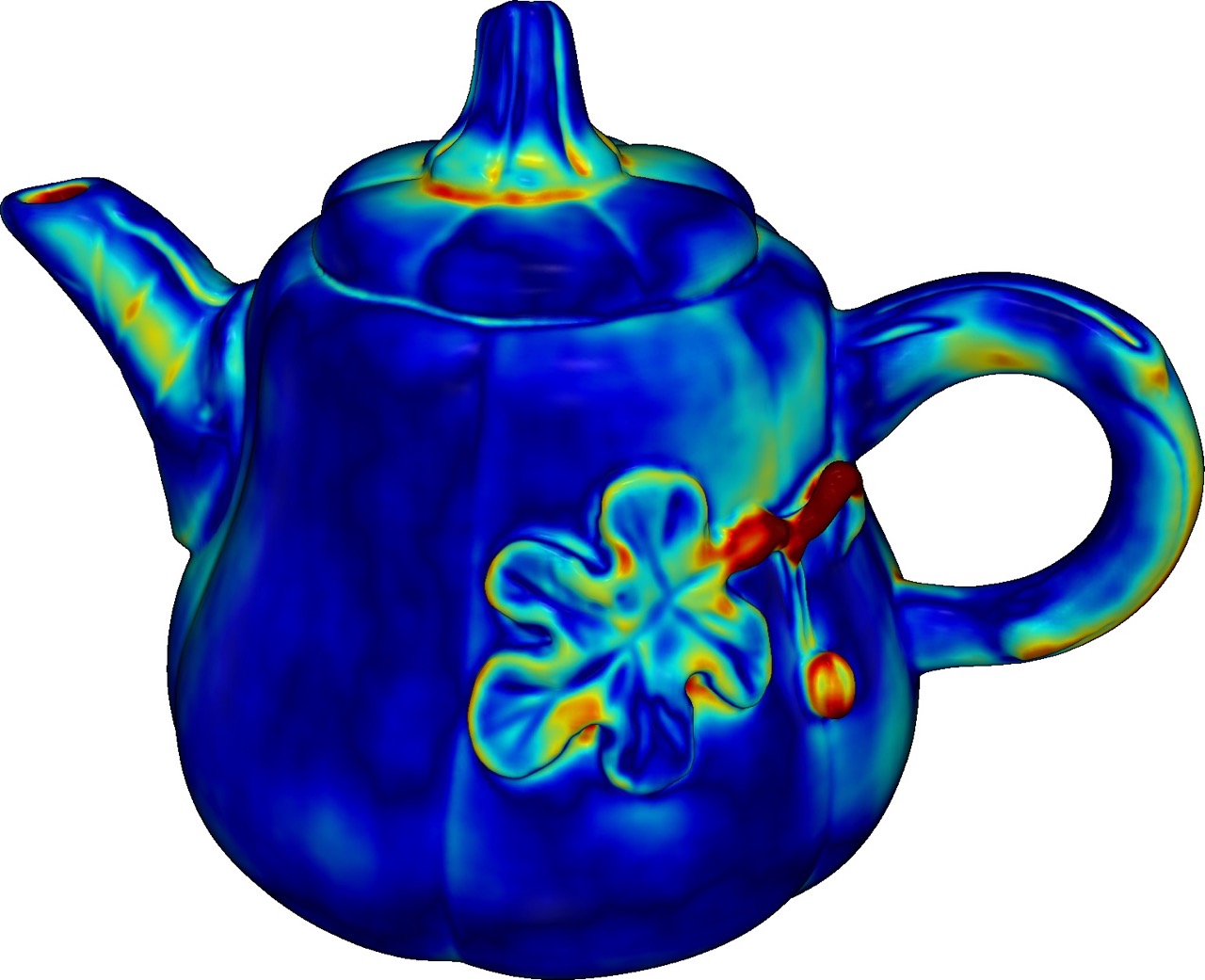} &
    \includegraphics[width=\figwidthMeshComparison\linewidth]{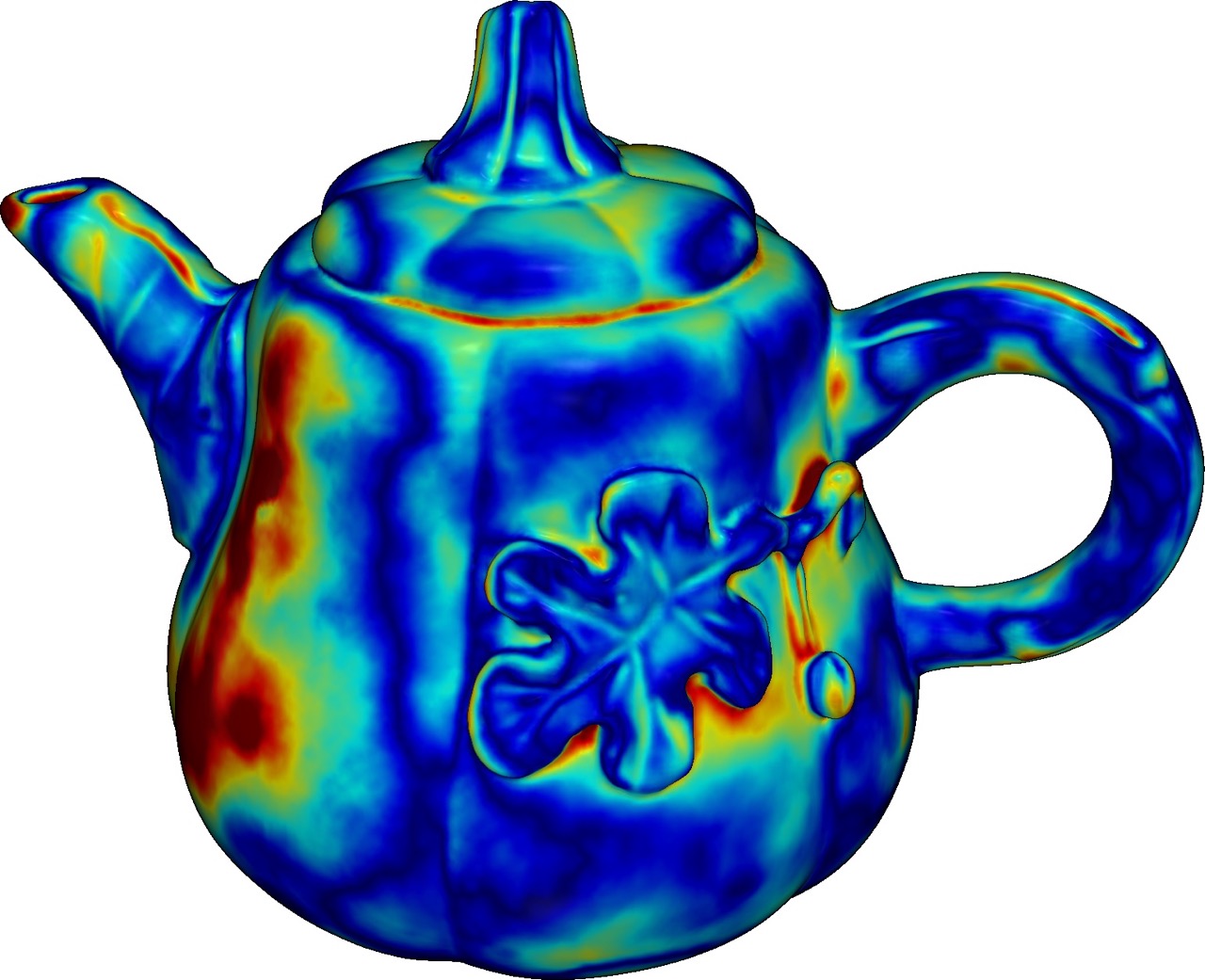}&
    \includegraphics[width=\figwidthMeshComparison\linewidth]{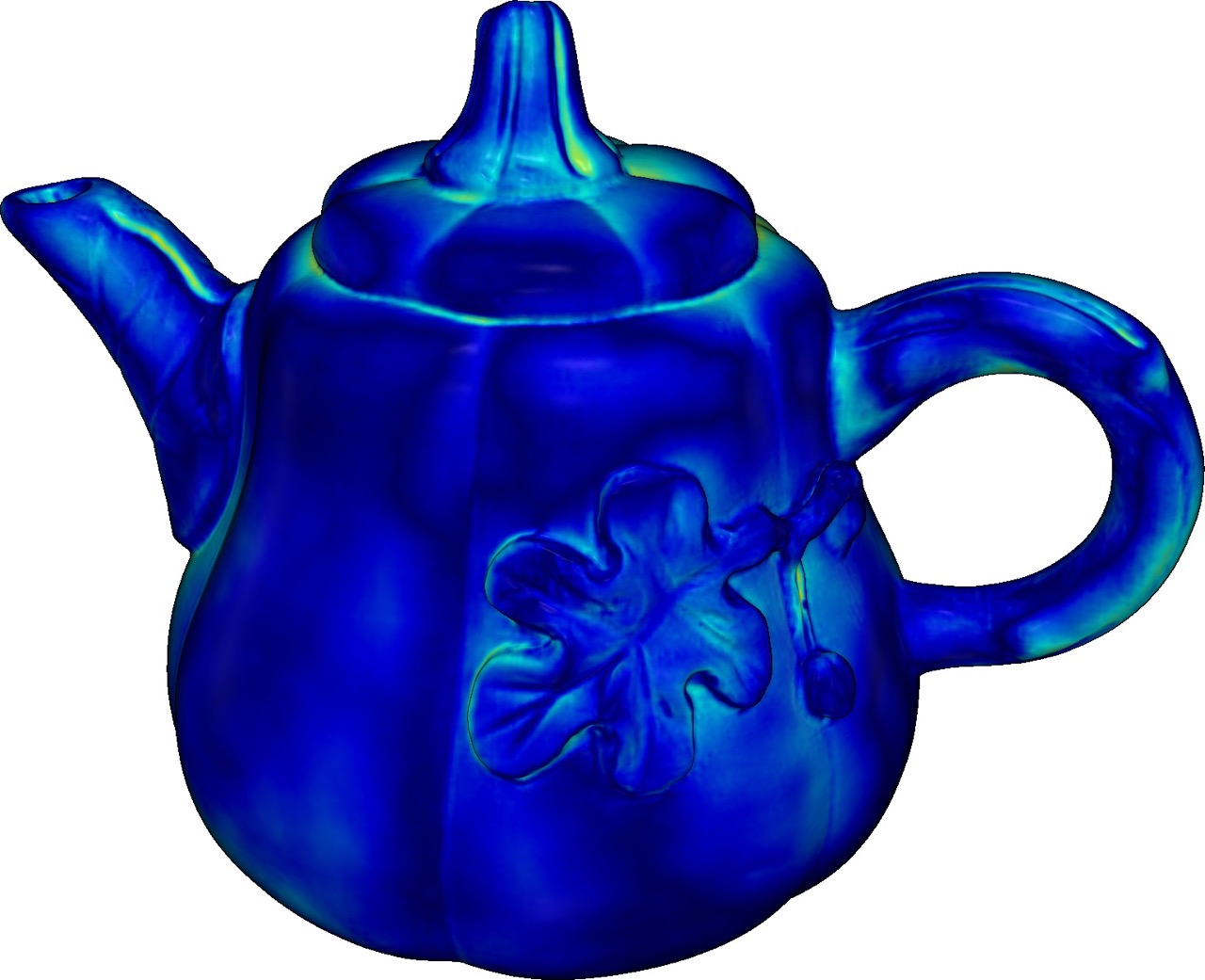} &
    \colorbar{0.05}{$\geq \SI{1}{\mm}$} 
    \end{tabular}
    }
    \caption{Qualitative comparison of recovered shapes of \emph{Buddha} and \emph{Pot2} and their error maps. Best viewed on screen.}
    \label{fig.mesh_comparison}
\end{figure*}

\paragraph{Evaluation metrics} 
We use L2 Chamfer distance (CD) and F-score with threshold $\thresholdFscore=\SI{0.5}{mm}$ to evaluate geometry accuracy~\cite{kaya2022uncertainty,knapitsch2017tanks}.
For CD and F-score, we only consider visible points from the input views by casting rays for all pixels and finding the first ray-mesh intersections (More details in \cref{sec.eval_metrics} in \suppl)

\paragraph{Results}
As reported in \Cref{tab.mvps_quan}, our method achieves the best geometry accuracy among all compared MVPS methods (See \cref{sec.benchmark_evaluation} in \suppl for normal accuracy).
We attribute this superior performance to two aspects: 1) the powerful expressive capability of multi-resolution hash encoding~\cite{muller2022ingp} and 2) the fine-grained normal estimation method \sdmunips, as discussed in \cref{sec.ablation_study}.
\Cref{fig.mesh_comparison} visualizes the recovered shapes of different methods.
Our method recovers fine-grained surface details on par with the ``GT'' meshes produced by a 3D scanner.

\Cref{tab.mvps_quan} also reports the average runtime measured on our machine with one RTX 4090Ti graphics card. 
The runtime of \rmvps and \bmvps is difficult to measure since both methods require intensive manual effort.
The code of \uanet is not publicly available, and several hours per object are required, as reported in \mvpsnet.
Both \psnerf and \mvas require hours per object since they optimize a dense MLP.
Our method is the fastest among neural implicit representation-based methods, with less than one minute per object.

On the other hand, \mvpsnet is highly efficient because its pipeline mainly consists of forward inferencing a trained neural network and an sPSR~\cite{kazhdan2013screened} step, the latter of which has been highly optimized over one decade.
Although our method is slower than sPSR~\cite{kazhdan2013screened}, we achieve better reconstruction quality within a reasonable time.

\subsection{Ablation study}
\label{sec.ablation_study}

\paragraph{Effect of normal maps}
Since multi-view normal integration is not bound to specific normal estimation methods, we investigate the effect of different normal estimation methods on normal integration methods.
We test three sets of normal maps, estimated by \sdps or \sdmunips, and the one rendered from the scanned mesh (\ie, ``GT'' normal maps provided by \diligentmv).
\Cref{tab.diff_normal} reports the L2-Chamfer distance and F-score averaged on five \diligentmv objects.
Our method consistently achieves better results than \psnerf and \mvas when using the same set of normal maps.
This verifies that our method can better fuse the information from 2.5D normal maps to 3D shapes.
Not surprisingly, a better normal estimation approach also improves the multi-view normal integration quality. 
Using the SoTA method \sdmunips for normal estimation, our method realizes the best reconstruction quality.

\paragraph{Effect of spatial encoding}
\Cref{tab.diff_normal} also compares the geometry accuracy between multi-resolution hash encoding (hash enc.) and positional encoding (pos. enc.).
Our results verify again that multi-resolution hash encoding surpasses positional encoding in expressiveness, resulting in enhanced geometry accuracy.
Consequently, we conclude that the keys to superior MVPS performance are high-quality normal estimation, an expressive surface representation, and an effective normal fusion approach.

\paragraph{Effect of SDF gradient computation}
\Cref{tab.sdf_gradient} compares the geometry accuracy and runtime using finite difference (FD), Pytorch's automatic differentiation (AD), and directional finite difference (DFD) to compute the SDF gradient.
All other parameter settings are kept the same.
We do not observe a significant difference regarding reconstruction quality.
However, DFD performs consistently faster than FD and AD by accelerating both the forward rendering and backward optimization.
Compared to FD, DFD avoids redundant SDF evaluations in forward rendering.
Compared to AD, DFD simplifies the computation graph by avoiding computing second-order derivatives in the backward pass.
Consequently, DFD is nearly twice as efficient as AD, and about three times faster than FD.
Unlike \neustwo requiring ReLU activation, the acceleration by DFD applies to any parametric SDF representation.

\begin{table}
\small
    \centering
    \caption{Quantitative evaluation using normal maps estimated by different photometric stereo methods as inputs. The metrics are averaged over five \diligentmv objects. Darker colors indicate better results.
    }
    \resizebox{\linewidth}{!}{
    \begin{tabular}{@{}lccc||ccc}
    \toprule
    & \multicolumn{3}{c||}{L2 Chamfer distance [\si{mm}] ($\downarrow$)} & \multicolumn{3}{c}{F-score  ($\tau = \SI{0.5}{mm}$) ($\uparrow$)} \\
         & \sdps & \sdmunips & GT & \sdps & \sdmunips & GT \\
       \midrule
        \mvas  & \cellcolor{orange!28} 0.307 & \cellcolor{orange!12} 0.332 &  \cellcolor{orange!52} 0.292 & \cellcolor{orange!20} 0.816 & \cellcolor{orange!12} 0.812 & \cellcolor{orange!52} 0.859\\
       \psnerf  & \cellcolor{orange!44} 0.293 & \cellcolor{orange!36} 0.296 &  \cellcolor{orange!76} 0.224 & \cellcolor{orange!44} 0.853 & \cellcolor{orange!28} 0.844 & \cellcolor{orange!68} 0.938 \\
       Ours (pos. enc.) & \cellcolor{orange!20} 0.301 & \cellcolor{orange!68} 0.228 & \cellcolor{orange!84} 0.195 & \cellcolor{orange!36} 0.849 & \cellcolor{orange!76} 0.942 & \cellcolor{orange!92} 0.977 \\
       Ours (hash enc.)  & \cellcolor{orange!60}  0.281  &  \cellcolor{orange!92} 0.194 &  \cellcolor{orange!100} 0.114  & \cellcolor{orange!60} 0.860 & \cellcolor{orange!84}  0.962 & \cellcolor{orange!100}0.998\\
    \bottomrule
    \end{tabular}
    }
    \label{tab.diff_normal}
\end{table}

\begin{table}[]
\footnotesize
    \centering
    \caption{Ablation study using different strategies for SDF gradients computation in volume rendering. We report the metrics averaged over five \diligentmv objects. \textbf{FD}: Finite difference, \textbf{AD}: Automatic differentiation, \textbf{DFD}: Directional finite difference.}
    \resizebox{0.8\linewidth}{!}{
    \begin{tabular}{@{}lrr||rrr@{}}
    \toprule
     & \multicolumn{2}{c||}{Geo. Acc.} & \multicolumn{3}{c}{Runtime (Sec)}
     \\
        & CD & F-Score & Forward & Backward & Total\\
       \midrule
       FD  & 0.192 & 0.963 & 47.7 & 70.3 & 124.6 \\
       AD  & 0.195 & 0.961 & 29.4 & 42.4 & 78.1\\
       DFD & 0.194 & 0.961 & 19.6 & 22.5 & 48.3\\
    \bottomrule
    \end{tabular}
    }
    \label{tab.sdf_gradient}
\end{table}

\subsection{Qualitative results on our data}
\label{sec.exp_real_world}

\begin{figure}
    \centering
    \includegraphics[width=0.48\linewidth]{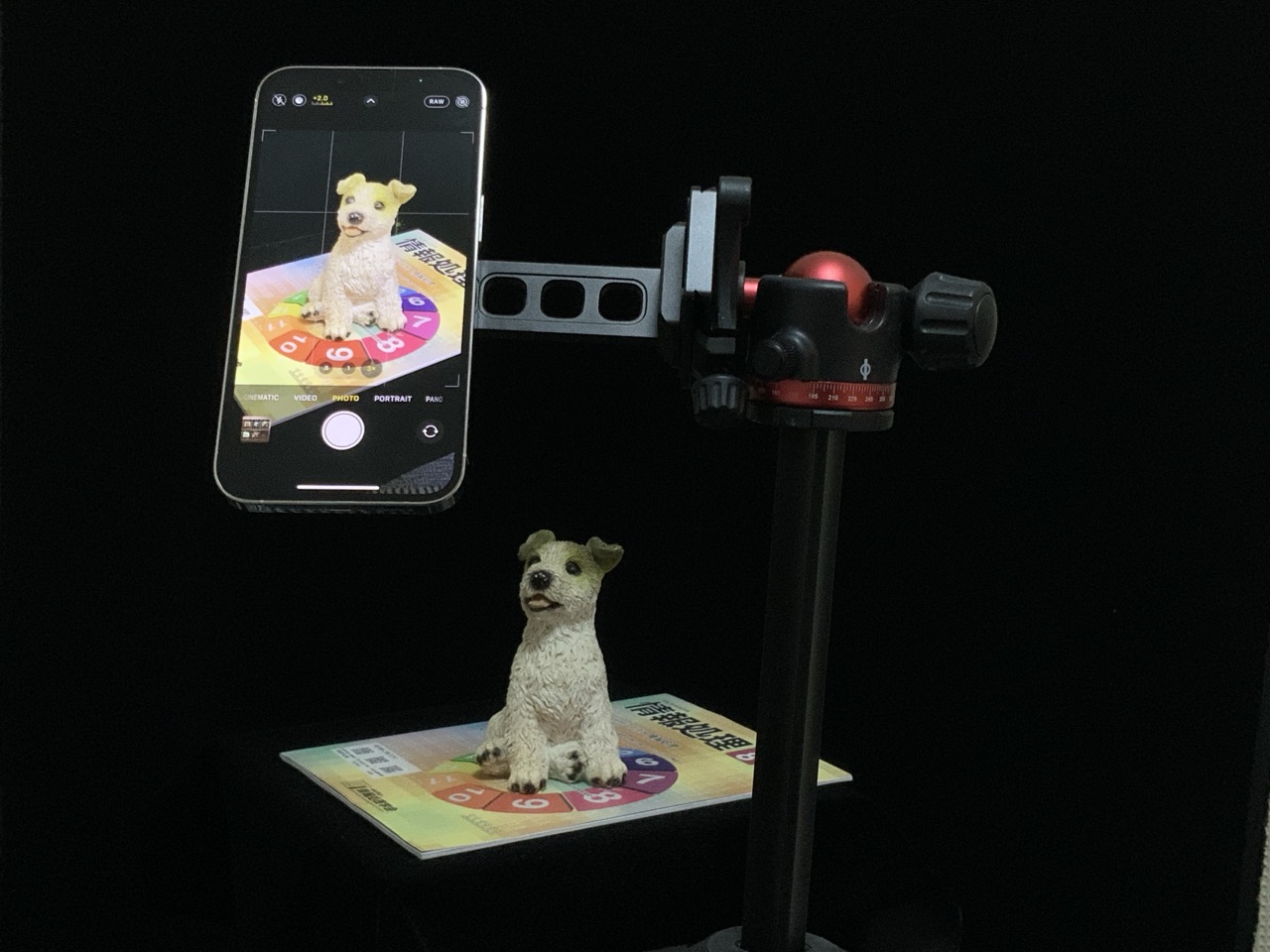}
    \includegraphics[width=0.48\linewidth]{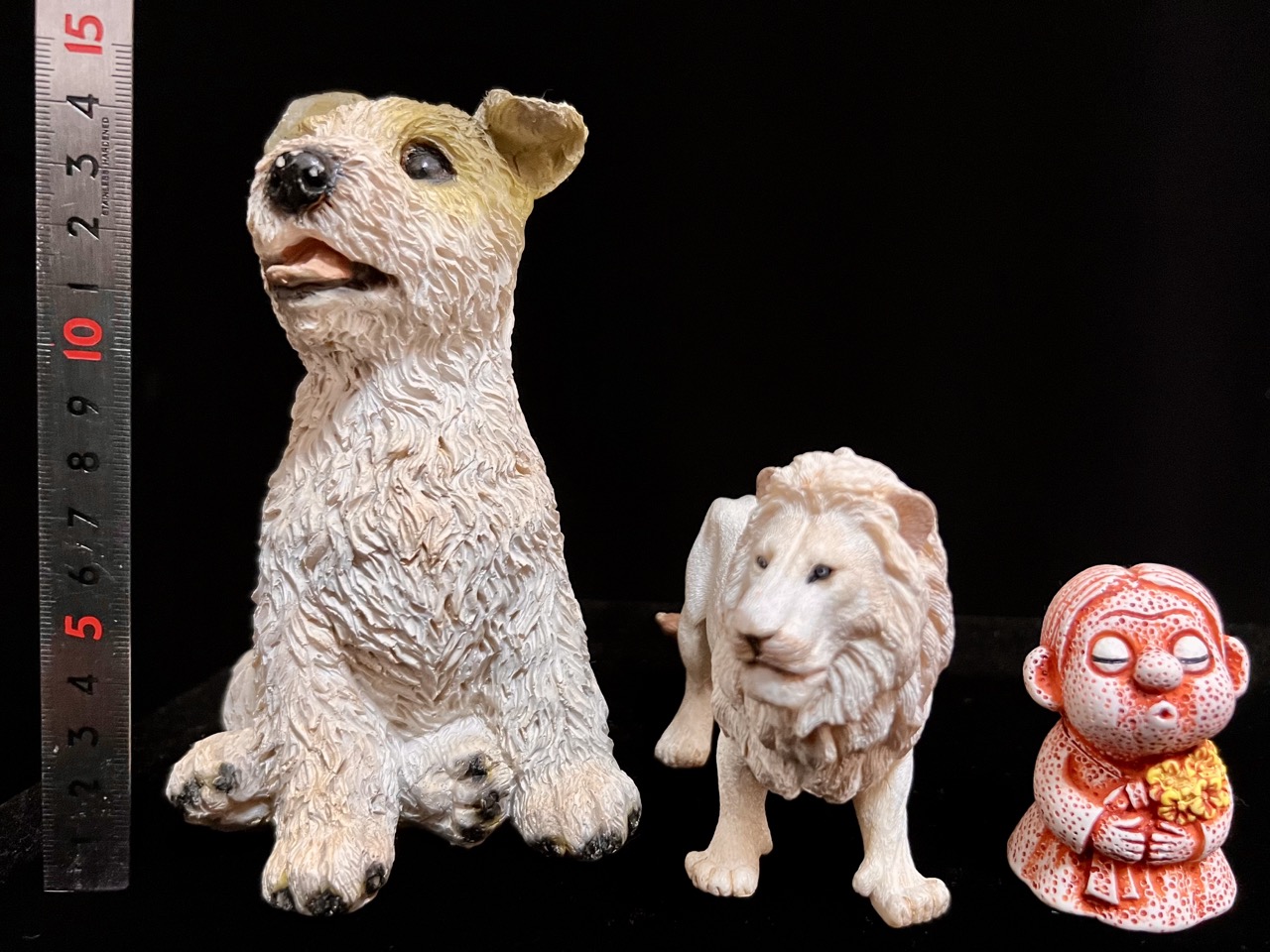}
    \caption{Our real-world data capture setup. The target object is placed on a turntable before a black background. We use an iPhone to take raw images and manually move a video light to illuminate the object from the camera side. Objects are \SI{6}{cm} to \SI{15}{cm} high.}
    \label{fig.real_world_data_capture}
\end{figure}

Since \diligentmv only captures objects with relatively simple geometry using low-resolution images ($0.3$MP), we also captured objects with more complex details using an iPhone ($12$MP) to demonstrate the performance of \ourmethod.
\Cref{fig.real_world_data_capture} shows our capture setup.
Thanks to the generality of \sdmunips, we can capture the objects under casual light conditions (\eg, within an apartment) without ensuring a darkroom setup.
Moreover, we manually move a video light during the capture without any photometric or geometric light calibration.
In total, we capture $18$ viewpoints $\times$ $13$ light conditions for each object ($1$ under ambient light for SfM). 
We use Metashape~\cite{metashape} to calibrate camera parameters, use SAM~\cite{sam} to create foreground masks, and apply \sdmunips to every $12$ image to infer the normal map in each view.

\begin{figure}
\newcommand{\figBwidth}{0.25}
\scriptsize
    \centering
    \begin{tabular}{@{}c@{}c@{}c@{}c@{}c@{}}
    Image & \neustwo & Ours & EinScan SE \\
     \includegraphics[width=\figBwidth\linewidth]{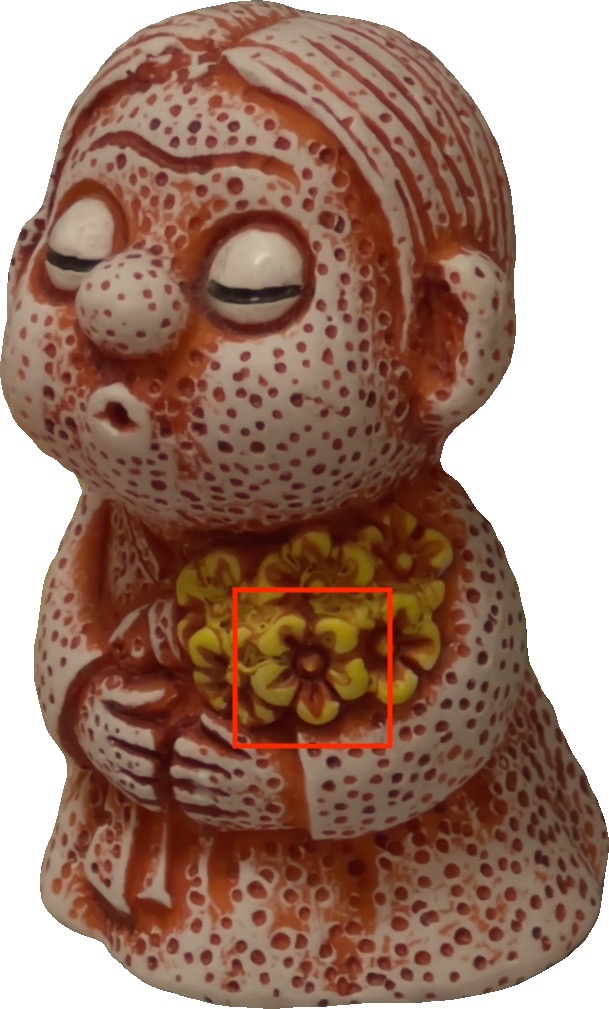}&  \includegraphics[width=\figBwidth\linewidth]{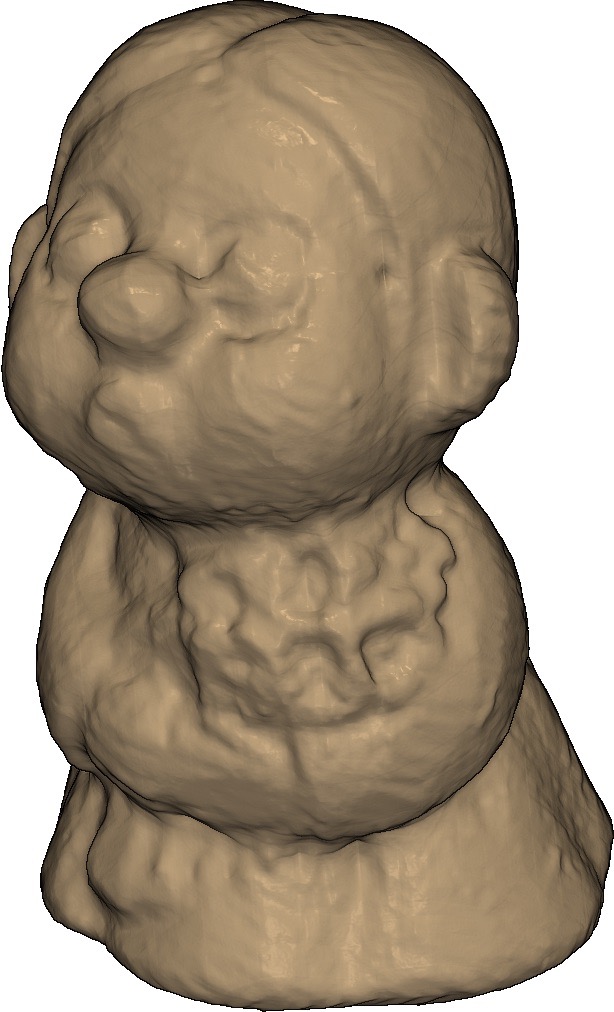}&
     \includegraphics[width=\figBwidth\linewidth]{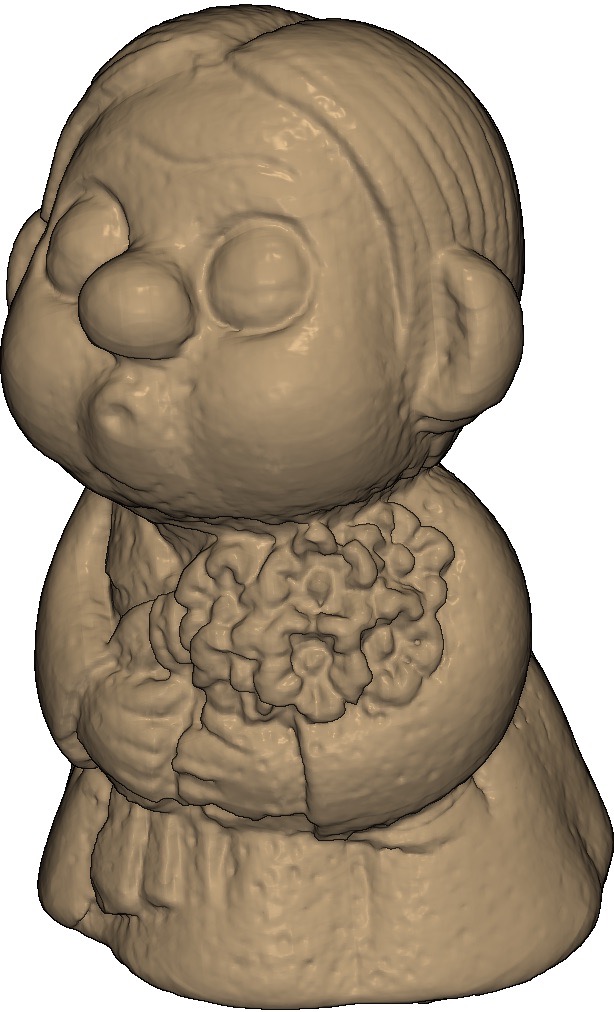}
     &
    \includegraphics[width=\figBwidth\linewidth]{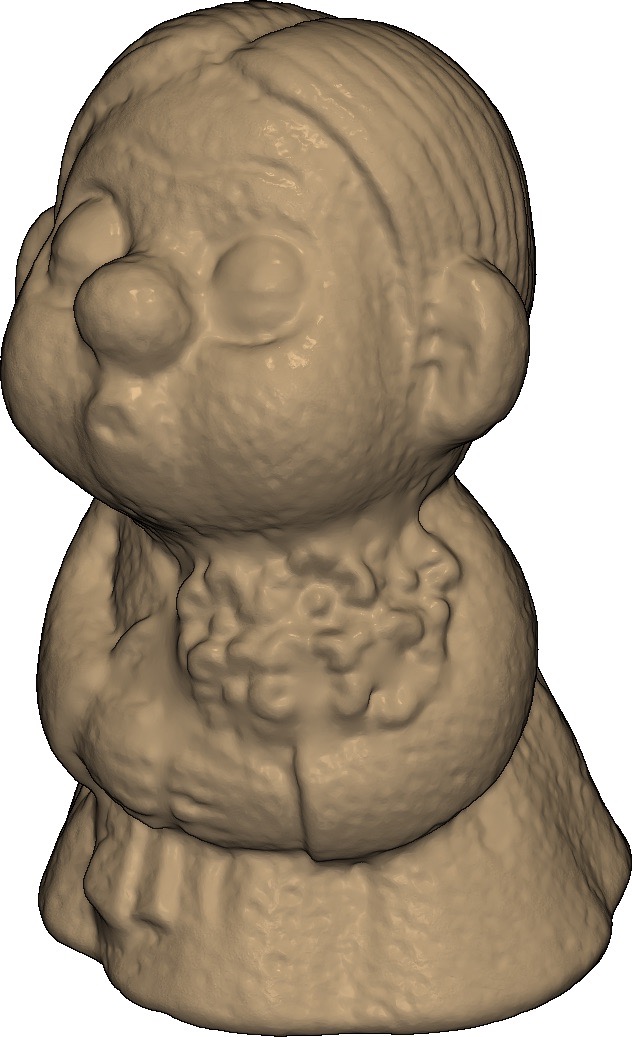}
     \\
    \includegraphics[width=\figBwidth\linewidth]{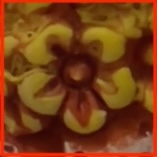}&  \includegraphics[width=\figBwidth\linewidth]{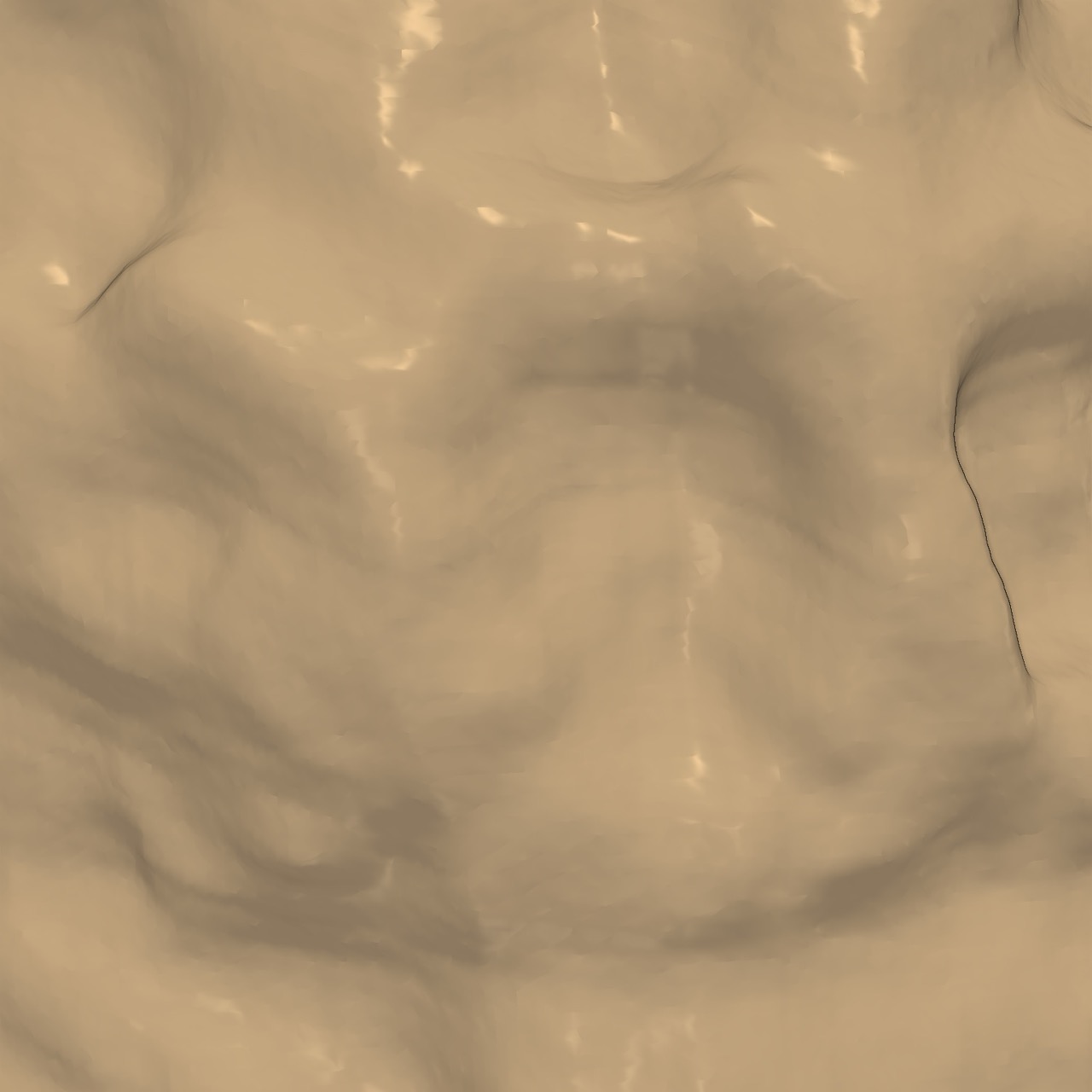}&
     \includegraphics[width=\figBwidth\linewidth]{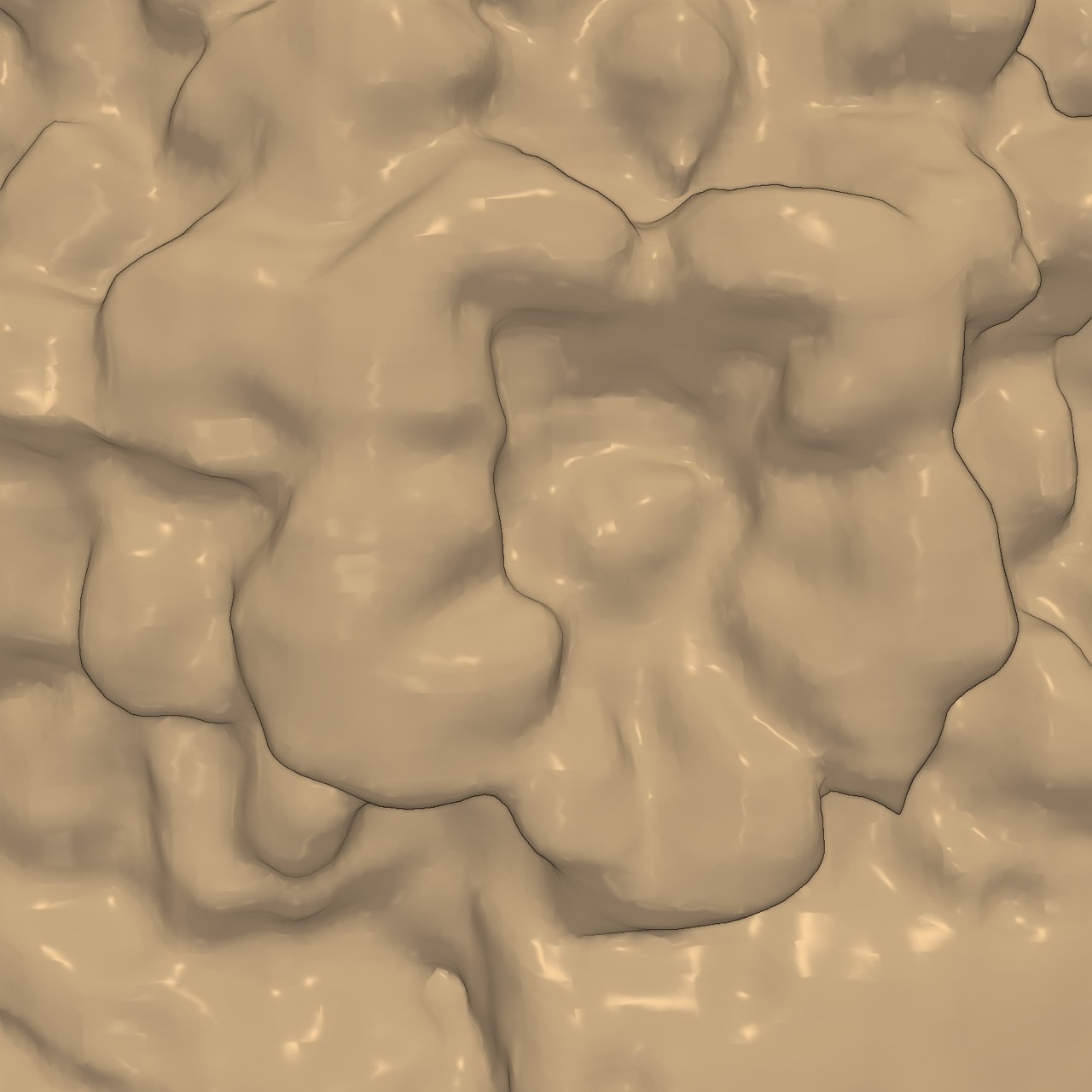}
     &
    \includegraphics[width=\figBwidth\linewidth]{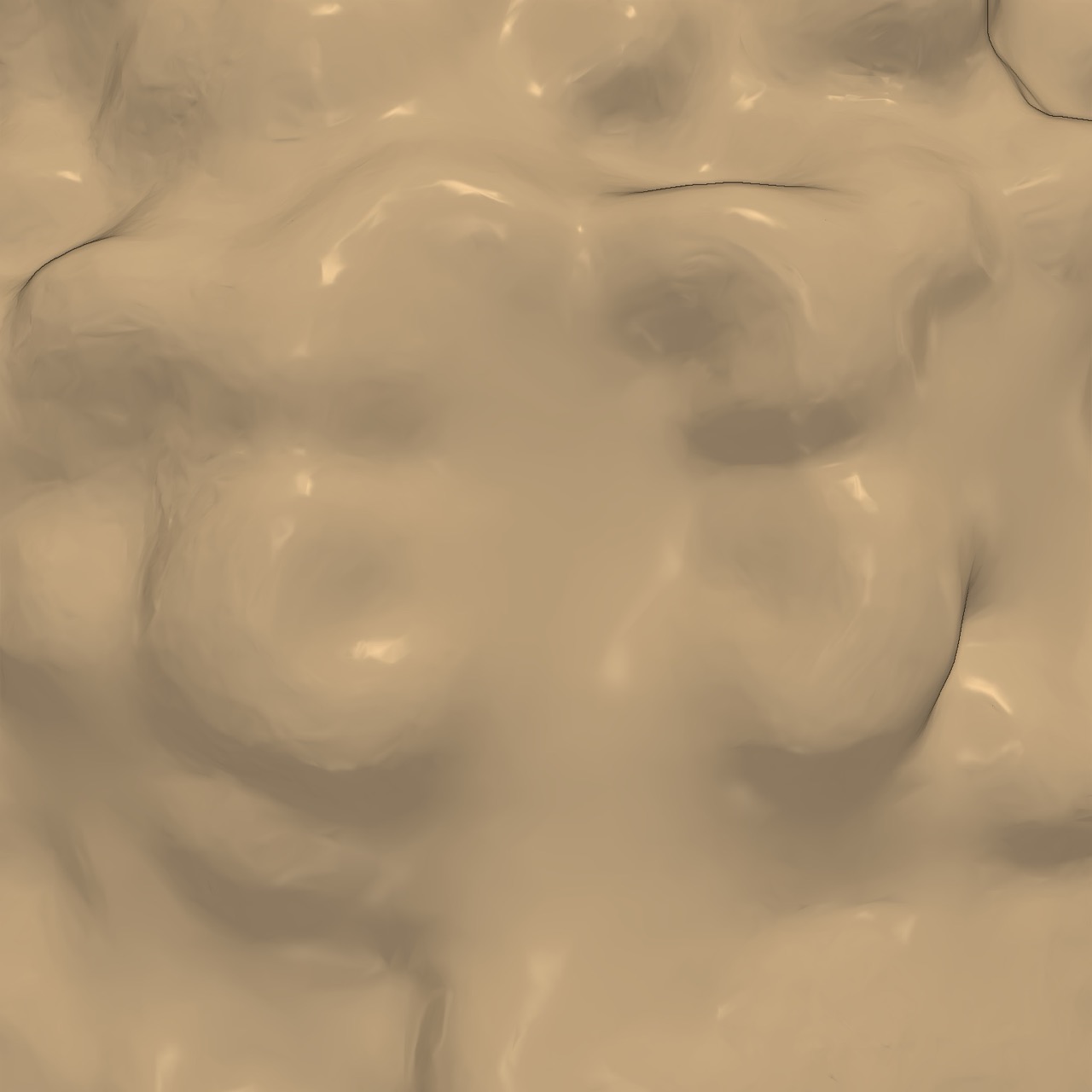}\\
     \includegraphics[width=\figBwidth\linewidth]{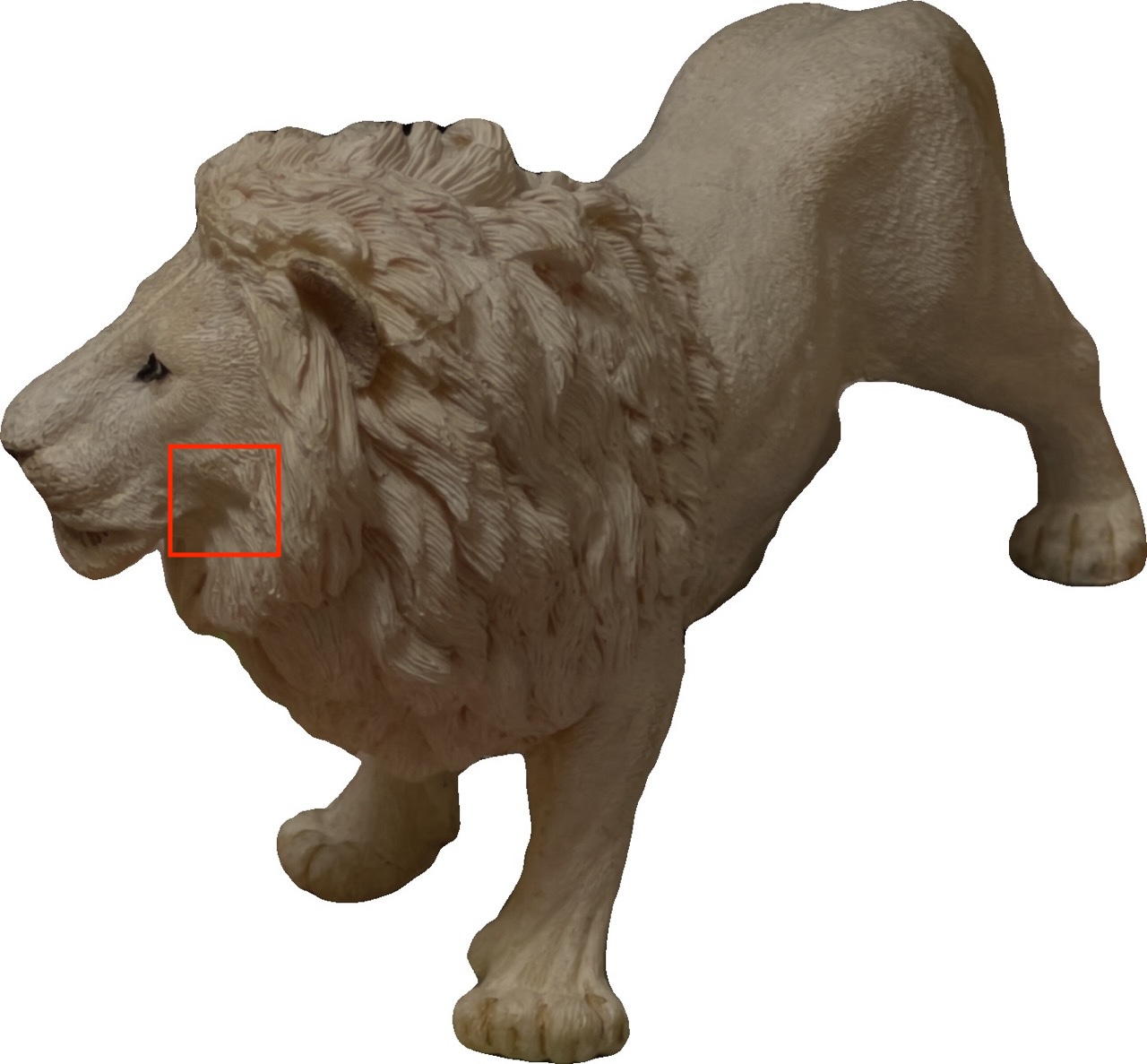}& 
     \includegraphics[width=\figBwidth\linewidth]{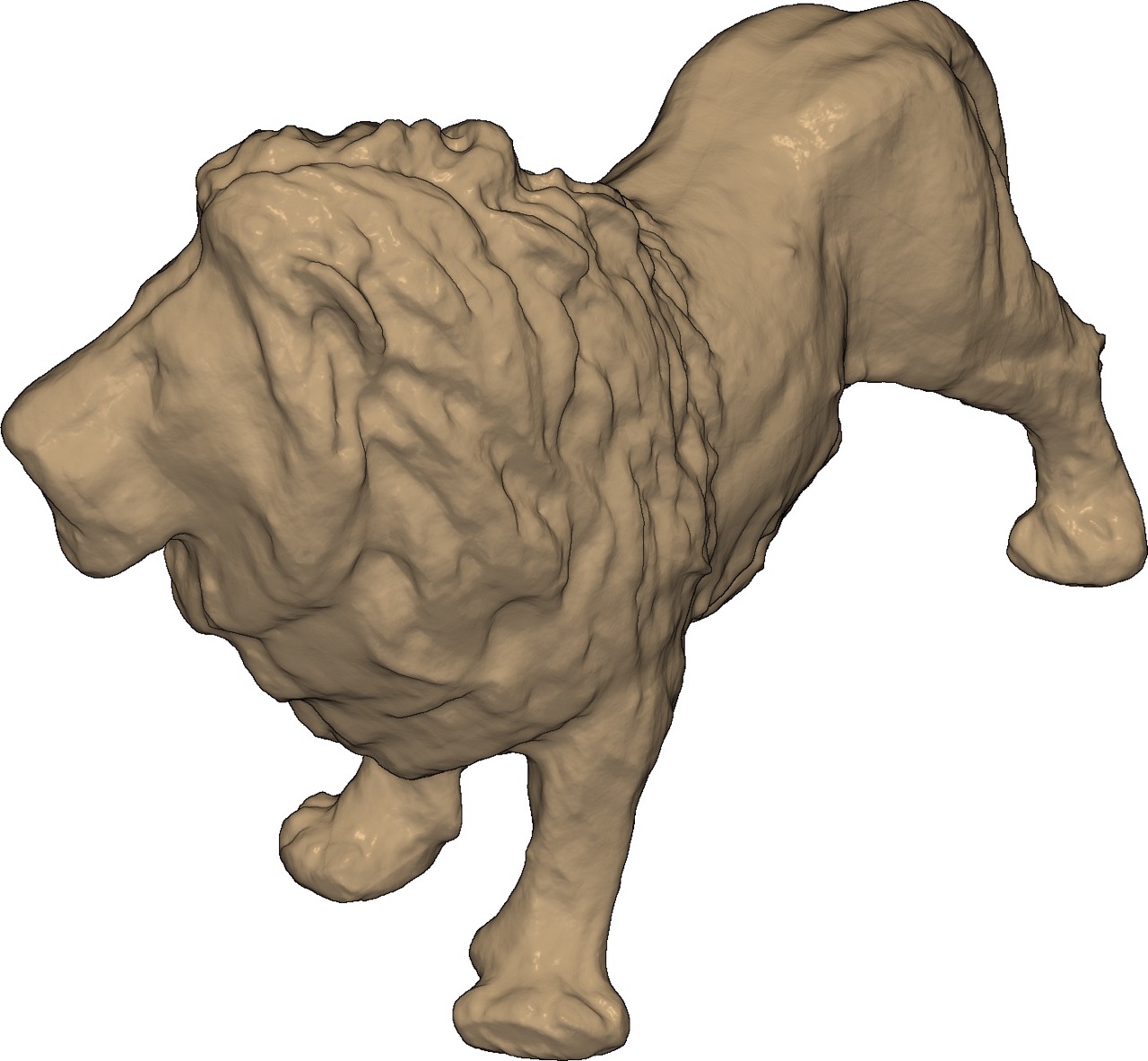}&
    \includegraphics[width=\figBwidth\linewidth]{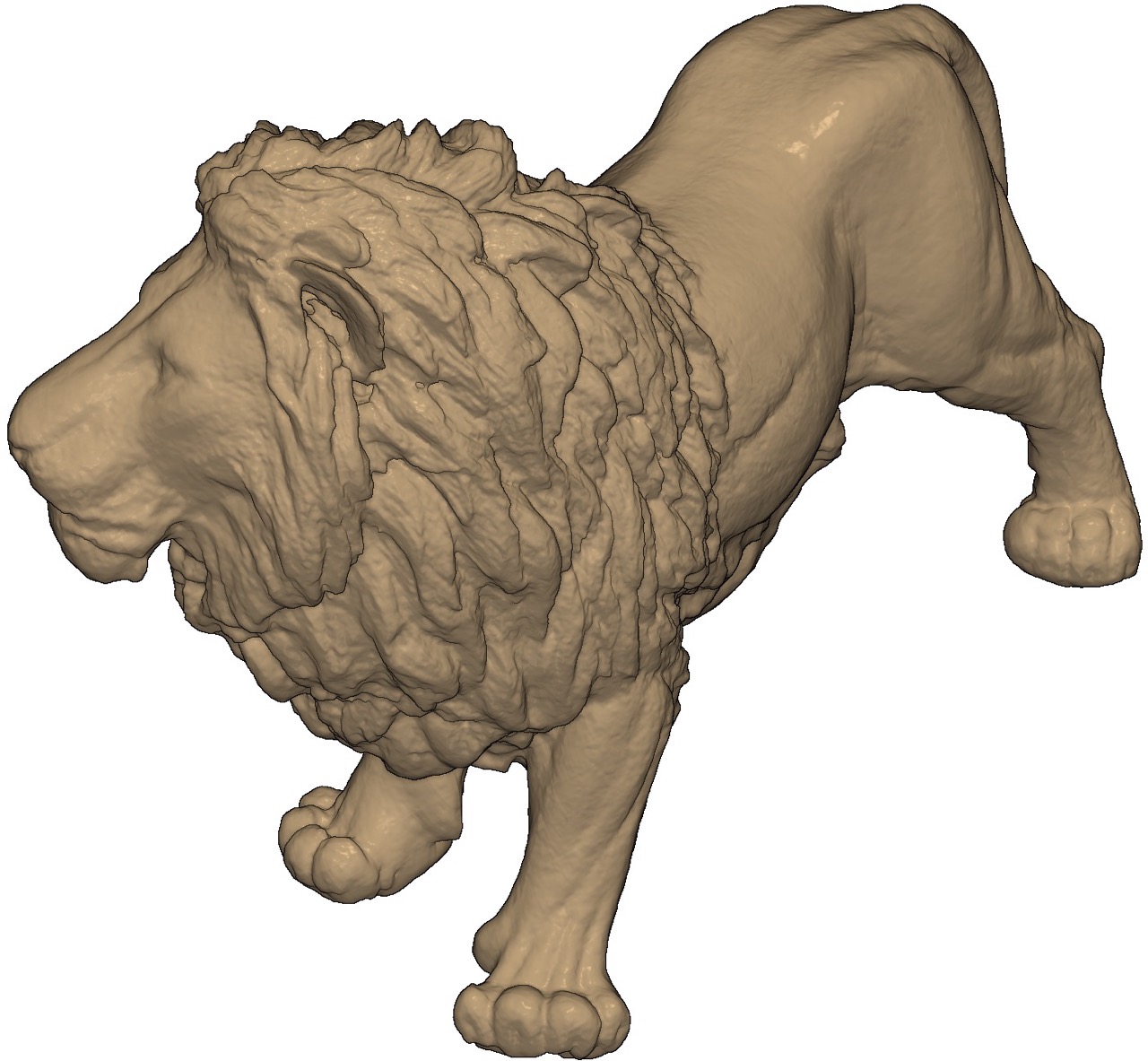}&
    \includegraphics[width=\figBwidth\linewidth]{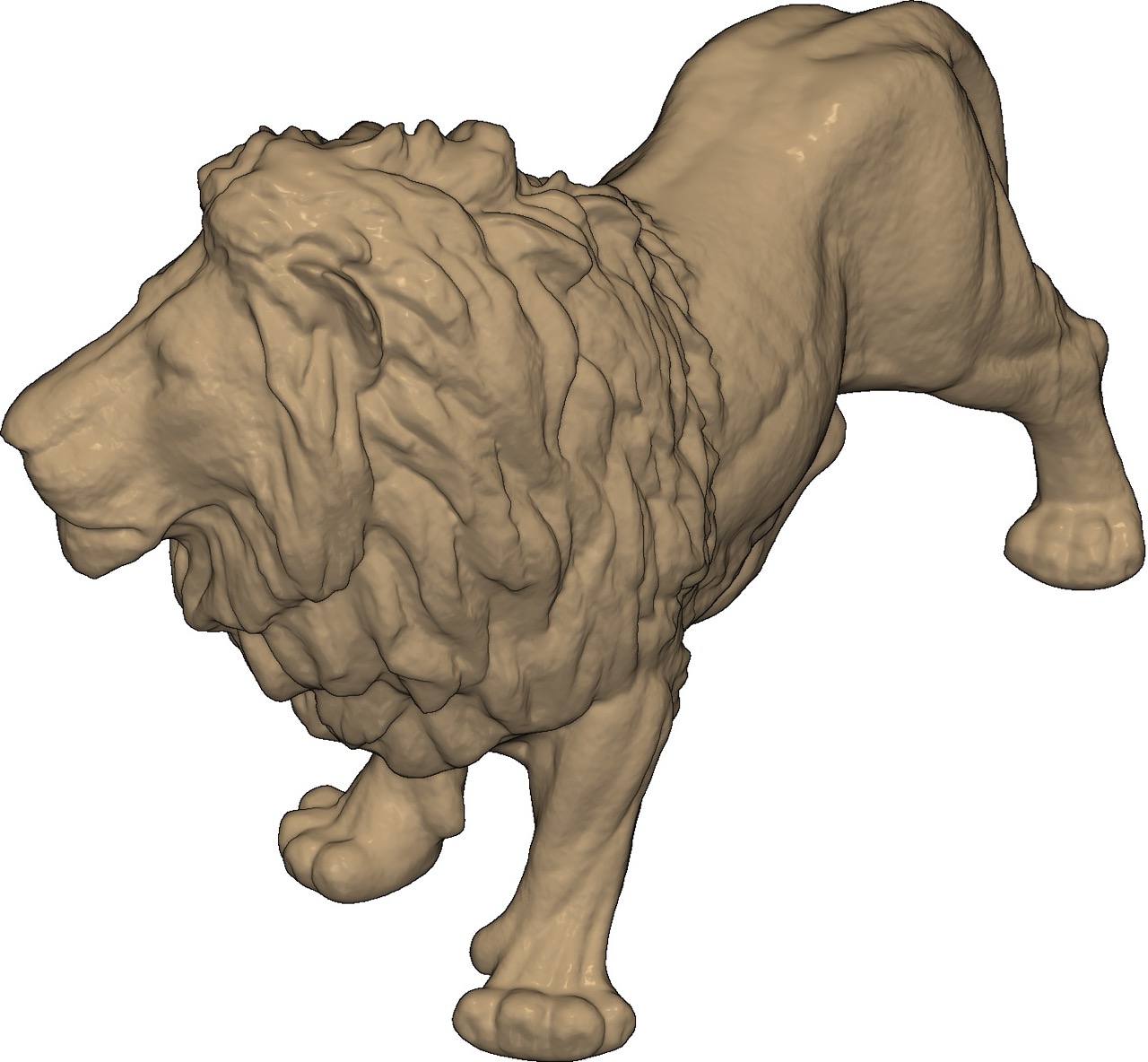}
    \\
    \includegraphics[width=\figBwidth\linewidth]{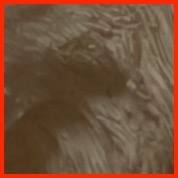}& 
     \includegraphics[width=\figBwidth\linewidth]{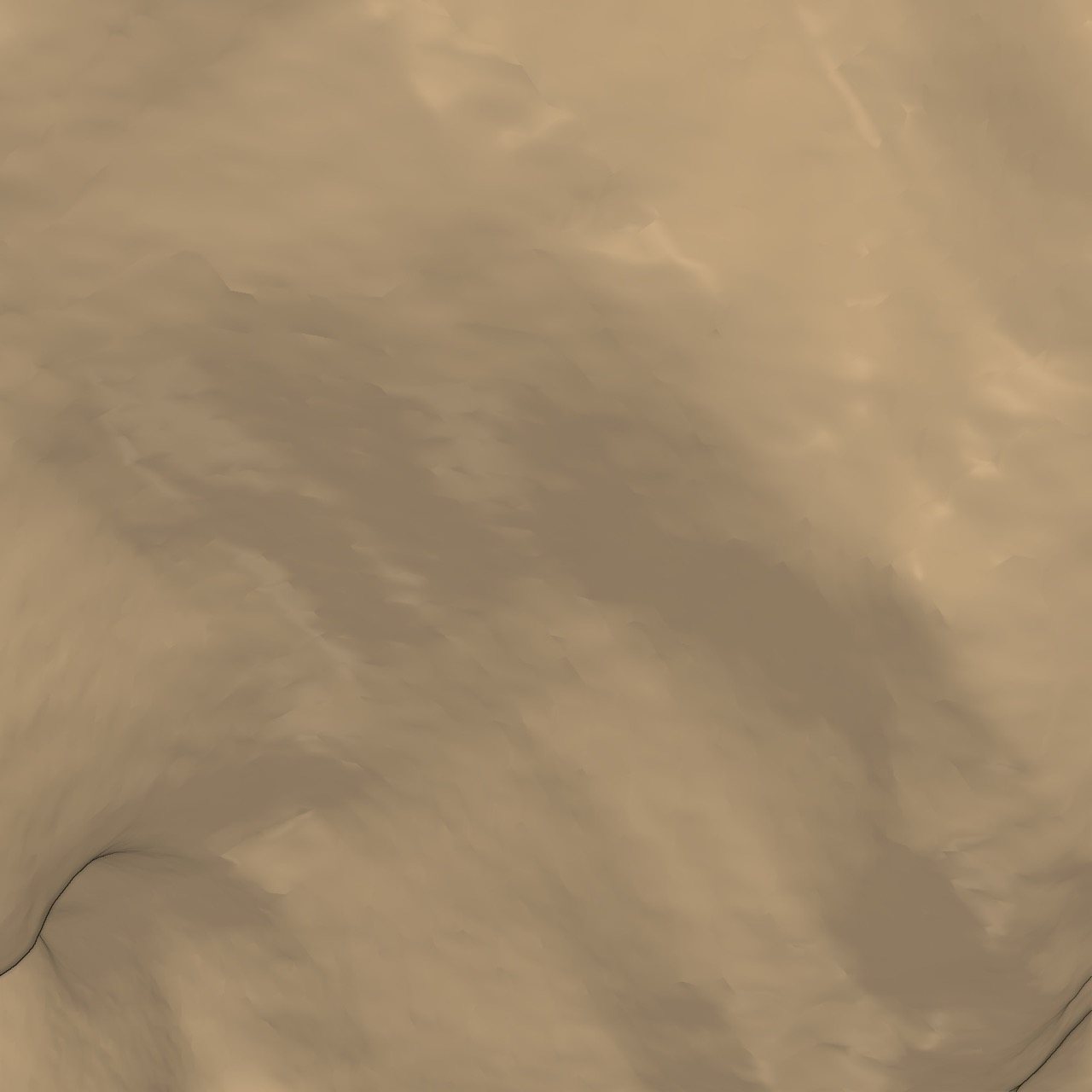}&
    \includegraphics[width=\figBwidth\linewidth]{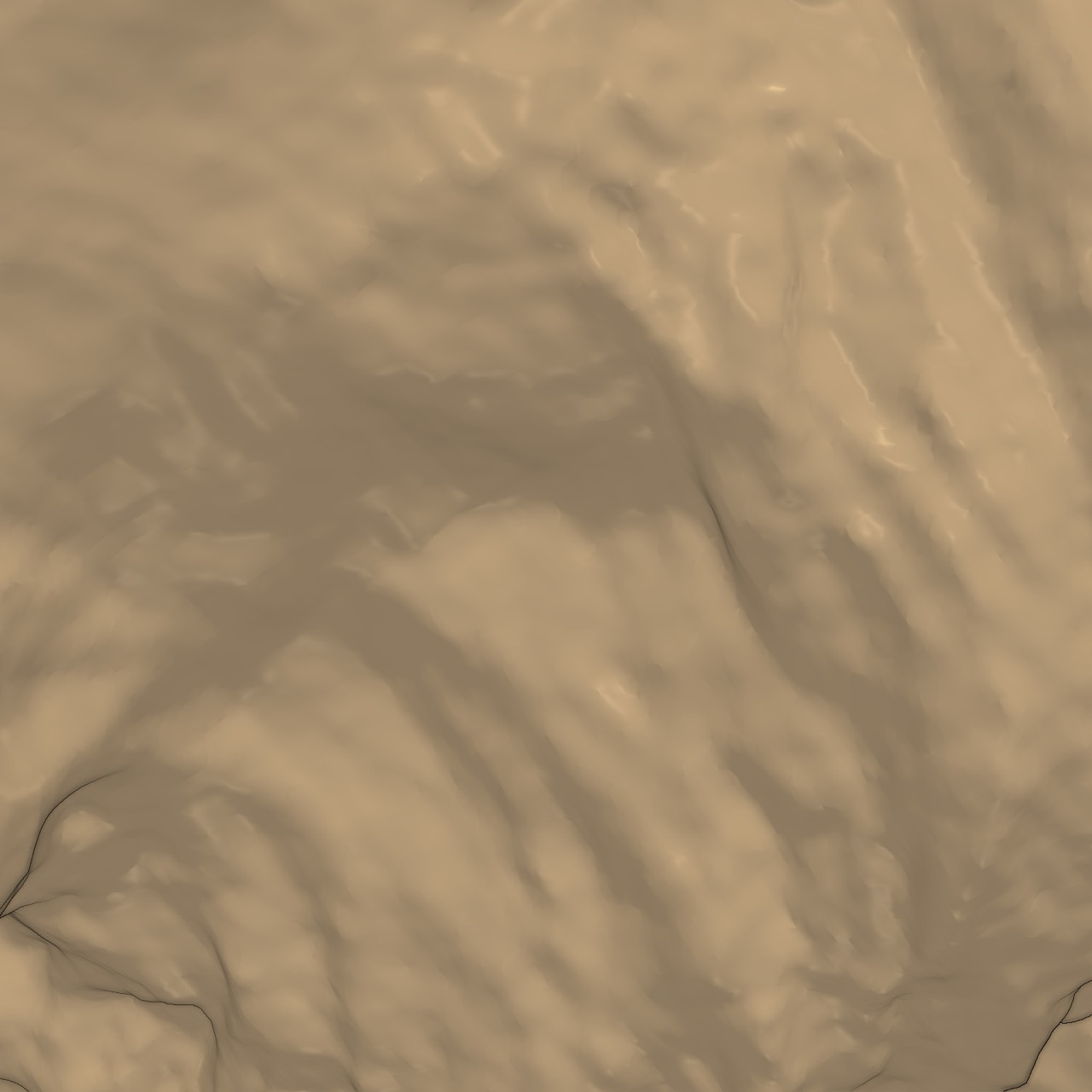}&
    \includegraphics[width=\figBwidth\linewidth]{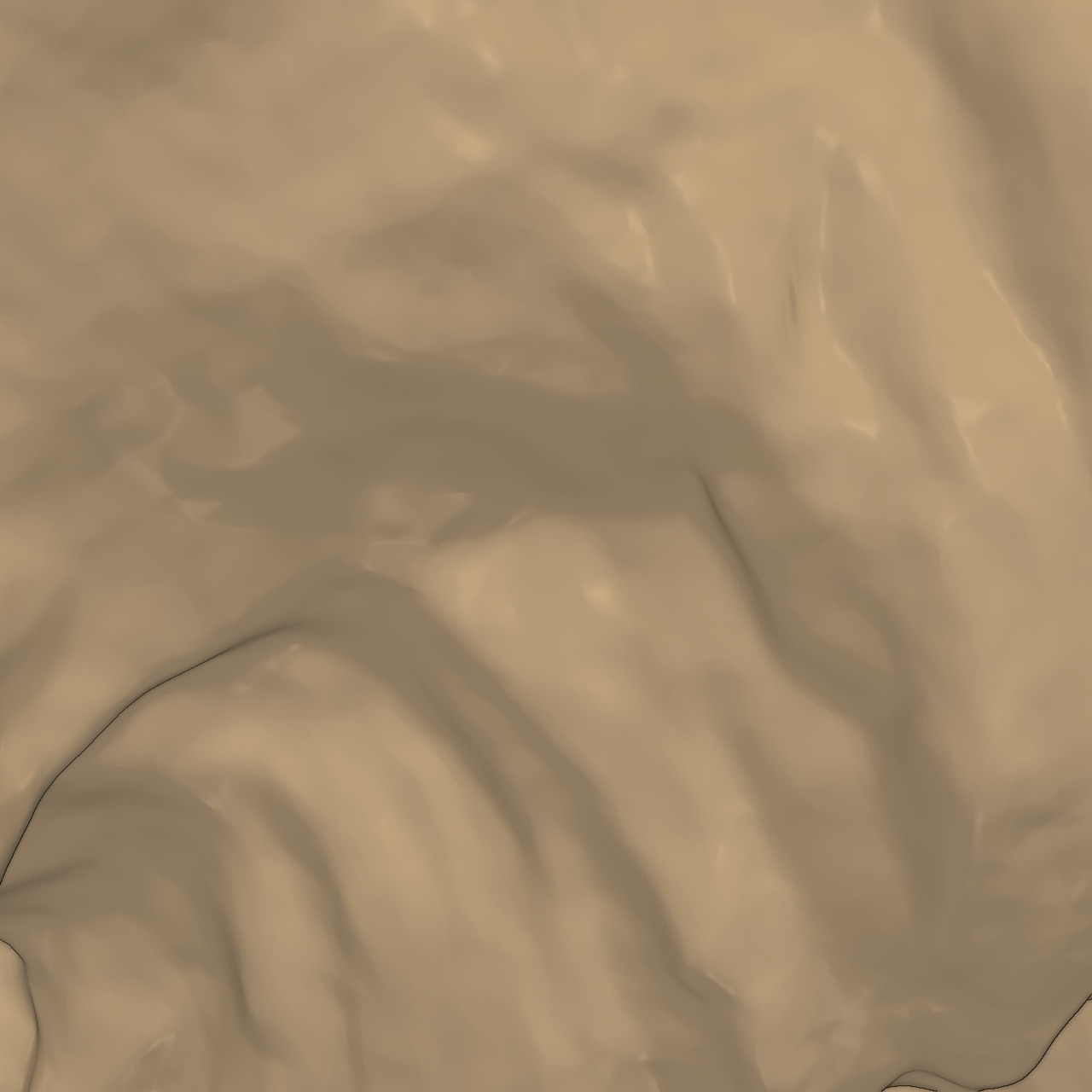}
    \end{tabular}
    \caption{Qualitative comparison using our objects. Our approach recovers better surface detail than multi-view stereo approaches and achieves competitive results with commercial 3D scanners.}
    \label{fig.real_world_comparison}
\end{figure}

\Cref{fig.real_world_comparison,fig.teaser1} display the outcomes from \neustwo, \ourmethod, and a structured-light based commercial 3D scanner EinScan SE.
Although \neustwo and our method use the same multi-resolution hash encoding, the surface details recovered from multi-view color images are smoothed out.
Our method, benefitting from the normal map inputs, can recover more fine-grained surface details that are comparable to the scanner.

We find that our scanner had difficulty capturing accurate surface detail in highly concave areas (\cref{fig.real_world_comparison} above). 
This problem probably arises because the scanner uses a wide baseline stereo camera (about \SI{15}{cm} apart) and struggles with objects as small as \SI{6}{cm} high. 
Small, concave surfaces cause occlusions for the camera, reducing the quality of the reconstruction. 
However, normal estimation does not suffer from occlusion because it is based on single-view observations. 
Therefore, fusing multi-view normal maps produces better results in concave areas.

\section{Discussion}
We have demonstrated \ourmethod, a fast, high-fidelity 3D reconstruction approach using multi-view normal maps.
\ourmethod outperforms other neural MVPS methods in speed and reconstruction quality on benchmark datasets.
Compared to MVS-based neural reconstruction methods, using normal maps can exploit the expressive power of multi-resolution hash encoding, and produce scanner-level surface detail.
Further, we have proven the effectiveness of directional finite difference in expediting training without sacrificing quality.

\ourmethod provides a promising direction to make high-quality 3D scanning more accessible. 
Unlike costly commercial 3D scanners, our approach is viable with everyday devices like smartphones and basic equipment such as a tripod, video light, and a turntable.

\begin{figure}
    \centering
    \includegraphics[width=\linewidth]{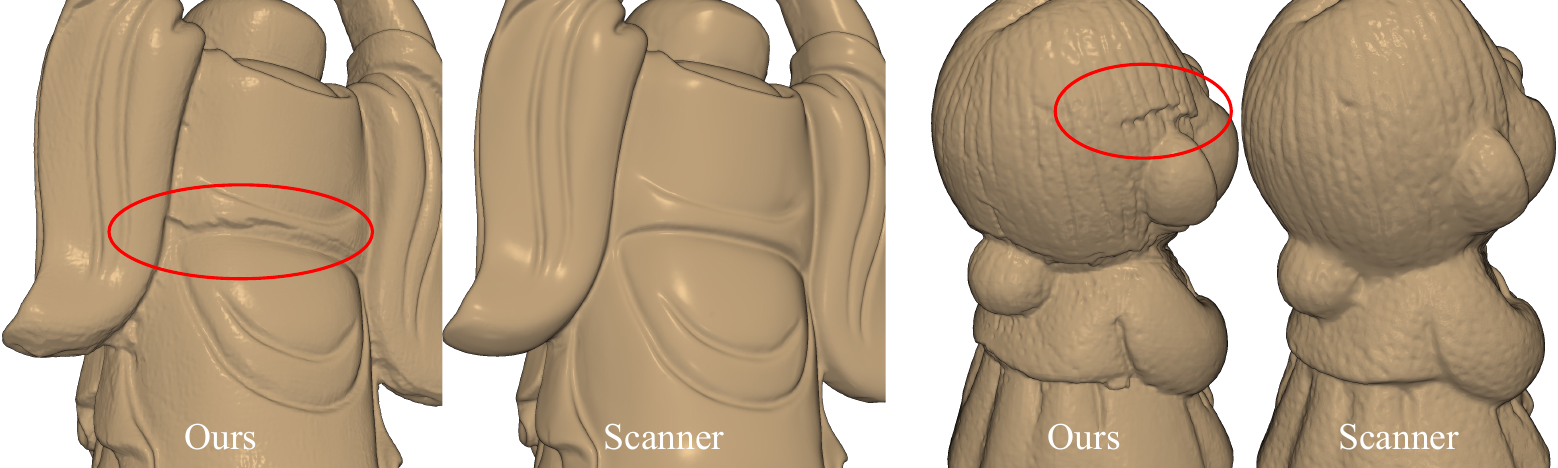}
    \caption{\textbf{Limitation.} Fault artifacts in red circles.}
    \label{fig.limitation}
\end{figure}

\vspace{-1em}
\paragraph{Limitation}
Our method may produce fault-like artifacts, as shown in \cref{fig.limitation}.
It exists no matter our SDF gradient computation approach (\ie, AD, FD, or DFD) and requires further investigation.

\paragraph{Future work}
Our work only uses multi-view normal maps, but incorporating additional clues like sparse point clouds estimated via SfM  might further improve reconstruction quality.
The directional finite difference can also be incorporated into the neural surface reconstruction methods like Neuralangelo~\cite{neuralangelo2023CVPR} to improve training efficiency.

We believe that advancements in MVPS require a more effective geometry evaluation approach and an expanded MVPS benchmark dataset. 
The ``ground truth'' meshes created by a 3D scanner might become insufficient to evaluate MVPS outcomes as normal estimation and integration techniques advance. 
Moreover, \diligentmv is becoming insufficient to fully demonstrate the potential advantages of MVPS approaches since its captured objects only have relatively simple geometry.

%% file: sections/X_suppl.tex

\begin{appendices}
\label{appendices}

This supplementary material includes a benchmark evaluation on normal maps in \cref{sec.benchmark_evaluation}, our implementation details in \cref{sec.implementation_details}, and more visualization of our captured objects in \cref{sec.exp_real_world_more}.

\section{Benchmark evaluation}
\label{sec.benchmark_evaluation}

In addition to mesh quality, \Cref{sec.normal_evaluation} compares the rendered normal map quality of different methods.
\Cref{sec.normal_render} compares different strategies for rendering the normal maps of a trained neural SDF.

\subsection{Normal evaluation}
\label{sec.normal_evaluation}
Normal maps are critical to the quality of physics-based rendering. 
Therefore, the quality of the normal maps rendered from a trained SDF or a recovered mesh is another indicator of the performance of the 3D reconstruction approach.
To evaluate normal accuracy, we use $15$ views of \diligentmv for training and the remaining $5$ views for testing.
Note that \psnerf and \mvas use the same training strategy, while other methods use all $20$ views for reconstruction.

\Cref{tab.normal_accuracy} reports the mean angular error averaged over all foreground pixels from the $5$ test views.
Since volume or surface rendering does not guarantee unit normal vectors, we normalize the rendered normal maps before evaluation.
On average, our method outperforms all compared methods in terms of normal accuracy.
Compared to neural rendering methods, our method achieves a mean angular error $24\%$ lower than \psnerf and $21\%$ lower than \mvas. 

\begin{figure*}
\small
\newcommand{\figWidthN}{0.16}
    \centering
    \begin{tabular}{@{}c@{}cc@{}c@{}c@{}c@{}}
    \multicolumn{2}{c}{Mesh rendering} & \multicolumn{3}{c}{Neural rendering} & \\
    \cmidrule(lr){1-2} \cmidrule(lr){3-5}
    \rmvps & \bmvps & \psnerf & \mvas & Ours & GT \\
    \includegraphics[width=\figWidthN\linewidth]{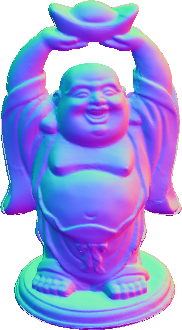}&
    \includegraphics[width=\figWidthN\linewidth]{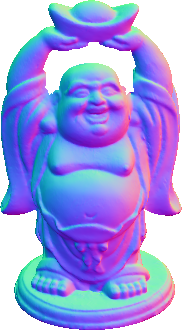}&
    \includegraphics[width=\figWidthN\linewidth]{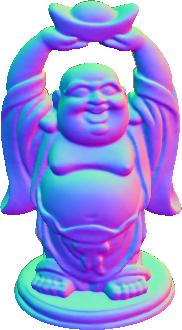}&
    \includegraphics[width=\figWidthN\linewidth]{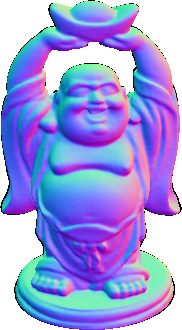}&
    \includegraphics[width=\figWidthN\linewidth]{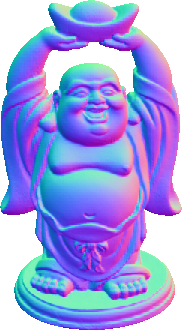}&
    \includegraphics[width=\figWidthN\linewidth]{images/normal_vis/normal_rendering/buddha/gt/normal_map_gt_epoch_49_view_01}
    \\
    \includegraphics[width=\figWidthN\linewidth]{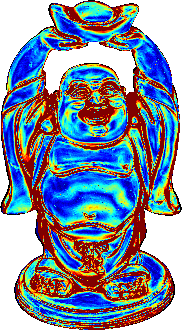}
    &
    \includegraphics[width=\figWidthN\linewidth]{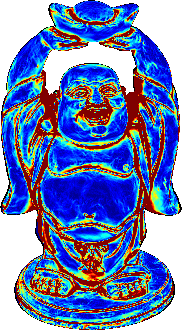}&
    \includegraphics[width=\figWidthN\linewidth]{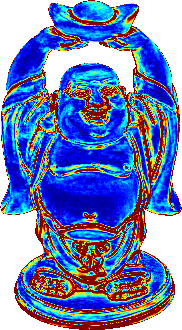}&
    \includegraphics[width=\figWidthN\linewidth]{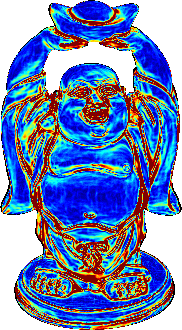}&
    \includegraphics[width=\figWidthN\linewidth]{images/normal_vis/normal_rendering/buddha/normals_dfd/00005000_0_0_ae_up_20}&
    \colorbar{0.14}{$\geq 20^\circ$}
    \\
    \includegraphics[width=\figWidthN\linewidth]{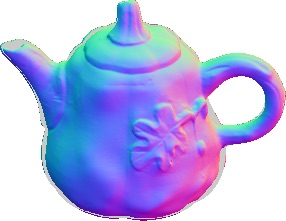}&
    \includegraphics[width=\figWidthN\linewidth]{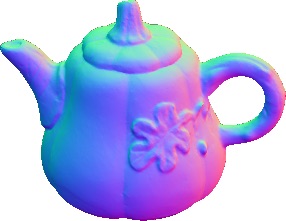}&
    \includegraphics[width=\figWidthN\linewidth]{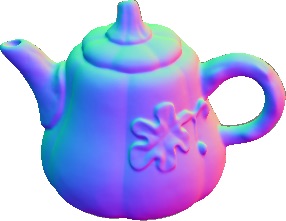}&
    \includegraphics[width=\figWidthN\linewidth]{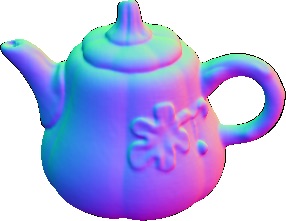}&
    \includegraphics[width=\figWidthN\linewidth]{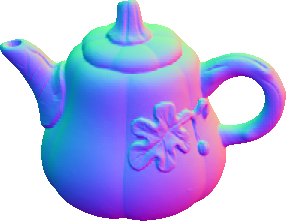}&
    \includegraphics[width=\figWidthN\linewidth]{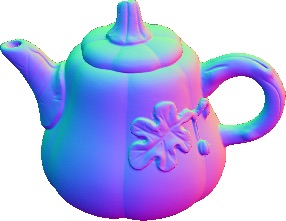}
    \\
    \includegraphics[width=\figWidthN\linewidth]{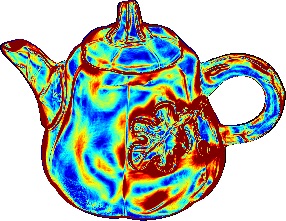}
    &
    \includegraphics[width=\figWidthN\linewidth]{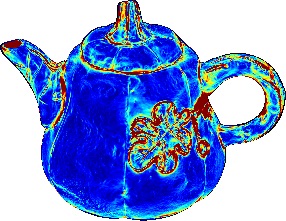}&
    \includegraphics[width=\figWidthN\linewidth]{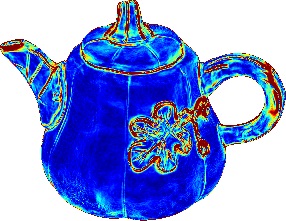}&
    \includegraphics[width=\figWidthN\linewidth]{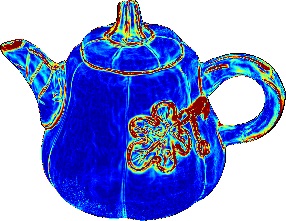}&
    \includegraphics[width=\figWidthN\linewidth]{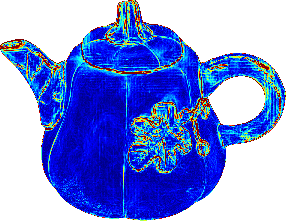}&
    \colorbar{0.05}{$\geq 20^\circ$}
    \\
    \includegraphics[width=\figWidthN\linewidth]{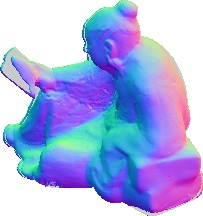}&
    \includegraphics[width=\figWidthN\linewidth]{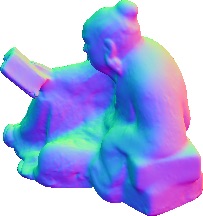}&
    \includegraphics[width=\figWidthN\linewidth]{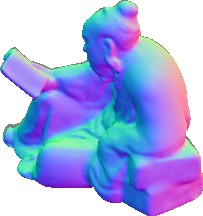}&
    \includegraphics[width=\figWidthN\linewidth]{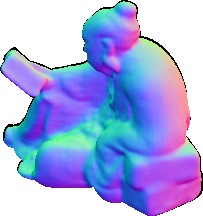}&
    \includegraphics[width=\figWidthN\linewidth]{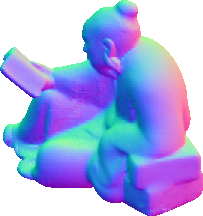}&
    \includegraphics[width=\figWidthN\linewidth]{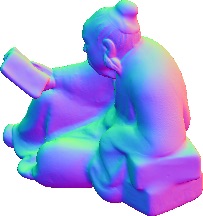}
    \\
    \includegraphics[width=\figWidthN\linewidth]{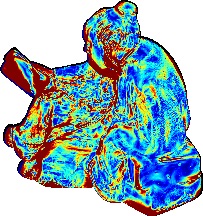}
    &
    \includegraphics[width=\figWidthN\linewidth]{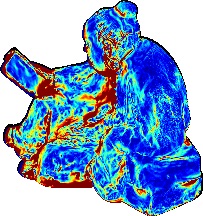}&
    \includegraphics[width=\figWidthN\linewidth]{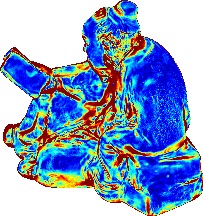}&
    \includegraphics[width=\figWidthN\linewidth]{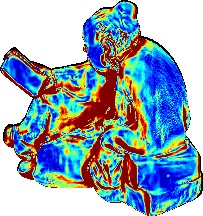}&
    \includegraphics[width=\figWidthN\linewidth]{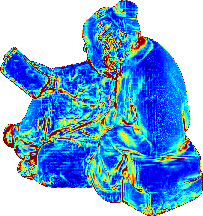}&
    \colorbar{0.05}{$\geq 20^\circ$}
    \end{tabular}
    \caption{Qualitative comparison of normal maps rendered from reconstructed meshes or neural representations and angular error maps.}
    \label{fig.normal_compare}
\end{figure*}

\Cref{fig.normal_compare} shows the rendered normal maps and the corresponding angular error maps.
Our method produces normal maps with the best high frequency detail, especially the eyes of \emph{Buddha} and the flower pattern on \emph{Pot2}.

\subsection{Normal map rendering at inference time}
\label{sec.normal_render}

At training time, we use DFD to compute the SDF gradients and accumulate the SDF gradients on the rays to obtain the normal vectors for the pixels.
At inference time, however, we are not restricted to DFD;
both FD and AD can be used.
Instead of volume rendering, we can also use surface rendering, \ie, we only compute the SDF gradient at the zero level set points.

\Cref{tab.normal_different_strategy} compares normal accuracy in terms of mean angular error and rendering time in terms of frames per second (FPS) using different strategies for rendering the normal map.
From \cref{tab.normal_different_strategy}, all volume rendering based methods produce close results.
Finite difference performs slightly better, but at the cost of rendering time.
It takes $64\%$ more time to produce $1.1\%$ better results.
Surface rendering, on the other hand, consistently performs worse than volume rendering at inference time, even though it is almost three times faster than volume rendering. 
There are two possible reasons for this: 1) the high frequency noise introduced by multi-resolution hash coding, and 2) the loss function is defined based on volume rendering, which only encourages volume rendering results to match the input normal maps. 

\begin{figure}
\small
\newcommand{\figWidthS}{0.2}
    \centering
    \begin{tabular}{c@{}c@{}c@{}c@{}c}
    VR-DFD & VR-AD & VR-FD & SR-AD & GT \\
    \includegraphics[width=\figWidthS\linewidth]{images/normal_vis/normal_rendering/buddha/normals_dfd/00005000_0_0_normalized}&
        \includegraphics[width=\figWidthS\linewidth]{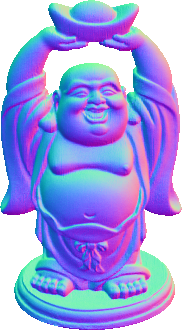}&
            \includegraphics[width=\figWidthS\linewidth]{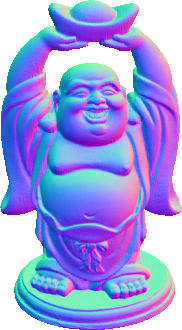}&
                \includegraphics[width=\figWidthS\linewidth]{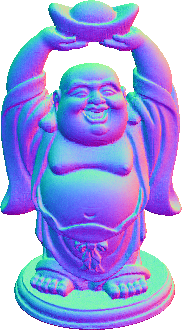}& 
                \includegraphics[width=\figWidthS\linewidth]{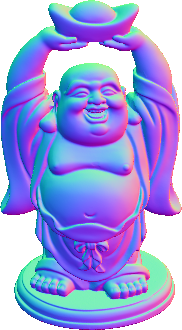}
                \\
    \includegraphics[width=\figWidthS\linewidth]{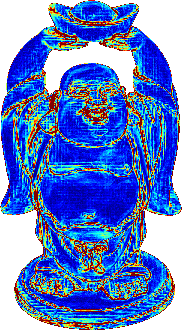}&
        \includegraphics[width=\figWidthS\linewidth]{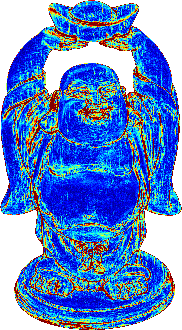}&
            \includegraphics[width=\figWidthS\linewidth]{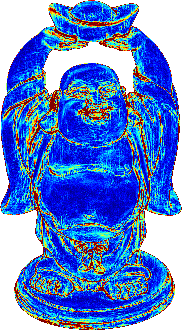}&
                \includegraphics[width=\figWidthS\linewidth]{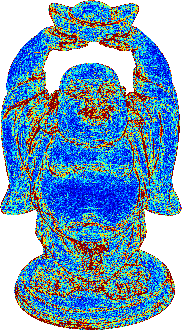}&  
    \colorbar{0.14}{$\geq 20^\circ$}
    \end{tabular}
    \caption{\textbf{(First row)} Normal maps rendered from the trained neural SDF using different rendering strategies. \textbf{(Second row)} Angular error maps. }
    \label{fig.normal_render_diff}
\end{figure}

\Cref{fig.normal_render_diff} shows the normal maps rendered from the same neural SDFs using different rendering strategies and corresponding angular error maps.
All volume rendering strategies produce visually similar normal maps.
This also validates the approximation accuracy of the DFD.
In contrast, surface rendering introduces noise into the rendered normal map.

\begin{table}[]
\scriptsize
    \centering
    \caption{Quantitative evaluation of rerendered normal maps. Mean angular error [deg.] averaged on $5$ test views are reported. Darker colors indicate better results.}
    \begin{tabular}{@{}lrrrrr||r@{}}
    \toprule
    Methods & Bear & Buddha & Cow & Pot2 & Reading & Average \\
    \midrule
    \rmvps & \cellcolor{orange!25!white}12.80 & \cellcolor{orange!25!white}13.67 & \cellcolor{orange!25!white}10.81 & \cellcolor{orange!25!white}14.99 & \cellcolor{orange!25!white}11.71 & \cellcolor{orange!25!white}12.80 \\
    \bmvps & \cellcolor{orange!40!white}3.80 & \cellcolor{orange!40!white}10.57 & \cellcolor{orange!100!white}2.83 & \cellcolor{orange!75!white}5.76 & \cellcolor{orange!100!white}6.90 & \cellcolor{orange!85!white}5.97 \\
    \psnerf & \cellcolor{orange!55!white}3.45 & \cellcolor{orange!75!white}10.25 & \cellcolor{orange!75!white}4.35 & \cellcolor{orange!40!white}5.94 & \cellcolor{orange!40!white}9.36 & \cellcolor{orange!40!white}6.67 \\
    \mvas & \cellcolor{orange!85!white}3.08 & \cellcolor{orange!85!white}9.90 & \cellcolor{orange!70!white}3.72 & \cellcolor{orange!85!white}5.07 & \cellcolor{orange!40!white}10.02 & \cellcolor{orange!70!white}6.36 \\
    Ours & \cellcolor{orange!100!white}2.56 & \cellcolor{orange!100!white}7.64 & \cellcolor{orange!85!white}3.10 & \cellcolor{orange!100!white}4.49 & \cellcolor{orange!85!white}7.40 & \cellcolor{orange!100!white}5.04 \\
    \bottomrule
    \end{tabular}
    \label{tab.normal_accuracy}
\end{table}

\begin{table}
\small
    \centering
    \caption{Normal map rendering accuracy and time using different strategies to render the normal maps. \textbf{VR}: Volume rendering. \textbf{SR}: Surface rendering.}
    \resizebox{\linewidth}{!}{
    \begin{tabular}{@{}llrrrrrr|r@{}}
    \toprule
    && \multicolumn{6}{c}{Mean angular error [deg.] ($\downarrow$)} & FPS
    \\
     & & Bear & Buddha & Cow & Pot2 & Reading & Average  &\\
     \midrule
    \multirow{3}{*}{VR} &DFD    & 2.56 & 7.64 &3.10&4.49&7.40& 5.04 & 1.8\\
        &AD     &2.63 & 7.50 &3.16&4.61&7.48& 5.08 & 1.5\\
        &FD     &2.57 & 7.29 &3.10&4.49&7.43& 4.98 & 1.1\\
    \midrule
    SR  &AD     &3.45 & 9.75 &4.08&5.89&8.80& 6.39 &3.3\\
    \bottomrule
    \end{tabular}
    }
    \label{tab.normal_different_strategy}
\end{table}

\section{Implementation Details}
\label{sec.implementation_details}

\subsection{Network architecture}
\label{sec.network}

\begin{figure}
    \centering
    \includegraphics[width=\linewidth]{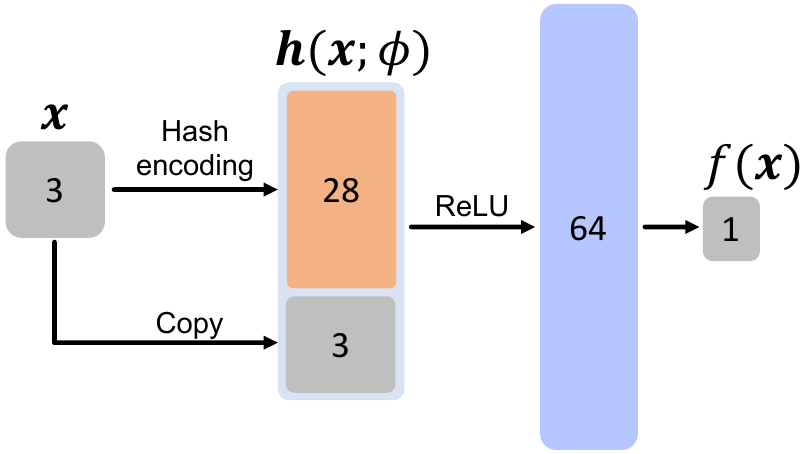}
    \caption{Neural SDF architecture.}
    \label{fig.network_architecture}
\end{figure}
\Cref{fig.network_architecture} shows our neural SDF architecture.
The input 3D coordinate \point is transformed by multi-resolution hash encoding into a $28$-dim feature vector $h(\point;\phi)$.
The feature vector is concatenated with the 3D vector \point and input to a one-layer MLP.
We use $64$ units and ReLU activation for the only hidden layer and no activation for the output layer.
Overall, the neural SDF can be written as 
\begin{equation}
    f(\point) = \V{w}_2 \max\left(\V{W}_1\begin{bmatrix}
        h(\point;\phi)\\ \point
    \end{bmatrix}+\V{b}_1, \V{0}\right) +b_2,
\end{equation}
where $\max(\cdot, \V{0})$ is the element-wise ReLU function.
We apply a geometric initialization~\cite{igr2020icml} to $\V{W}_1$ and $\V{w}_2$ so that the initial zero level set of the neural SDF approximates a sphere with radius $b_2$.
We set $b_2=0.7$ as the initial value.
Despite its simplicity, the neural SDF can represent highly complex geometry, as shown in our reconstruction results.

\subsection{Patch-based ray marching details}
\label{sec.ray_marching}
Given a patch of pixels, we perform ray marching for the center ray using \nerfacc package and compute the points on the remaining rays by finding the ray-plane intersections.
We maintain an occupancy grid so that the empty regions on the rays are skipped for efficient training.

Denote \cameraCenter the ray origin of a patch of rays, $\viewDirection_i$ the center ray's unit direction, and $\viewDirection_j$ the unit direction of any remaining rays in the same patch.
Suppose we have sampled a point from the center ray at a distance $t_i$ from the ray origin.
According to the geometry, the corresponding point to be sampled on ray $\viewDirection_j$ should satisfy
\begin{equation}
    (\cameraCenter + t_i \viewDirection_i)^\top \V{m} = (\cameraCenter + t_j \viewDirection_j)^\top \V{m}.
    \label{eq.plane_marching}
\end{equation}
where $\V{m}$ is the marching plane normal.
In our case, $\V{m}$ is perpendicular to the image plane where the patch is located.
Rearrange \cref{eq.plane_marching} yields
\begin{equation}
    t_j = \frac{t_i \viewDirection_i^\top \V{m}}{\viewDirection_j^\top \V{m}},
\end{equation}

After patch-based ray marching, we obtain the sampled points for a batch of patches of pixels, stored as a tensor in the shape of \texttt{(num\_samples, patch\_height, patch\_width, 3)}, alone with a 1D tensor in the shape of \texttt{(num\_samples,)} indicating to which \emph{patch} each sample belong to.
\texttt{num\_samples} is the total number of points sampled from the center rays of all the patches.
We then compute the SDF gradients using DFD and obtain a tensor in the shape of \texttt{(num\_samples, patch\_height, patch\_width, 3)}.

However, \nerfacc does not support patch-based volume rendering, \ie, accumulating the SDF gradients of shape \texttt{(num\_samples, patch\_height, patch\_width, 3)} into a tensor of shape \texttt{(num\_patches, patch\_height, patch\_width, 3)}.
Using a for-loop to process each patch significantly slows down the volume rendering procedure.
To address this, we modify \nerfacc's CUDA code to handle patch-based volume rendering in parallel.
As a result, patch-based volume rendering is as fast as
pixel-based volume rendering, assuming the same amount of pixels.

\subsection{Occupancy grid}
\label{sec.occ_grid}
For efficient training, we periodically update a binary occupancy grid.
Specifically, we update the $128^3$ occupancy grid for every $8$ batch.
The value of each grid is determined by the SDF value $f(\point)$ at that grid:
\begin{equation}
    occ(\point) = \begin{cases}
    1, \quad \text{if} \quad \frac{1}{1 + \exp(-kf(\point))} < \thresholdOcc; \\
    0, \quad \text{otherwise.}
    \end{cases}
\end{equation}
We empirically set $k=80$ and use a $\thresholdOcc = 0.1$ threshold.

\subsection{Evaluation metrics}
\label{sec.eval_metrics}
\textbf{L2 Chamfer distance} measures the distance from one set of points to another.
Given two sets of points \pointsetOne and \pointsetTwo, we first define the distance from one point to another set of points as
\begin{equation}
	\begin{aligned}
	&d_{\pointOne \to \pointsetTwo} = \min_{\pointTwo \in \pointsetTwo} \norm{\pointOne - \pointTwo}_2 \quad \text{and} \\ 
	&d_{\pointTwo \to \pointsetOne} = \min_{\pointOne \in \pointsetOne} \norm{\pointOne - \pointTwo}_2.
	\end{aligned}
\end{equation}
The Chamfer distance \chamferDist is then defined as
\begin{align}
	\chamferDist = \frac{1}{2\abs{\pointsetOne}} \sum_{\pointOne \in \pointsetOne} d_{\pointOne \to \pointsetTwo} + \frac{1}{2\abs{\pointsetTwo}} \sum_{\pointTwo \in \pointsetTwo} d_{\pointTwo \rightarrow \pointsetOne}.
\end{align}

\noindent
\textbf{F-score} is the geometric mean of the precision and recall of the recovered surfaces to the GT surfaces.
Precision and recall are defined based on the distances from a point to a set of points as
\begin{equation}
	\begin{aligned}
	\precision &=  \frac{1}{\abs{\pointsetOne}} \sum_{\pointOne \in \pointsetOne}[d_{\pointOne \rightarrow \pointsetTwo} < \fscoreThreshold]  \quad \text{and}\\
	\recall &=  \frac{1}{\abs{\pointsetTwo}} \sum_{\pointTwo \in \pointsetTwo}[d_{\pointTwo \rightarrow \pointsetOne} < \fscoreThreshold].
	\end{aligned}
\end{equation}
Here, $[\cdot]$ is the Iverson bracket, and \fscoreThreshold is the distance threshold for a point to be considered close enough to a point set.
F-score is then
\begin{align}
	\fscore = \frac{2\precision \recall}{\precision + \recall}.
\end{align}
We set $\fscoreThreshold=\SI{0.5}{mm}$ in our evaluations.

To compute Chamfer distance and F-score, we find the points from the resulting mesh that are visible to the input views.
Specifically, we cast rays for pixels within the mask and find their first intersection with the mesh.
This strategy avoids the randomness~\cite{yang2022psnerf} by sampling points randomly from the mesh and eliminates the effect of invisible regions.

\section{More visualization on our objects}
\label{sec.exp_real_world_more}
\begin{figure}
\newcommand{\figwidthS}{0.24}
    \centering
    \begin{tabular}{c@{}c@{}c@{}c}
        Image & \neustwo & Ours & EinScan SE \\
    \includegraphics[width=\figwidthS\linewidth]{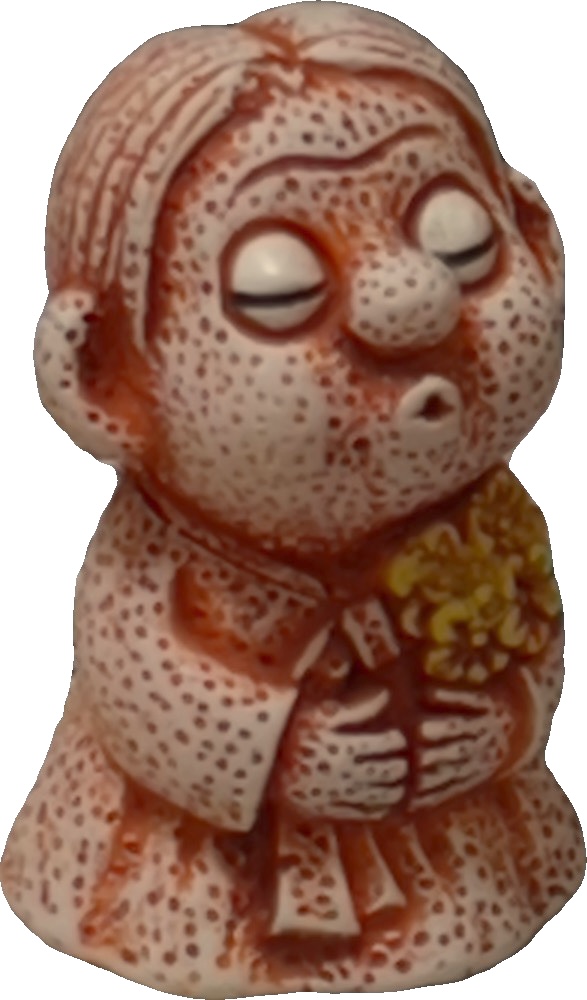}&
        \includegraphics[width=\figwidthS\linewidth]{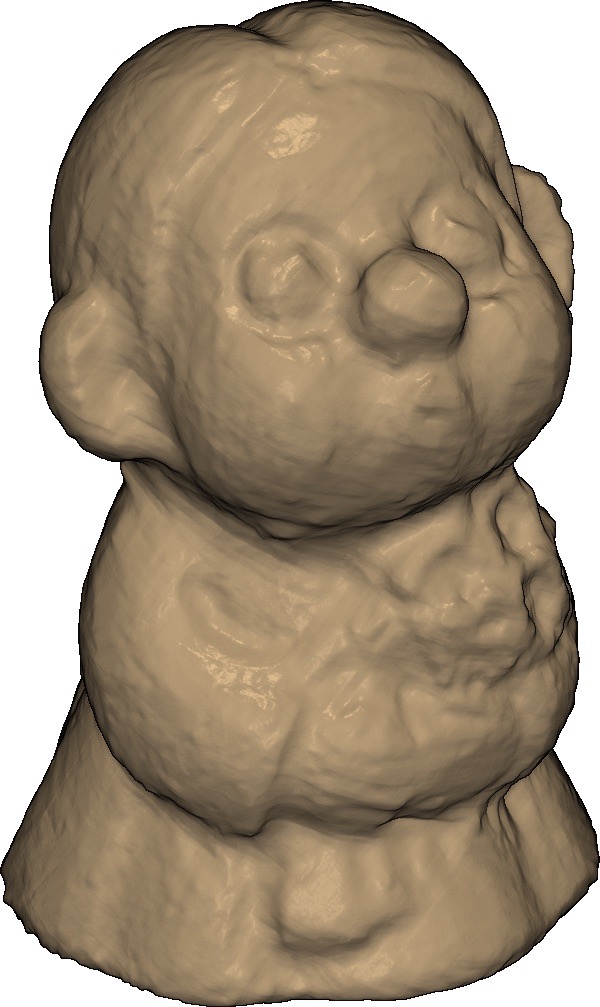}&
                \includegraphics[width=\figwidthS\linewidth]{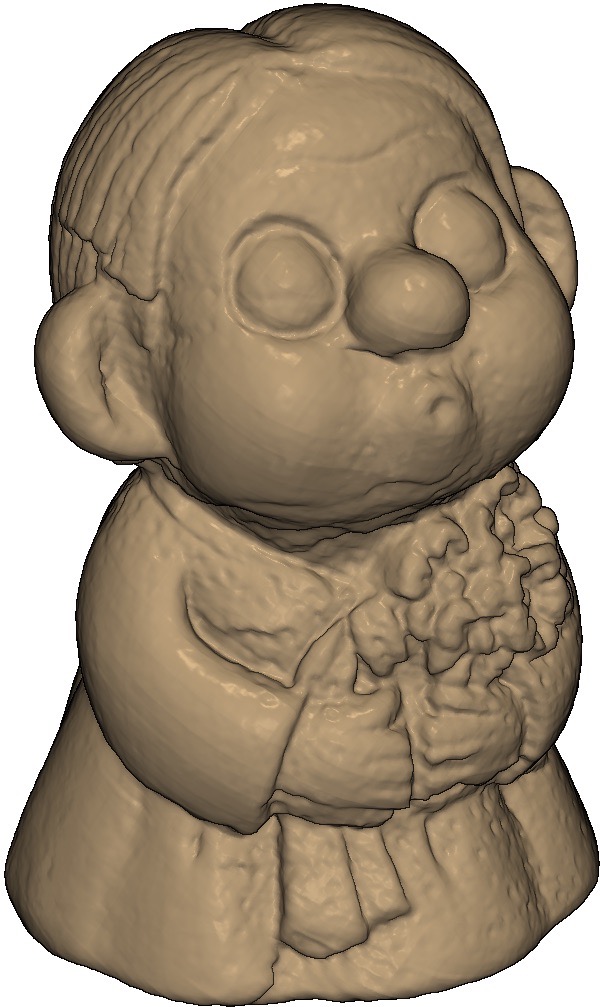}&
                        \includegraphics[width=\figwidthS\linewidth]{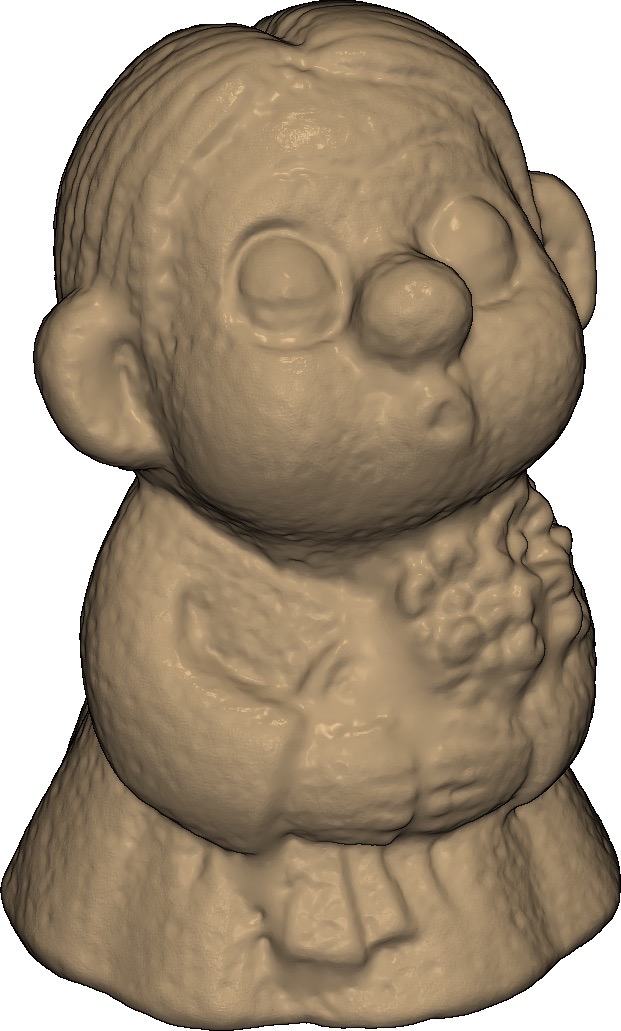}\\
    \includegraphics[width=\figwidthS\linewidth]{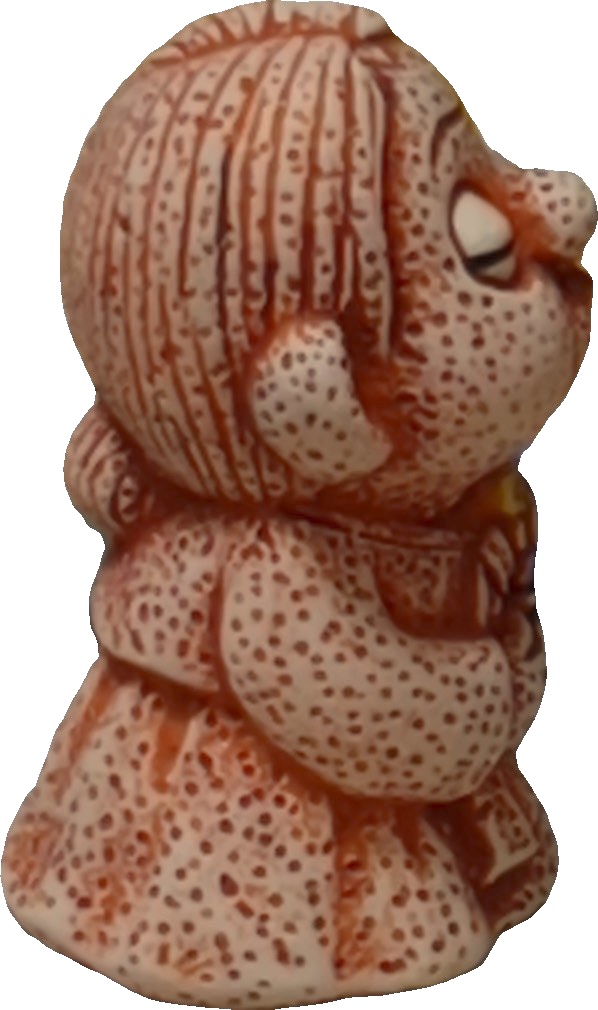}&
        \includegraphics[width=\figwidthS\linewidth]{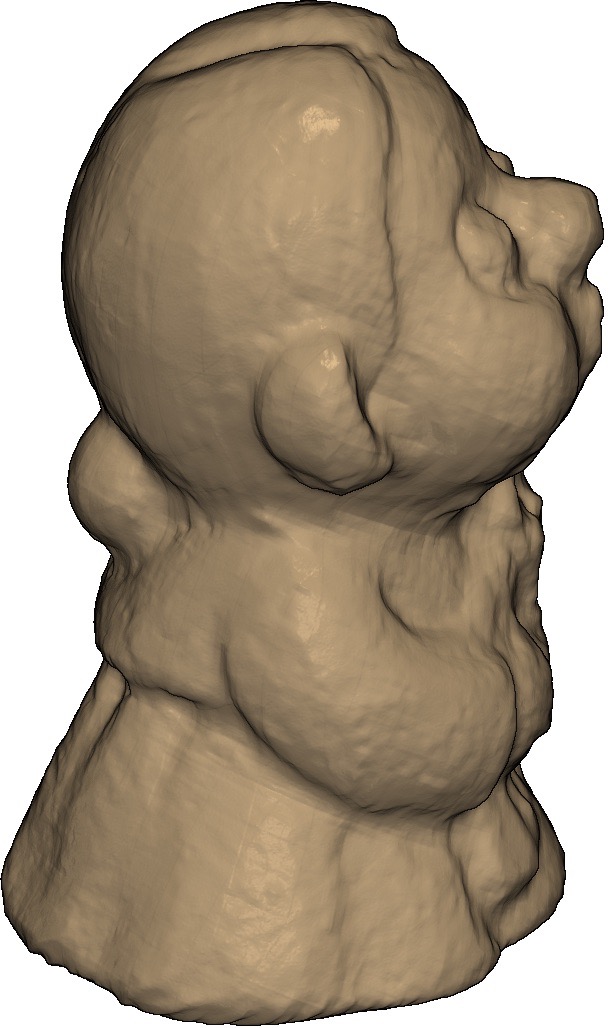}&
                \includegraphics[width=\figwidthS\linewidth]{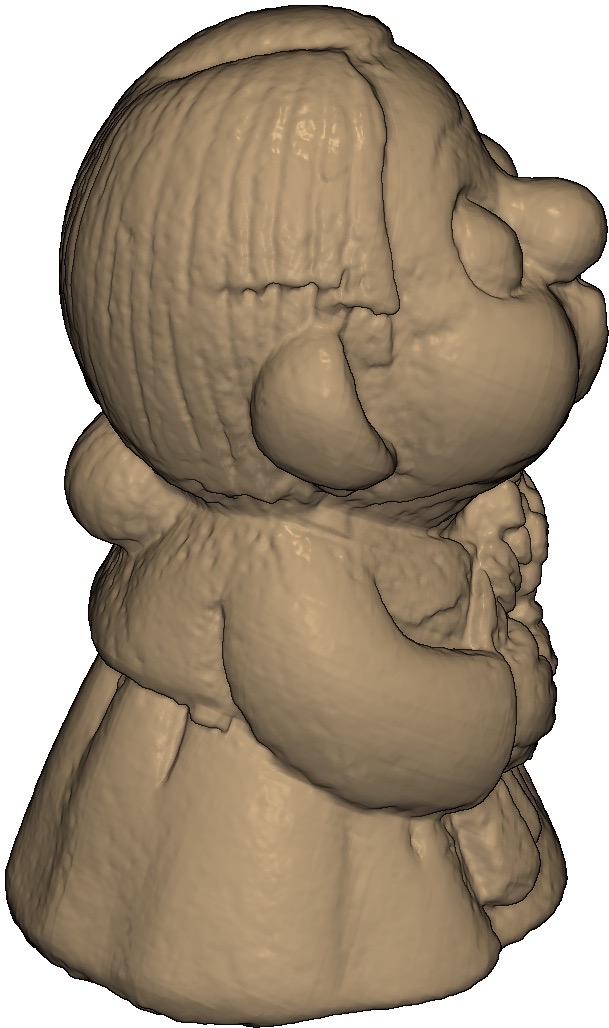}&
                        \includegraphics[width=\figwidthS\linewidth]{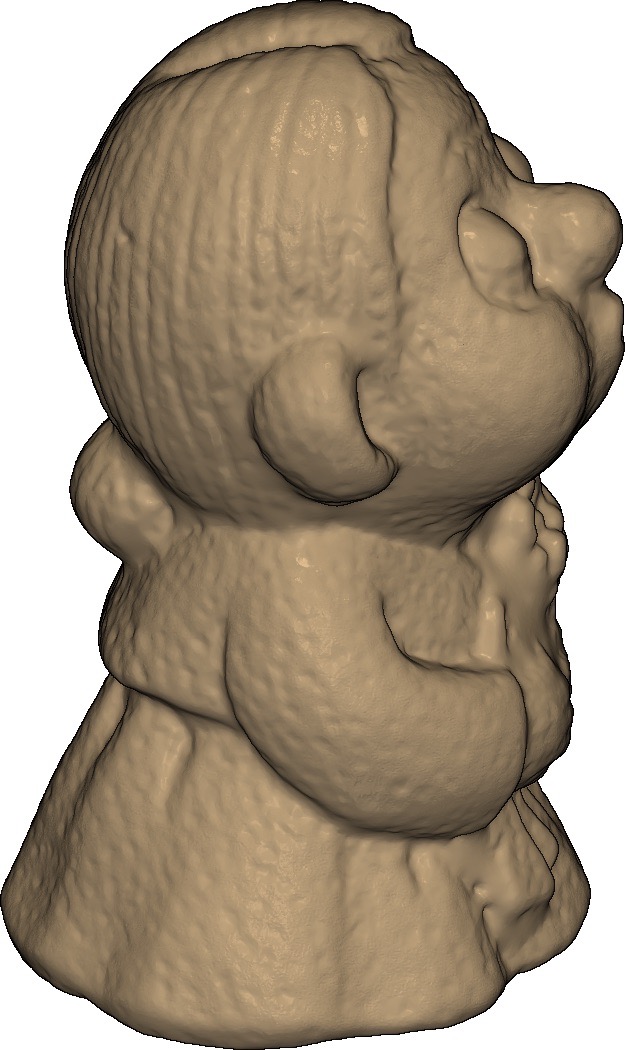}\\
    \includegraphics[width=\figwidthS\linewidth]{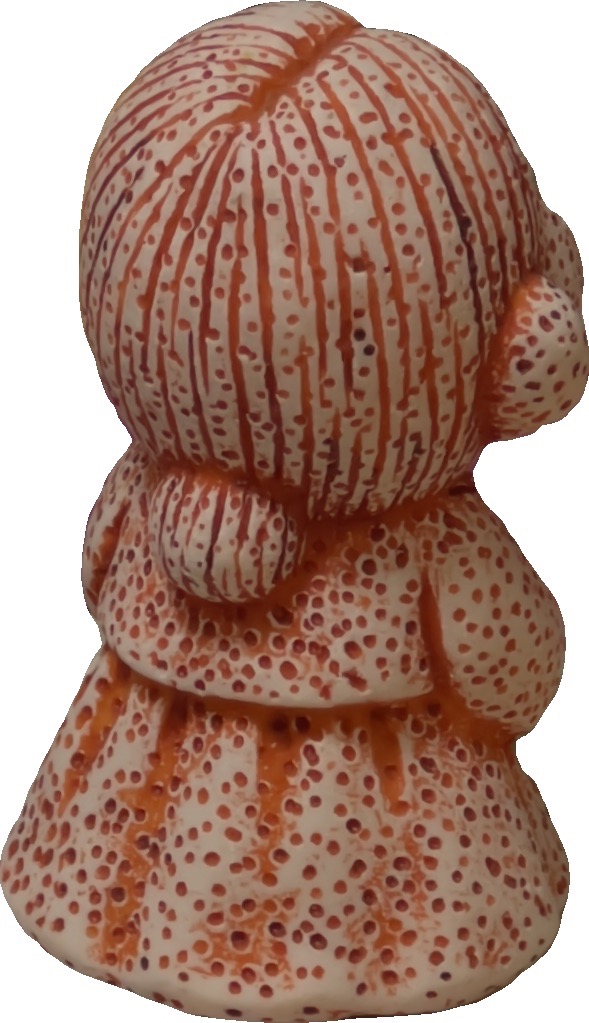}&
        \includegraphics[width=\figwidthS\linewidth]{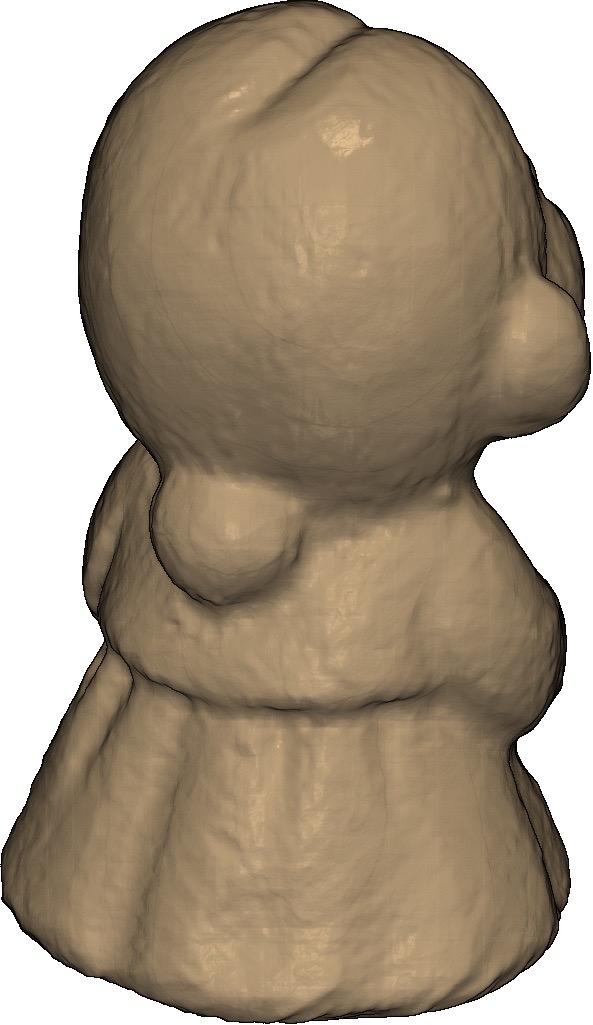}&
                \includegraphics[width=\figwidthS\linewidth]{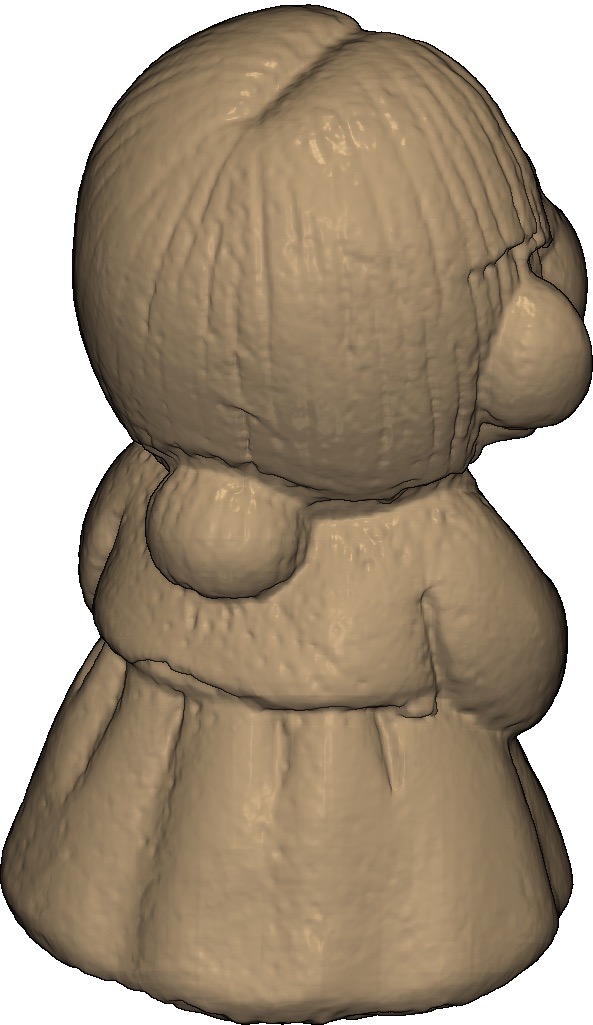}&
                        \includegraphics[width=\figwidthS\linewidth]{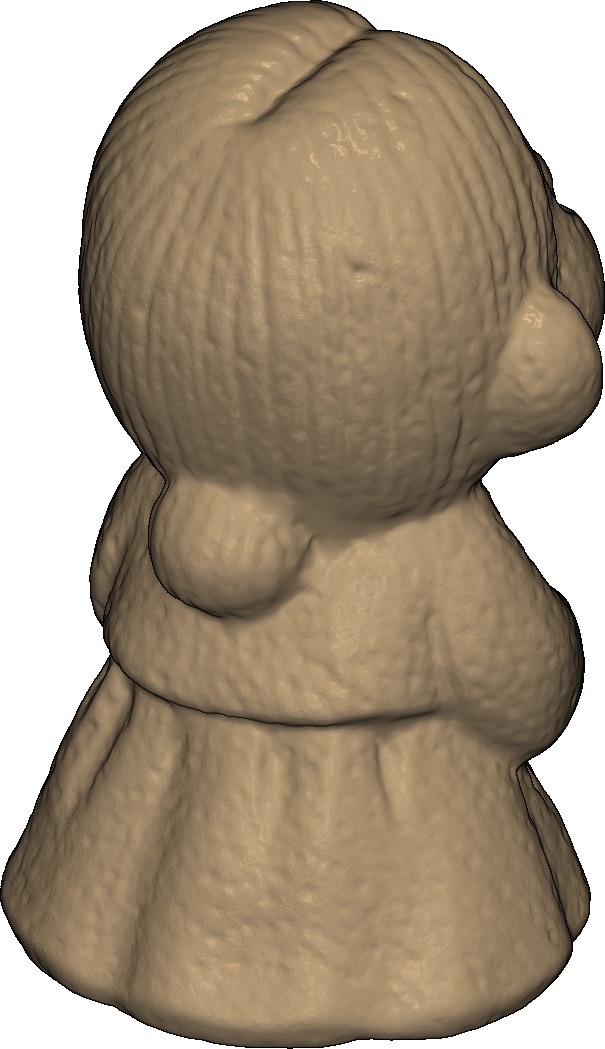}\\
    \includegraphics[width=\figwidthS\linewidth]{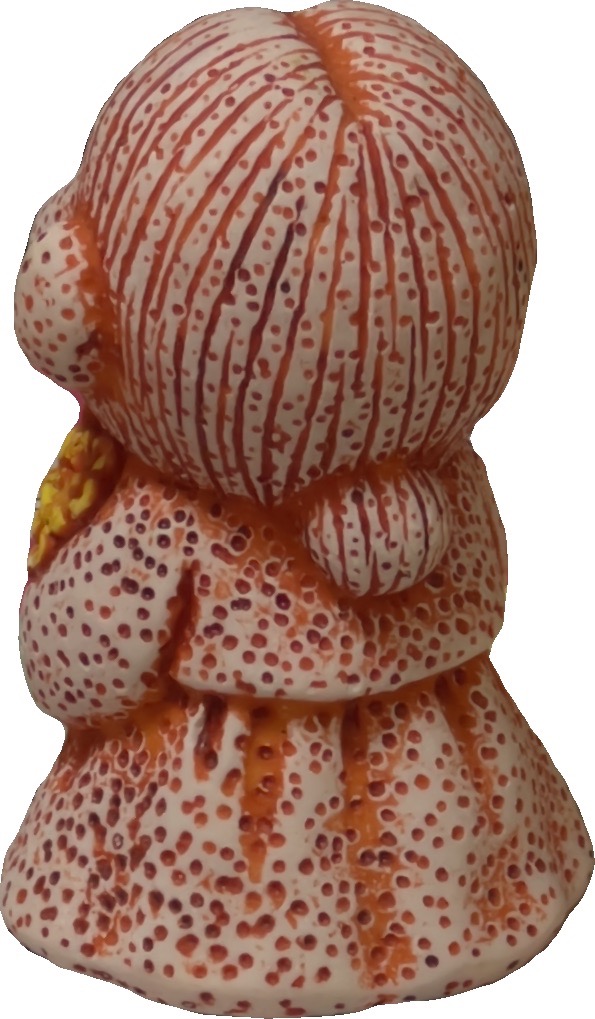}&
        \includegraphics[width=\figwidthS\linewidth]{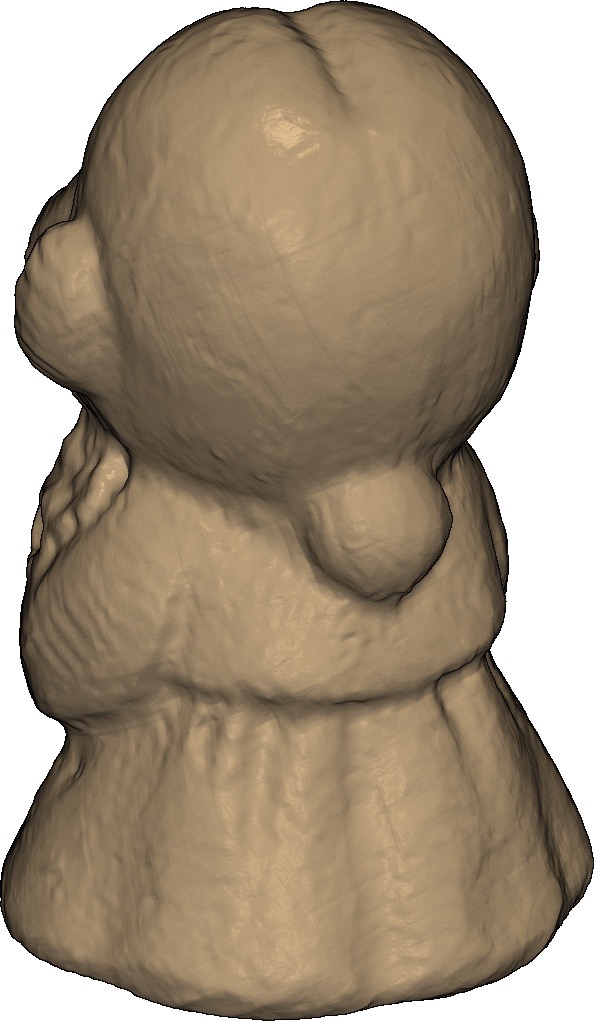}&
            \includegraphics[width=\figwidthS\linewidth]{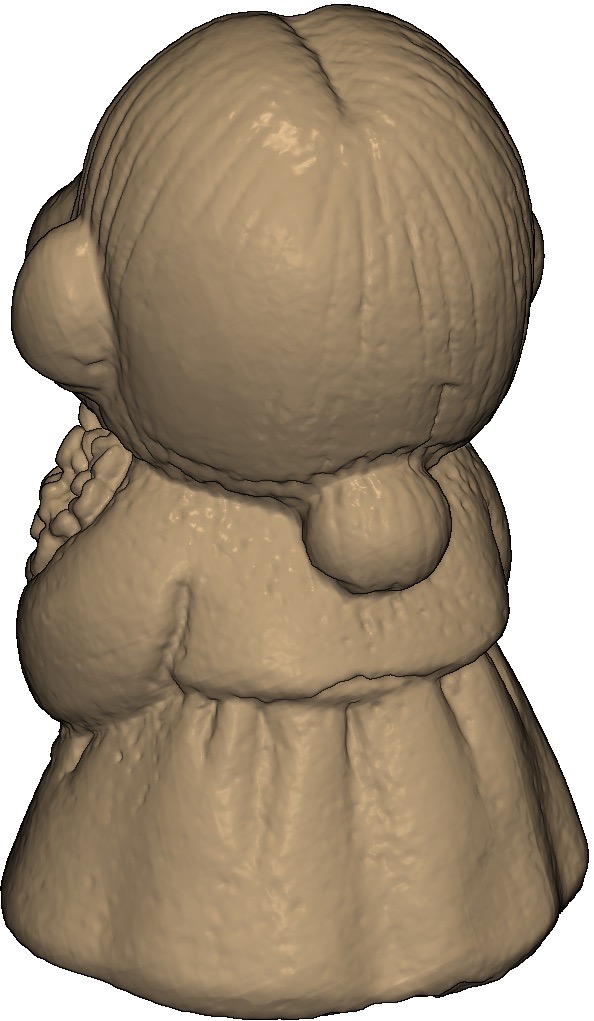}&
                \includegraphics[width=\figwidthS\linewidth]{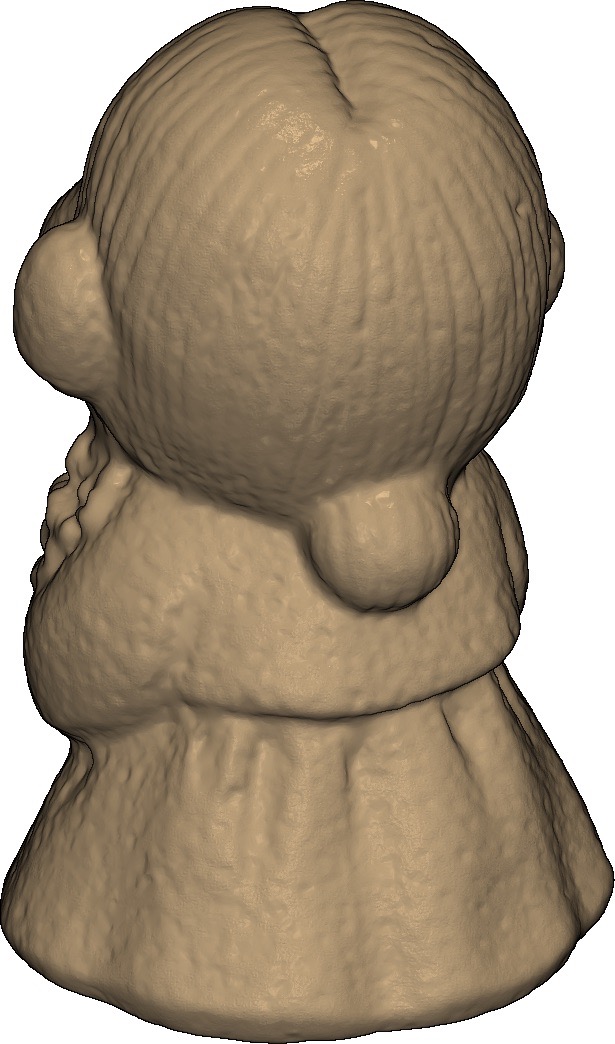}\\
    \end{tabular}
    \caption{Qualitative comparison on our captured object \emph{Girl}.}
    \label{fig.real_world_A}
\end{figure}

\begin{figure}
\newcommand{\figwidthS}{0.24}
    \centering
    \begin{tabular}{c@{}c@{}c@{}c}
        Image & \neustwo & Ours & EinScan SE \\
    \includegraphics[width=\figwidthS\linewidth]{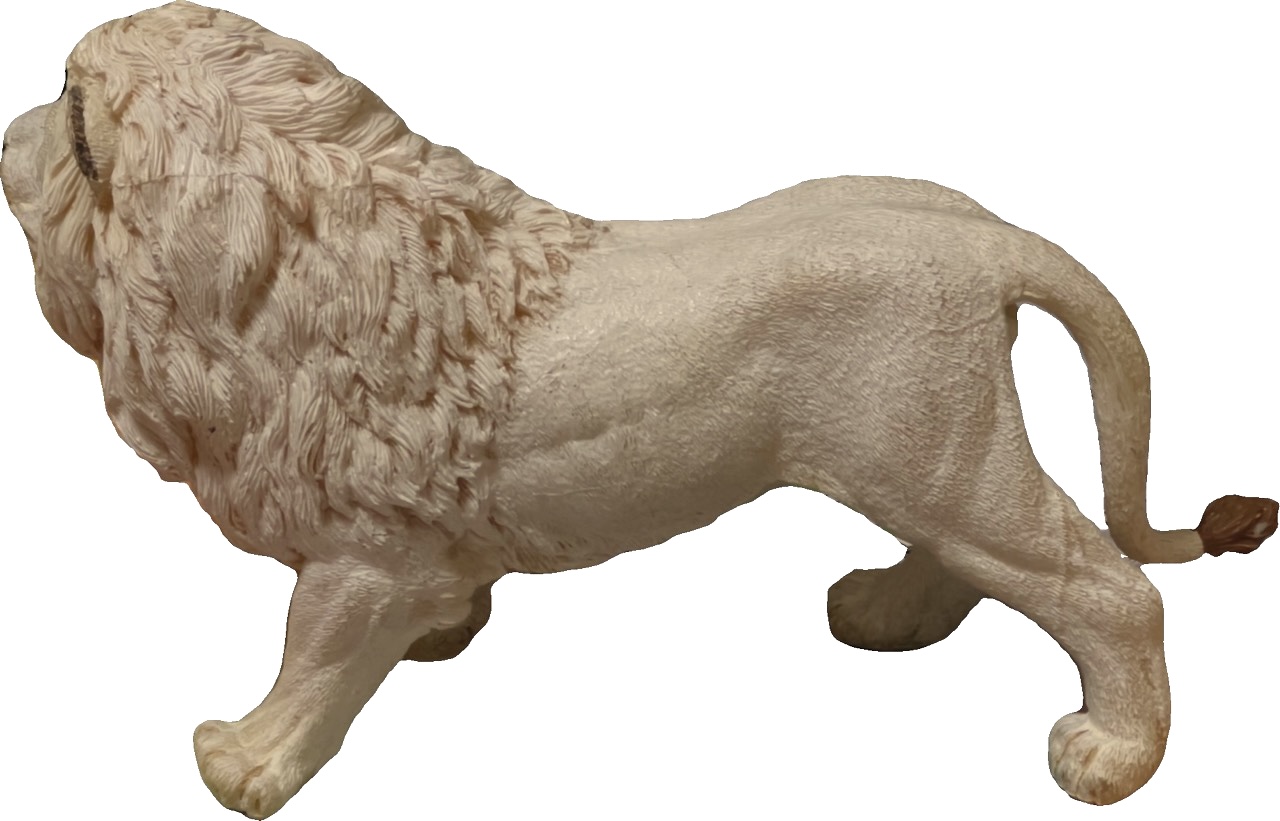}&
        \includegraphics[width=\figwidthS\linewidth]{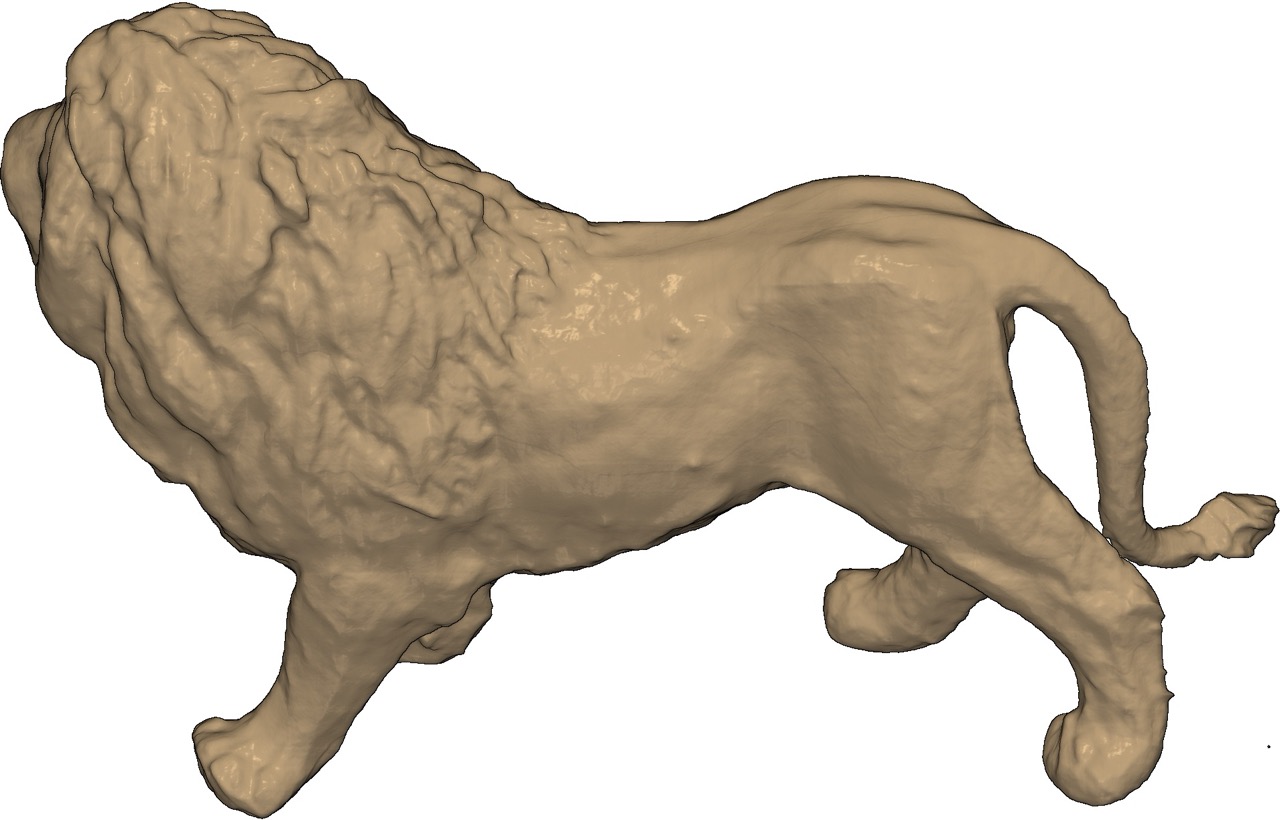}&
                \includegraphics[width=\figwidthS\linewidth]{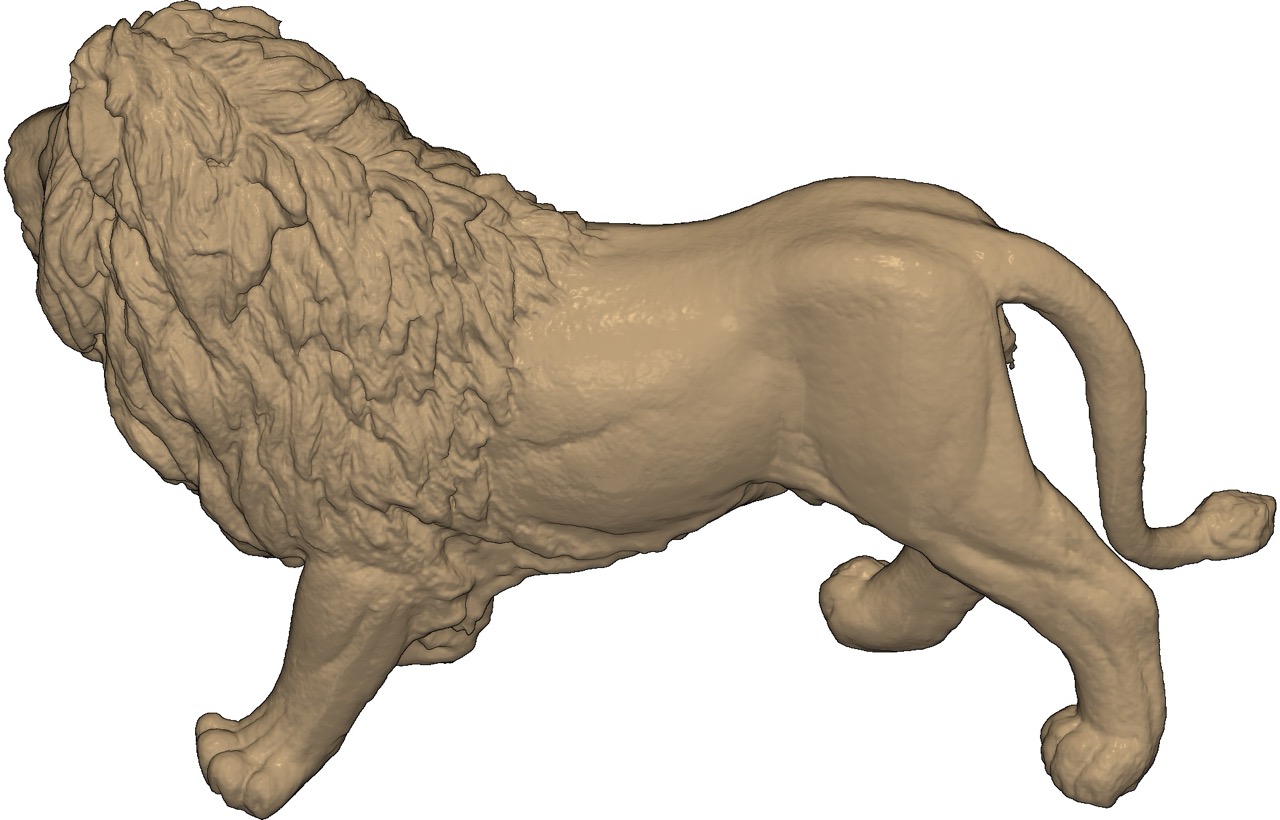}&
                        \includegraphics[width=\figwidthS\linewidth]{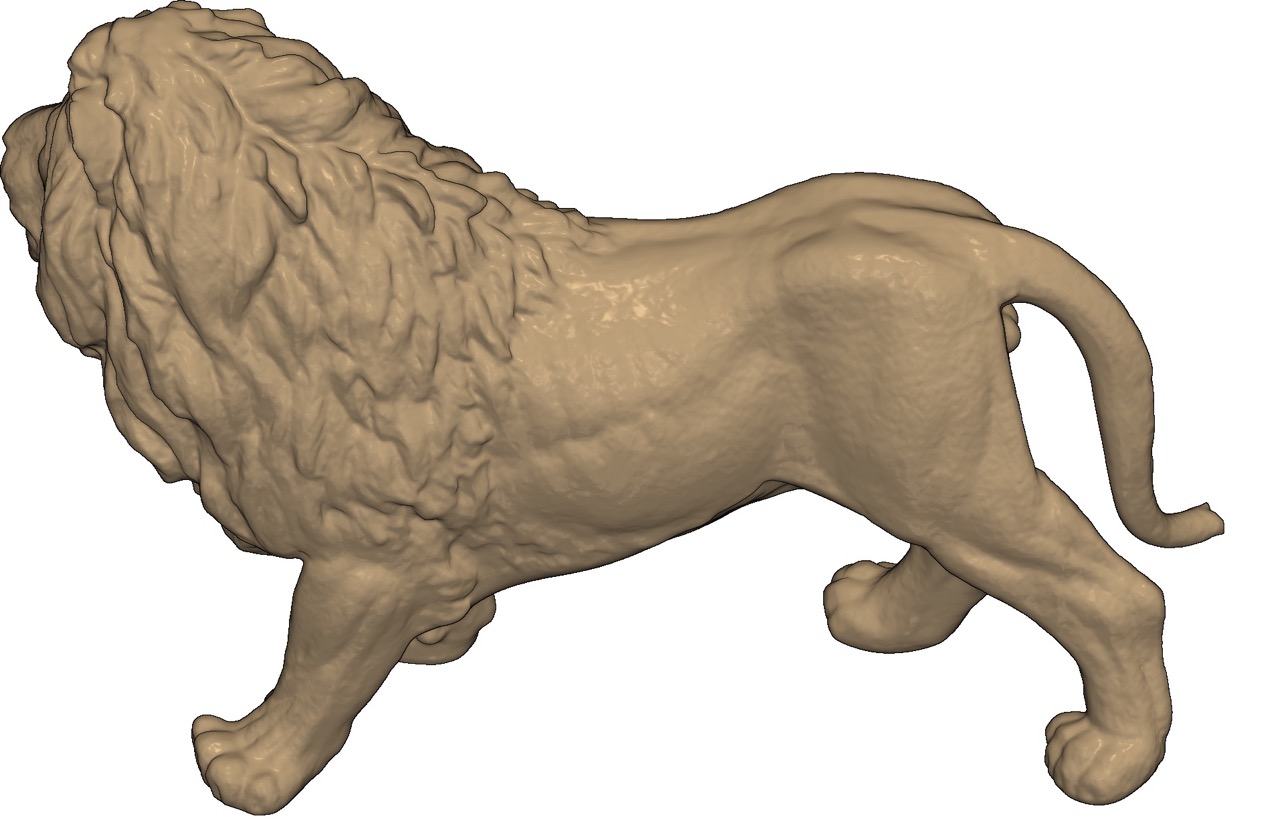}\\
    \includegraphics[width=\figwidthS\linewidth]{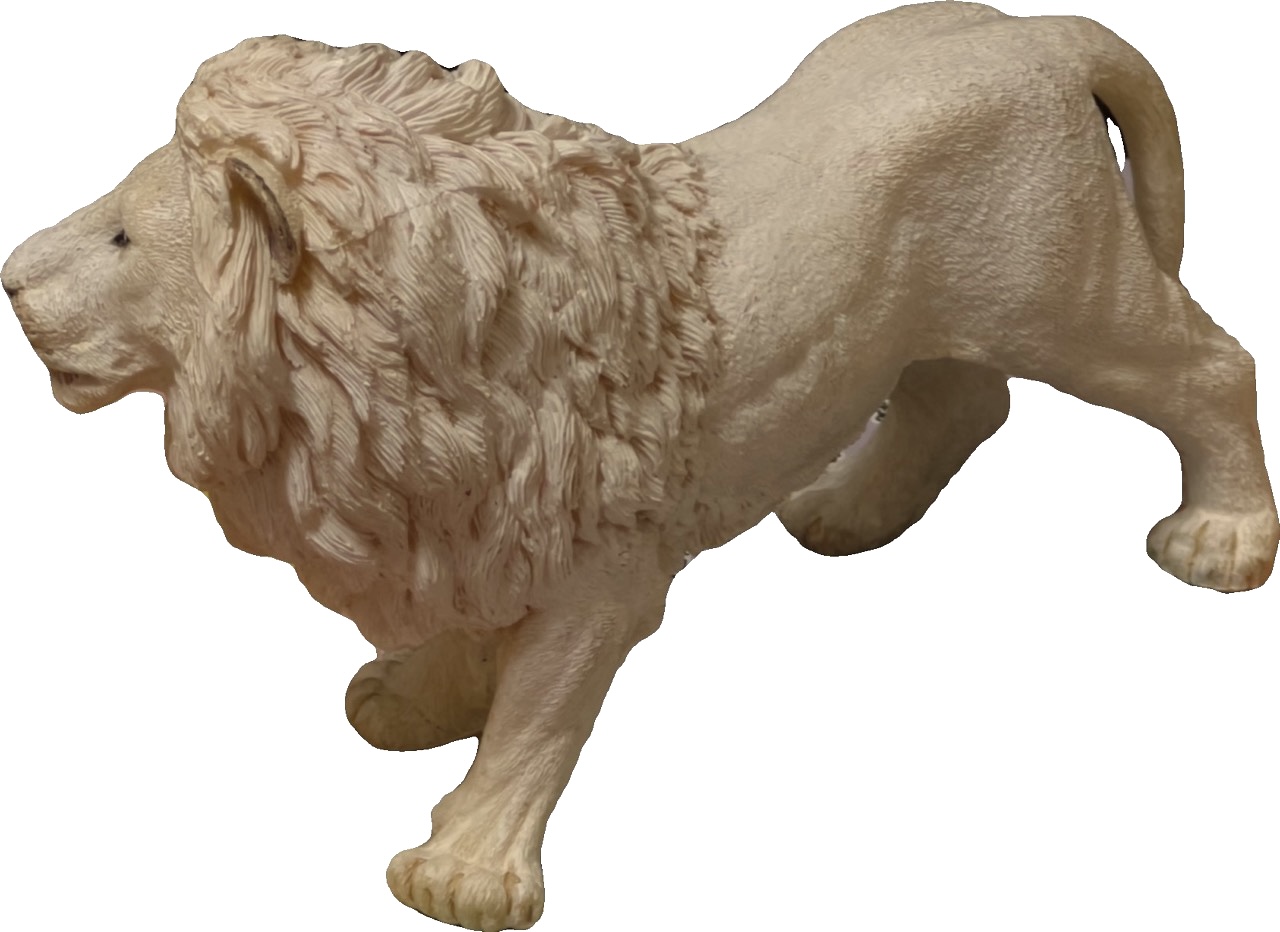}&
        \includegraphics[width=\figwidthS\linewidth]{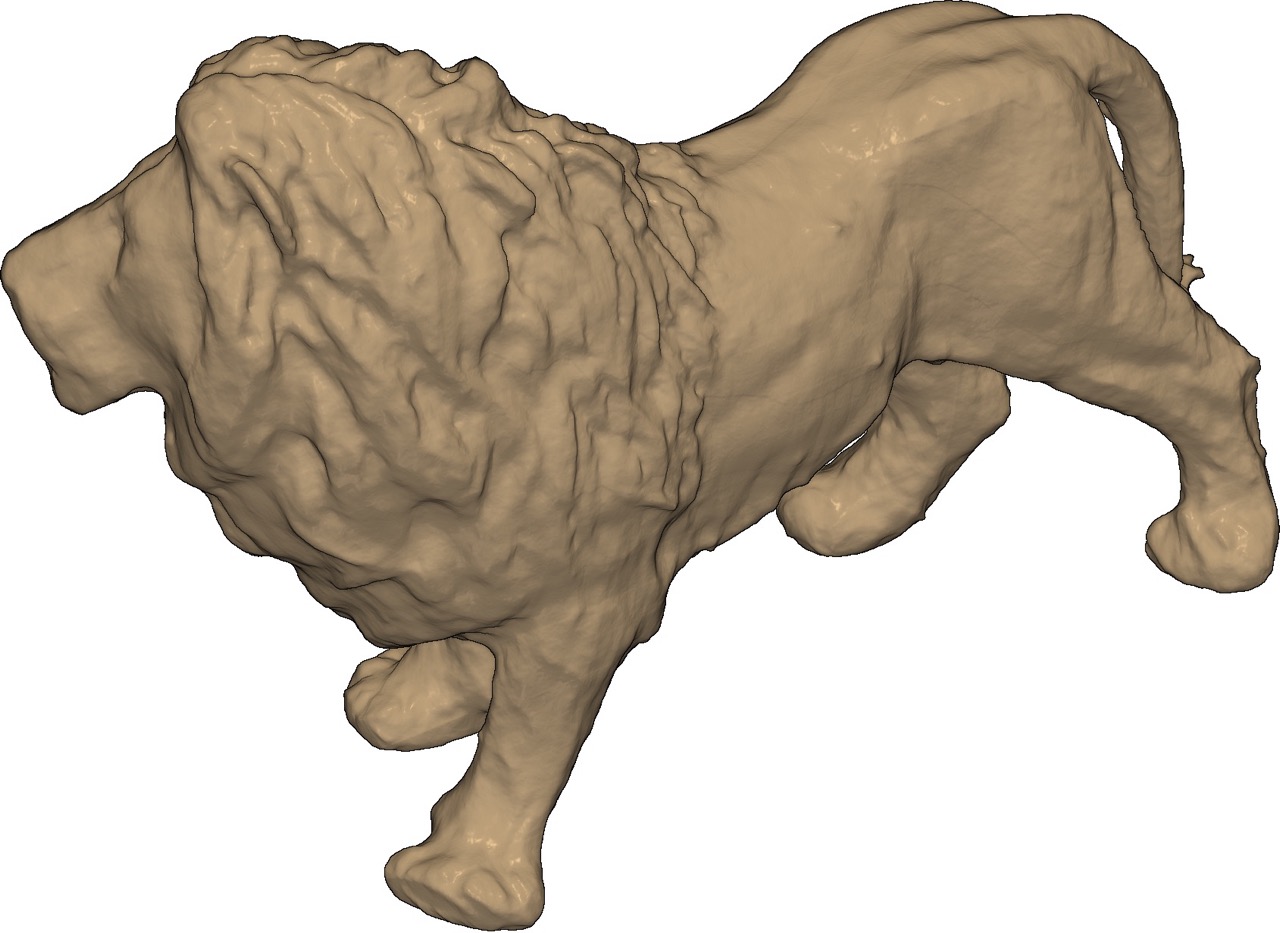}&
                \includegraphics[width=\figwidthS\linewidth]{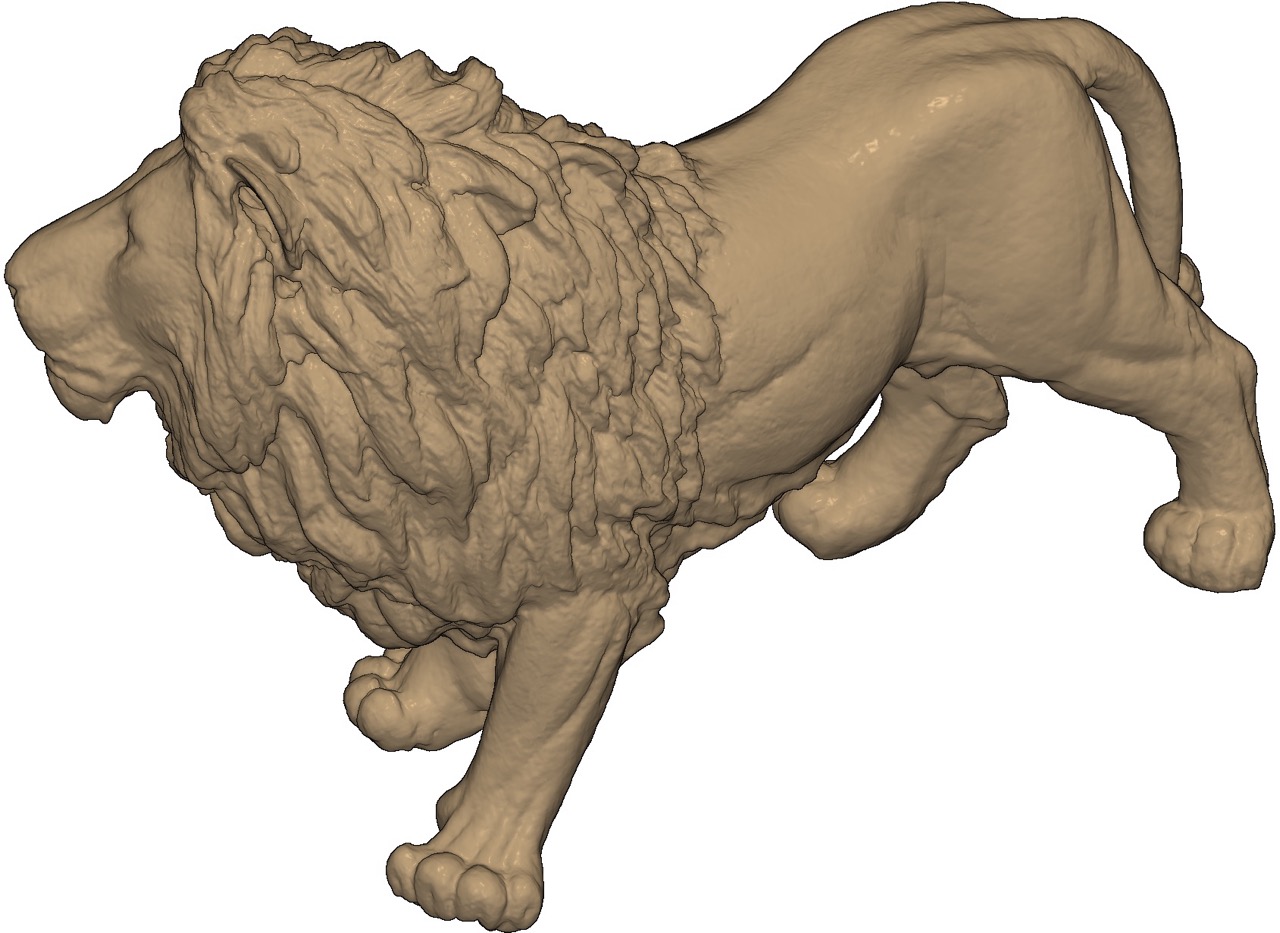}&
                        \includegraphics[width=\figwidthS\linewidth]{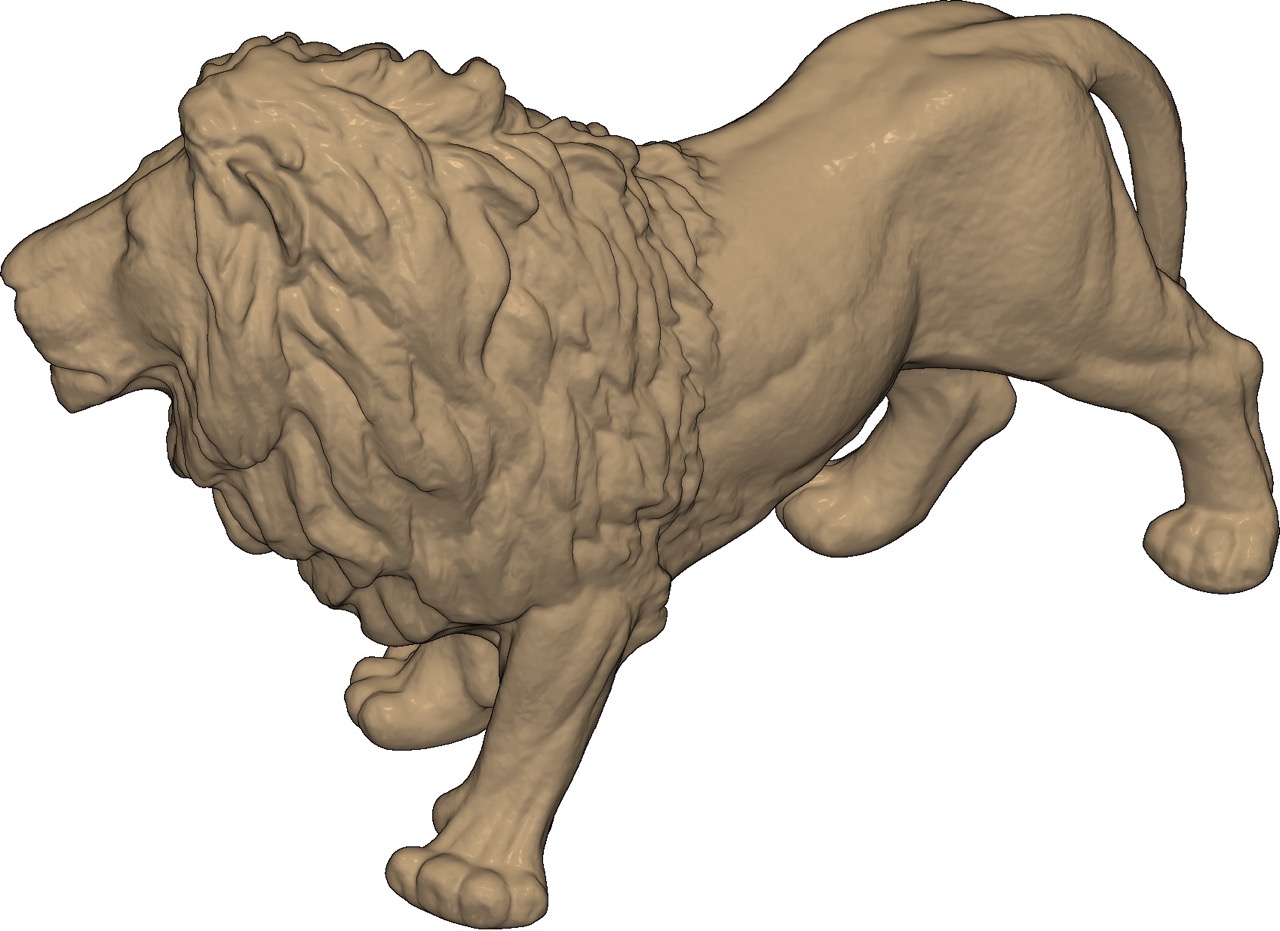}\\
    \includegraphics[width=\figwidthS\linewidth]{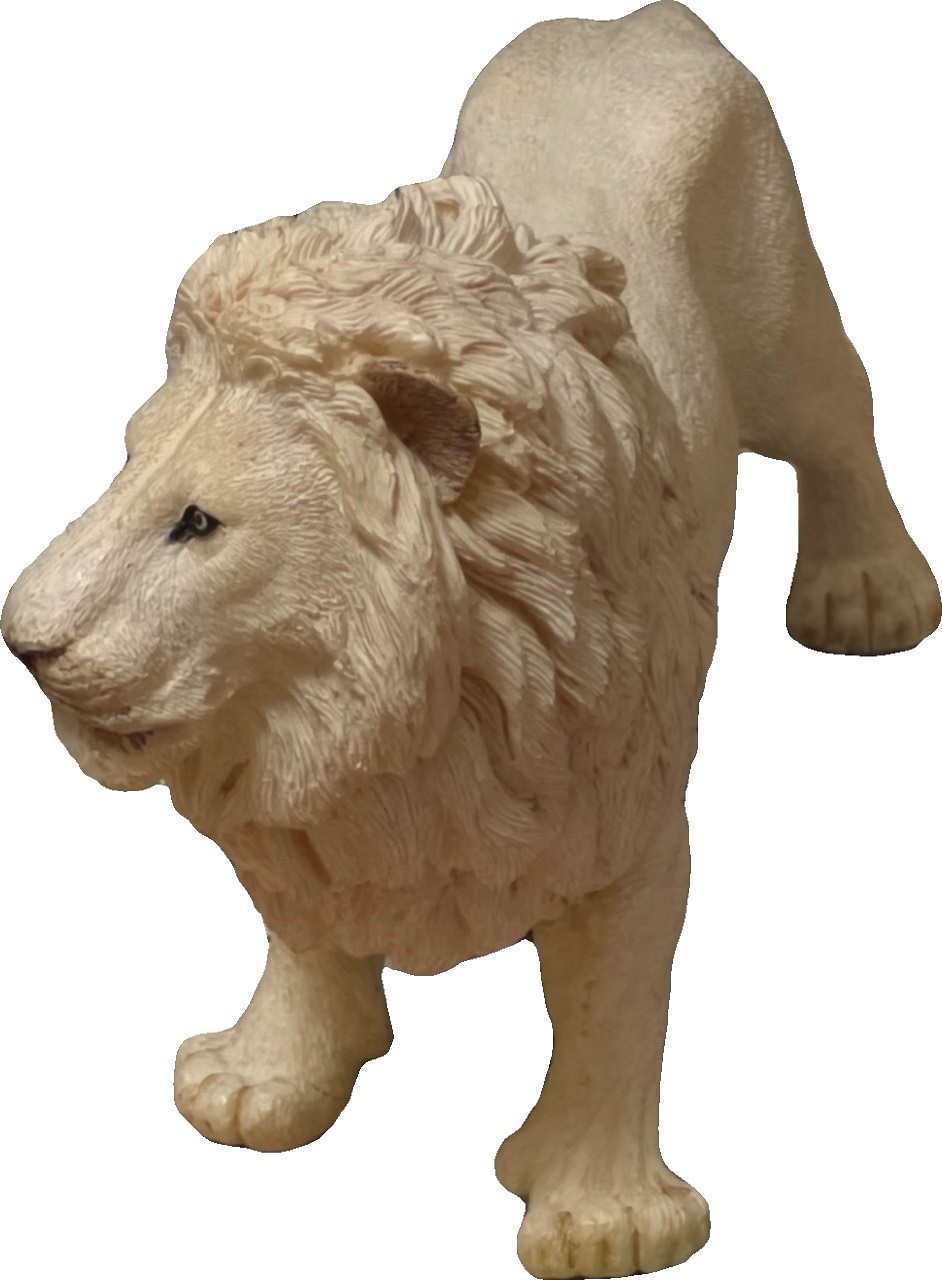}&
        \includegraphics[width=\figwidthS\linewidth]{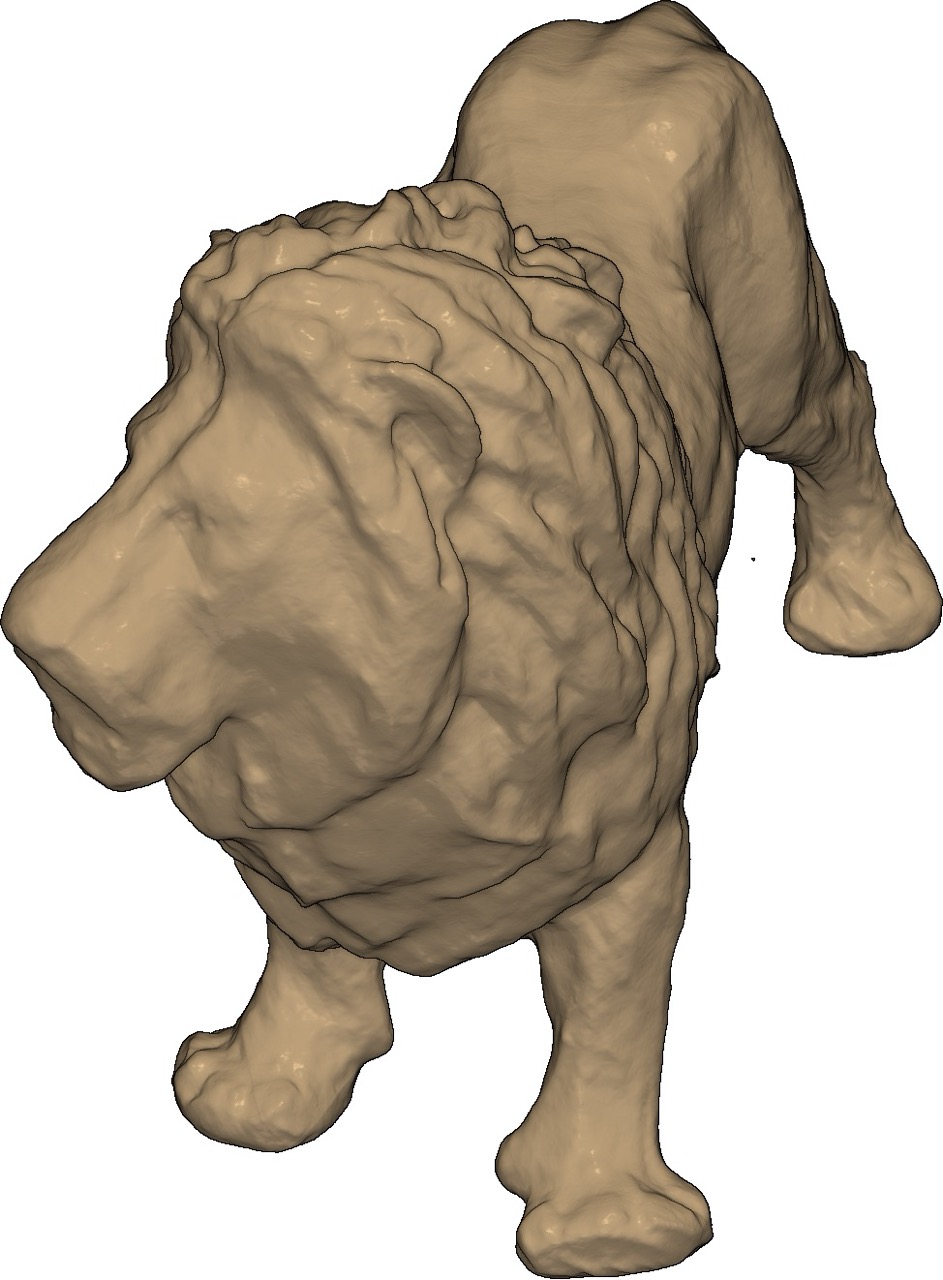}&
                \includegraphics[width=\figwidthS\linewidth]{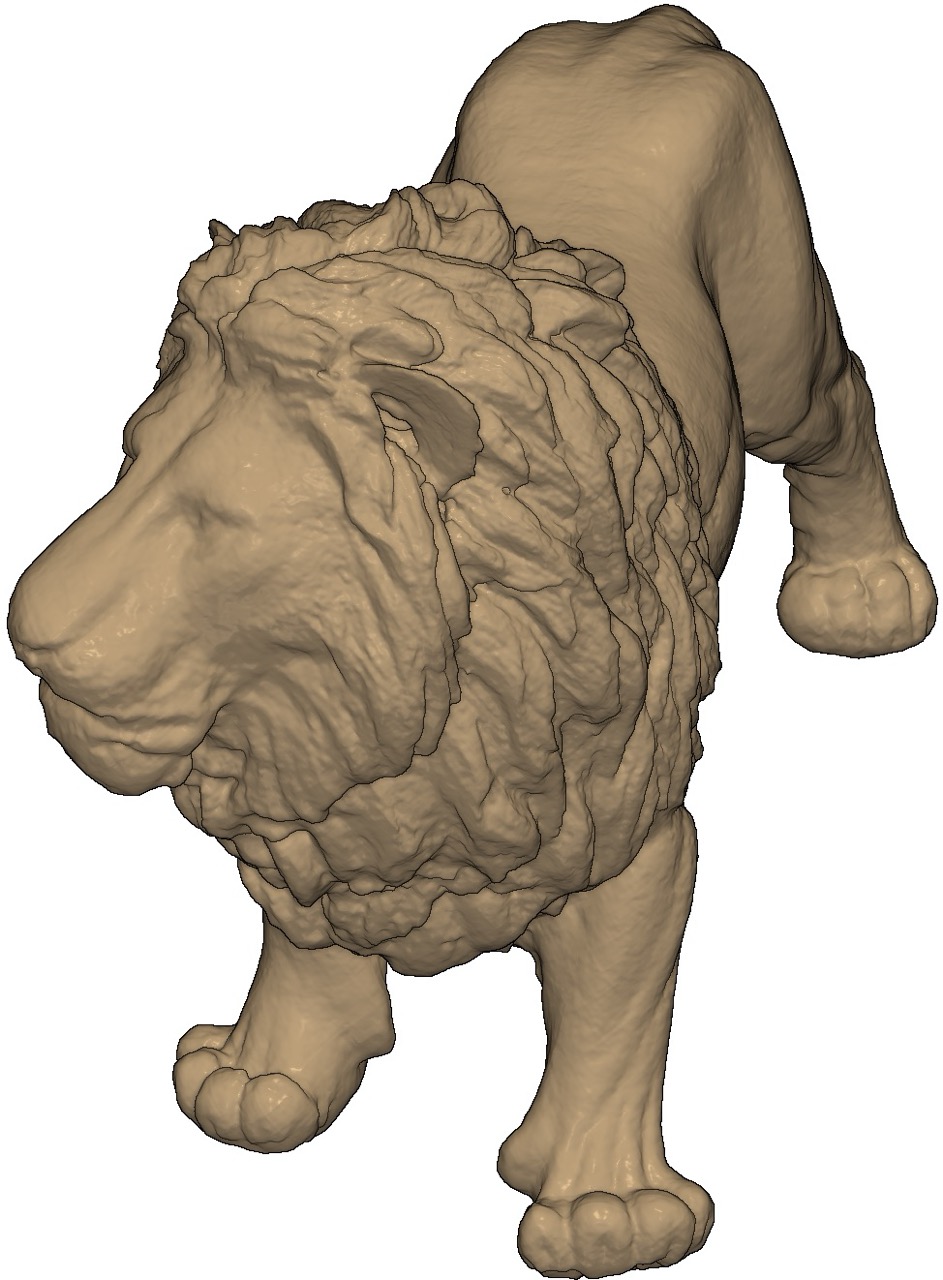}&
                        \includegraphics[width=\figwidthS\linewidth]{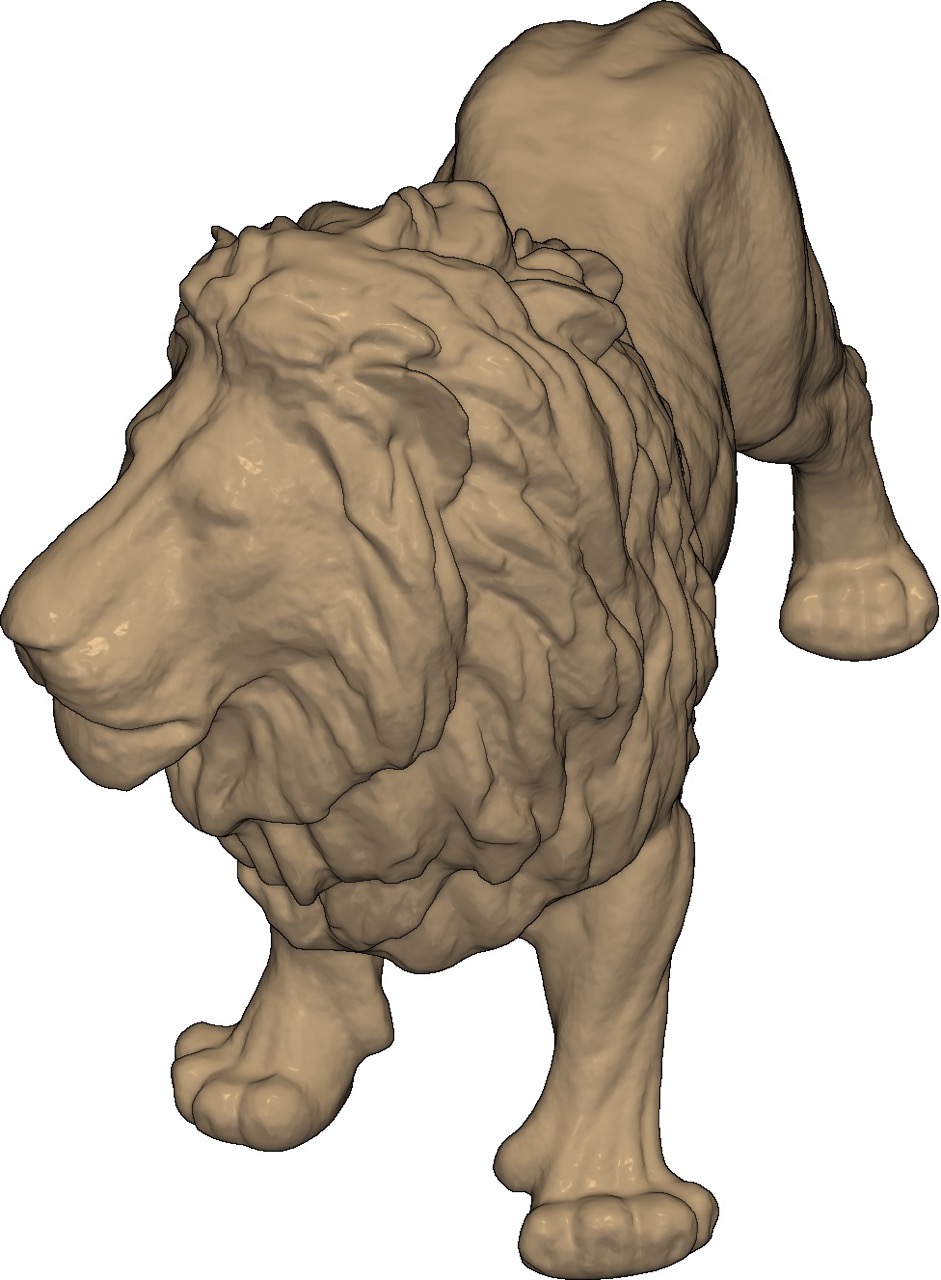}\\
    \includegraphics[width=\figwidthS\linewidth]{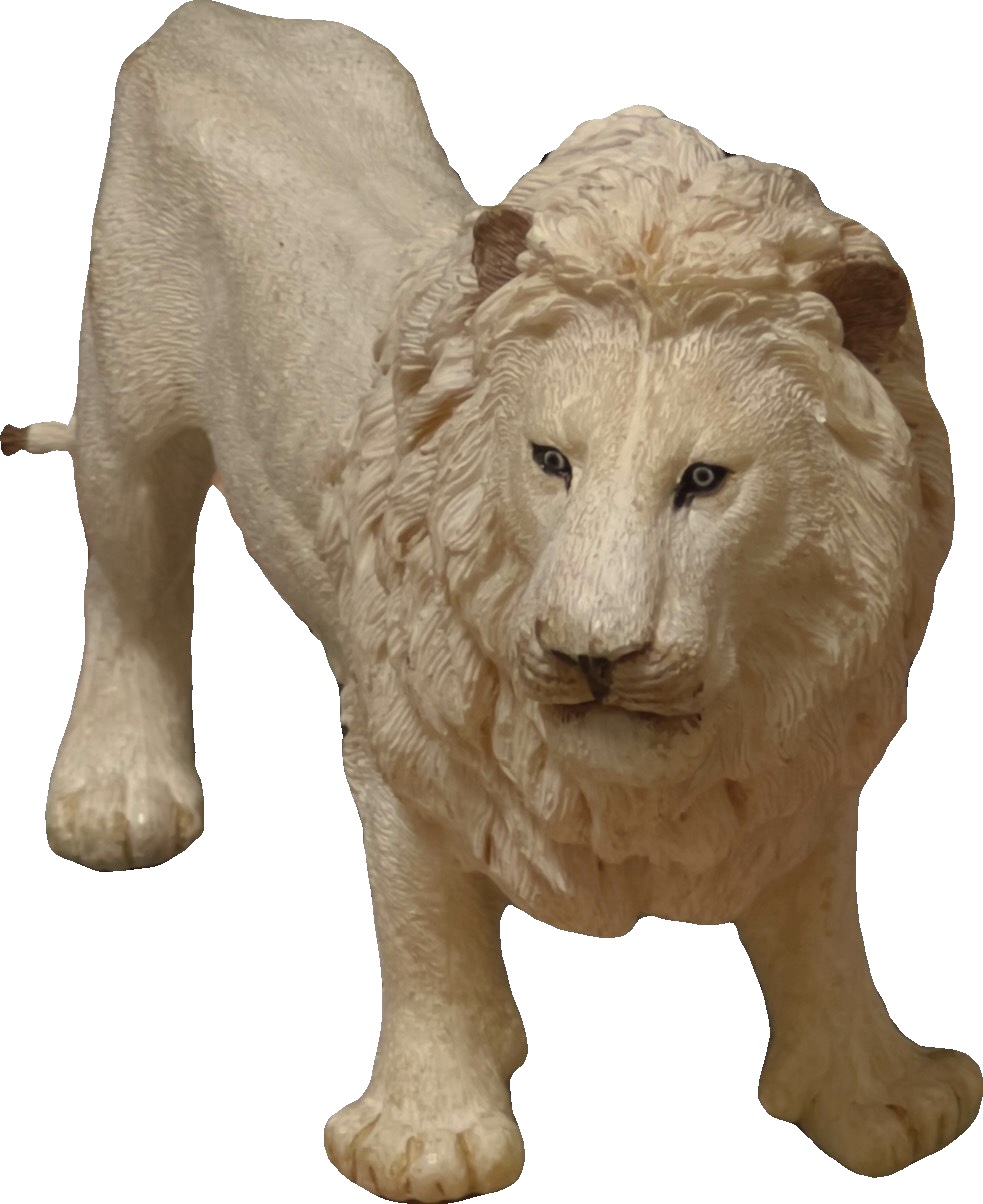}&
        \includegraphics[width=\figwidthS\linewidth]{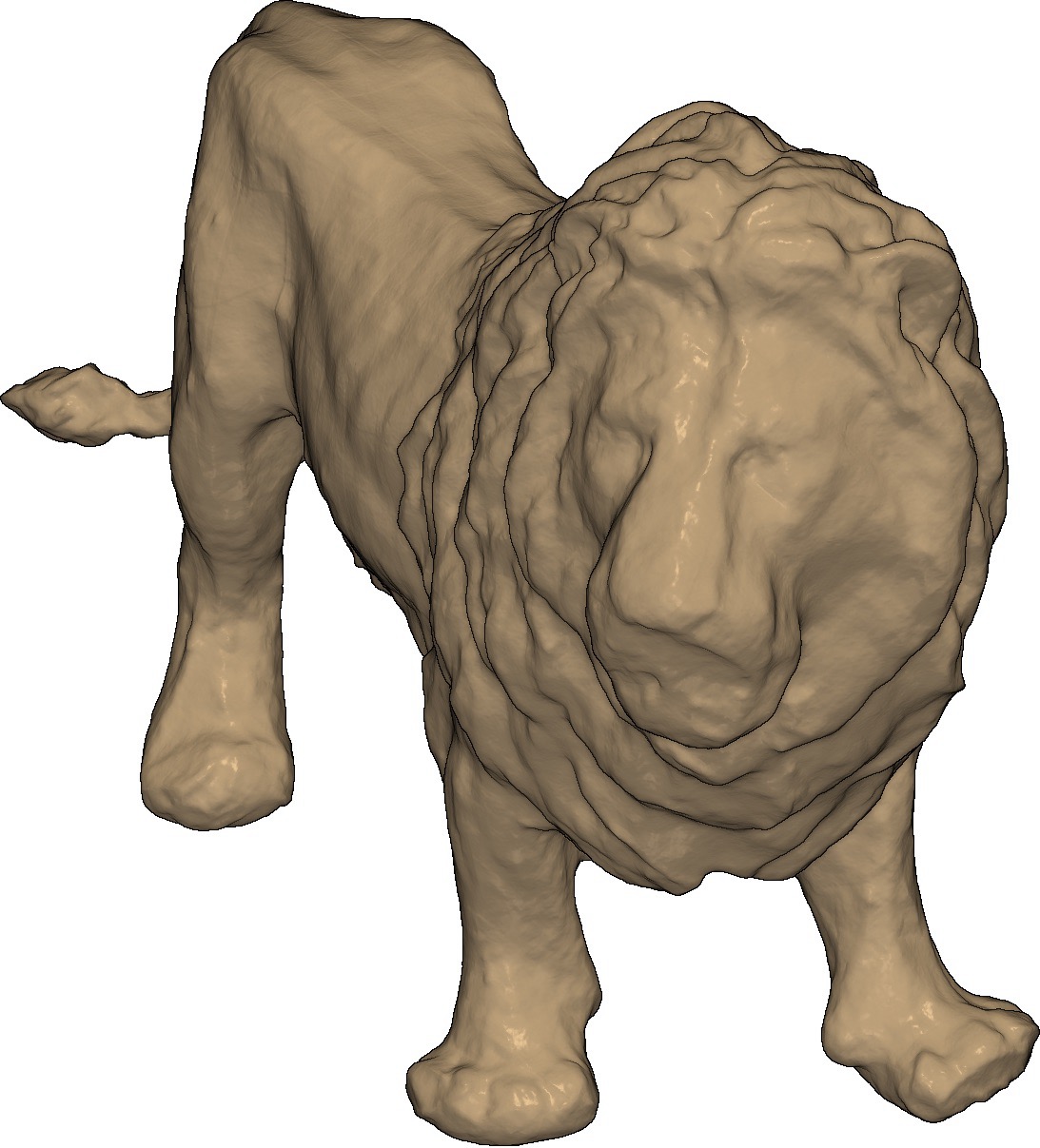}&
                \includegraphics[width=\figwidthS\linewidth]{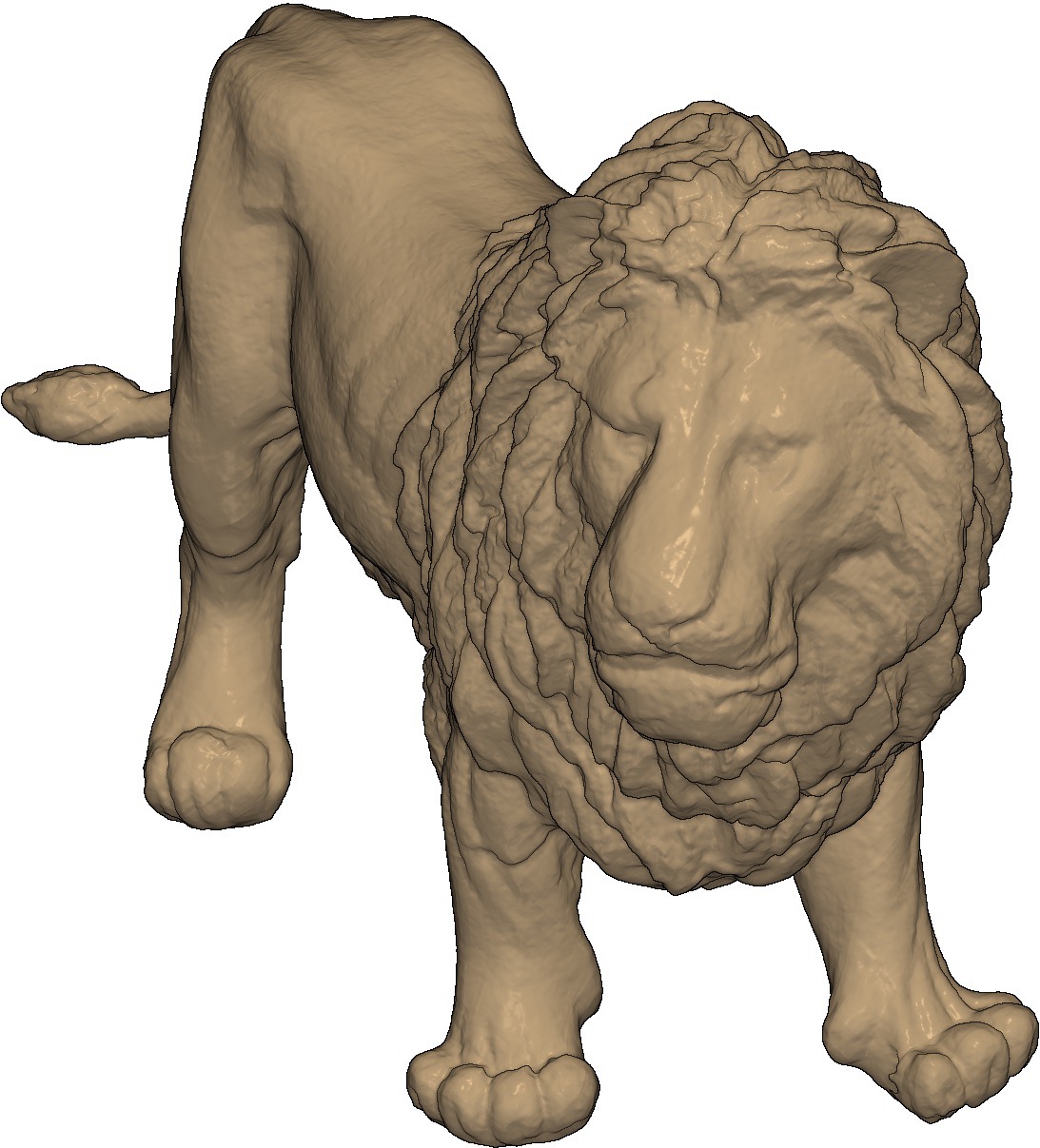}&
                        \includegraphics[width=\figwidthS\linewidth]{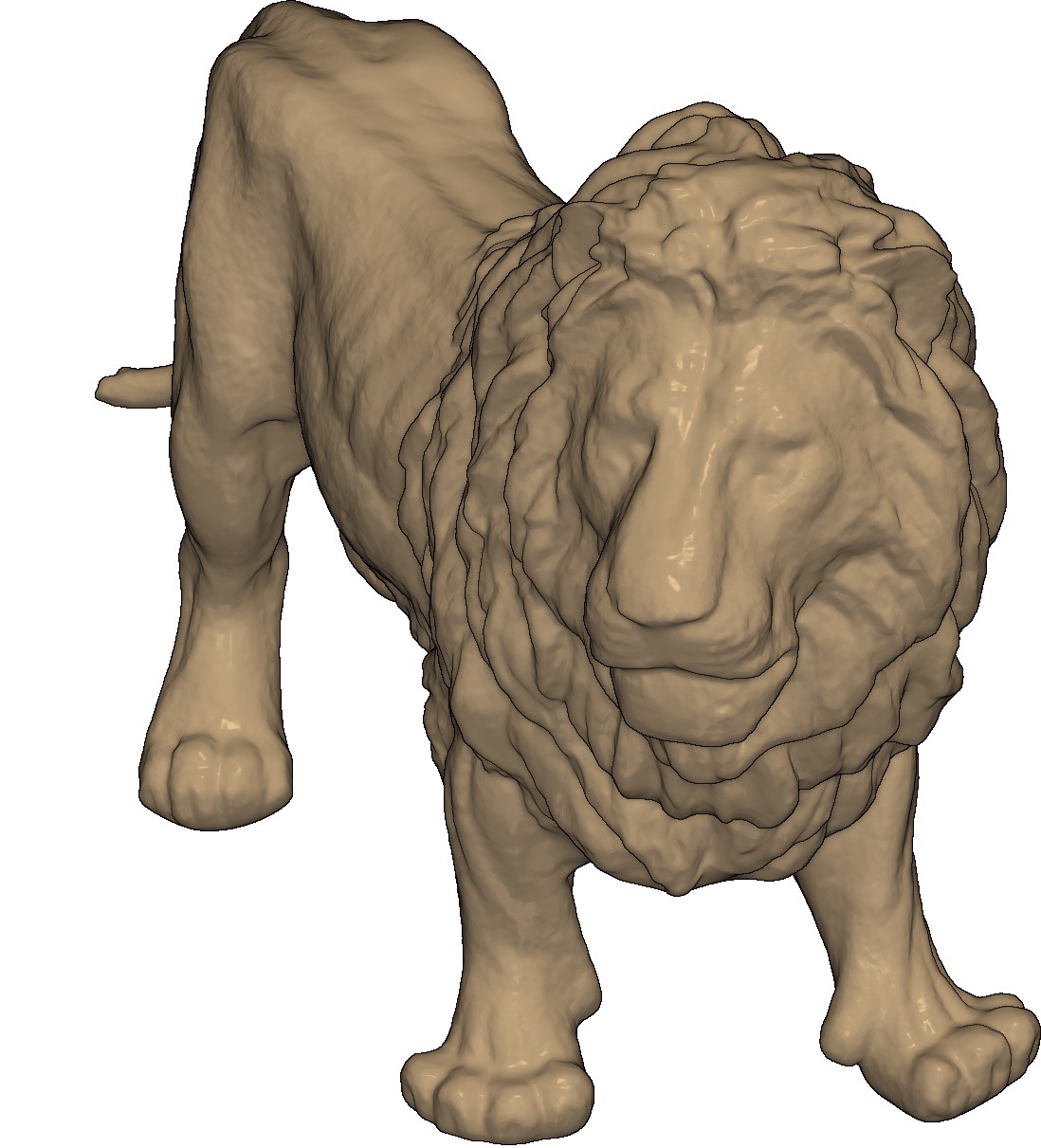}\\
    \includegraphics[width=\figwidthS\linewidth]{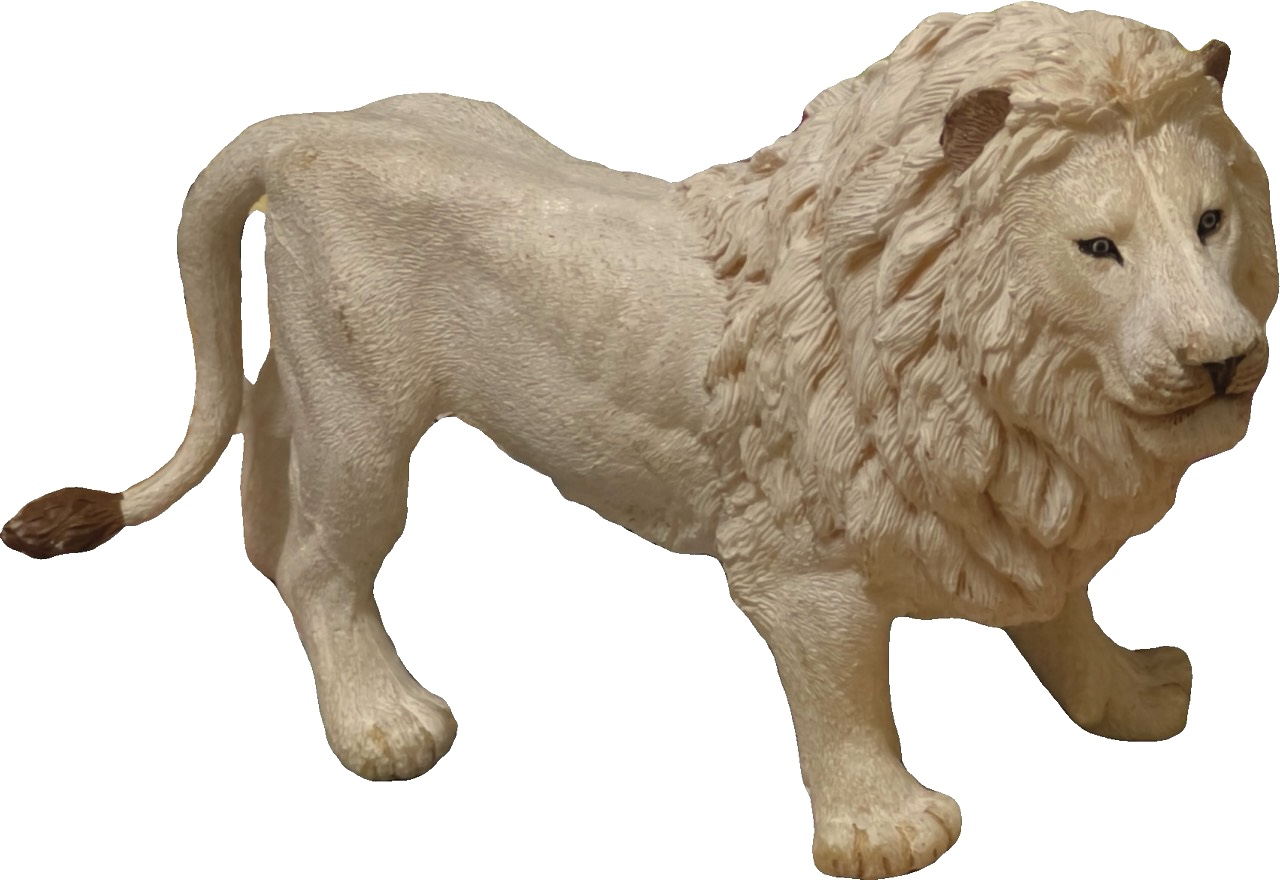}&
        \includegraphics[width=\figwidthS\linewidth]{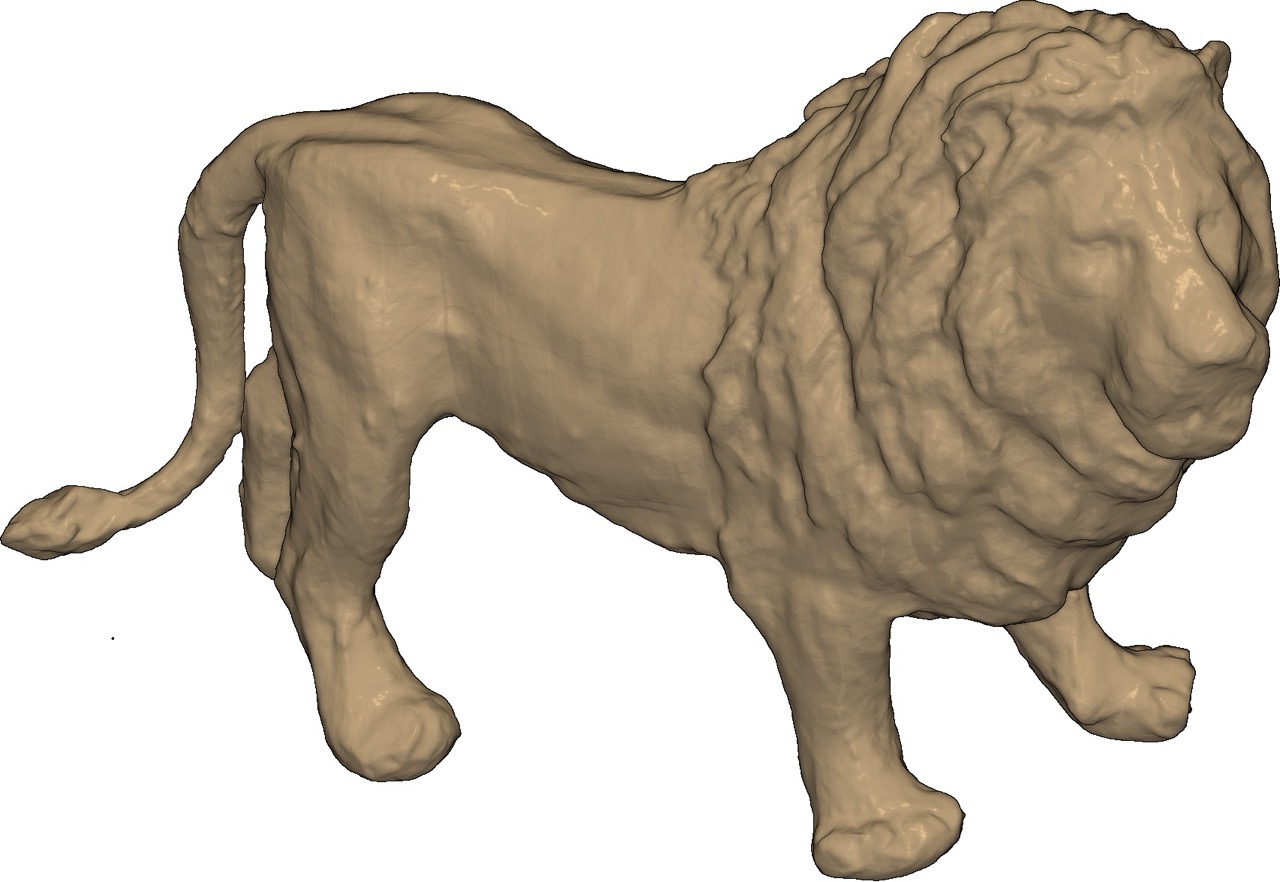}&
            \includegraphics[width=\figwidthS\linewidth]{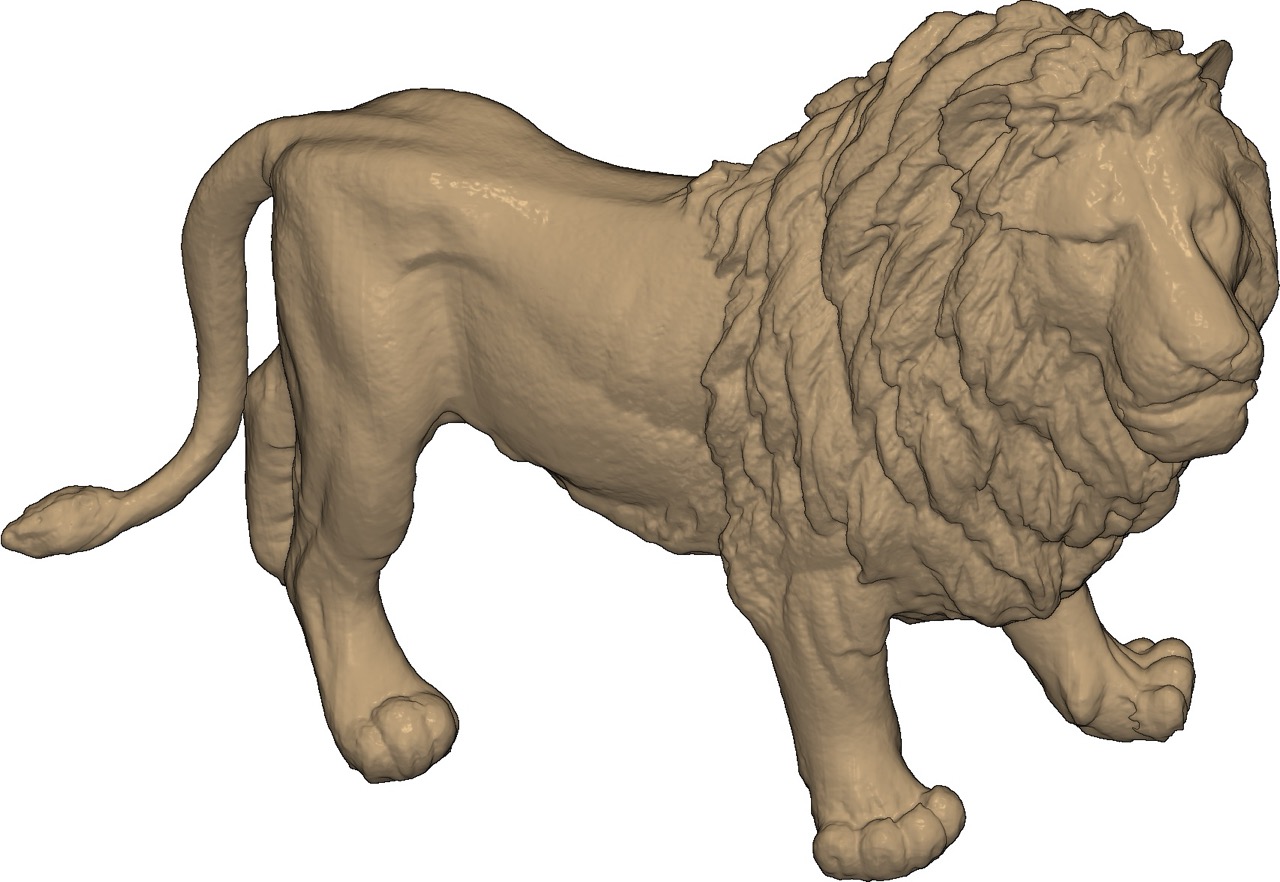}&
                \includegraphics[width=\figwidthS\linewidth]{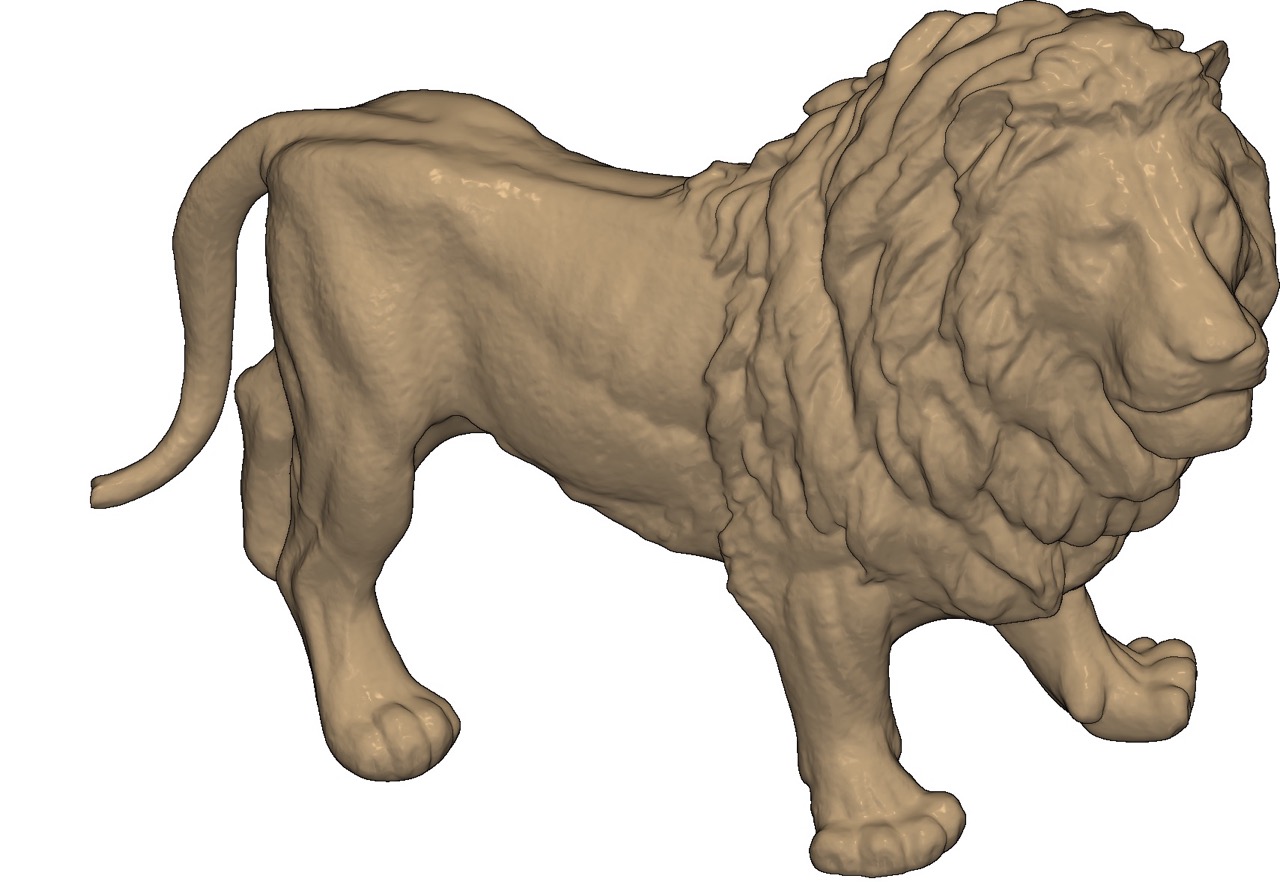}\\
    \includegraphics[width=\figwidthS\linewidth]{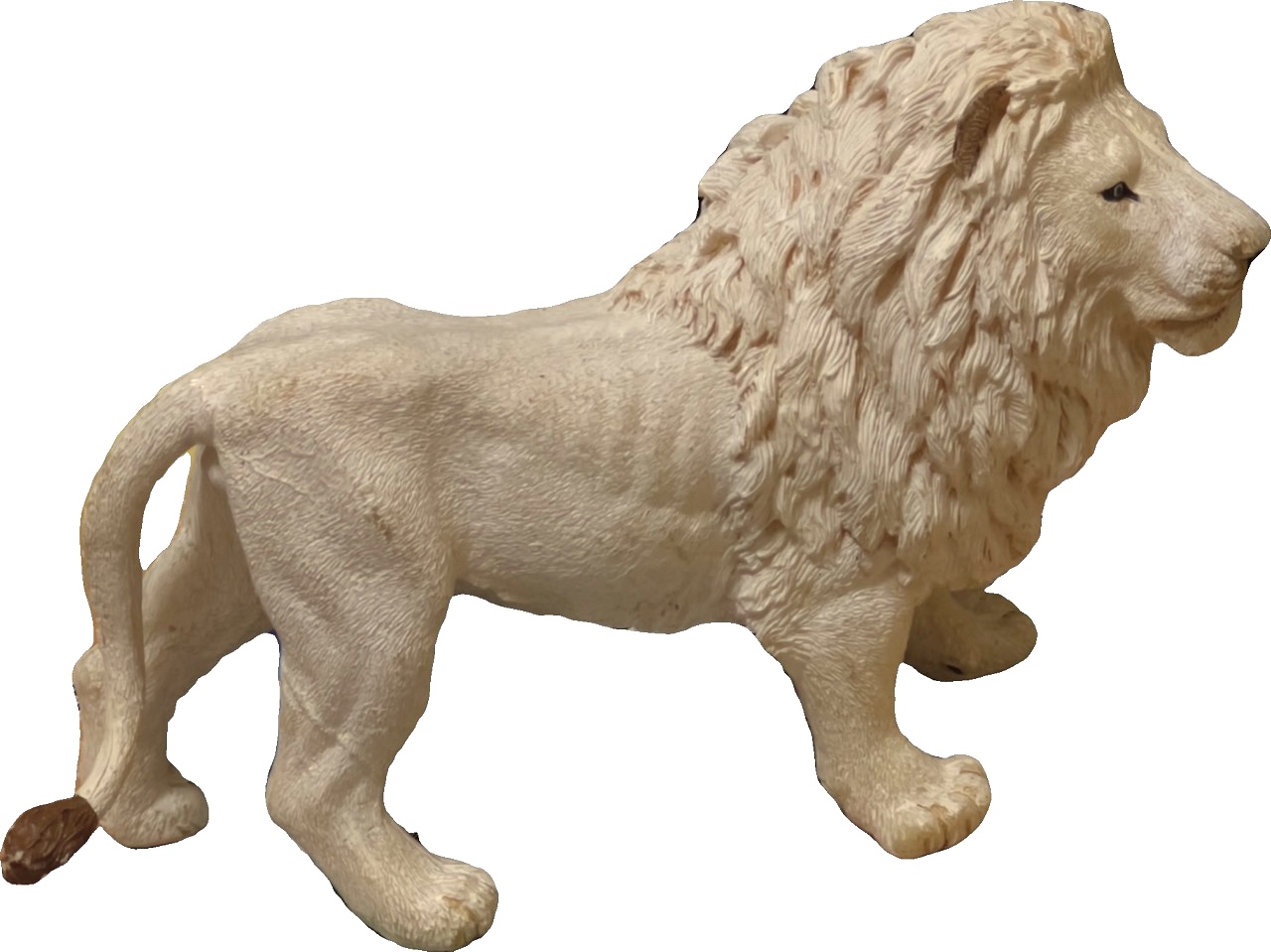}&
        \includegraphics[width=\figwidthS\linewidth]{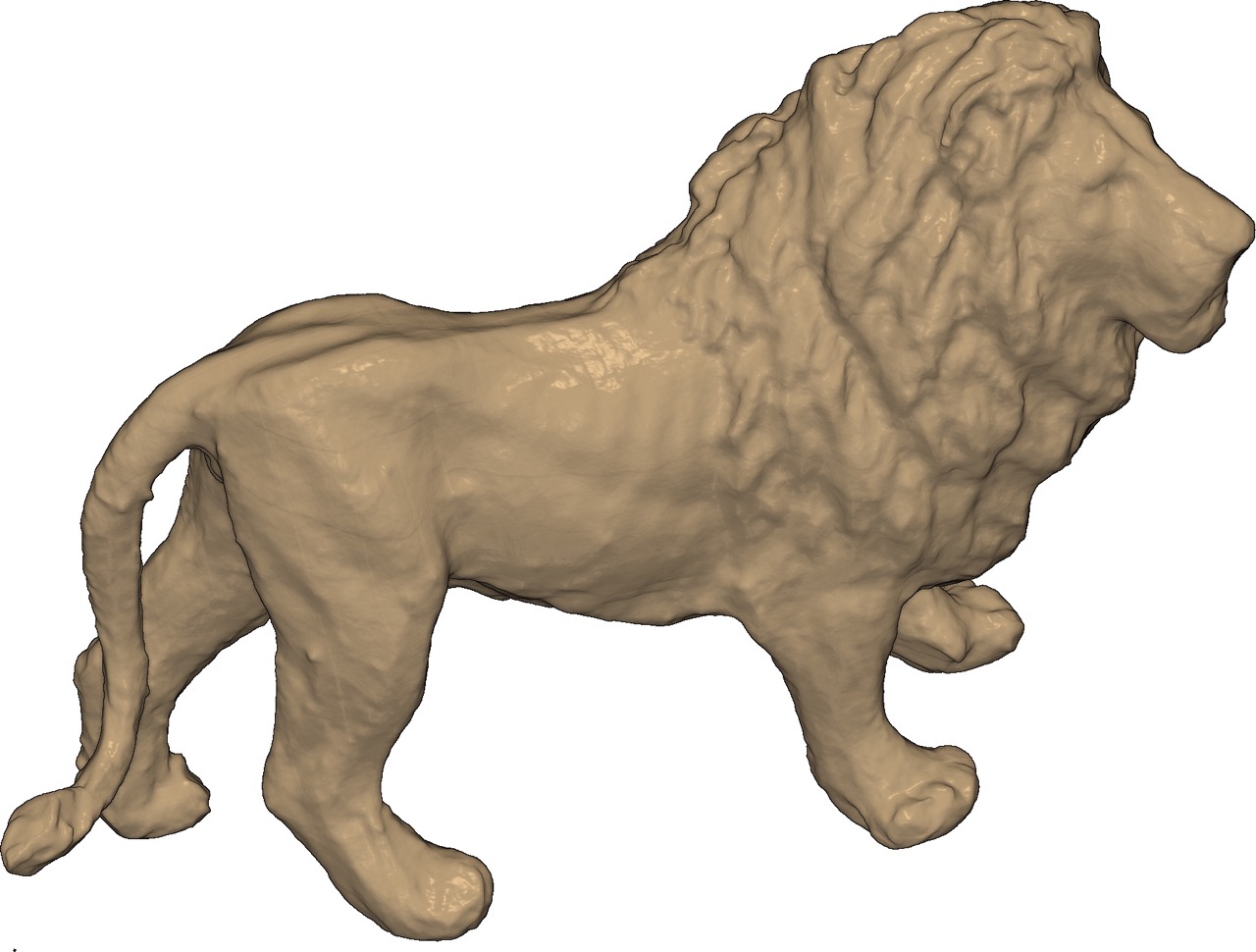}&
                \includegraphics[width=\figwidthS\linewidth]{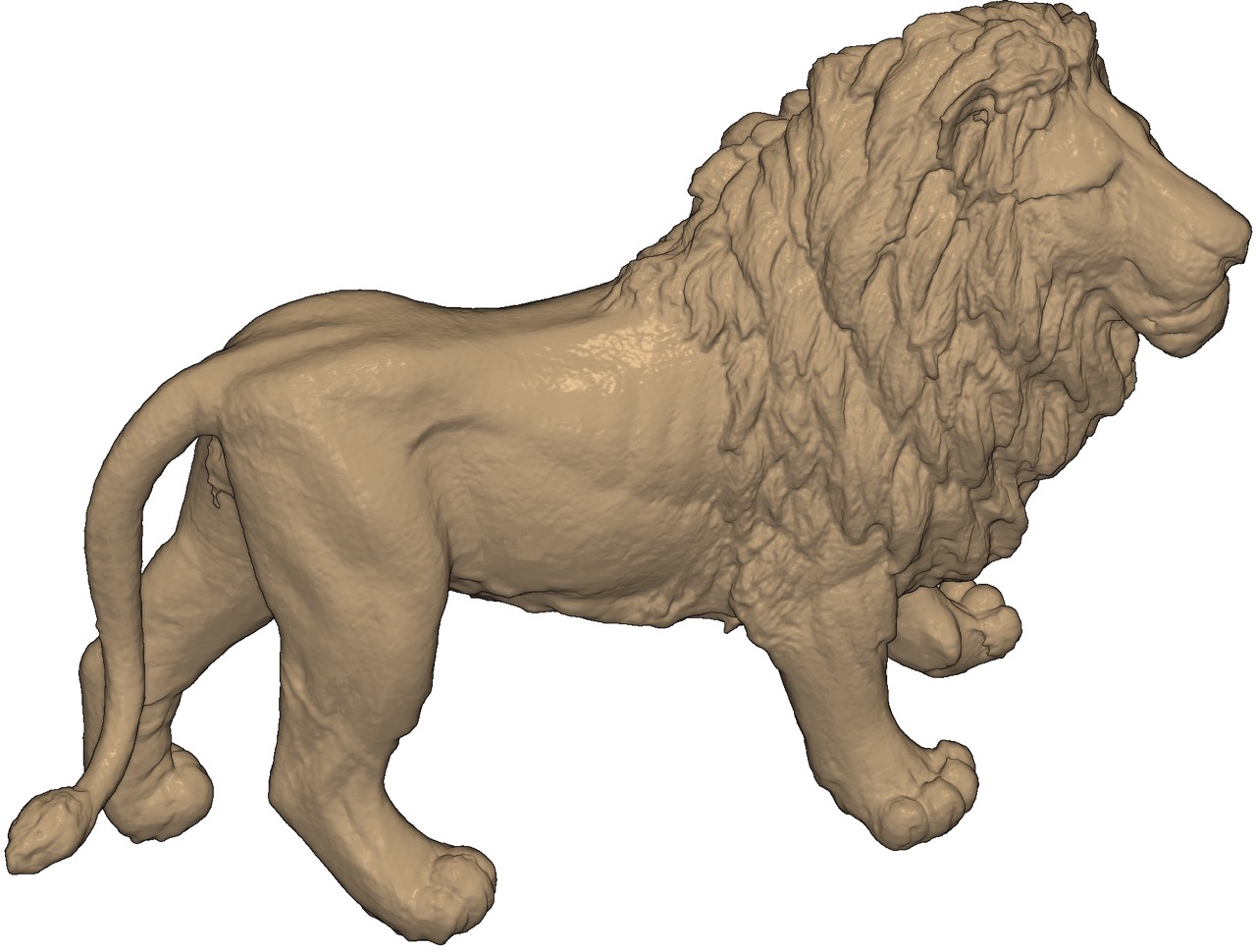}&
                        \includegraphics[width=\figwidthS\linewidth]{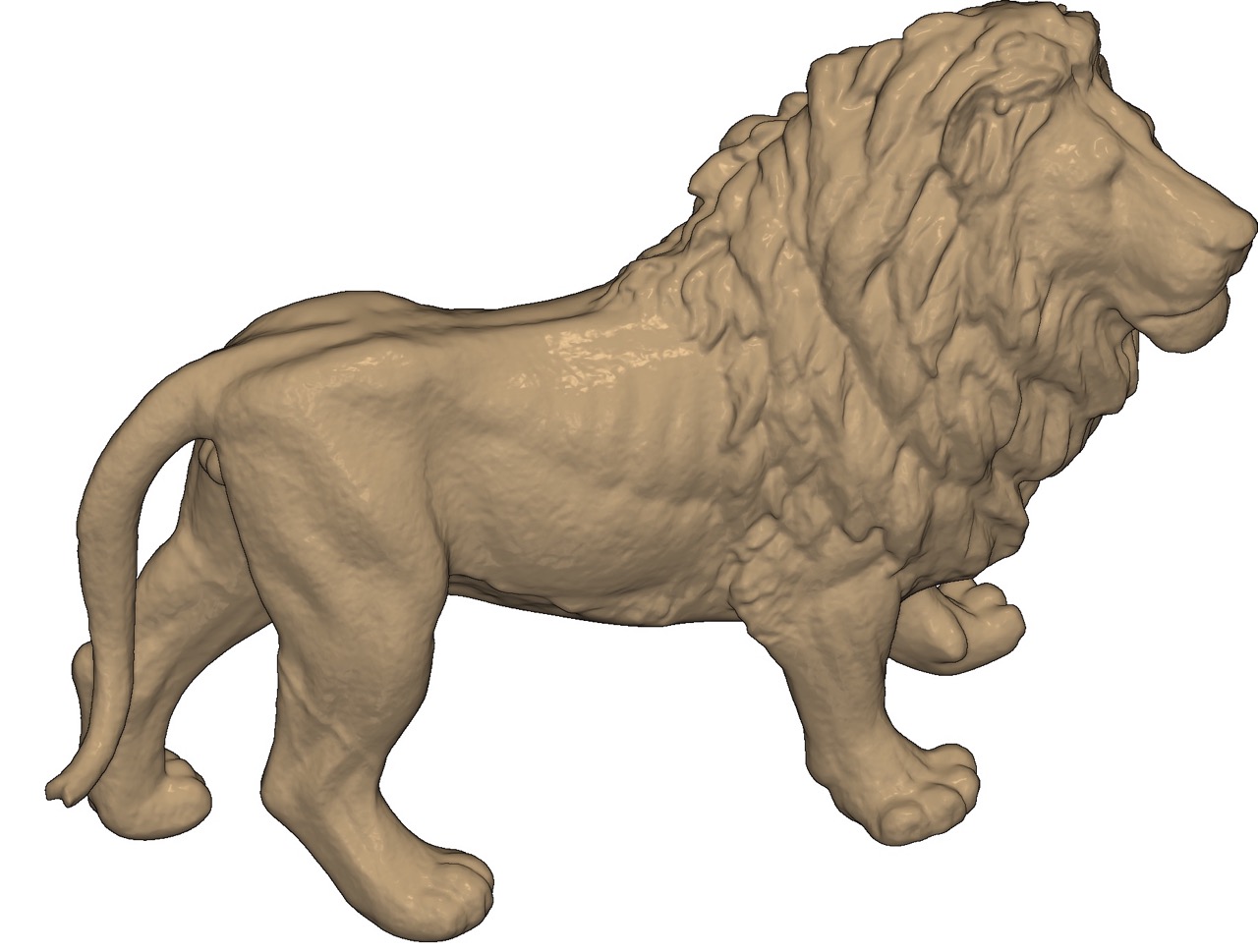}\\
    \includegraphics[width=\figwidthS\linewidth]{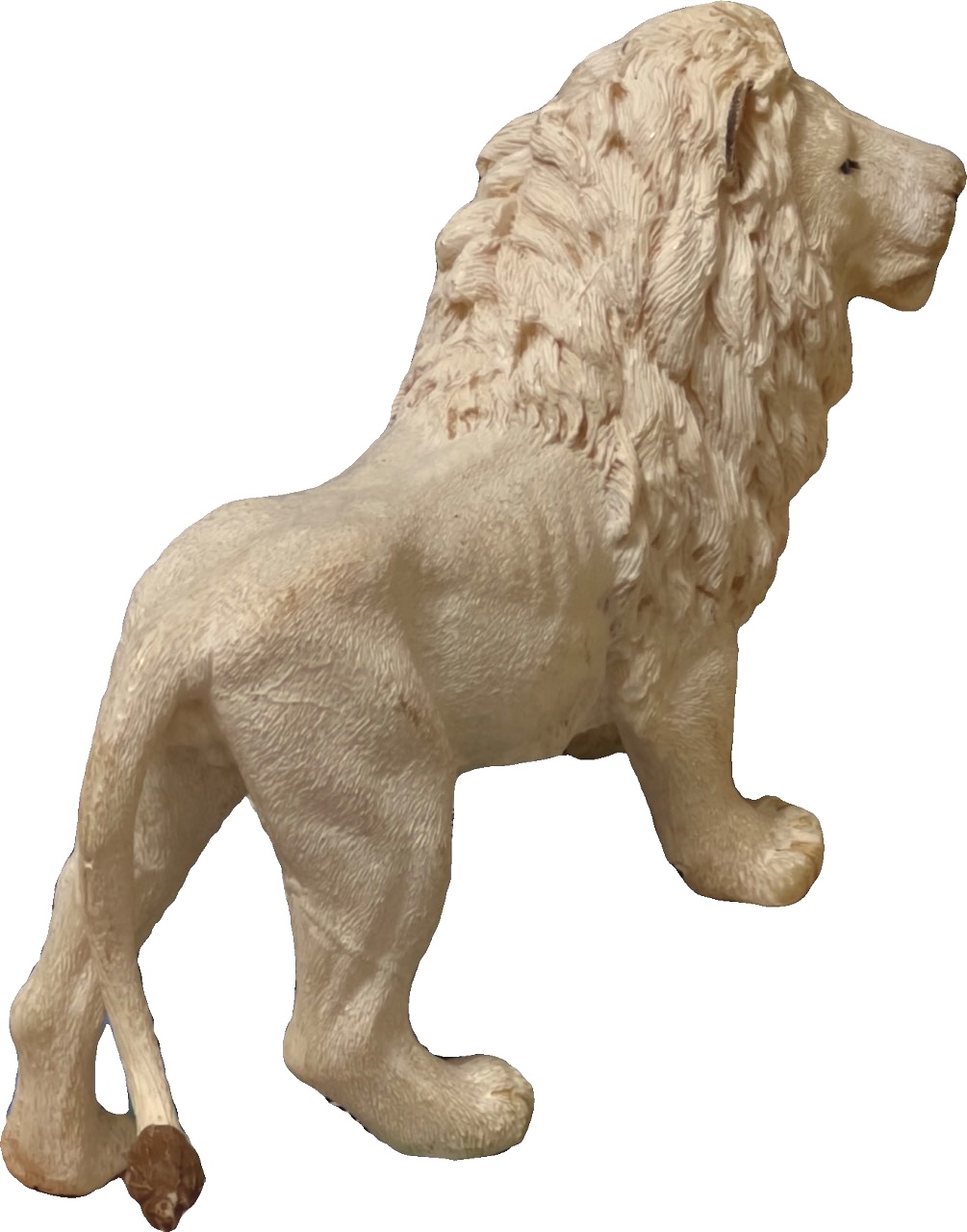}&
        \includegraphics[width=\figwidthS\linewidth]{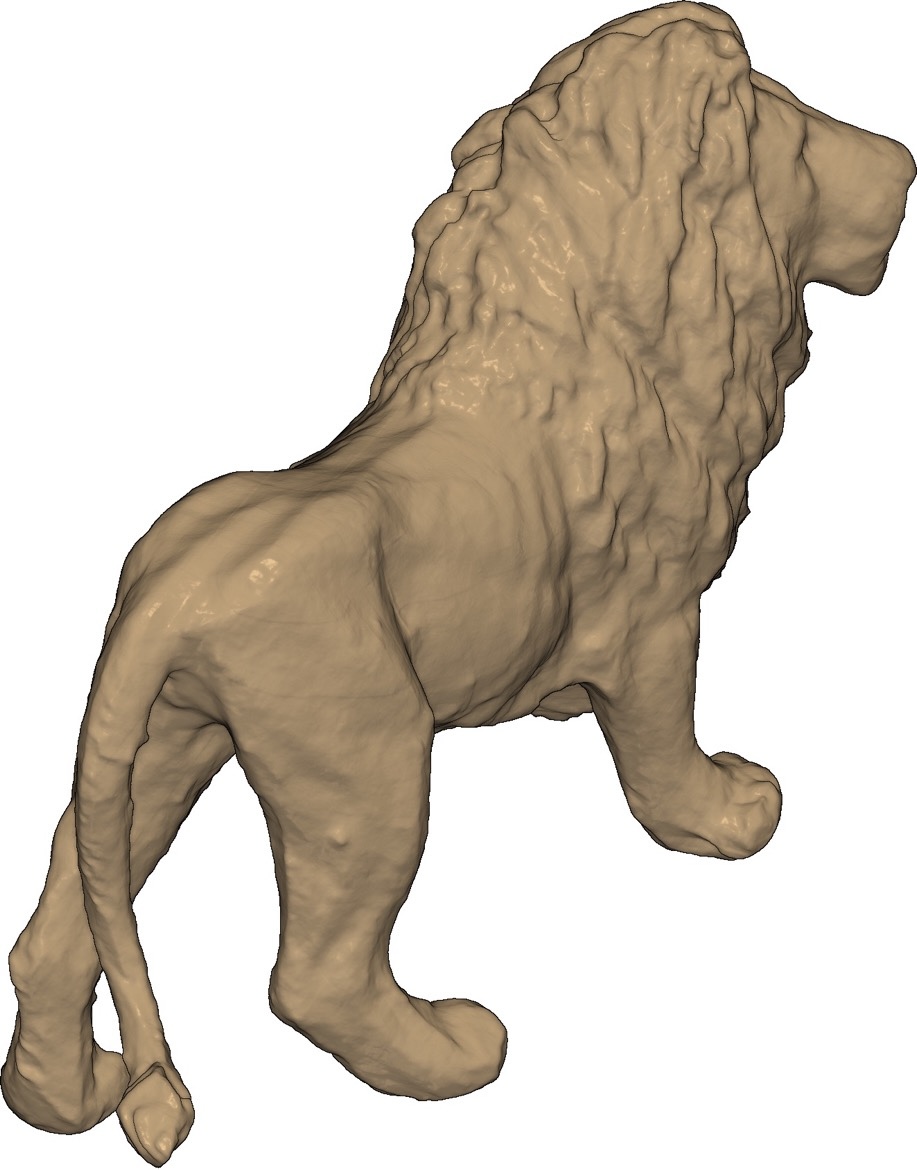}&
                \includegraphics[width=\figwidthS\linewidth]{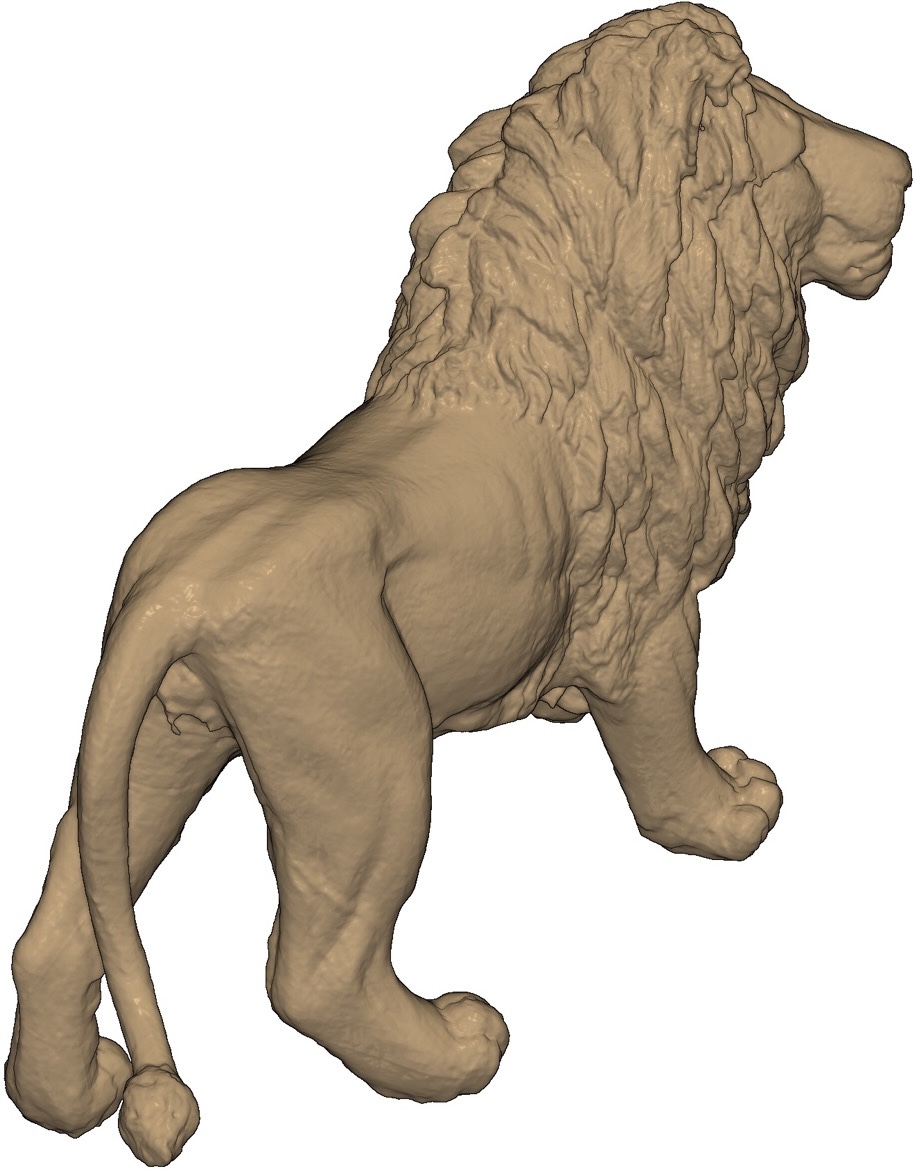}&
                        \includegraphics[width=\figwidthS\linewidth]{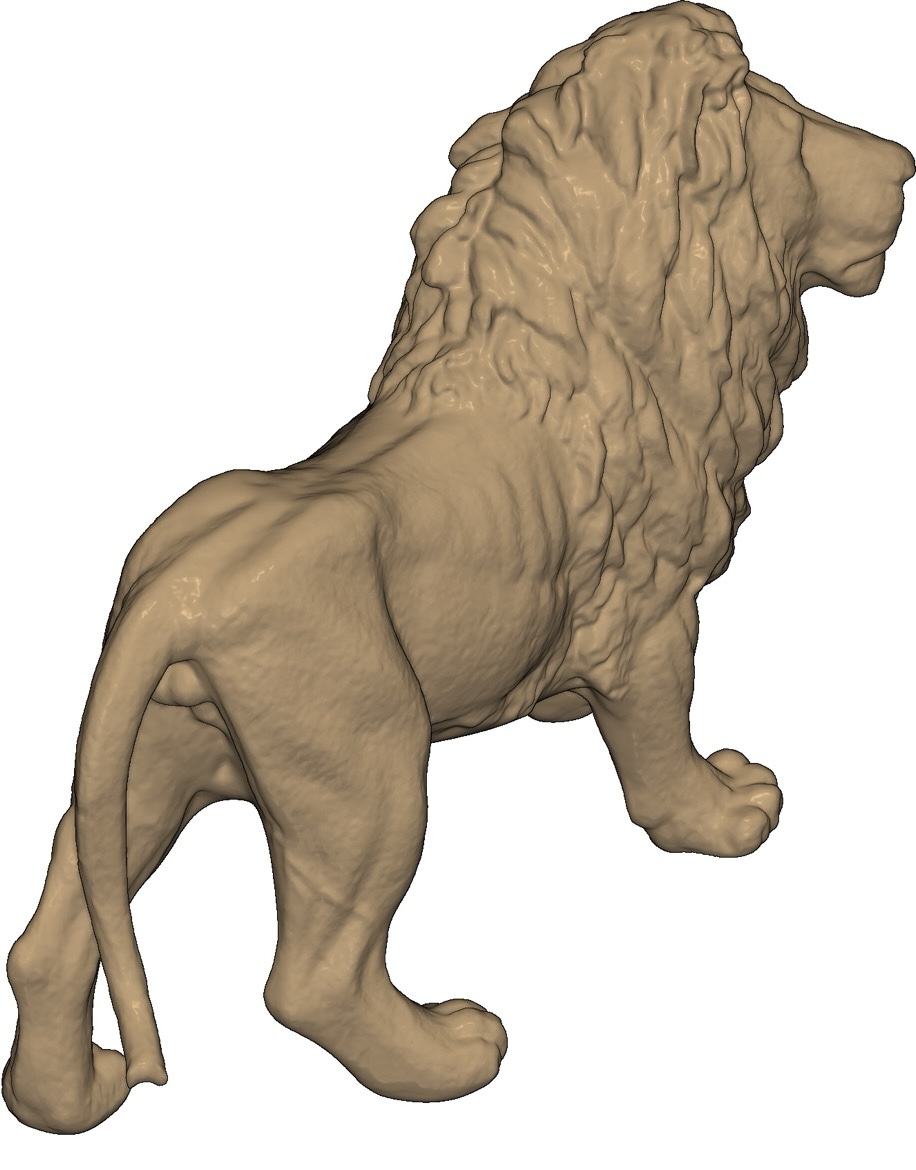}\\
        \includegraphics[width=\figwidthS\linewidth]{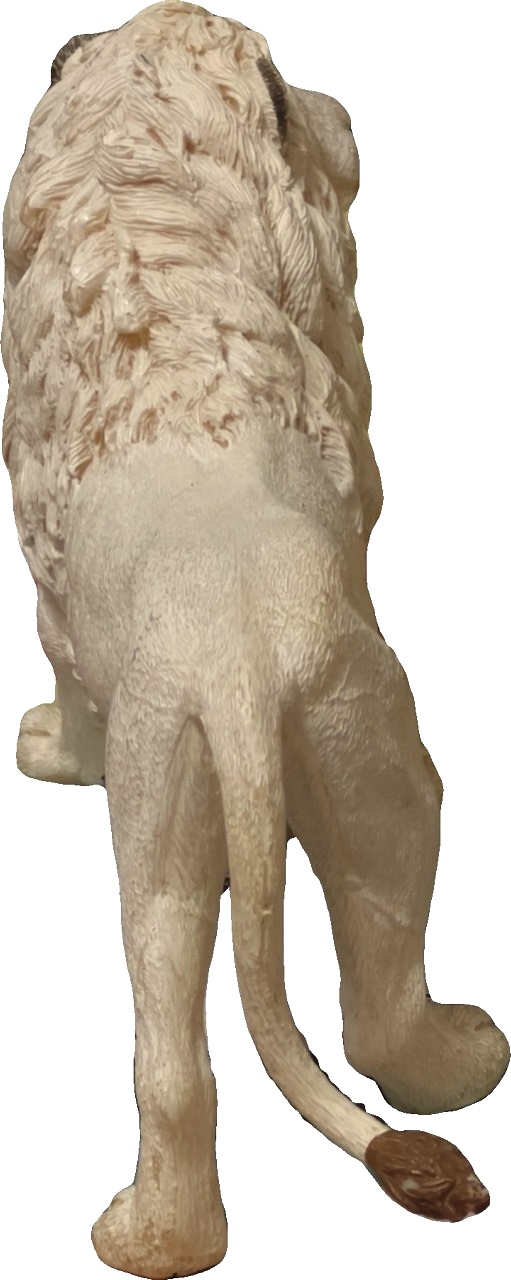}&
        \includegraphics[width=\figwidthS\linewidth]{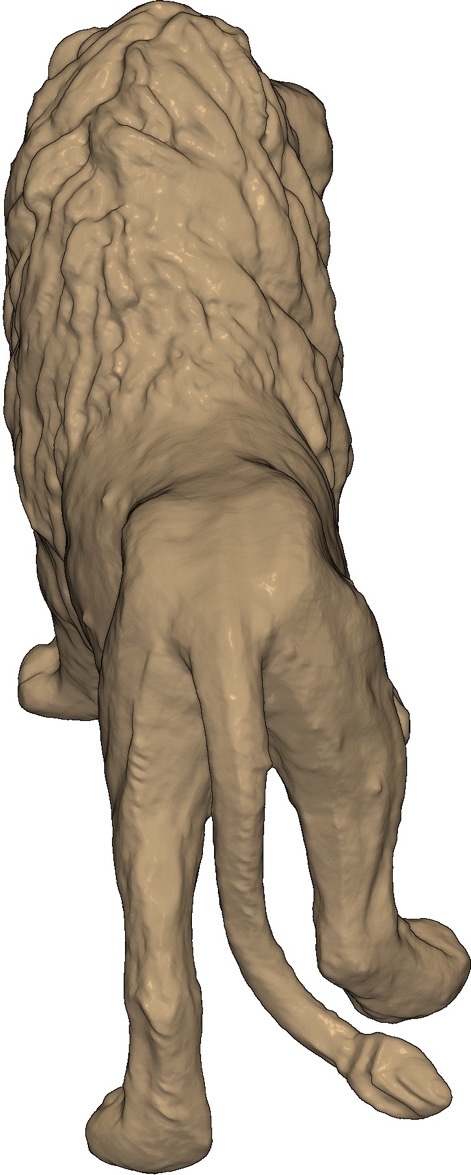}&
                \includegraphics[width=\figwidthS\linewidth]{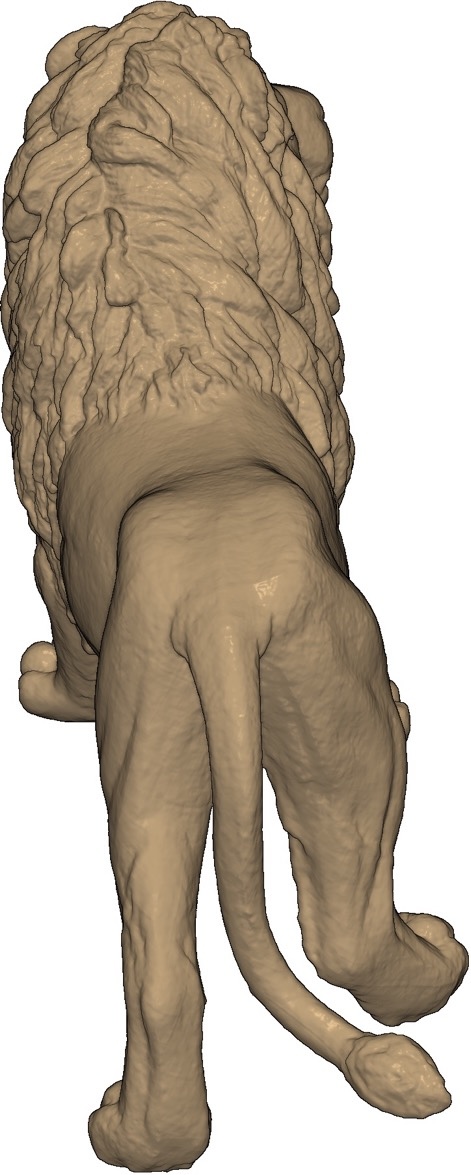}&
                        \includegraphics[width=\figwidthS\linewidth]{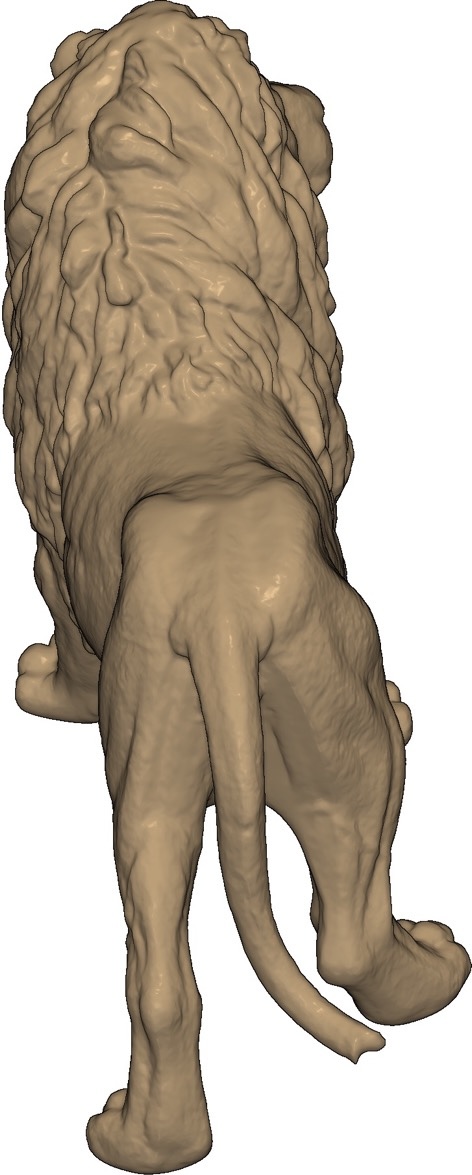}\\
                \includegraphics[width=\figwidthS\linewidth]{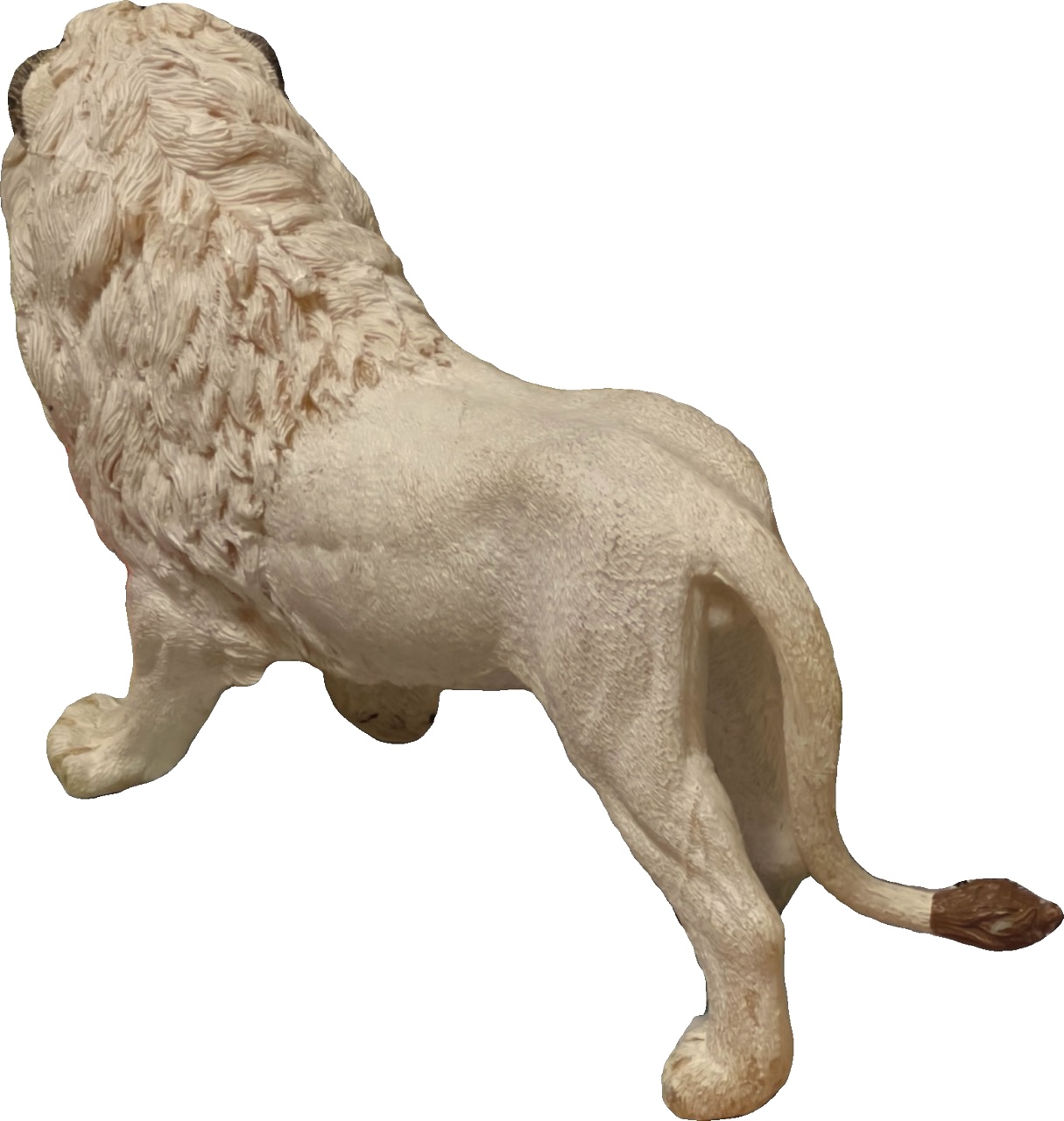}&
        \includegraphics[width=\figwidthS\linewidth]{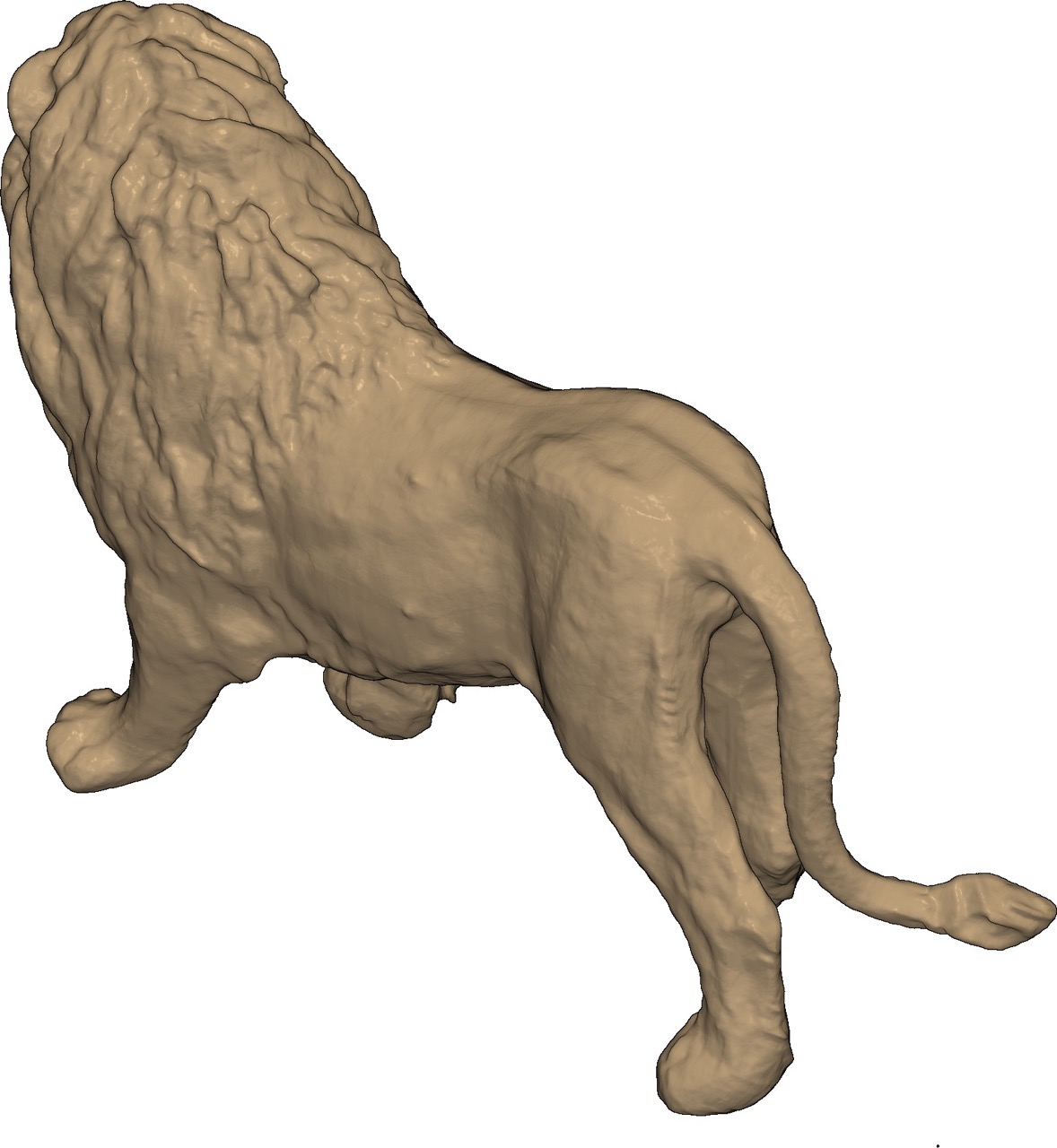}&
                \includegraphics[width=\figwidthS\linewidth]{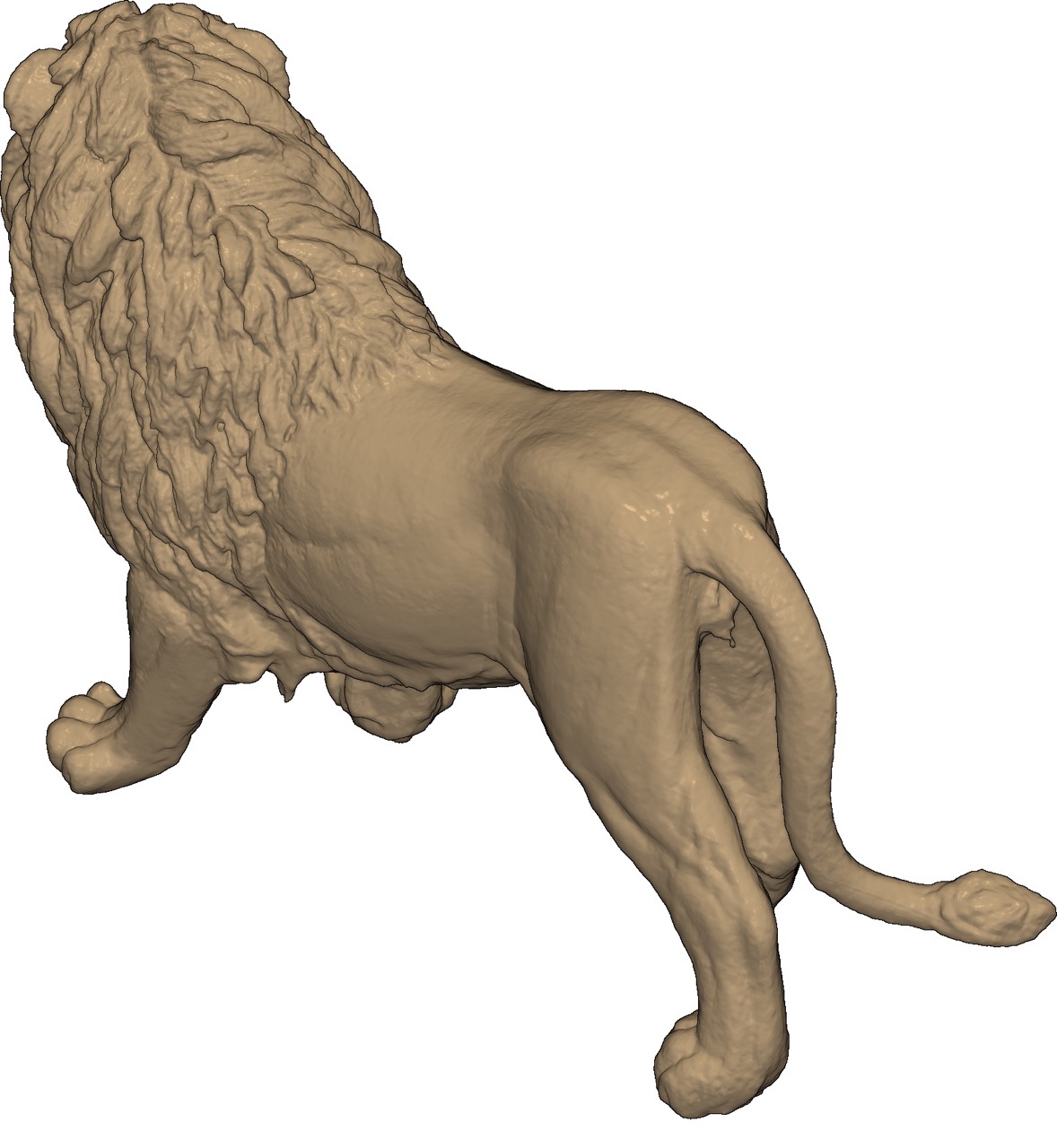}&
                        \includegraphics[width=\figwidthS\linewidth]{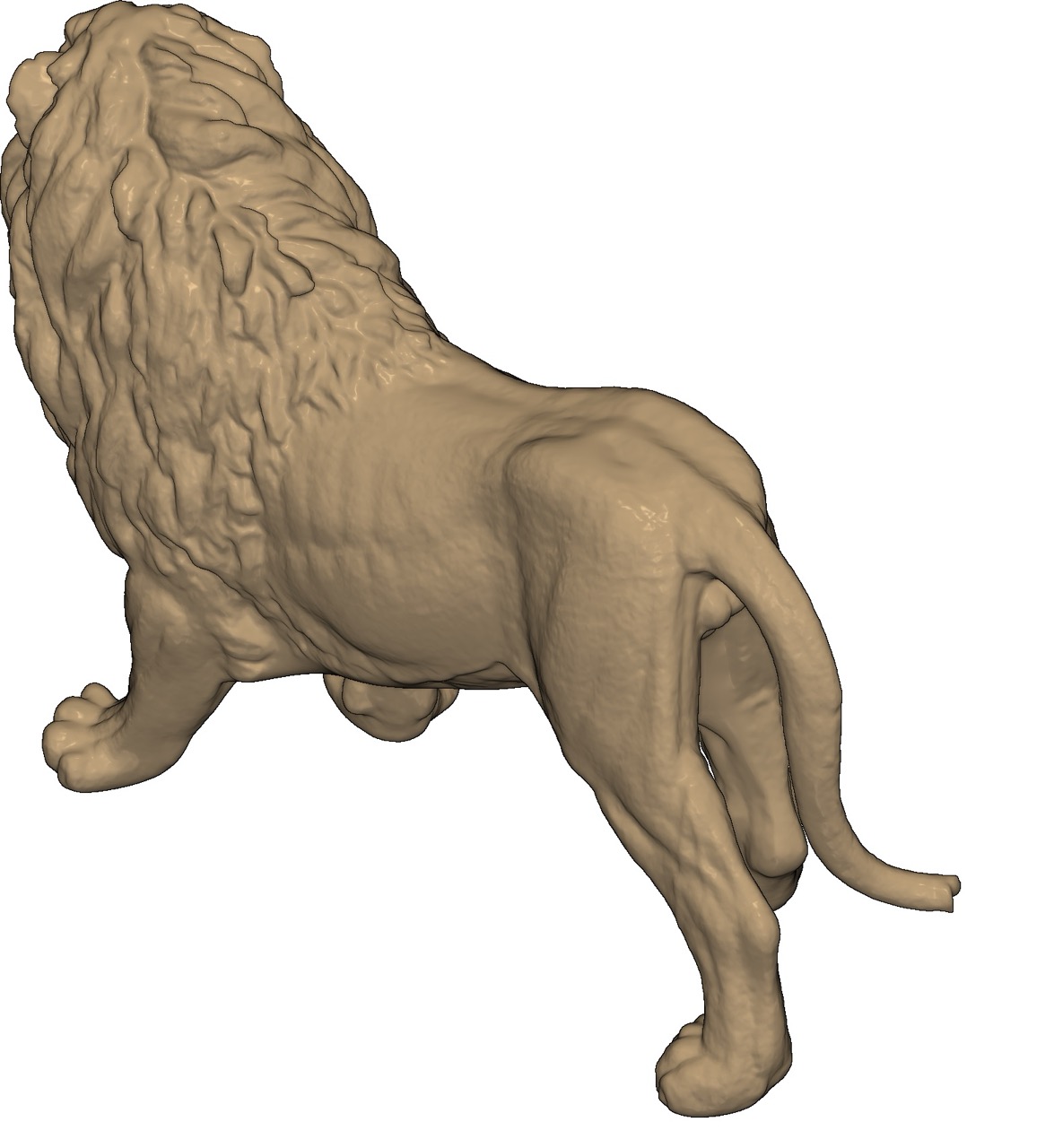}\\
    \end{tabular}
    \caption{Qualitative comparison on our captured object \emph{Lion}.}
    \label{fig.real_world_B}
\end{figure}

\begin{figure}
\newcommand{\figwidthS}{0.24}
    \centering
    \begin{tabular}{c@{}c@{}c@{}c}
        Image & \neustwo & Ours & EinScan SE \\
    \includegraphics[width=\figwidthS\linewidth]{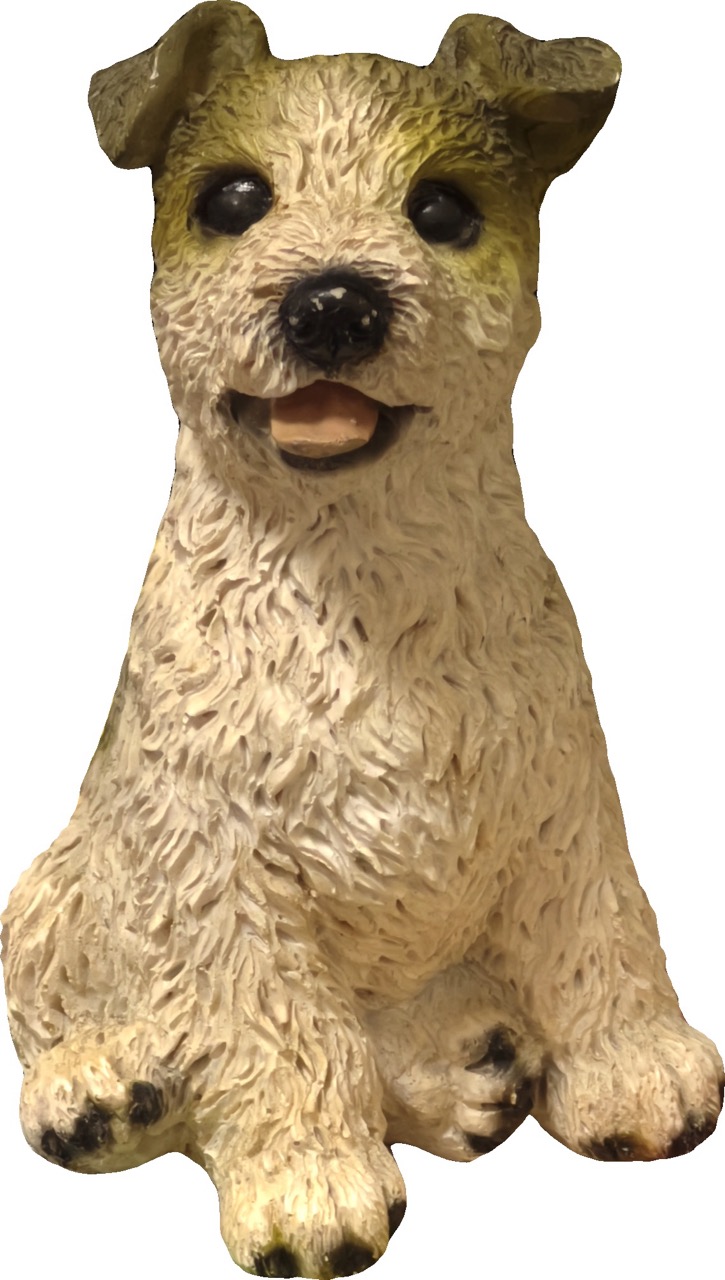}&
        \includegraphics[width=\figwidthS\linewidth]{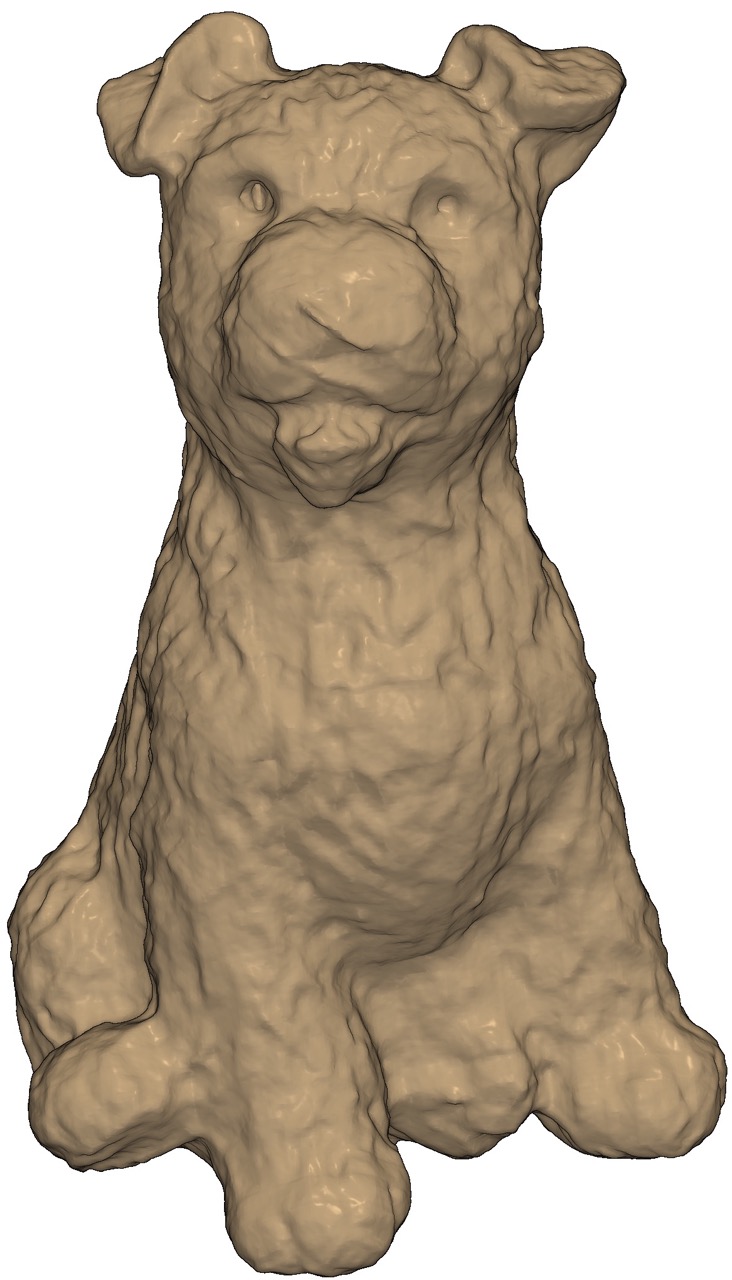}&
                \includegraphics[width=\figwidthS\linewidth]{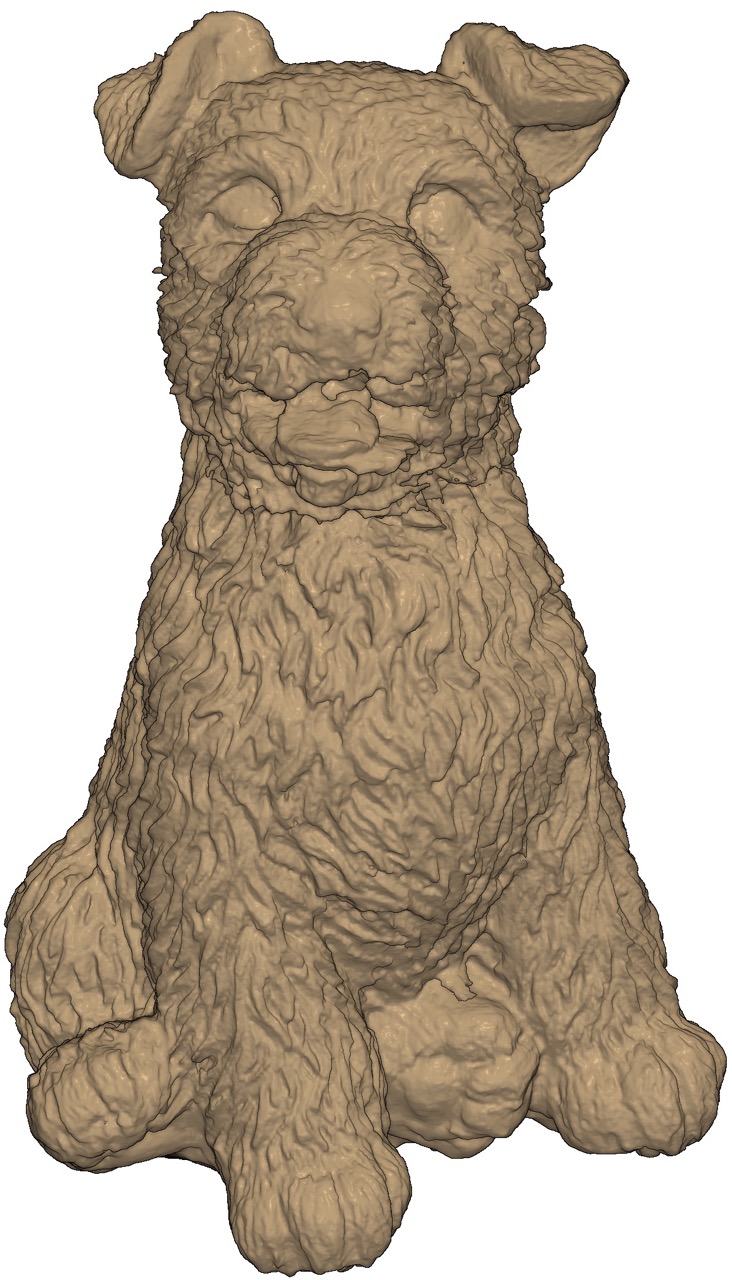}&
                        \includegraphics[width=\figwidthS\linewidth]{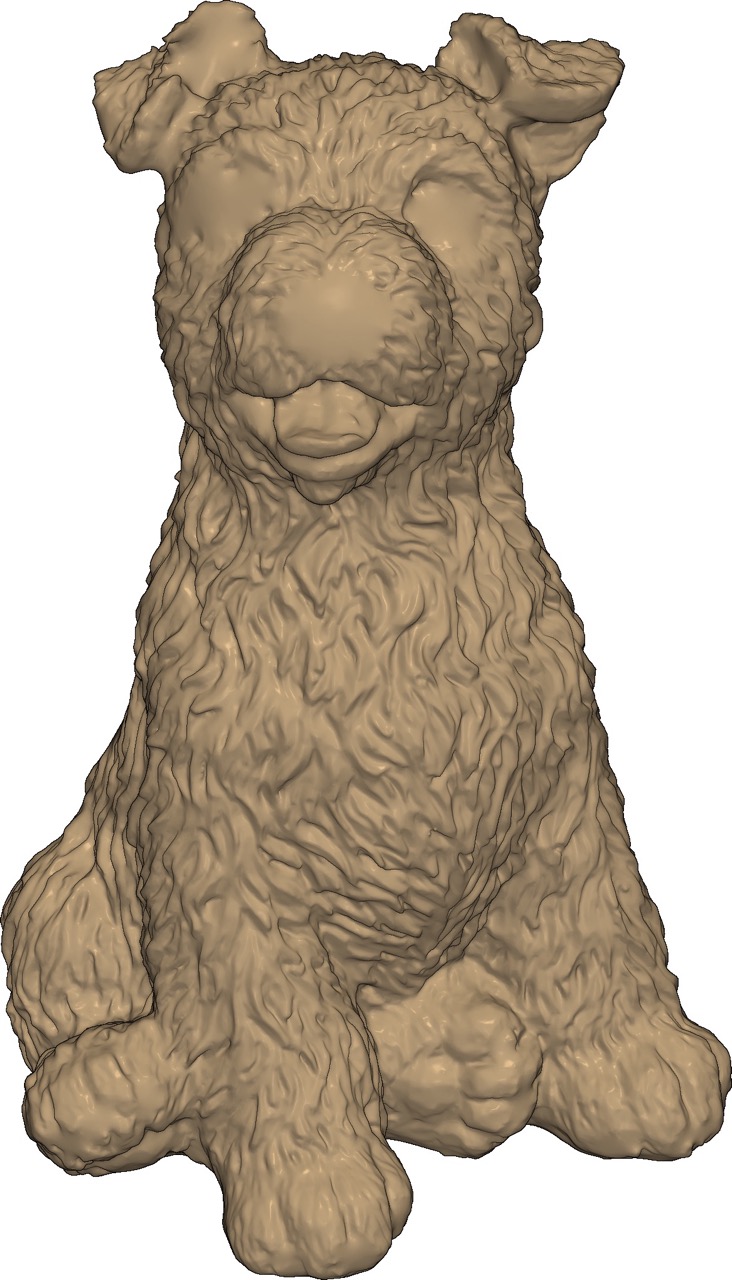}\\
        \includegraphics[width=\figwidthS\linewidth]{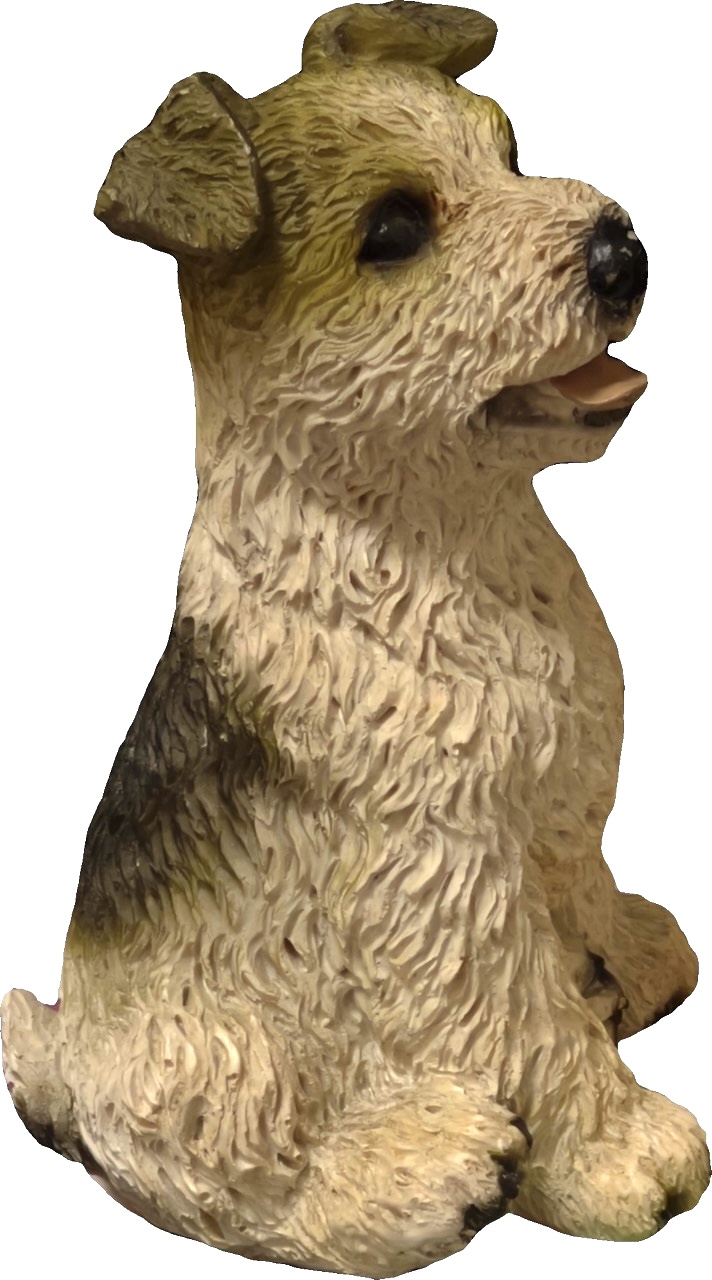}&
        \includegraphics[width=\figwidthS\linewidth]{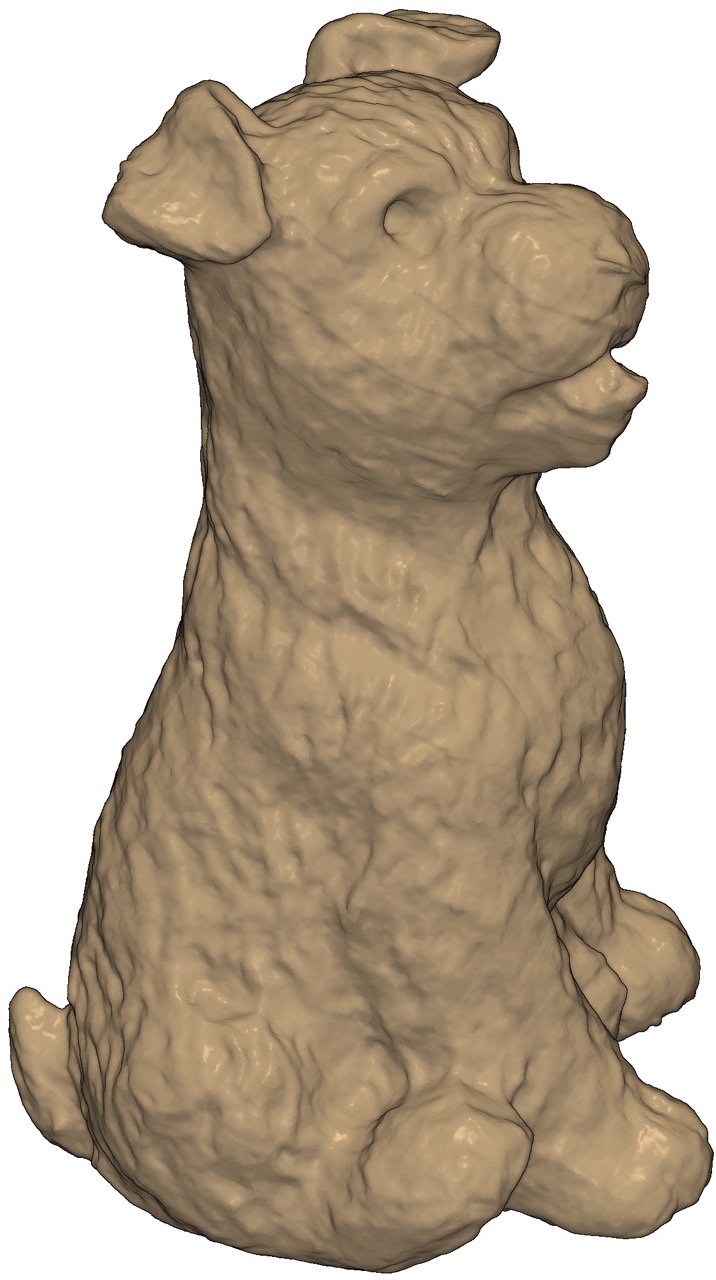}&
                \includegraphics[width=\figwidthS\linewidth]{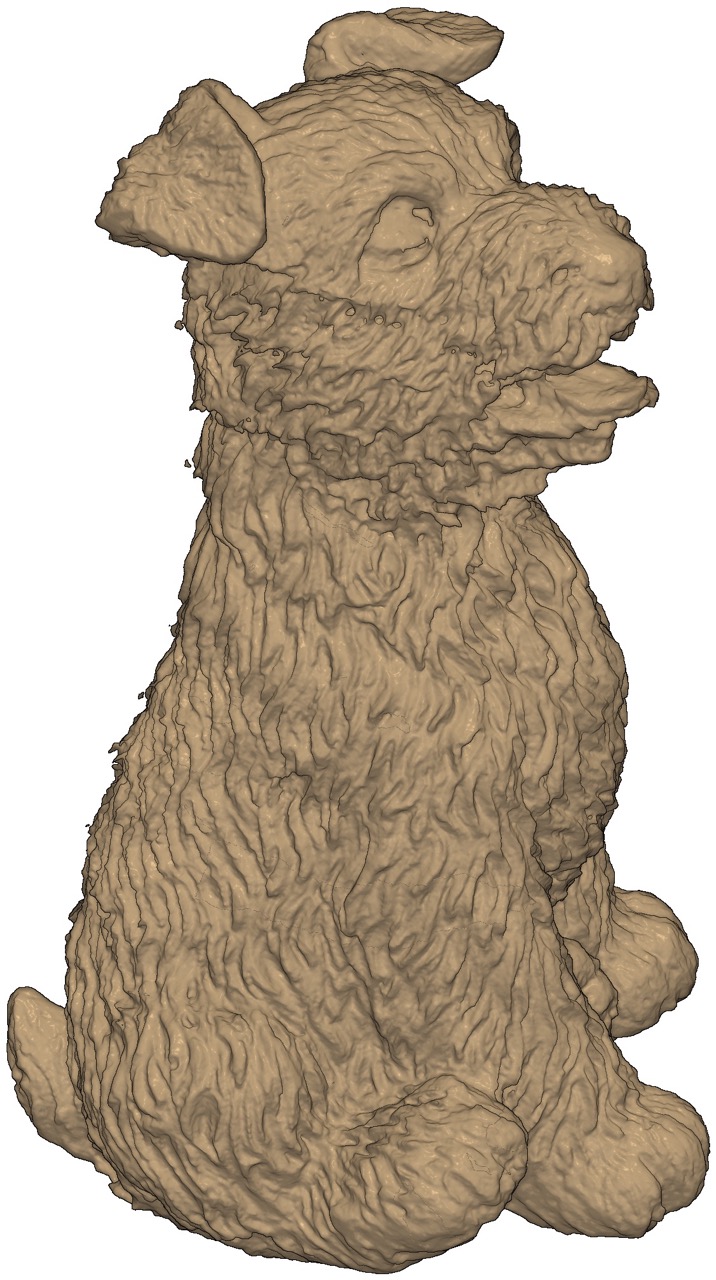}&
                        \includegraphics[width=\figwidthS\linewidth]{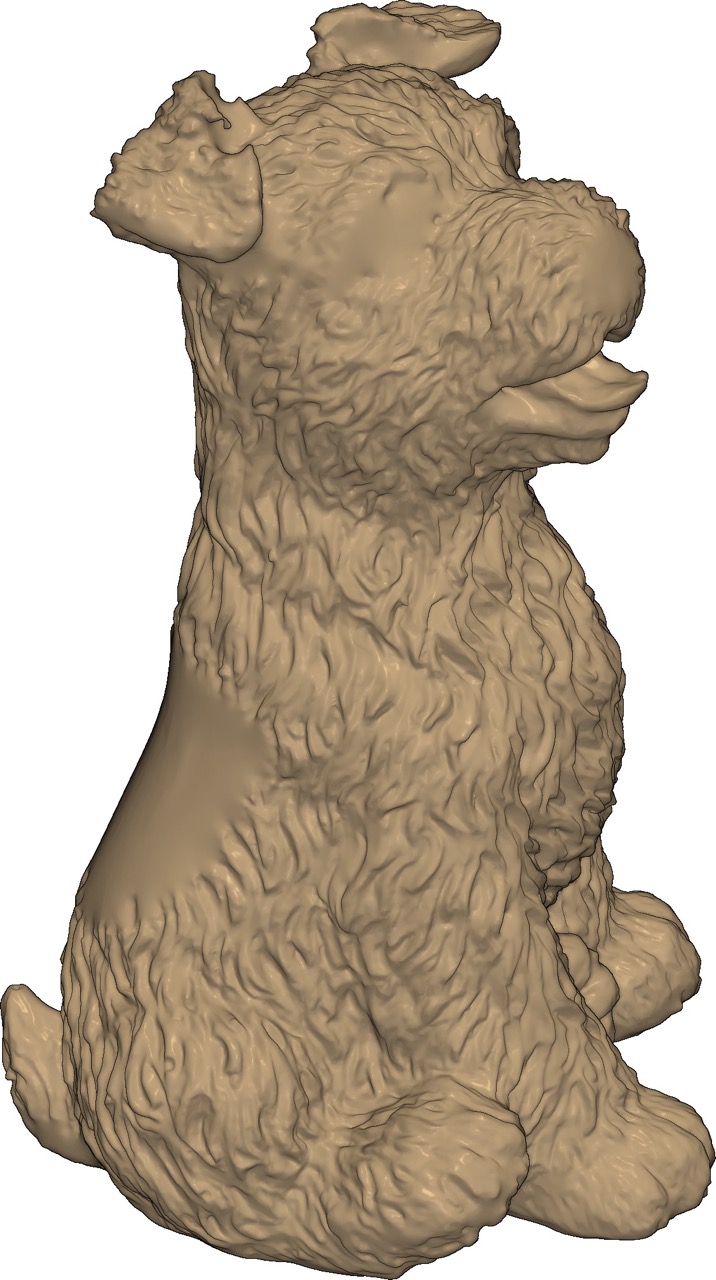}\\
        \includegraphics[width=\figwidthS\linewidth]{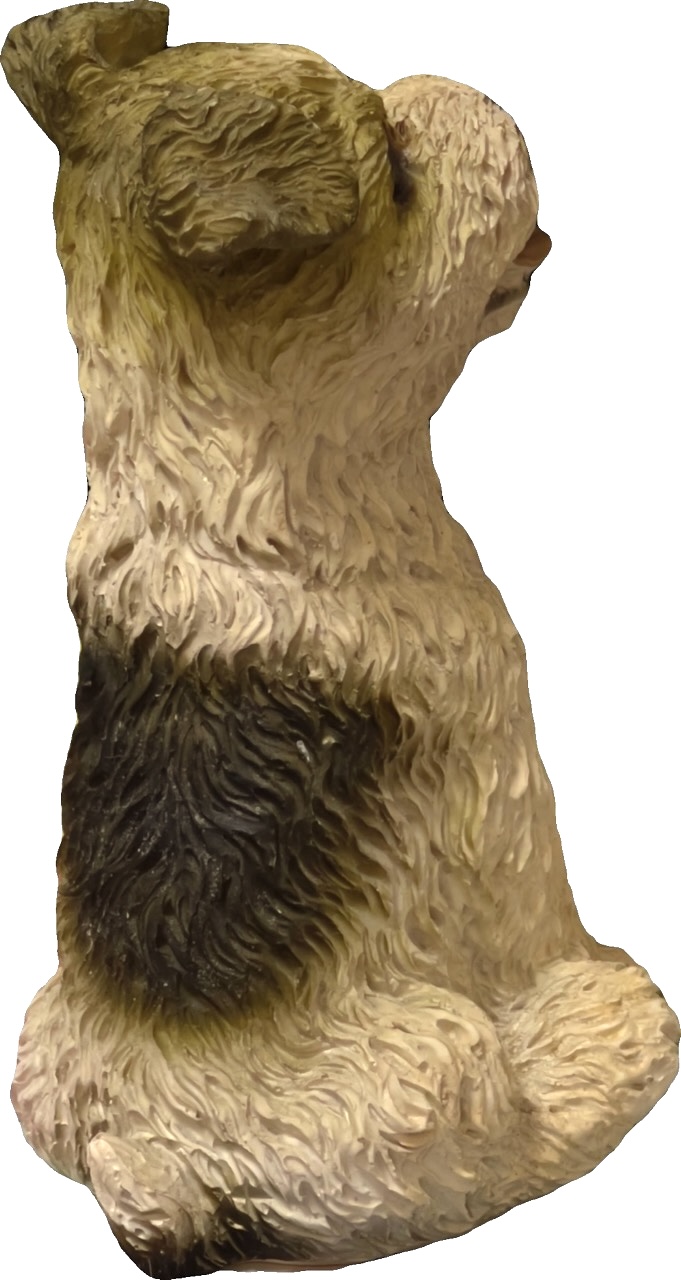}&
        \includegraphics[width=\figwidthS\linewidth]{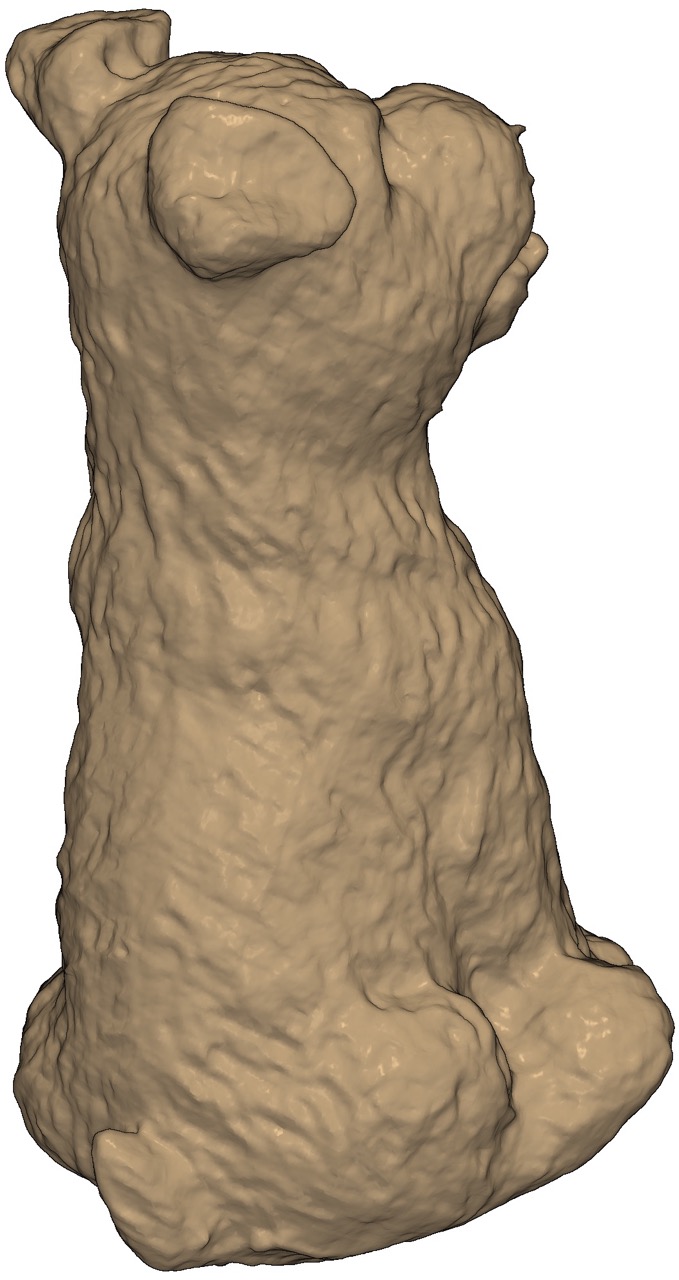}&
                \includegraphics[width=\figwidthS\linewidth]{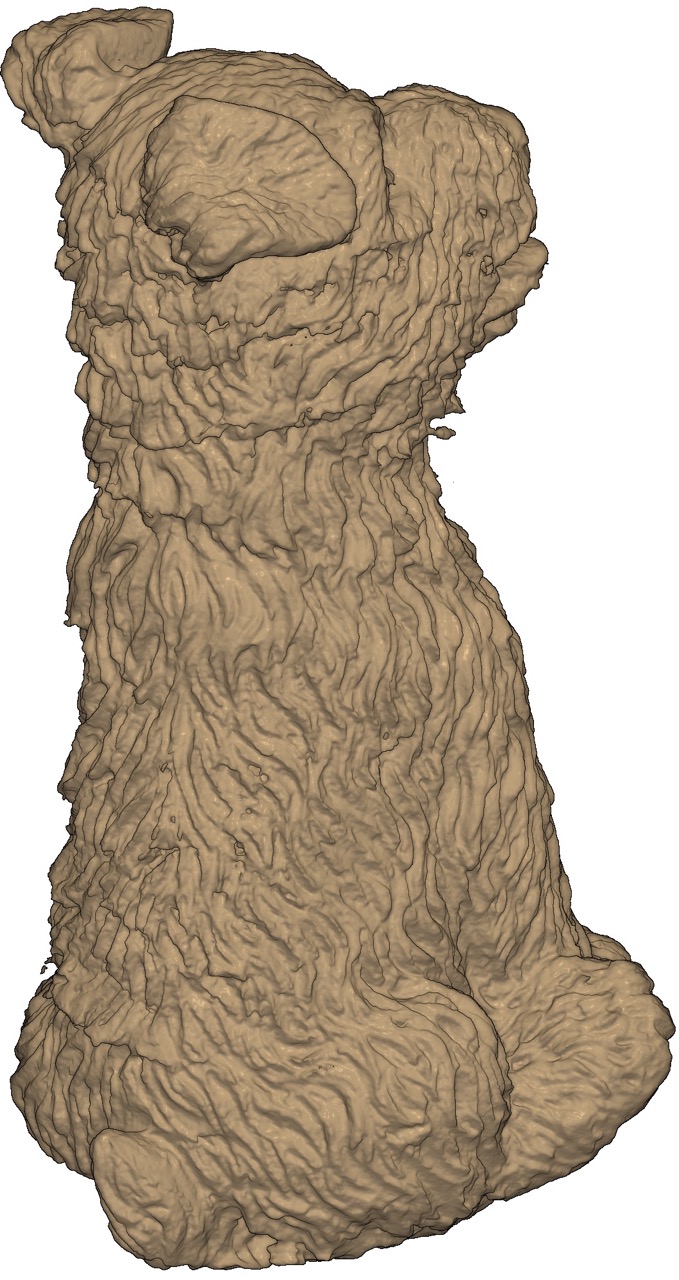}&
                        \includegraphics[width=\figwidthS\linewidth]{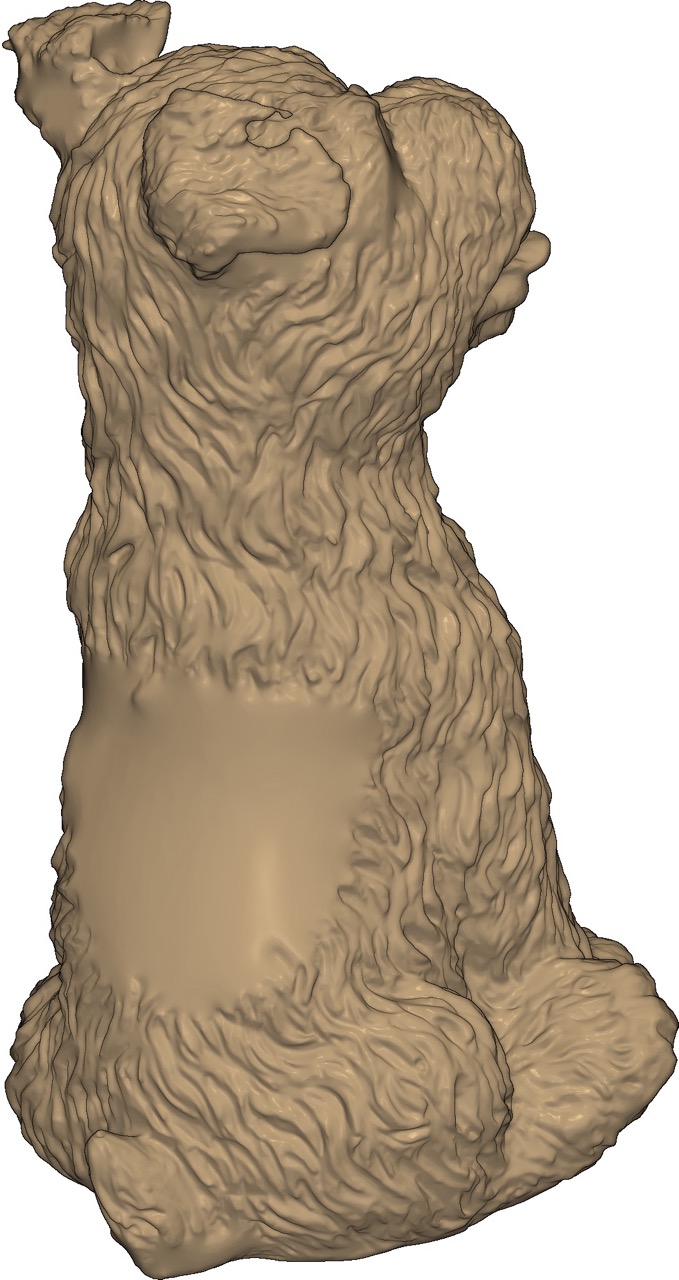}\\
            \includegraphics[width=\figwidthS\linewidth]{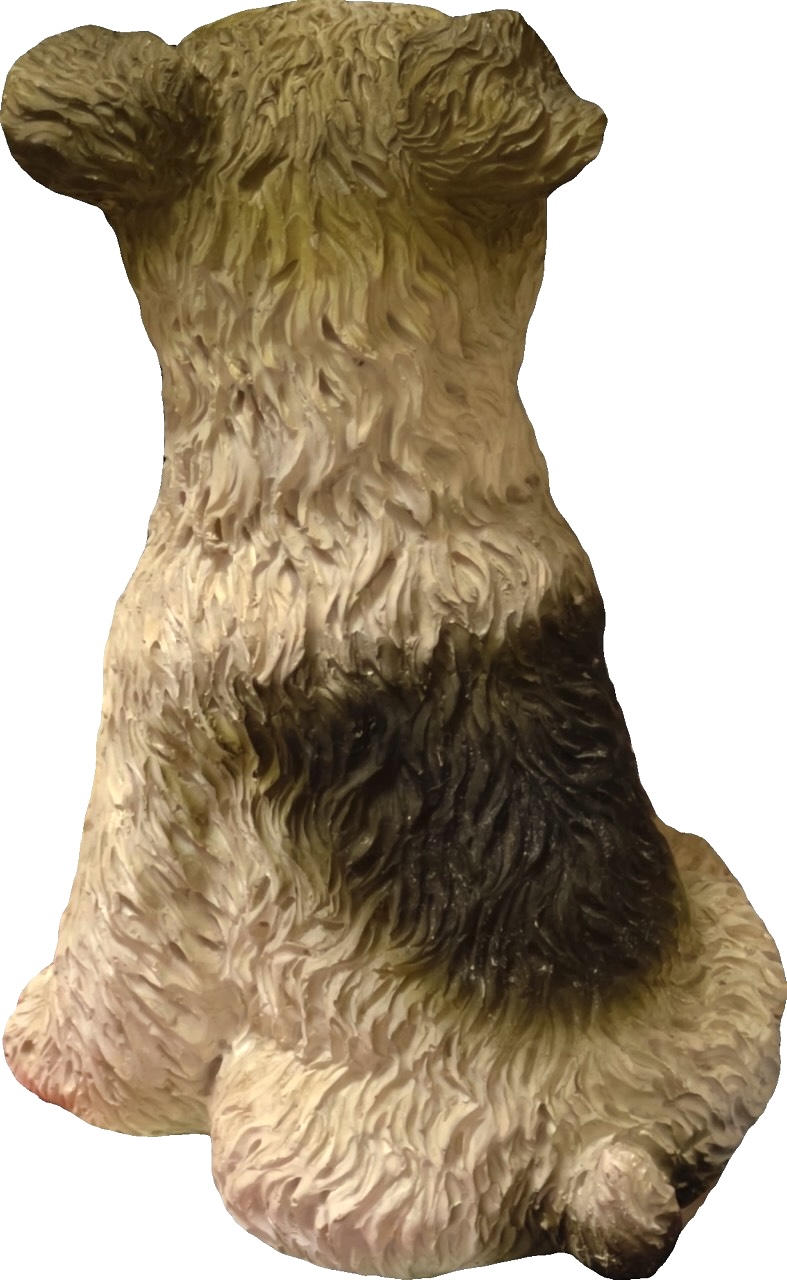}&
        \includegraphics[width=\figwidthS\linewidth]{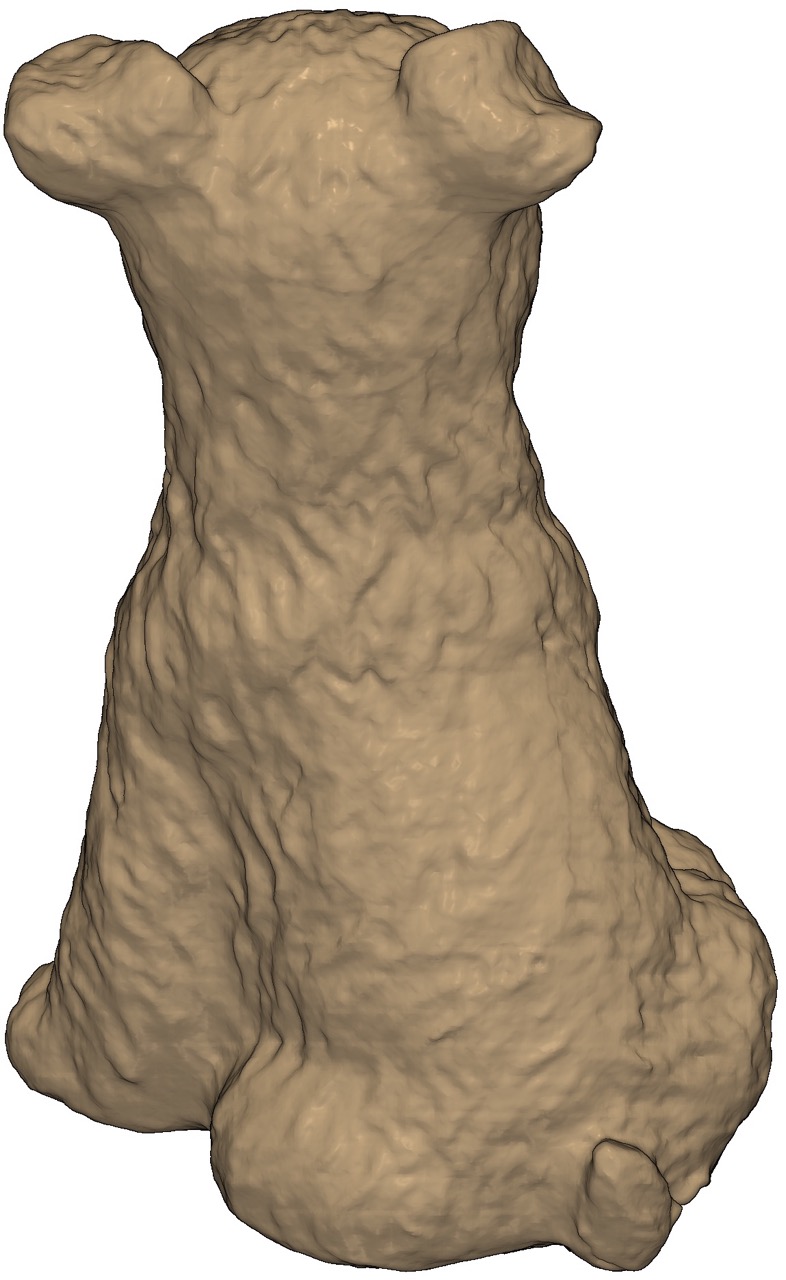}&
                \includegraphics[width=\figwidthS\linewidth]{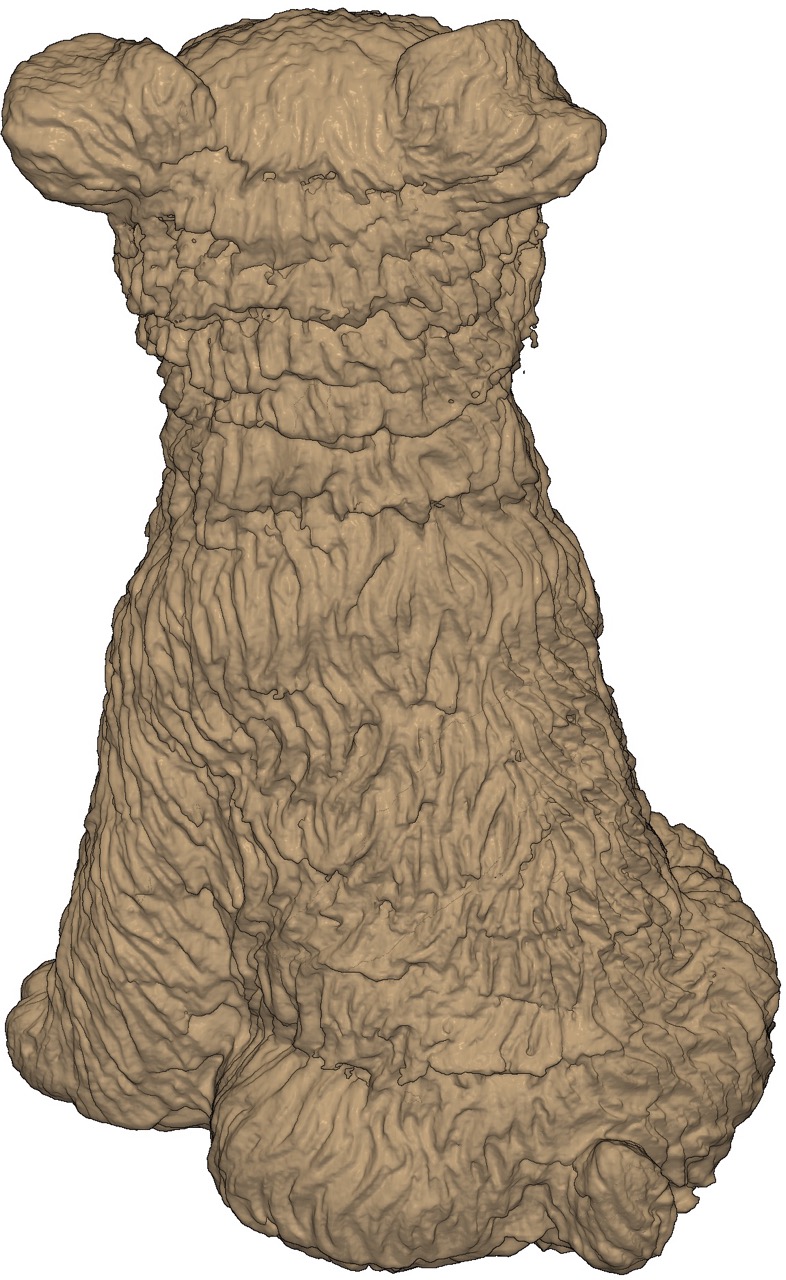}&
                        \includegraphics[width=\figwidthS\linewidth]{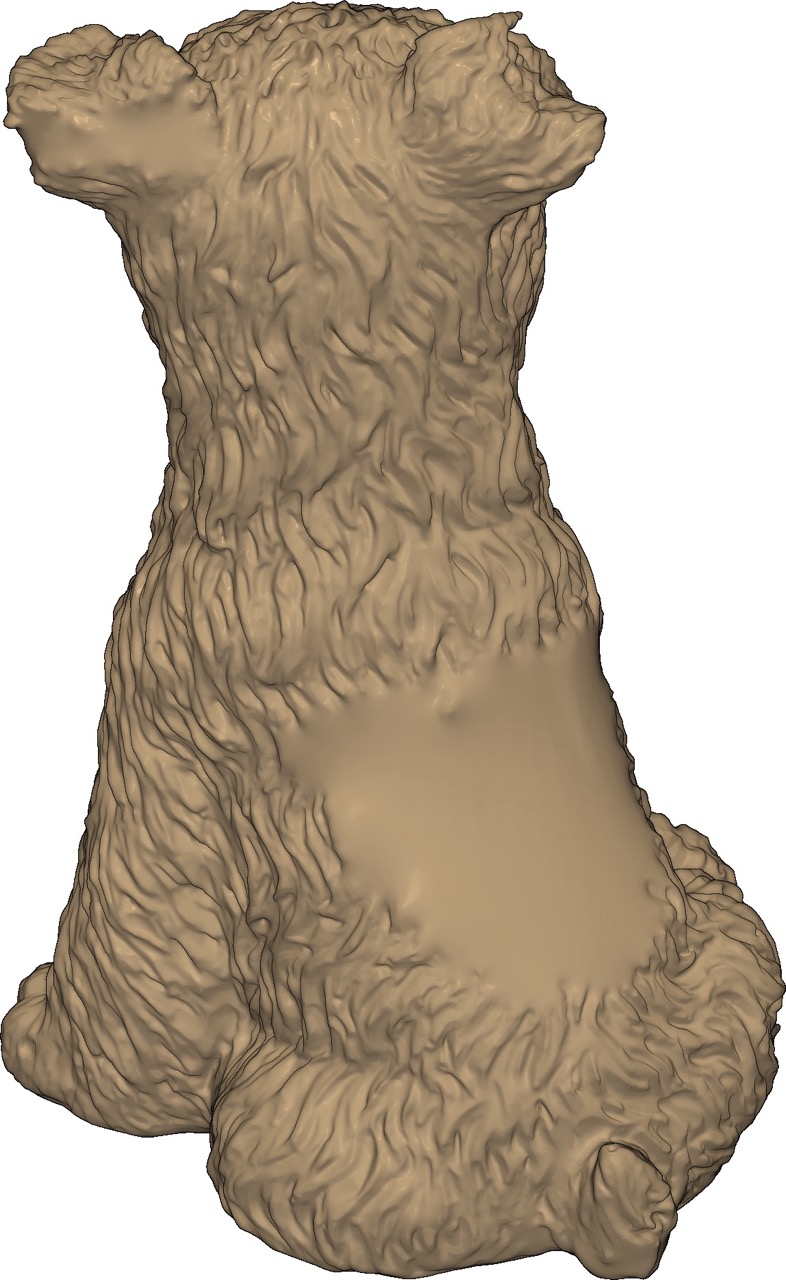}\\
            \includegraphics[width=\figwidthS\linewidth]{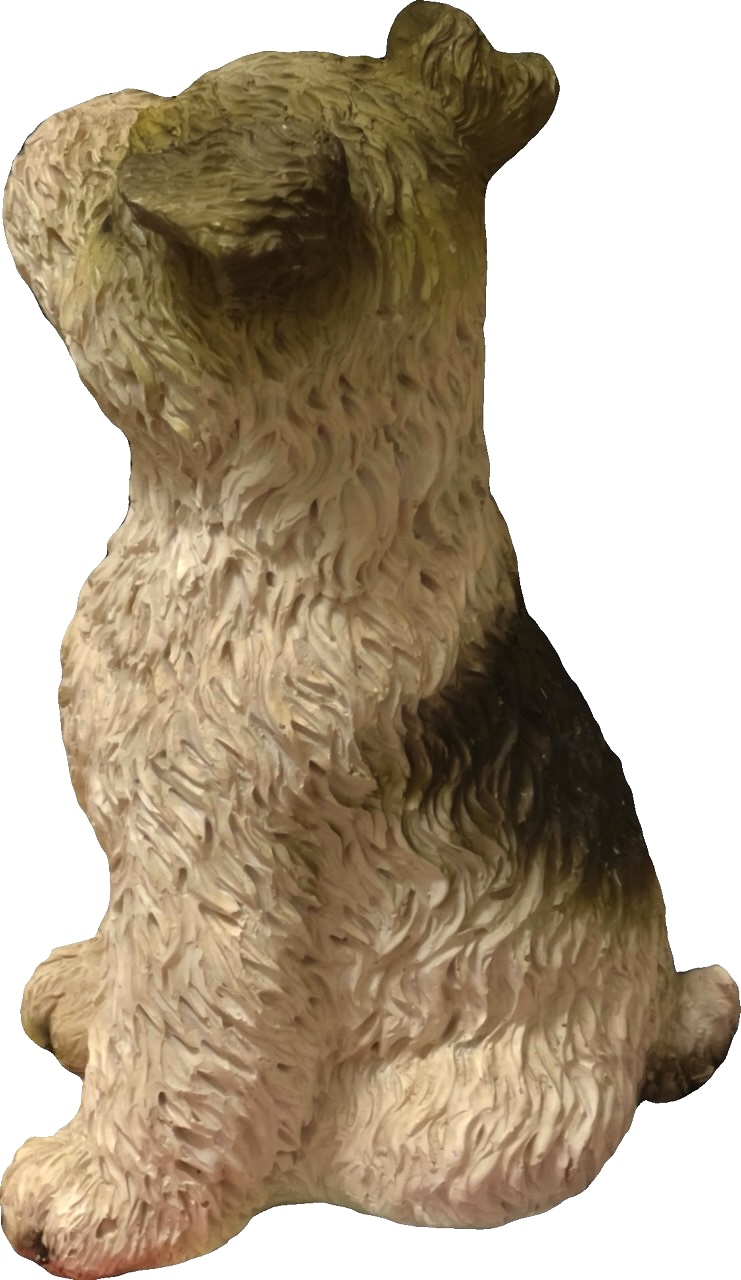}&
        \includegraphics[width=\figwidthS\linewidth]{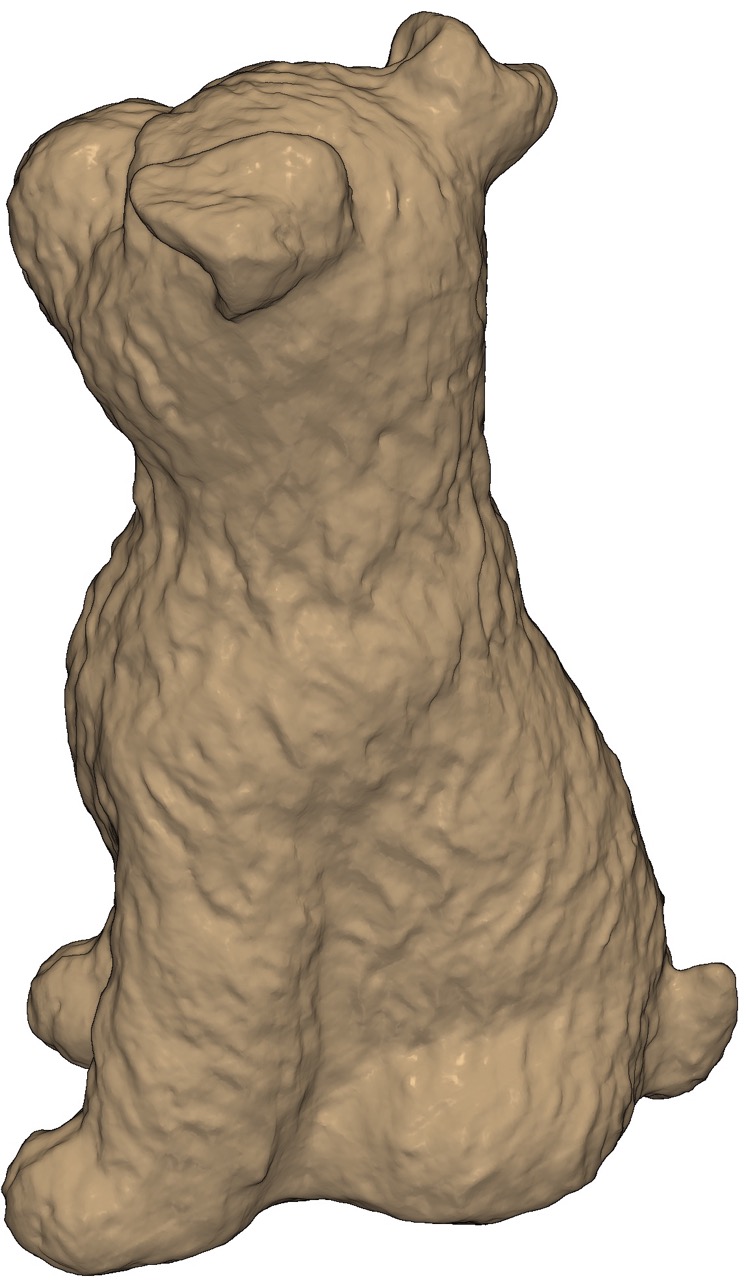}&
                \includegraphics[width=\figwidthS\linewidth]{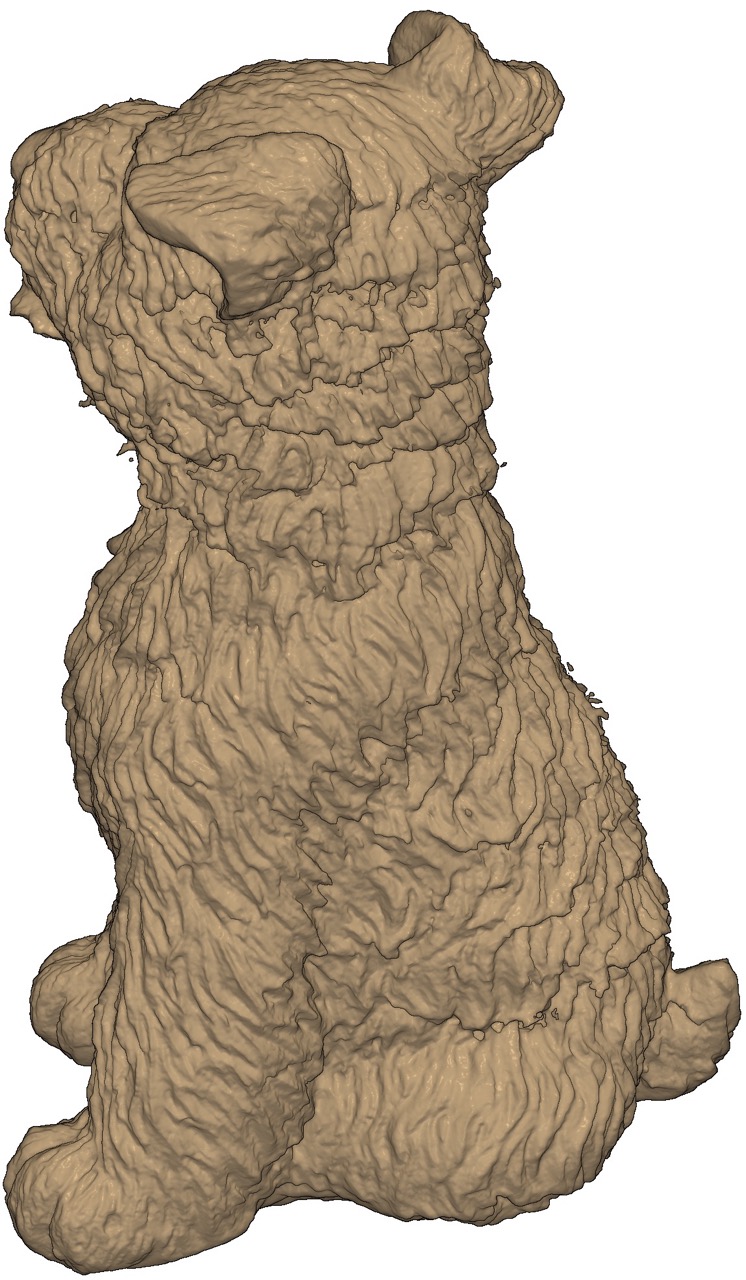}&
                        \includegraphics[width=\figwidthS\linewidth]{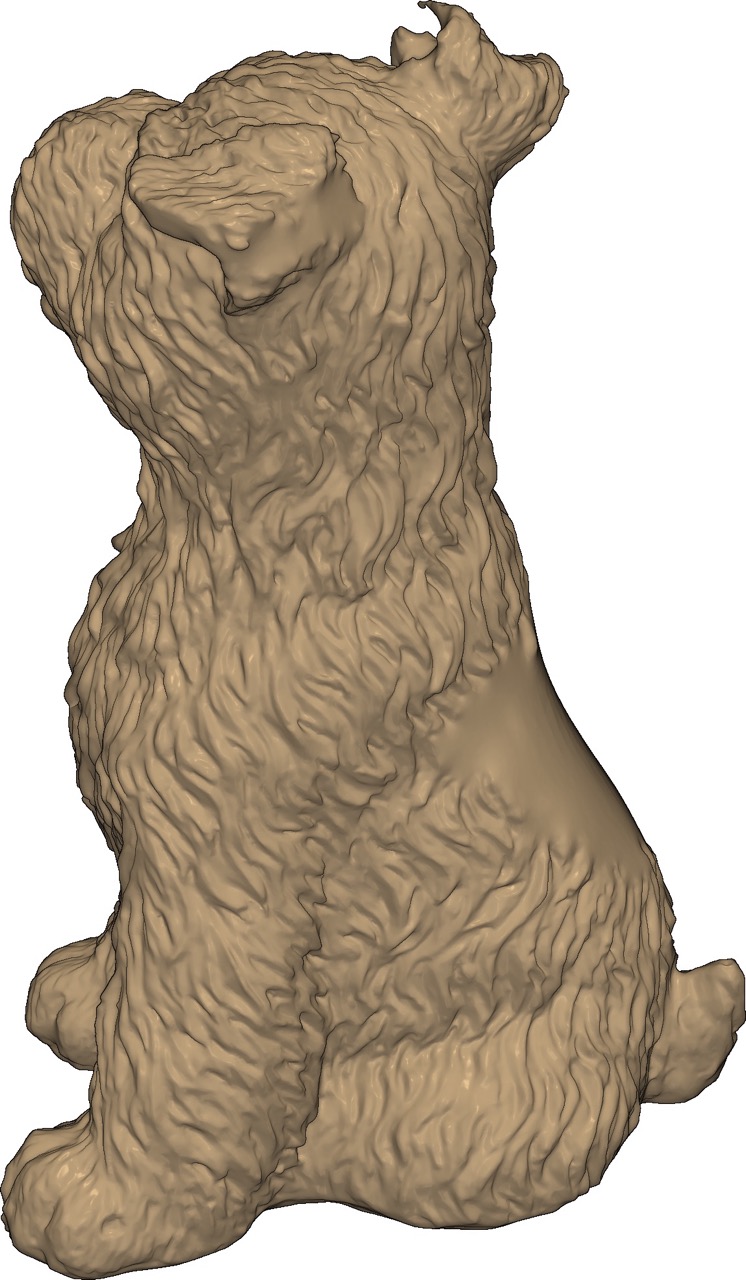}\\
            \includegraphics[width=\figwidthS\linewidth]{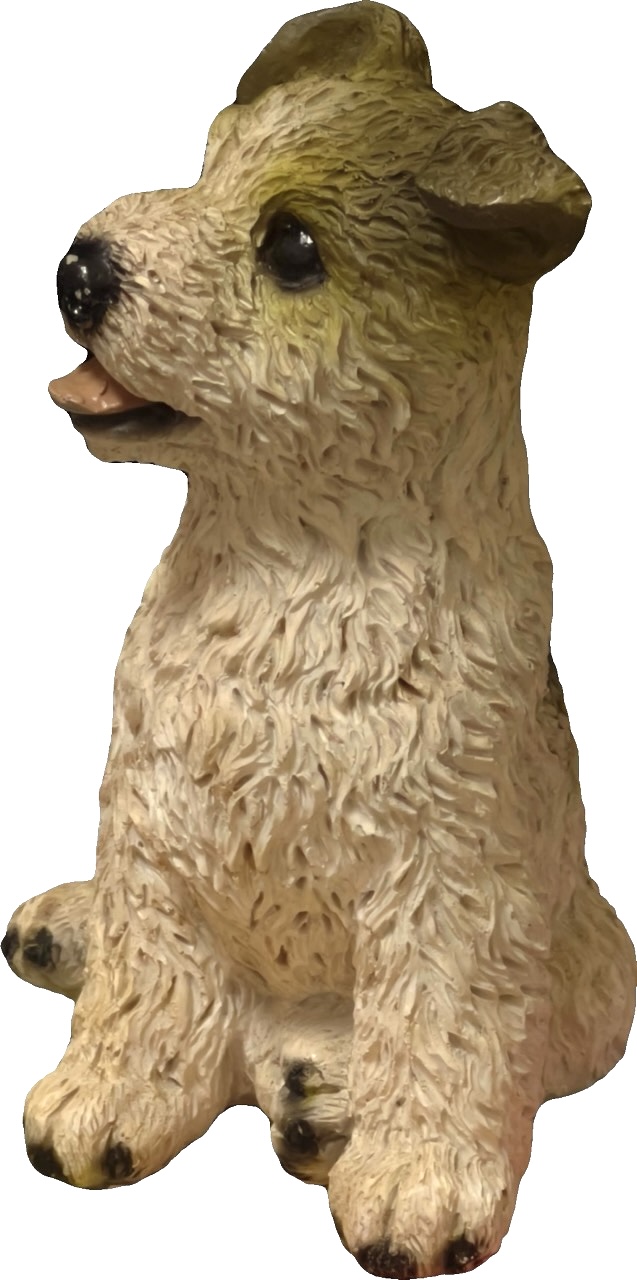}&
        \includegraphics[width=\figwidthS\linewidth]{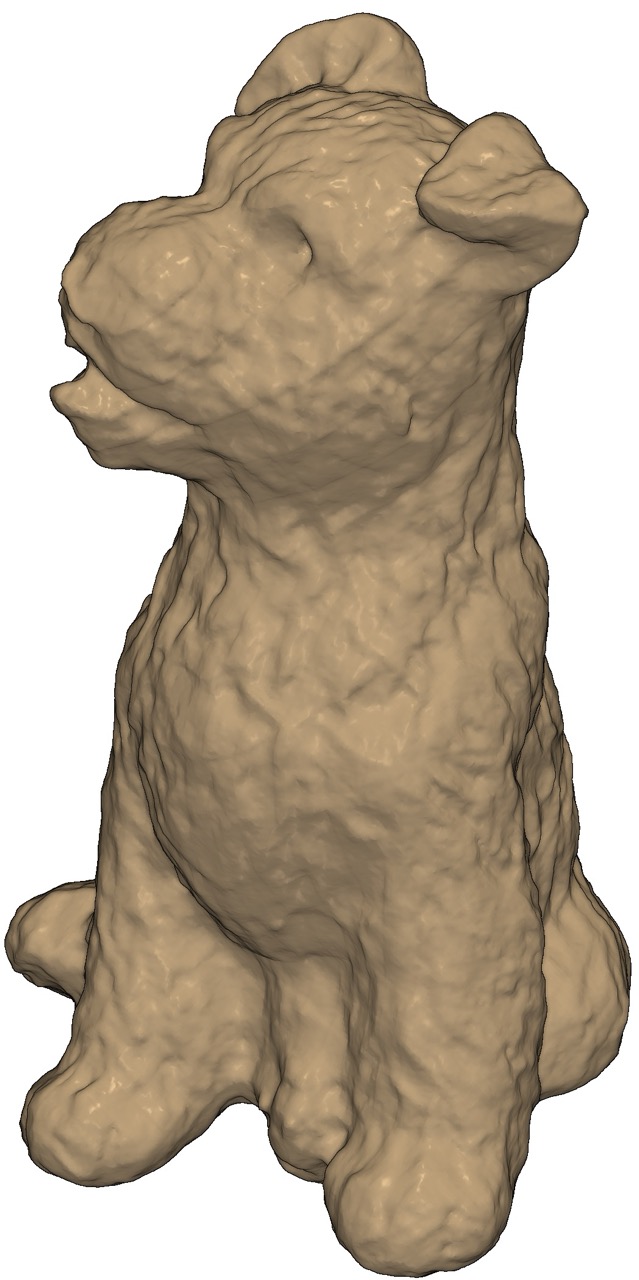}&
                \includegraphics[width=\figwidthS\linewidth]{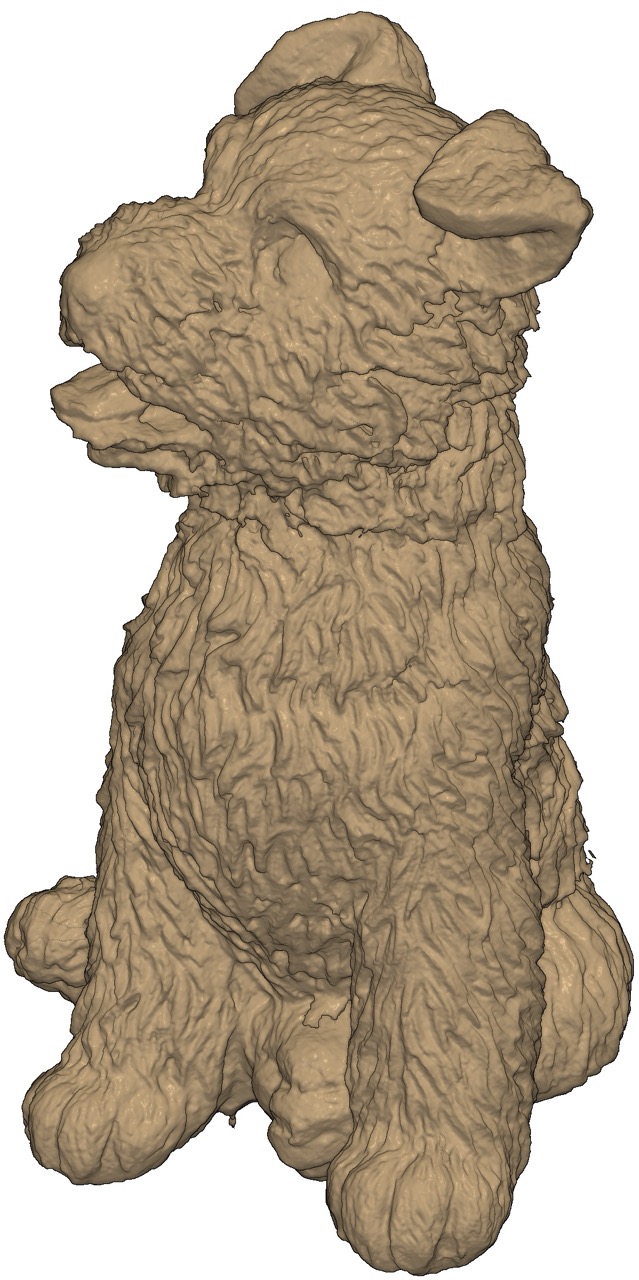}&
                        \includegraphics[width=\figwidthS\linewidth]{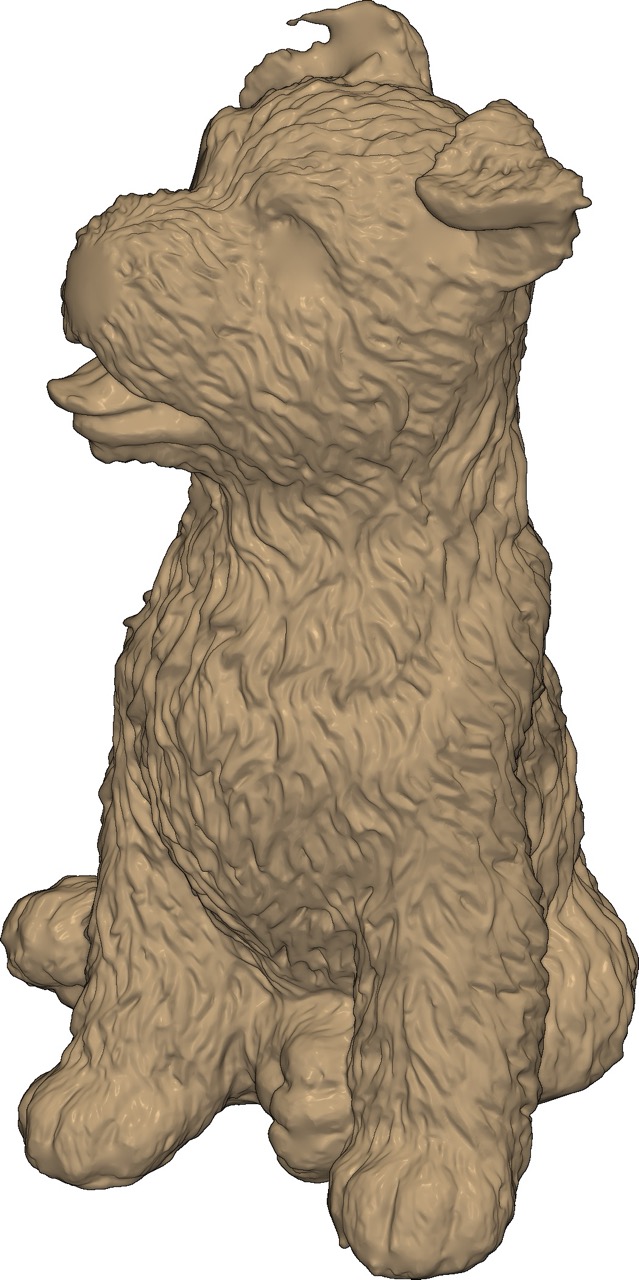}\\
    \end{tabular}
    \caption{Qualitative comparison on our captured object \emph{Dog}.}
    \label{fig.real_world_C}
\end{figure}

\Cref{fig.real_world_A,fig.real_world_B,fig.real_world_C} show more visualization results on our captured objects.
Overall, our method produces better surface detail than the MVS method \neustwo, and comparable results to the structured light based scanner.
The scanner struggles with dark or concave regions.
As shown in \cref{fig.real_world_C}, the details of the dark region of \emph{dog} are almost missing in the mesh reconstructed by the scanner.
Our method can sometimes produce fault artifacts, as seen in the hair part of \cref{fig.real_world_A} and the neck part in \cref{fig.real_world_C}.

\end{appendices}